\documentclass{article}
\usepackage{tabularx}

\PassOptionsToPackage{authoryear}{natbib}
\usepackage{arxiv}

\usepackage[american]{babel}
\usepackage[utf8]{inputenc} %
\usepackage[T1]{fontenc}    %
\usepackage{hyperref}
\usepackage{url}
\usepackage{hyperref}           %
\usepackage{booktabs}       %
\usepackage{longtable}      %
\usepackage{amsfonts}       %
\usepackage{nicefrac}       %
\usepackage{microtype}      %
\usepackage{subcaption}     %
\usepackage[table]{xcolor}         %

\usepackage{amsmath}
\usepackage{amssymb}
\usepackage{mathtools}
\usepackage{amsthm}
\usepackage{adjustbox}
\usepackage{etoc}
\usepackage{float}
\usepackage{researchpack}

\usepackage{microtype}
\usepackage{graphicx}
\usepackage[dvipsnames]{xcolor}
\usepackage{booktabs}
\usepackage{hyperref}
\usepackage{url}
\usepackage{multirow,rotating,booktabs}
\usepackage{thm-restate}
\usepackage{thmtools}
\usepackage[capitalise,nameinlink]{cleveref}
\usepackage{xspace}
\usepackage{enumitem}
\usepackage{natbib}

\usepackage{amsmath,amsfonts,bm}

\def\eqref#1{equation~\ref{#1}}

\def\1{\bm{1}}

\def\vc{{\bm{c}}}

\def\vk{{\bm{k}}}

\def\vs{{\bm{s}}}

\def\vx{{\bm{x}}}
\def\vy{{\bm{y}}}

\DeclareMathAlphabet{\mathsfit}{\encodingdefault}{\sfdefault}{m}{sl}
\SetMathAlphabet{\mathsfit}{bold}{\encodingdefault}{\sfdefault}{bx}{n}

\DeclareMathOperator*{\argmax}{arg\,max}

\definecolor{acoolcolor}{HTML}{815b9b}

\hypersetup{
    colorlinks=true,
    citecolor=acoolcolor,
    linkcolor=acoolcolor,
    urlcolor=acoolcolor,
}

\theoremstyle{definition}

\crefname{assumption}{Assumption}{Assumptions}
\Crefname{assumption}{Assumption}{Assumptions}

\newcommand{\SPL}{\texttt{SPL}\xspace}
\newcommand{\DIMACS}{\texttt{DIMACS}\xspace}
\newcommand{\nat}{\texttt{NAT}\xspace}
\newcommand{\pysat}{\texttt{PySAT}\xspace}
\newcommand{\cpmpy}{\texttt{CPMpy}\xspace}
\newcommand{\sympy}{\texttt{SymPy}\xspace}
\newcommand{\rsbench}{\texttt{rsbench}\xspace}
\newcommand{\autobench}{\texttt{auto-nesy-bench}\xspace}
\newcommand{\dcpbench}{\texttt{dcpbench}\xspace}
\newcommand{\satbench}{\texttt{satbench}\xspace}
\newcommand{\mc}{\textsc{MC}\xspace}
\newcommand{\frel}{\textsc{Rel}\xspace}
\newcommand{\con}{\textsc{Con}\xspace}
\newcommand{\mcr}{\textsc{MC-R}\xspace}
\newcommand{\HF}[1]{\href{https://huggingface.co/#1}{\texttt{#1}}}

\usepackage{pifont}
\newcommand{\cmark}{\ding{51}}
\newcommand{\xmark}{\ding{55}}
\newcommand{\CMARK}{\textcolor{teal}{\cmark}\xspace}
\newcommand{\XMARK}{\textcolor{WildStrawberry}{\xmark}\xspace}

\usepackage{fvextra}

\DefineVerbatimEnvironment{MyVerbatim}{Verbatim}{
  breaklines=true,
  breakanywhere=true
}

\usepackage{tikz}
\usetikzlibrary{positioning, shapes.callouts, arrows.meta, calc, backgrounds, fit, decorations.pathreplacing}
\definecolor{nodeTeal}{RGB}{0, 153, 168}
\definecolor{nodeOrange}{RGB}{245, 130, 32}
\definecolor{nodePurple}{RGB}{175, 160, 210}
\definecolor{darkSlate}{RGB}{70, 90, 100}
\definecolor{badgeRed}{RGB}{215, 38, 38}
\definecolor{bgLight}{RGB}{248, 249, 250}

\title{Auto-Formalizing Neuro-Symbolic Predictors}

\hypersetup{
    pdftitle={Auto-Formalizing Neuro-Symbolic Predictors},
    pdfauthor={Samuele Bortolotti, Weixin Chen, Han Zhao, Andrea Passerini, Stefano Teso, Antonio Vergari},
    pdfkeywords={Neuro-Symbolic AI, Auto-Formalization, Constraint Learning, Large Language Models, Probabilistic Logic, Semantic Probabilistic Layers, Semantic Loss, Probabilistic Machine Learning},
}

\author{
\hspace{-2em}
Samuele Bortolotti\textsuperscript{1,}
\thanks{
\begin{tabular}{@{}l@{}}
Corresponding author: \href{mailto:samuele.bortolotti@unitn.it}{\texttt{samuele.bortolotti@unitn.it}}. \\
{\scriptsize$\dagger$} Work started when Samuele Bortolotti was visiting the
\href{https://april-tools.github.io/}{APRIL} lab, University of Edinburgh.\\
{\scriptsize$\MZero$} Equal contribution.\\
{\scriptsize$\MOne$} Shared supervision.
\end{tabular}
}\textsuperscript{, $\MZero$},
Weixin Chen\textsuperscript{2, $\MZero$},
Han Zhao\textsuperscript{2},
Andrea Passerini\textsuperscript{1},
Stefano Teso\textsuperscript{1, $\MOne$},
Antonio Vergari\textsuperscript{3, $\MOne$}
\\
\textsuperscript{1}University of Trento, Italy
\quad
\textsuperscript{2}University of Illinois Urbana-Champaign, USA
\quad
\textsuperscript{3}University of Edinburgh, UK
\\
\texttt{\{samuele.bortolotti,andrea.passerini,stefano.teso\}@unitn.it}
\\
\texttt{\{weixinc2,hanzhao\}@illinois.edu}
\quad
\texttt{avergari@ed.ac.uk}
}

\everypar=\expandafter{\the\everypar\looseness=-1 }  %
\begin{document}

\maketitle

\begin{abstract}
Neuro-Symbolic (NeSy) predictors incorporate prior knowledge into the prediction process of neural networks, ensuring that outputs satisfy specified constraints, making them particularly suitable for high-stakes applications where compliance with domain knowledge is essential.
A key bottleneck in this paradigm is the acquisition of symbolic constraints: encoding domain knowledge into logical formulas remains a manual and expert-intensive process.
In this work, we investigate the extent to which auto-formalization via LLMs can systematically translate textual knowledge into symbolic knowledge that can be plugged into NeSy predictors.
To this end, we introduce \autobench, a new benchmark for evaluating constraint formalization and its impact on downstream accuracy of NeSy predictors. %
Through an extensive evaluation across several domains, we find that LLMs can formalize constraints to a meaningful extent, generating formulas that are often similar to those provided by human experts. Moreover, when the generated formulas are syntactically valid, they can lead to high-quality downstream predictions. The code and benchmark are available at \url{https://unitn-sml.github.io/auto-nesy-bench/}.
\end{abstract}

\section{Introduction}

Neuro-Symbolic (NeSy) AI combines the learning capabilities of neural networks with the reasoning capabilities of symbolic systems, making it a promising paradigm for building reliable AI models~\citep{de2021statistical, garcez2022neural}.
A prominent and well-studied class of NeSy models is that of \textit{NeSy predictors}~\citep{manhaeve2018deepproblog, badreddine2022logic, chen2026npc}, which incorporate prior knowledge into the prediction process to ensure that outputs comply with domain-specific constraints~\citep{giunchiglia2020coherent, giunchiglia2024ccn, chen2026vlc}, making them suitable for high-stakes applications~\citep{giunchiglia2022deep, giunchiglia2022learning, dash2022review}.

Such models typically assume that the \emph{constraints are provided beforehand}, requiring a domain expert to identify the relevant knowledge and formalize it into a representation that can be incorporated into the predictor.
This requirement persists even in recent frameworks such as \texttt{DeepLog}~\citep{derkinderen2026deeplog} and \texttt{ULLER}~\citep{van2024uller}, which aim to make NeSy predictors more accessible to practitioners, as the formalization of logical constraints still falls on the end user.
This raises a natural question: \textit{what if an end user wants to deploy a NeSy predictor but has no expert available to formalize the required knowledge?}
We address this gap by leveraging Large Language Models (LLMs) as domain experts to formalize natural language into symbolic constraints.

Recent work has shown that LLMs can generate constraint programs directly from natural language descriptions~\citep{tsouros2023holy, michailidis2024constraint}. 
Existing benchmarks target either Boolean satisfiability (SAT), such as \satbench~\citep{wei2025satbench}, or CP, such as \dcpbench~\citep{michailidis2025cp}, but do not evaluate how generated constraints affect learning and downstream NeSy performance.
To address this gap, we introduce \autobench, a challenging NeSy benchmark that extends the well-established \rsbench~\citep{bortolotti2024benchmark} with prompts, data instances, output annotations, and evaluation metrics in which the task is to formalize a logical constraint in conjunctive normal form (CNF) out of a textual description (see \cref{fig:nesy-autoform-overview}). 

We therefore evaluate both the quality of constraints generated by modern and frontier LLMs on \satbench, \dcpbench, and \autobench, as well as their downstream impact when integrated into state-of-the-art NeSy predictors, such as \textit{semantic probabilistic layers}~\citep[\SPL]{ahmed2022semantic}, or employed as soft constraints using \textit{semantic loss}~\citep[\SL]{xu2018semantic} on \autobench.
To instantiate this evaluation, the LLM formalizes the constraints in propositional logic (PL) in one of several formalisms ranging from a textual representation of a CNF such as \DIMACS~\citep{johnson1996cliques} or natural language (\nat) %
to Python programs using \pysat~\citep{imms-sat18}, \cpmpy~\citep{guns2019increasing} and \sympy~\citep{sympy2017meurer}, %
each one coming with different challenges (see~\cref{sec:constraint-formalization}).
Our results show that high-performing LLMs can formalize NeSy tasks, but performance depends on the level of detail in the constraint description. When the generated formulas are syntactically valid, formalization errors can have little impact on downstream predictions.

\textbf{Our contributions}: 
We introduce \autobench, a benchmark for evaluating auto-formalization in NeSy tasks.  
We investigate the following research questions: (\textbf{RQ1}) Can LLMs formalize logical constraints from natural language descriptions? and (\textbf{RQ2}) How do NeSy models perform when provided with LLM-generated constraints? To address \textbf{RQ1}, we analyze the problem from multiple perspectives, examining which types of (a) LLMs, (b) input formats, and (c) output formats lead to the best results, as well as which configurations fail. For \textbf{RQ2}, we evaluate two representative state-of-the-art NeSy approaches, a layer and a loss, using LLM-generated constraints. %

\definecolor{nodeTeal}{RGB}{0, 153, 168}
\definecolor{nodeOrange}{RGB}{245, 130, 32}
\definecolor{nodePurple}{RGB}{175, 160, 210}
\definecolor{darkSlate}{RGB}{70, 90, 100}
\definecolor{badgeRed}{RGB}{215, 38, 38}
\definecolor{bgLight}{RGB}{248, 249, 250}
\definecolor{badgeGreen}{RGB}{20, 130, 90}
\definecolor{formatBg}{RGB}{240, 245, 247}

\begin{figure}
    \centering
    \scalebox{.9}{
    \begin{tikzpicture}[
        font=\sffamily\small,
        tealnode/.style={circle, draw=nodeTeal, line width=1.5pt, fill=white, minimum size=4mm, inner sep=0pt},
        orangenode/.style={circle, draw=nodeOrange, line width=1.5pt, fill=white, minimum size=4mm, inner sep=0pt},
        purplenode/.style={circle, draw=nodePurple, line width=1.5pt, fill=white, minimum size=4mm, inner sep=0pt},
        block/.style={rectangle, draw=darkSlate, line width=1pt, fill=bgLight, rounded corners=4pt, inner sep=4pt, align=center},
        myarrow/.style={->, >=Stealth, line width=1.1pt, color=darkSlate},
        formatbox/.style={rectangle, draw=darkSlate, densely dotted, line width=0.7pt, fill=formatBg, rounded corners=2pt, minimum height=1.55cm, inner xsep=3pt, inner ysep=3pt, align=left, font=\ttfamily\fontsize{6}{7}\selectfont},
        formatline/.style={line width=0.7pt, color=darkSlate},
        speech/.style={rectangle callout, draw=darkSlate, line width=1pt, fill=white, align=left, inner sep=6pt, font=\scriptsize, rounded corners=2pt}
    ]

    \node (scene) at (1,5.3) {\includegraphics[width=2.5cm]{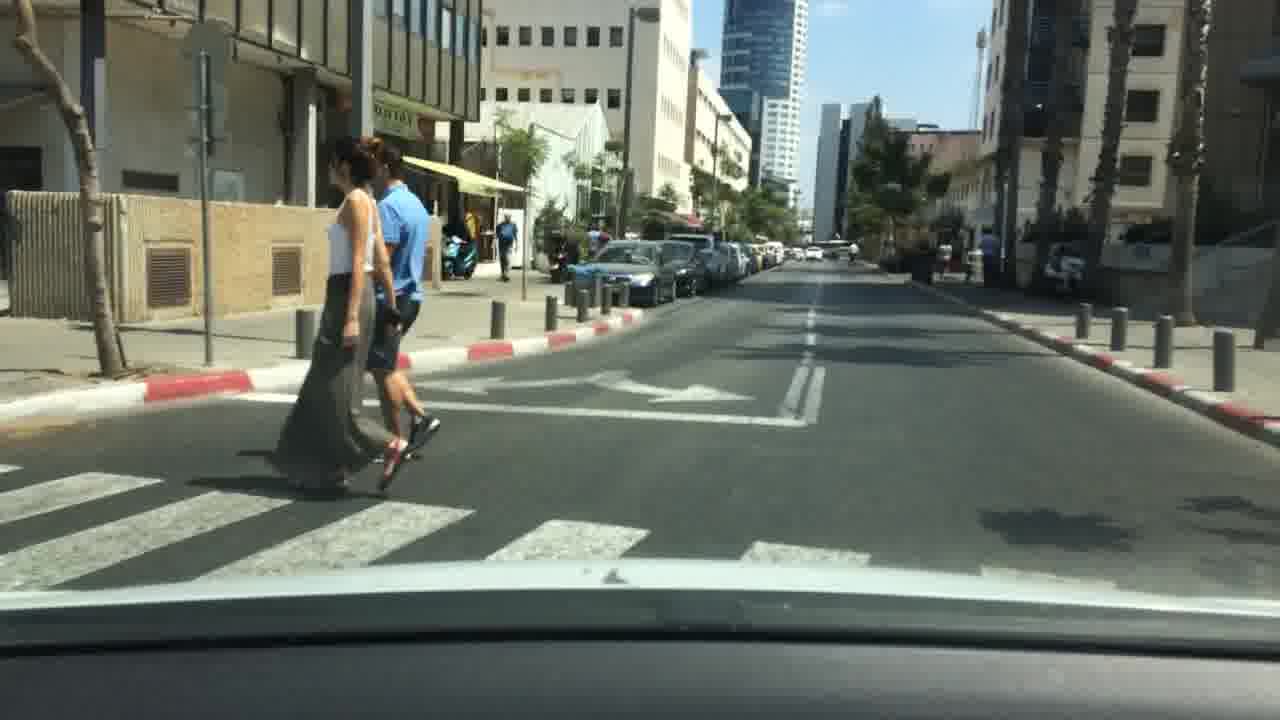}};
    \coordinate (sceneLeft) at ($(scene.center)+(-1.25,0)$);
    \node (stop) at (11,5.3) [rectangle, fill=badgeRed, text=white, font=\sffamily\bfseries\scriptsize, inner sep=3pt, rounded corners=2pt] {not forward};
    \coordinate (braceStopY) at (0,4.85);
    \draw [decorate, decoration={brace, amplitude=4pt, mirror}]
        (stop.west |- braceStopY) --
        (stop.east |- braceStopY)
        node (prediction) [midway, below=4pt, font=\scriptsize, inner xsep=0pt, outer xsep=0pt]
        {$\argmax_{\vy} \frac{1}{Z_\vx}{q_\theta(\vy \mid \vx)\,c_\BK(\vy)}$};

    \coordinate (formatRowY) at (0,7.75);
    \node (fmt-sympy) at (prediction.east |- formatRowY) [formatbox, anchor=east] {
        {\bfseries\scriptsize SymPy}\\[2pt]
        \strut to\_cnf(\\
        \strut\hspace*{0.3em}\textasciitilde green\\
        \strut\hspace*{0.3em}| \textasciitilde clear\\
        \strut\hspace*{0.3em}| forward)
    };
    \node (fmt-cpmpy) [formatbox, left=0.1cm of fmt-sympy] {
        {\bfseries\scriptsize CPMpy}\\[2pt]
        \strut Model(\\
        \strut\hspace*{0.3em}\textasciitilde green\\
        \strut\hspace*{0.3em}| \textasciitilde clear\\
        \strut\hspace*{0.3em}| forward)
    };
    \node (fmt-pysat) [formatbox, left=0.1cm of fmt-cpmpy] {
        {\bfseries\scriptsize PySAT}\\[2pt]
        \strut cnf = CNF()\\
        \strut cnf.append(\\
        \strut\hspace*{0.3em}[-1,-2,3])\\
        \strut
    };
    \node (fmt-nat) [formatbox, left=0.1cm of fmt-pysat] {
        {\bfseries\scriptsize NAT}\\[2pt]
        \strut (NOT green OR\\
        \strut\hspace*{0.3em}NOT clear OR\\
        \strut\hspace*{0.3em}forward)\\
        \strut
    };
    \node (fmt-dimacs) [formatbox, left=0.1cm of fmt-nat] {
        {\bfseries\scriptsize DIMACS}\\[2pt]
        \strut p cnf 3 1\\
        \strut -1 -2 3 0\\
        \strut\\
        \strut
    };

    \node (llm) at (9.205,9.5) [block, minimum width=1.8cm] {\textbf{LLM}};
    \draw[myarrow] (llm.north east) .. controls +(0.65,0) and +(0.65,0) ..
        node[right=1pt, font=\tiny\bfseries, text=darkSlate, align=left] {Self-Verification} (llm.south east);

    \node (prompt) [speech, anchor=north west, text width=3.3cm,
        callout relative pointer={(0,-0.23cm)}]
        at (sceneLeft |- llm.north) {
        \textit{``The car can move forward if there is a green light and the road is clear.''}\\[3pt]
        \textbf{Variables}: \texttt{1: green} \\
        \phantom{\textbf{Variables}: }\texttt{2: clear}\\
        \phantom{\textbf{Variables}: }\texttt{3: forward}
    };
    \node (girl) [anchor=north, inner sep=0pt, outer sep=0pt]
        at ($(prompt.south)+(0,-0.35)$) {\includegraphics[width=1cm]{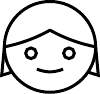}};
    \draw[myarrow] (prompt.east |- llm.west) -- (llm.west);

    \coordinate (formatInY) at ($(fmt-dimacs.north)+(0,0.3)$);
    \draw[formatline] (llm.south) -- (llm.south |- formatInY);
    \draw[formatline] (fmt-dimacs.north |- formatInY) -- (fmt-sympy.north |- formatInY);
    \foreach \fmt in {dimacs,nat,pysat,cpmpy,sympy} {
        \draw[myarrow, line width=0.7pt] (fmt-\fmt.north |- formatInY) -- (fmt-\fmt.north);
    }
    \node[font=\tiny\bfseries, text=darkSlate, anchor=east] at ($(llm.south)!0.55!(llm.south |- formatInY)+(-0.1,0)$) {Auto-Formalization};

    \node (s-q1) at (7.1,5.6) [orangenode] {};
    \node (s-q2) at (7.1,5.0) [orangenode] {};
    \node (s-out) at (8.1,5.3) [purplenode] {};
    \draw[line width=0.6pt, darkSlate!70] (s-q1) -- (s-out);
    \draw[line width=0.6pt, darkSlate!70] (s-q2) -- (s-out);
    \begin{scope}[on background layer]
        \node [draw=darkSlate!40, dashed, line width=0.8pt, fill=white, rounded corners=3pt, fit=(s-q1) (s-q2) (s-out), inner sep=5pt] (splbox) {};
    \end{scope}

    \coordinate (conversionIn) at ($(fmt-dimacs.south)+(0.3,0)$);
    \coordinate (compilationOut) at ($(fmt-dimacs.south)+(-0.3,0)$);
    \coordinate (conversionY) at ($(fmt-dimacs.south)+(0,-0.3)$);
    \foreach \fmt in {nat,pysat,cpmpy,sympy} {
        \draw[formatline] (fmt-\fmt.south) -- (fmt-\fmt.south |- conversionY);
    }
    \draw[myarrow, line width=0.7pt]
        (fmt-sympy.south |- conversionY) -- (conversionIn |- conversionY) -- (conversionIn);
    \coordinate (compilationY) at ($(fmt-dimacs.south)+(0,-0.65)$);
    \draw[myarrow] (compilationOut) -- (compilationOut |- compilationY)
        -- node (compilationLabel) [midway, above=1pt, inner sep=0pt, font=\tiny\bfseries, text=darkSlate]
            {Knowledge Compilation}
        (splbox.north |- compilationY) -- (splbox.north);
    \coordinate (conversionLabelX) at ($(fmt-pysat.south)!0.5!(fmt-sympy.south)$);
    \node (conversionLabel) [anchor=base, inner sep=0pt, font=\tiny\bfseries, text=darkSlate]
        at (conversionLabelX |- compilationLabel.base) {Conversion};

    \node (i1) at (3.6,5.3) [tealnode] {};
    \node (h1) at (4.6,5.8) [orangenode] {};
    \node (h2) at (4.6,5.3) [orangenode] {};
    \node (h3) at (4.6,4.8) [orangenode] {};
    \node (o1) at (5.6,5.55) [purplenode] {};
    \node (o2) at (5.6,5.05) [purplenode] {};
    \foreach \h in {h1,h2,h3} {\draw[line width=0.5pt, color=darkSlate!60] (i1) -- (\h);}
    \foreach \h in {h1,h2,h3} {
        \draw[line width=0.5pt, color=darkSlate!60] (\h) -- (o1);
        \draw[line width=0.5pt, color=darkSlate!60] (\h) -- (o2);
    }
    \begin{scope}[on background layer]
        \node [draw=darkSlate!40, dashed, line width=0.8pt, fill=white, rounded corners=3pt, fit=(i1) (h1) (h3) (o1) (o2), inner sep=4pt] (percbox) {};
    \end{scope}
    
    \draw[myarrow] (scene) -- (percbox.west);

    \node at ($(percbox.east)!0.5!(splbox.west)$)
    [font=\huge\bfseries, text=darkSlate] {$\times$};

    \draw[myarrow] (splbox.east) -- (stop);

    \coordinate (bracePercY) at (0,4.35);
    
    \draw [decorate, decoration={brace, amplitude=4pt, mirror}]
        (percbox.south west |- bracePercY) --
        (percbox.south east |- bracePercY)
        node[midway, below=4pt, font=\scriptsize]
        {$q_\theta(\vy \mid \vx)$};
    
    \draw [decorate, decoration={brace, amplitude=4pt, mirror}]
        (splbox.south west |- bracePercY) --
        (splbox.south east |- bracePercY)
        node[midway, below=4pt, font=\scriptsize]
        {$c_\BK(\vy)$};
    
    \end{tikzpicture}
    }
    \caption{\textbf{Illustration of our \autobench auto-formalization framework}:
    given %
    a natural language description of a constraint and the corresponding variables, an LLM generates %
    \DIMACS, \pysat, \cpmpy, \sympy, or natural language (\nat) formalizations. %
    Non-\DIMACS outputs are converted to \DIMACS and later compiled, via knowledge compilation, into circuits that can be used in NeSy predictors (\SPL), thus ensuring that the output satisfies the specified constraints by design, following \cref{eq:spl}.
    The five boxes show equivalent encodings of $(\texttt{green} \land \texttt{clear}) \Rightarrow \texttt{forward}$. %
    } %
    \label{fig:nesy-autoform-overview}
\end{figure}

\section{NeSy Predictors}

Among NeSy approaches \citep{de2021statistical, garcez2022neural}, here
we focus on \emph{NeSy predictors}, classifiers designed for reliability \citep{giunchiglia2022deep, chen2025rnpc, chen2026vlc}.
Given an input $\vx \in \bbR^n$, these models predict a (multi-)label $\vy \in \{0,1\}^m$ while leveraging \emph{prior knowledge} $\BK$, 
which is usually expressed in the form of structural \citep{di2020efficient} or safety requirements \citep{hoernle2022multiplexnet}.
The model then produces a predictive distribution $p_\theta(\vy \mid \vx; \BK)$ that assigns lower \citep{xu2018semantic, van2022anesi}, or even provably zero \citep{manhaeve2018deepproblog, ahmed2022semantic}, probability to outputs $\vy$ that violate the constraint $\BK$. 

As an example, consider~\cref{fig:nesy-autoform-overview}, where a NeSy predictor determines whether a car can move forward from input observation $\vx$, such as an image of the road. Let \texttt{green}, \texttt{clear}, and \texttt{forward} denote Boolean variables indicating whether the traffic light is green, the road is clear, and the car moves forward, respectively. 
We impose the domain constraint $\BK: (\texttt{green} \land \texttt{clear}) \Rightarrow \texttt{forward}$, stating that the car can move forward when the traffic light is green and the road is clear. 
While the neural network may assign nonzero probability to output combinations that violate $\BK$, a NeSy predictor incorporates $\BK$ into the prediction process. It \textit{assigns zero probability to invalid combinations} and renormalizes the remaining probability mass. For example, $\texttt{green}=1$, $\texttt{clear}=1$, and $\texttt{forward}=0$ is ruled out because it violates $\BK$, while all combinations satisfying $\BK$ remain possible.

Existing NeSy predictors differ along two main dimensions. First, they vary in the underlying logical formalism they support, \ie propositional logic \citep{ahmed2022semantic, giunchiglia2024ccn}, first-order logic~\citep{manhaeve2018deepproblog, zhou2019abductive, badreddine2022logic, pojer2024generalized, pojer2025neuro}, and satisfiability modulo theory over linear real arithmetic~\citep{kurscheidt2025pal, kurscheidt2026theory}. Second, they differ in how they enforce constraint satisfaction, relying on paradigms such as probabilistic logic \citep{lippi2009prediction, marra2021neural, pryor2022neupsl, skryagin2022neural, maene2023softunification} or fuzzy logic \citep{diligenti2017semantic, donadello2017logic, badreddine2022logic}. 

In this work, we focus primarily on \emph{semantic probabilistic layers} (\SPL) \citep{ahmed2022semantic}, a representative state-of-the-art NeSy predictor for PL. \SPL is a predictive layer that can be placed on top of a neural network to enforce logical constraints. It combines the unconstrained predictive distribution $q_\theta(\vy \mid \vx)$ produced by the neural network with a constraint distribution $c_\BK(\vy)$, represented as a differentiable probabilistic circuit~\citep{darwiche2001decomposable,vergari2019tractable,choi2020pc} obtained through knowledge compilation \citep{darwiche2002knowledge}. The resulting predictive distribution is given by the tractable product of experts:
\[
    \label{eq:spl}
    \textstyle
    p_\theta(\vy \mid \vx; \BK)
    =
    \frac{1}{Z_{\vx}} q_\theta(\vy \mid \vx)\, c_\BK(\vy),
    \qquad
    Z_{\vx} := \sum_{\vy} q_\theta(\vy \mid \vx)\, c_\BK(\vy),
\]
where $Z_{\vx}$ is the partition function, \ie the weighted model count induced by $\BK$~\citep{chavira2008wmc}, which can be computed exactly with circuits \citep{vergari2021compositional}. 
By doing this, \SPL \emph{provably allocates zero probability mass to invalid output combinations}, making it reliable in high-stakes scenarios.
However, to work \SPL requires that the constraint $\BK$ is given as an input as a well-formed CNF.
As this might not always be possible or easy, we investigate next how to \textit{auto-formalize} $\BK$ out of a textual description that we assume to be more easily accessible.
An alternative to \SPL is to use \textit{soft constraints}. A representative NeSy approach is the \textit{semantic loss} (\SL)~\citep{xu2018semantic}, which  minimizes the negative log-partition function in \cref{eq:spl}, see ~\cref{sec:semantic-loss}.

\section{Auto-formalizing Neuro-Symbolic Predictors}
\label{sec:autoformalization-pipeline}

Our goal is to use an LLM as an expert that translates natural language %
domain knowledge into symbolic constraints that can be later used by a NeSy predictor. The resulting pipeline, shown in~\cref{fig:nesy-autoform-overview}, consists of two stages: first, the LLM formalizes the constraints; then, the generated formula is incorporated into the NeSy predictor and used during learning and inference.

\textbf{LLMs as constraint experts}. %
Recently, reasoning models have been developed to improve performance on complex tasks by generating intermediate reasoning steps before producing a final answer~\citep{wei2022chain, guo2025deepseek}. 
Following~\citet{michailidis2025cp}, we treat an LLM as a black-box function $h:\Lambda^* \rightarrow \Lambda^*$, where $\Lambda$ is the model vocabulary and $\Lambda^*$ denotes the set of finite token sequences. As shown in~\cref{fig:nesy-autoform-overview}, given a prompt, the LLM produces an output sequence $\vs$ without requiring access to its weights, logits, hidden states, or training data.

\subsection{Constraint Formalization}
\label{sec:constraint-formalization}

Learning symbolic constraints from data has been extensively studied as \emph{constraint acquisition}~\citep{de2018learning, tsouros2024learning, tsouros2025generalizing}. Existing approaches typically learn constraints from labeled satisfying and violating assignments or through interaction with an oracle or solver~\citep{de2016learning, bessiere2017constraint, freuder1992partial}. 
In our setting, instead of learning from a collection of examples, the LLM receives a single natural language description of the domain knowledge and acts as the constraint-acquisition mechanism~\citep{michailidis2024constraint}.%

We focus on PL and use CNF as the target language for constraints. 
A CNF can be expressed through several equivalent representations, including high-level modeling languages and libraries such as \cpmpy, %
\sympy%
, and %
\pysat. %
For example, as depicted in~\cref{fig:nesy-autoform-overview}, the constraint $(\texttt{green} \land \texttt{clear}) \Rightarrow \texttt{forward}$ can be expressed in any of these representations, despite representing the same logical relation. %
We consider these different representations because they reflect interfaces practitioners actually use to specify constraints, from low-level solver formats to high-level modeling libraries, allowing us to test whether LLMs auto-formalization ability depends on the target syntax. 

Among these, Python-based libraries (\pysat, \cpmpy, \sympy) are of particular interest, as LLMs have repeatedly been shown to be especially capable of generating correct Python code~\citep{chen2021evaluating, austin2021program, li2022competition}, making it a natural interface to test for auto-formalization.
We also consider direct generation of Boolean formulas in natural language (\nat) and \DIMACS CNF, \ie the common standard for SAT solvers~\citep{johnson1996cliques}. As shown in~\cref{fig:nesy-autoform-overview}, all these representations are translated into \DIMACS, to unify evaluation (see~\cref{sec:results-auto-form}).%

\textbf{Prompting}. %
Each prompt frames the LLM as an expert in SAT solving and constraint modeling, specifies the syntax of the target representation (\DIMACS, \nat, \pysat, \cpmpy, or \sympy), provides the grounded variables, and requires the generated formula to be contained within a \texttt{<cnf>...</cnf>} block.
Unlike~\citet {michailidis2025cp}, we also investigate how performance varies with the level of detail provided in the prompt (\cref{sec:results-auto-form}), thereby assessing the extent to which auto-formalization depends on the user's expertise. Complete prompts are provided in~\cref{app:prompts}.

\textbf{Symbol grounding}. 
A key challenge in translating natural language descriptions into logical constraints is identifying relevant symbols and consistently mapping them to logical variables, a problem known as \emph{symbol grounding}~\citep{harnard1990grounding}.
Recent work has shown that widely used open weight LLMs can struggle to correctly identify symbols in data~\citep{wust2025bongard, debole2026concept}. Furthermore, comparing formulas that use different symbols requires an additional alignment step~\citep{bortolotti2025shortcuts}, making evaluation substantially more challenging.
Since our goal is to study constraint formalization rather than symbol identification, we provide the LLM with the \textit{complete set of primary variables for each task}. Each variable is specified by its name %
and meaning.\footnote{Different formats require additional information, \eg \DIMACS index. Details are given in~\cref{app:prompts}.} %
As illustrated in~\cref{fig:nesy-autoform-overview}, this grounded set of variables serves as the interface between the natural language description and the target representation, allowing us to evaluate whether the LLM correctly captures the intended constraint independently of symbol identification.
The LLM may still introduce auxiliary variables, provided that they are explicitly defined in terms of the primary variables.

\textbf{Self-verification}. 
Following~\citet{michailidis2025cp}, we allow the LLM to iteratively correct its output over multiple stages. When a syntax error is detected, we provide a correction prompt containing the erroneous formula and corresponding error message. The LLM has to return either \texttt{[[OK]]} if no correction is needed, or \texttt{[[FIXED]]} followed by the corrected formula in a \texttt{<cnf>...</cnf>} block. Before and after each correction step, we automatically fix formatting issues that do not require LLM reasoning, such as incorrect \DIMACS headers or missing Python imports (see~\cref{app:errors}).

\textbf{Formalization evaluation}. 
Existing benchmarks, such as \satbench and \dcpbench, primarily rely on satisfiability results %
to evaluate the quality of the generated formulas. %
In \satbench, this amounts to checking whether the generated formula
has the same satisfiability status as the ground-truth formula, \ie whether both are SAT or UNSAT. %
In \dcpbench, the evaluation instead tests whether %
a generated CP solution is feasible and satisfies the ground-truth %
constraint.
These metrics are not suitable for the full auto-formalization pipeline, as we are interested not only in whether a formula is SAT, but also in \textit{which solutions it admits}. This is because NeSy predictors such as \SPL assign probability mass only to satisfiable assignments, preventing the final distribution from assigning mass to UNSAT assignments.  %
To capture this, we evaluate the agreement between the valid assignments of the generated formula and those of the ground-truth formula using model counting (\mc)~\citep{chavira2008wmc}. 
We obtain the entries of the confusion matrix by counting the valid assignments satisfying both the ground-truth $\phi^*$ and the generated formula $\hat{\phi}$:
\[
\textstyle
\text{TP}=\text{MC}(\phi^*\!\land\hat{\phi}),\;
\text{FP}=\text{MC}(\neg\phi^*\!\land\hat{\phi}),\;
\text{FN}=\text{MC}(\phi^*\!\land\neg\hat{\phi}),\;
\text{TN}=\text{MC}(\neg\phi^*\!\land\neg\hat{\phi}),
\]
\ie true positive, false positive, false negative and true negative, respectively. We compute these quantities exactly for tasks with at most $20$ variables and use \texttt{ApproxMC}~\citep{chakraborty2016approxmc} for larger instances (see~\cref{app:metrics} for details), then
derive F1, precision, and recall. %
Recall is particularly important in our setting as \SPL uses the  constraints to restrict the space of possible outputs. If a generated constraint incorrectly excludes a valid assignment, %
the predictor cannot assign probability mass to that assignment, making the corresponding configuration impossible to predict~\citep{grivas2022low,grivas2024taming}.
Finally, we report the proportion of generated formulas containing syntax errors.%

\subsection{From Auto-Formalization To NeSy Predictors}

Once generated, each CNF formula is embedded into an \SPL~\citep{ahmed2022semantic} and, as an alternative, used as a \SL~\citep{xu2018semantic}.
CNFs are compiled into sentential decision diagrams (SDDs)~\citep{darwiche2011sdd} via the \texttt{PySDD} Python library. For \SPL, the SDD is loaded as a differentiable probabilistic circuit using \texttt{cirkit}~\citep{The_APRIL_Lab_cirkit_2024}, while \SL uses \texttt{KLay}~\citep{maene2025klay}. %

\textbf{Training and inference}.
\SPL is trained by minimizing the negative log-likelihood of the target label $\vy$ using maximum likelihood: $\calL = -\mathbb{E}\left[\log p_\theta(\vy \mid \vx; \BK)\right]$.
In contrast, \SL uses a standard classifier and incorporates $\BK$ through a penalty loss: $\calL_{\mathrm{SL}} =-\log \sum_{\vy \models \BK} \prod_{i} p (Y_i \mid \vx)$, \ie the negative log-probability that the network predictions satisfy $\BK$.
At test time, we perform maximum a posteriori inference over the predictive distribution, \ie $\textstyle \hat{\vy} = \arg\max_{\vy} p_\theta(\vy \mid \vx;\BK)$ for \SPL and $\textstyle \hat{\vy} = \arg\max_{\vy} p_\theta(\vy \mid \vx)$ for \SL.
The quality of the LLM-generated constraints can affect the final predictor in two ways: incorrect constraints may \textit{rule out valid predictions}, while incomplete or weak constraints may \textit{allow invalid predictions}. 
These effects are hard in \SPL, where the formula determines the feasible predictions, but act as a training bias in \SL. Thus, we evaluate not only whether LLMs generate valid constraints, but also whether those constraints yield correct predictions.

\textbf{Metrics}. In addition to F1, accuracy, and recall, we introduce three metrics tailored to our setting: formula relationship (\frel), consistency (\con), and model-count ratio (\mcr).
\frel characterizes the relationship between the generated formula $\hat{\phi}$ and the ground-truth formula $\phi^*$, distinguishing four cases: \texttt{equal} 
($\hat{\phi}=\phi^*$), \texttt{permissive} ($\hat{\phi}\Leftarrow\phi^*$), \texttt{strict} ($\hat{\phi}\Rightarrow\phi^*$), and \texttt{incomparable} ($\hat{\phi}\parallel\phi^*$). %
A \texttt{permissive} formula admits all valid assignments but also some invalid ones, while a \texttt{strict} formula admits only valid assignments, but excludes some valid ones. \texttt{Incomparable} formulas exhibit both types of errors. %
\con measures the proportion of assignments %
consistent with %
the ground-truth constraint, 
while \mcr measures the relative number of satisfying assignments admitted by the generated formula.
This is useful for \texttt{strict} and \texttt{permissive} cases, as it quantifies the extent to which the set of satisfying assignments admitted by the generated formula differs from that of the ground-truth formula, \eg if a \texttt{permissive} formula has $\mcr=1.75$, then $\hat{\phi}$ admits $1.75$ times as many satisfying assignments as the ground-truth. %
Further details are reported in~\cref{app:metrics}.

\section{Auto-NeSy-Bench}
\label{sec:auto-formalization}

The proposed pipeline raises two questions: \emph{(\textbf{RQ1}) can an LLM correctly formalize the intended domain knowledge?} and \emph{(\textbf{RQ2}) how do LLM-generated constraints affect the behavior of the resulting NeSy predictor?} Existing constraint-generation benchmarks address the first question in standalone CP settings, such as \dcpbench, or SAT settings, such as \satbench, while NeSy benchmarks typically assume pre-existing background knowledge. Thus, neither evaluates the full pipeline from natural language descriptions to formal constraints and their downstream impact on NeSy predictors.
To address this gap, we introduce \autobench, a benchmark designed to evaluate LLM-based auto-formalization and its downstream impact on NeSy predictors. We first describe the existing benchmarks and then introduce \autobench. Additional details are in~\cref{app:datasets}. %

\textbf{SATBench}. \satbench~\citep{wei2025satbench} evaluates LLMs logical reasoning through natural language description of puzzles generated from CNF formulas. The original benchmark contains $2{,}100$ instances, including both satisfiable and unsatisfiable problems. We repurpose its paired natural language descriptions and reference CNF formulas to evaluate auto-formalization. We exclude UNSAT instances because, when equipped with \SPL, they assign zero probability to every possible configuration, resulting in an uninformative downstream predictor. After excluding these instances and applying the consistency checks described in \cref{app:datasets_satbench}, we retain $1{,}041$ SAT instances.

\textbf{DCPBench-Open}. We repurpose \dcpbench%
~\citep{dcpbenchopen, michailidis2026dcpbenchopen} for auto-formalization, %
creating a new version of the original benchmark consisting of $164$ instances containing natural language descriptions of combinatorial problems collected from multiple sources. %
These problems are designed for CP, where the goal is to construct a constraint-based model rather than enumerate the set of solutions. Consequently, the benchmark includes problems involving both discrete and continuous variables, as well as optimization problems.
Optimization problems are not suitable for our NeSy predictor setting, as they seek a single optimal solution rather than a set of valid solutions. Similarly, instances involving continuous variables fall outside the scope of PL. We therefore discard such instances during preprocessing. We also remove constraints that lead to excessive circuit sizes during knowledge compilation, as well as instances whose formulas fail to convert into CNF.
Additionally, our formulation differs from the original use of the dataset in that, for instance-specific problems, we explicitly ground the relevant instance in the constraint formalization. After this preprocessing, we obtain $85$ instances, which we use to evaluate the auto-formalization capabilities of the LLMs.

\textbf{Auto-NeSy-Bench}. \autobench is an extension of 
\rsbench~\citep{bortolotti2024benchmark} which originally contained $8$ datasets spanning mathematical reasoning and high-stakes applications, and was designed to study the quality of concepts in NeSy predictors~\citep{marconato2023not}. We expand \rsbench to $18$ representative NeSy problems adapted from established NeSy datasets, including \texttt{sudoku}~\citep{augustine2022visual}, \texttt{road-r}~\citep{giunchiglia2023road}, \texttt{sushi}~\citep{kamishima2003nantonac}, and \texttt{warcraft}~\citep{vlastelica2019differentiation}. Unlike existing SAT and CP benchmarks, which typically provide a single fixed problem description, \autobench pairs each task with natural language descriptions at multiple levels of granularity, together with annotated examples, so that it can be used to evaluate every stage of the auto-formalization pipeline%
, from constraint extraction through downstream NeSy prediction. \Cref{app:datasets} details how each problem was adapted and describes the granularity levels of the natural language descriptions and the corresponding dataset statistics.

\section{Experimental Results}

We try to answer our \textbf{RQ1} and \textbf{RQ2} by evaluating several open-weight LLMs families: Qwen(-Coder)~\citep{yang2025qwen3, cao2026qwen3codernexttechnicalreport}, %
Gemma-3/4~\citep{gemma_2025, team2026gemma}, Mistral-Nemo~\citep{jiang2023mistral7b}, DeepSeek-R1-Distill~\citep{liu2024deepseek}, Olmo~\citep{groeneveld2024olmo}, Phi~\citep{abdin2025phi}, and GPT-OSS~\citep{openai2025gptoss120bgptoss20bmodel}. Sizes vary from 8.2B to 117B parameters. Every LLM has been evaluated \emph{unquantized}, except for GPT-OSS that comes natively quantized. %

\subsection{%
RQ1: Auto-formalization of Logical Constraints%
}
\label{sec:results-auto-form}

We structure our analysis along three main axes. First, we compare different LLMs to assess their \textit{auto-formalization} capability. Second, we investigate how \textit{prompt} configuration affects performance. Finally, we compare different \textit{output formats} to identify the most effective for formalizing constraints.

\textbf{Settings}. For each benchmark, we consider two prompting strategies, \ie zero-shot (ZS) and zero-shot chain-of-thought (ZS-CoT), and five output formats, \ie \DIMACS, \nat, \pysat, \cpmpy, and \sympy, yielding ten configurations per LLM and benchmark. For \autobench, we additionally evaluate two prompt variants with different levels of detail. Due to computational constraints, we evaluate a broader range of LLMs on \autobench, our novel benchmark, while evaluating a subset of LLMs on \satbench and \dcpbench. This allows us to thoroughly evaluate our new benchmark while also providing validation across existing benchmarks. 
Due to space constraints, we defer additional experiments and \SPL ablations using LLM-generated formulas from less detailed prompts to~\cref{app:additional_results}, along with an analysis of \SL with soft constraints in~\cref{sec:semantic-loss}.

\begin{table}[!t]
\centering
\caption{
\textbf{Model choice substantially affects auto-formalization performance across benchmarks.}
We compare various models on all benchmarks, reporting F1, precision, recall, and syntax error rate (mean $\pm$ standard deviation). LLM coverage varies %
slightly due to resource constraints (see~\cref{app:additional_results} for the full results). Best results are in bold, and second-best results are underlined. Ordered by size. %
}
\label{tab:average-problems}
\scriptsize
\setlength{\tabcolsep}{5pt}
\scalebox{0.95}{
\begin{tabular}{llrrrr}
\toprule
\textsc{Dataset} & \textsc{Model} & \textsc{F1} ($\uparrow$) & \textsc{Precision} ($\uparrow$) & \textsc{Recall} ($\uparrow$) & \textsc{Syntax ER} ($\downarrow$) \\
\midrule
\multirow[c]{11}{*}{\autobench}
 & \texttt{qwen3-8b} & $0.372 \pm 0.202$ & $0.402 \pm 0.192$ & $0.423 \pm 0.200$ & $30.6\% \pm 25.5\%$ \\
 & \texttt{mistral-nemo} & $0.053 \pm 0.074$ & $0.051 \pm 0.073$ & $0.154 \pm 0.093$ & $62.8\% \pm 22.9\%$ \\
 & \texttt{phi4-reasoning-plus} & $0.423 \pm 0.177$ & $0.438 \pm 0.167$ & $0.440 \pm 0.191$ & $49.4\% \pm 16.9\%$ \\
 & \texttt{gpt-oss-20b} & $0.505 \pm 0.163$ & $0.564 \pm 0.124$ & $0.556 \pm 0.149$ & $30.0\% \pm 10.9\%$ \\
 & \texttt{gemma3-27b} & $0.040 \pm 0.040$ & $0.048 \pm 0.053$ & $0.164 \pm 0.134$ & $55.6\% \pm 29.7\%$ \\
 & \texttt{gemma4-31b} & $\underline{0.706 \pm 0.233}$ & $\underline{0.732 \pm 0.203}$ & $\underline{0.782 \pm 0.197}$ & $\mathbf{3.9\% \pm 2.5\%}$ \\ & \texttt{olmo3-32b-think} & $0.569 \pm 0.126$ & $0.609 \pm 0.105$ & $0.677 \pm 0.110$ & $15.0\% \pm 9.0\%$ \\
 & \texttt{qwen3-32b} & $0.655 \pm 0.138$ & $0.678 \pm 0.127$ & $0.712 \pm 0.132$ & $14.4\% \pm 11.2\%$ \\
 & \texttt{deepseek-r1-llama-70b} & $0.412 \pm 0.148$ & $0.435 \pm 0.164$ & $0.527 \pm 0.159$ & $28.3\% \pm 22.7\%$ \\
 & \texttt{qwen3-coder-next} & $0.478 \pm 0.192$ & $0.507 \pm 0.218$ & $0.576 \pm 0.147$ & $20.0\% \pm 13.2\%$ \\
 & \texttt{gpt-oss-120b} & $\mathbf{0.786 \pm 0.137}$ & $\mathbf{0.796 \pm 0.137}$ & $\mathbf{0.838 \pm 0.101}$ & $\underline{7.8\% \pm 7.1\%}$ \\
\midrule
\multirow[c]{4}{*}{\dcpbench}
 & \texttt{gpt-oss-20b} & $\underline{0.193 \pm 0.061}$ & $\underline{0.196 \pm 0.060}$ & $0.245 \pm 0.053$ & $43.1\% \pm 14.4\%$ \\
 & \texttt{deepseek-r1-llama-70b} & $0.078 \pm 0.037$ & $0.077 \pm 0.037$ & $0.202 \pm 0.077$ & $39.5\% \pm 16.2\%$ \\
 & \texttt{qwen3-coder-next} & $0.139 \pm 0.083$ & $0.142 \pm 0.088$ & $\underline{0.269 \pm 0.147}$ & $\underline{36.8\% \pm 23.0\%}$ \\
 & \texttt{gpt-oss-120b} & $\mathbf{0.324 \pm 0.098}$ & $\mathbf{0.340 \pm 0.092}$ & $\mathbf{0.400 \pm 0.094}$ & $\mathbf{18.0\% \pm 17.4\%}$ \\
\midrule
\multirow[c]{3}{*}{\satbench}
 & \texttt{gemma4-31b} & $\underline{0.936 \pm 0.136}$ & $\underline{0.934 \pm 0.138}$ & $\underline{0.944 \pm 0.126}$ & $\underline{0.1\% \pm 0.3\%}$ \\
 & \texttt{qwen3-coder-next} & $0.648 \pm 0.188$ & $0.642 \pm 0.188$ & $0.682 \pm 0.192$ & $13.5\% \pm 11.6\%$ \\
 & \texttt{gpt-oss-120b} & $\mathbf{0.963 \pm 0.008}$ & $\mathbf{0.959 \pm 0.008}$ & $\mathbf{0.981 \pm 0.006}$ & $\mathbf{0.0\% \pm 0.1\%}$ \\
\bottomrule
\end{tabular}
}
\end{table}

\begin{table}[!t]
\centering
\caption{\textbf{Detailed descriptions and CoT prompts improve auto-formalization performance across all benchmarks.} %
We compare detailed and natural descriptions on \autobench and zero-shot (ZS) and zero-shot-chain-of-thought (ZS-CoT) prompts on \autobench, \dcpbench, and \satbench, reporting F1, precision, recall, and syntax error rate, with mean $\pm$ standard deviation. Best results are highlighted in bold. See~\cref{app:additional_results} for the complete results. %
}
\label{tab:input-comparison}
\scriptsize
\setlength{\tabcolsep}{5pt}
\scalebox{0.97}{
\begin{tabular}{lllrrrr}
\toprule
\textsc{Dataset} & \textsc{Variant} & \textsc{Config.} & \textsc{F1} ($\uparrow$) & \textsc{Precision} ($\uparrow$) & \textsc{Recall} ($\uparrow$) & \textsc{Syntax ER} ($\downarrow$) \\
\midrule
\multirow[c]{4}{*}{\autobench} & \multirow[c]{2}{*}{Description} & Non-Detailed & $0.330 \pm 0.206$ & $0.357 \pm 0.218$ & $0.464 \pm 0.221$ & $\mathbf{25.3\% \pm 24.9\%}$ \\
 &  & Detailed & $\mathbf{0.455 \pm 0.277}$ & $\mathbf{0.478 \pm 0.279}$ & $\mathbf{0.532 \pm 0.263}$ & $28.9\% \pm 25.7\%$ \\
\cmidrule(lr){2-7}
 & \multirow[c]{2}{*}{Prompting} & ZS & $0.437 \pm 0.262$ & $0.463 \pm 0.268$ & $0.515 \pm 0.243$ & $\mathbf{28.2\% \pm 24.7\%}$ \\
 &  & ZS-CoT & $\mathbf{0.472 \pm 0.290}$ & $\mathbf{0.494 \pm 0.289}$ & $\mathbf{0.548 \pm 0.281}$ & $29.6\% \pm 26.6\%$ \\
\midrule
\multirow[c]{2}{*}{\dcpbench} & \multirow[c]{2}{*}{Prompting} & ZS & $0.176 \pm 0.119$ & $0.181 \pm 0.124$ & $\mathbf{0.285 \pm 0.132}$ & $\mathbf{33.7\% \pm 20.6\%}$ \\
 &  & ZS-CoT & $\mathbf{0.191 \pm 0.113}$ & $\mathbf{0.197 \pm 0.118}$ & $0.272 \pm 0.113$ & $35.0\% \pm 20.3\%$ \\
\midrule
\multirow[c]{2}{*}{\satbench} & \multirow[c]{2}{*}{Prompting} & ZS & $0.805 \pm 0.240$ & $0.801 \pm 0.241$ & $0.826 \pm 0.233$ & $5.6\% \pm 10.9\%$ \\
 &  & ZS-CoT & $\mathbf{0.893 \pm 0.124}$ & $\mathbf{0.889 \pm 0.125}$ & $\mathbf{0.912 \pm 0.112}$ & $\mathbf{3.5\% \pm 7.0\%}$ \\
\bottomrule
\end{tabular}
}
\end{table}

\textbf{Can LLMs formalize logical constraints?} 
We average each LLM contribution across all output formats and prompting strategies. To make the comparison fair, we consider \textit{only} the detailed prompt for \autobench. \cref{tab:average-problems} shows that model choice substantially affects auto-formalization performance.
On \autobench, \texttt{gpt-oss-120b} performs best, achieving an F1 of $0.786$ (precision $0.796$, recall $0.838$), followed by \texttt{gemma4-31b} with an F1 of $0.706$. %
These two LLMs also exhibit the lowest syntax error rates, at $7.8\%$ and $3.9\%$, respectively. In contrast, \texttt{gemma-3-27b} performs the worst in F1, achieving only $0.040$ and a syntax error rate of $55.6\%$, indicating that a large fraction of its generated formalizations cannot be successfully parsed. No LLM reaches perfect performance, indicating that \autobench remains a non-trivial challenge for these popular open-weight LLMs.
On \satbench, \texttt{gpt-oss-120b} achieves near-optimal performance, with an F1 of $0.963$, followed by \texttt{gemma4-31b} with an F1 of $0.936$. %
In contrast, performance drops substantially on \dcpbench, where \texttt{gpt-oss-120b} achieves an F1 of only $0.324$. %
Thus, \autobench offers a challenging yet substantially more tractable setting than \dcpbench.
Across datasets and models, recall is generally higher than precision, suggesting that generated formulas tend to be permissive, admitting some invalid assignments rather than excluding valid ones. %

\textbf{How impactful is the prompt?} 
To answer this question, we
vary (i) the granularity of the prompt, and (ii) the introduction of CoT. To do that, we averaged each model and output modality.
We evaluate (i) on \autobench, as it is the \emph{only} benchmark providing this level of granularity. %
As shown in \cref{tab:input-comparison} (Description), detailed descriptions substantially improve auto-formalization performance: increasing F1 from $0.330$ to $0.455$, precision from $0.357$ to $0.478$ and recall from $0.464$ to $0.532$. This improvement is unlikely to be attributed to easier parsing, as detailed prompts exhibit a higher syntax error rate, \ie $28.9\%$ vs $25.3\%$ for non detailed descriptions.
To evaluate (ii), we compare ZS and ZS-CoT prompting across all three benchmarks. As shown in \cref{tab:input-comparison} (Prompting), CoT improves F1: from $0.437$ to $0.472$ on \autobench, $0.176$ to $0.191$ on \dcpbench, and $0.805$ to $0.893$ on \satbench. However, its impact on recall and syntax errors varies across datasets: recall decreases slightly on \dcpbench, while syntax errors increase on \autobench and \dcpbench but decrease on \satbench. Overall, CoT consistently improves F1, but its benefits are dataset dependent.
Full prompts are reported in~\cref{app:prompts}.

\textbf{How important is the LLM output format?}
We average each model-prompting configuration, using \textit{only} the detailed \autobench prompt for a fair comparison. As shown in \cref{tab:output-comparison}, code-based formats (\pysat, \cpmpy, \sympy) outperform direct \DIMACS and \nat generation on \autobench and \dcpbench. On \autobench, \cpmpy achieves the highest F1 ($0.575$), precision ($0.586$), and recall ($0.622$), while \sympy ranks second in all three metrics. Both outperform \DIMACS and \nat in F1 ($0.364$ and $0.331$), with \pysat at $0.453$. On \dcpbench, \cpmpy has again the highest F1 ($0.247$) and precision ($0.251$), while \DIMACS has the highest recall ($0.358$); all code-based formats outperform \nat and \DIMACS in F1. In contrast, \nat performs best on \satbench, achieving an F1 of $0.920$, followed by \sympy at $0.878$.
Performance does not correlate with syntax errors: \DIMACS achieves the lowest error ($20.5\%$, $21.3\%$, and $0.4\%$).

\begin{table*}[t]
\centering
\scriptsize
\caption{
\textbf{Code-based output formats generally achieve higher auto-formalization performance on \autobench and \dcpbench, whereas the \nat performs best on \satbench.}
We compare five output formats reporting F1, precision, recall, and syntax error, with mean $\pm$ standard deviation. Best results are in bold, and second-best results are underlined (see~\cref{app:additional_results} for full results).}
\label{tab:output-comparison}
\begin{tabular}{llrrrr}
\toprule
\textsc{Dataset} & \textsc{Format} & \textsc{F1} ($\uparrow$) & \textsc{Precision} ($\uparrow$) & \textsc{Recall} ($\uparrow$) & \textsc{Syntax ER} ($\downarrow$) \\
\midrule
\multirow[c]{5}{*}{\autobench}
 & \DIMACS & $0.364 \pm 0.223$ & $0.394 \pm 0.231$ & $0.467 \pm 0.227$ & $\mathbf{20.5\% \pm 18.8\%}$ \\
 & \nat & $0.331 \pm 0.173$ & $0.361 \pm 0.178$ & $0.457 \pm 0.199$ & $31.8\% \pm 22.5\%$ \\
 & \pysat & $0.453 \pm 0.300$ & $0.500 \pm 0.310$ & $0.498 \pm 0.284$ & $31.1\% \pm 26.3\%$ \\
 & \cpmpy & $\mathbf{0.575 \pm 0.299}$ & $\mathbf{0.586 \pm 0.303}$ & $\mathbf{0.622 \pm 0.280}$ & $\underline{30.3\% \pm 28.2\%}$ \\
 & \sympy & $\underline{0.549 \pm 0.280}$ & $\underline{0.550 \pm 0.281}$ & $\underline{0.614 \pm 0.265}$ & $30.8\% \pm 29.4\%$ \\
\midrule
\multirow[c]{5}{*}{\dcpbench}
 & \DIMACS & $0.147 \pm 0.094$ & $0.157 \pm 0.106$ & $\mathbf{0.358 \pm 0.114}$ & $\mathbf{21.3\% \pm 16.6\%}$ \\
 & \nat & $0.109 \pm 0.045$ & $0.114 \pm 0.051$ & $0.218 \pm 0.051$ & $45.4\% \pm 14.9\%$ \\
 & \pysat & $\underline{0.207 \pm 0.135}$ & $\underline{0.212 \pm 0.139}$ & $0.251 \pm 0.133$ & $\underline{28.4\% \pm 21.2\%}$ \\
 & \cpmpy & $\mathbf{0.247 \pm 0.120}$ & $\mathbf{0.251 \pm 0.123}$ & $\underline{0.295 \pm 0.128}$ & $34.9\% \pm 18.4\%$ \\
 & \sympy & $0.206 \pm 0.112$ & $0.209 \pm 0.117$ & $0.272 \pm 0.124$ & $41.8\% \pm 20.7\%$ \\
\midrule
\multirow[c]{5}{*}{\satbench}
 & \DIMACS & $0.699 \pm 0.311$ & $0.693 \pm 0.313$ & $0.726 \pm 0.300$ & $\mathbf{0.4\% \pm 0.7\%}$ \\
 & \nat & $\mathbf{0.920 \pm 0.085}$ & $\mathbf{0.914 \pm 0.089}$ & $\mathbf{0.944 \pm 0.068}$ & $\underline{1.0\% \pm 1.4\%}$ \\
 & \pysat & $0.873 \pm 0.138$ & $0.869 \pm 0.139$ & $0.888 \pm 0.135$ & $7.0\% \pm 11.6\%$ \\
 & \cpmpy & $0.875 \pm 0.149$ & $0.872 \pm 0.149$ & $0.886 \pm 0.148$ & $9.4\% \pm 13.5\%$ \\
 & \sympy & $\underline{0.878 \pm 0.129}$ & $\underline{0.875 \pm 0.130}$ & $\underline{0.901 \pm 0.118}$ & $4.9\% \pm 6.6\%$ \\
\bottomrule
\end{tabular}
\end{table*}

\subsection{RQ2: Downstream Impact of LLM-Generated Constraints}
\label{sec:results-spl}

The generated formula alone does not reveal its impact on a downstream NeSy predictor: a formula may differ substantially from the ground truth yet have little effect on prediction, or achieve a high F1 score while remaining unusable in practice. %
To address this, we embed LLM-generated formulas into an \SPL and evaluate the resulting models, closing the loop between auto-formalization and downstream NeSy predictors. 
We performed this evaluation on \autobench using \texttt{gpt-oss-120b}, \texttt{olmo3-32b-think} and \texttt{qwen3-8b}, representing large, medium, and small models, respectively, all of which perform well on \autobench (see ~\cref{tab:average-problems}). For each LLM, we use its best-performing configuration: \cpmpy output with a detailed CoT prompt (see~\cref{tab:input-comparison} and~\cref{tab:output-comparison}). 
For each (LLM, task) pair, we train an \SPL using the LLM-generated formula as prior knowledge and compare it with the same backbone trained using the ground-truth constraint.

\begin{table}[t]
\centering
    \caption{\textbf{Generated formulas, when available, achieve comparable downstream \SPL performance to expert-provided formulas on \autobench}. We report mean $\Delta$ \textsc{F1}/$\Delta$ Recall (\textsc{Rec}) relative to the ground-truth; standard deviations across $5$ seeds are omitted (see \cref{app:additional_results}). \frel denotes the relation between predicted and ground-truth formulas: \texttt{equal} ($=$), \texttt{strict} ($\Rightarrow$), \texttt{permissive} ($\Leftarrow$), or \texttt{incomparable} ($\parallel$), indicating whether the predicted formula satisfying assignments are equal to, a subset of, a superset of, or neither. \mcr measures the ratio of satisfying assignments of the predicted formula to that of the ground-truth formula. Missing cells indicate: $\dagger$~degenerate formula (\eg UNSAT), $\star$~no formula generated (no CNF), or $\S$~run failure (crash) (see~\cref{tab:spl-coverage-gaps}). 
    }
    \label{tab:spl}
    \scalebox{0.7}{
\begin{tabular}{lcccccccccccccc}
\toprule
& \multicolumn{2}{c}{\textsc{G-T}} 
& \multicolumn{4}{c}{\texttt{gpt-oss-120b}} 
& \multicolumn{4}{c}{\texttt{olmo3-32b-think}} 
& \multicolumn{4}{c}{\texttt{qwen3-8b}} \\
 \cmidrule(lr){2-3} \cmidrule(lr){4-7} \cmidrule(lr){8-11} \cmidrule(lr){12-15}
\textsc{Task} 
& \textsc{F1} $(\uparrow)$ & \textsc{Rec}  $(\uparrow)$
& $\Delta$\textsc{F1} & $\Delta$\textsc{Rec} & \frel & \mcr
& $\Delta$\textsc{F1} & $\Delta$\textsc{Rec} & \frel & \mcr
& $\Delta$\textsc{F1} & $\Delta$\textsc{Rec} & \frel & \mcr \\
\midrule
\texttt{bdd-oia} & $0.58$ & $0.57$ & $+0.01$ & $+0.01$ & $=$ & $1.00$ & $\dagger$ & $\dagger$ & $\dagger$ & $\dagger$ & $+0.00$ & $+0.01$ & $=$ & $1.00$ \\
\rowcolor[HTML]{EFEFEF}
\texttt{bdd-oia-2} & $0.74$ & $0.74$ & $+0.01$ & $+0.01$ & $=$ & $1.00$ & $+0.01$ & $+0.01$ & $=$ & $1.00$ & $\star$ & $\star$ & $\star$ & $\star$ \\
\texttt{cebab} & $0.47$ & $0.46$ & $+0.00$ & $+0.01$ & $=$ & $1.00$ & $-0.01$ & $-0.01$ & $=$ & $1.00$ & $\star$ & $\star$ & $\star$ & $\star$ \\
\rowcolor[HTML]{EFEFEF}
\texttt{chx} & $0.54$ & $0.52$ & $-0.06$ & $-0.04$ & $=$ & $1.00$ & $\S$ & $\S$ & $\S$ & $\S$ & $\dagger$ & $\dagger$ & $\dagger$ & $\dagger$ \\
\texttt{cifar10} & $0.87$ & $0.87$ & $-0.07$ & $-0.07$ & $=$ & $1.00$ & $-0.06$ & $-0.06$ & $=$ & $1.00$ & $-0.06$ & $-0.06$ & $=$ & $1.00$ \\
\rowcolor[HTML]{EFEFEF}
\texttt{cifar100} & $0.35$ & $0.41$ & $-0.08$ & $-0.09$ & $=$ & $1.00$ & $-0.09$ & $-0.10$ & $=$ & $1.00$ & $-0.08$ & $-0.09$ & $=$ & $1.00$ \\
\texttt{cle4evr} & $1.00$ & $1.00$ & $-0.00$ & $-0.00$ & $=$ & $1.00$ & $-0.00$ & $-0.00$ & $=$ & $1.00$ & $-0.00$ & $-0.00$ & $=$ & $1.00$ \\
\rowcolor[HTML]{EFEFEF}
\texttt{fashion} & $0.92$ & $0.92$ & $-0.00$ & $+0.00$ & $=$ & $1.00$ & $-0.00$ & $-0.00$ & $=$ & $1.00$ & $-0.00$ & $+0.00$ & $=$ & $1.00$ \\
\texttt{kand-logic} & $0.70$ & $0.70$ & $-0.09$ & $-0.09$ & $=$ & $1.00$ & $\star$ & $\star$ & $\star$ & $\star$ & $\dagger$ & $\dagger$ & $\dagger$ & $\dagger$ \\
\rowcolor[HTML]{EFEFEF}
\texttt{kand-logic-2} & $0.81$ & $0.81$ & $-0.05$ & $-0.05$ & $=$ & $1.00$ & $-0.04$ & $-0.04$ & $=$ & $1.00$ & $-0.11$ & $-0.11$ & $\Rightarrow$ & $0.02$ \\
\texttt{mn-add-bin} & $0.98$ & $0.98$ & $-0.00$ & $+0.00$ & $=$ & $1.00$ & $\dagger$ & $\dagger$ & $\dagger$ & $\dagger$ & $+0.00$ & $+0.00$ & $=$ & $1.00$ \\
\rowcolor[HTML]{EFEFEF}
\texttt{mn-add} & $0.98$ & $0.98$ & $-0.01$ & $-0.01$ & $=$ & $1.00$ & $-0.00$ & $+0.00$ & $=$ & $1.00$ & $-0.00$ & $-0.00$ & $=$ & $1.00$ \\
\texttt{mn-mul-bin} & $0.98$ & $0.98$ & $-0.00$ & $-0.00$ & $=$ & $1.00$ & $-0.00$ & $-0.00$ & $=$ & $1.00$ & $\S$ & $\S$ & $\S$ & $\S$ \\
\rowcolor[HTML]{EFEFEF}
\texttt{mn-mul} & $0.98$ & $0.98$ & $-0.00$ & $-0.00$ & $=$ & $1.00$ & $-0.01$ & $-0.01$ & $=$ & $1.00$ & $-0.00$ & $-0.01$ & $=$ & $1.00$ \\
\texttt{road-r} & $0.31$ & $0.30$ & $-0.09$ & $-0.07$ & $\parallel$ & $1.03$ & $\S$ & $\S$ & $\S$ & $\S$ & $\dagger$ & $\dagger$ & $\dagger$ & $\dagger$ \\
\rowcolor[HTML]{EFEFEF}
\texttt{sudoku} & $0.79$ & $0.79$ & $-0.02$ & $-0.02$ & $=$ & $1.00$ & $\star$ & $\star$ & $\star$ & $\star$ & $+0.05$ & $+0.05$ & $=$ & $1.00$ \\
\texttt{sushi} & $0.29$ & $0.30$ & $-0.02$ & $+0.00$ & $=$ & $1.00$ & $+0.01$ & $+0.02$ & $=$ & $1.00$ & $+0.01$ & $+0.01$ & $=$ & $1.00$ \\
\rowcolor[HTML]{EFEFEF}
\texttt{warcraft} & $0.87$ & $0.86$ & $-0.03$ & $-0.02$ & $\Leftarrow$ & $1.75$ & $-0.01$ & $-0.01$ & $\Leftarrow$ & $1.75$ & $\S$ & $\S$ & $\S$ & $\S$ \\
\bottomrule
\end{tabular}
}

\end{table}

\textbf{Discussion.} Across models, \cref{tab:spl} shows that LLM-generated formulas are, in most cases, both \emph{usable}, \ie the auto-formalization pipeline produces a syntactically valid and executable formula, and \emph{exactly correct} ($=$), \ie they yield the same set of valid assignments. Empirically, an \SPL equipped with LLM-generated constraints achieves performance comparable to that obtained with the ground-truth formula, with average differences in $\Delta$ F1 and $\Delta$ recall below $0.03$ and $0.02$, respectively. 
The main exception is \texttt{kand-logic-2} with \texttt{qwen3-8b}, 
whose generated formula captures 
only $2\%$ of valid assignments (\mcr $=0.02$). In contrast, over-approximations ($\Leftarrow$), such as \texttt{warcraft} %
, have little effect on downstream predictions. This suggests that, in-distribution, the negative effect of overly permissive formulas can be mitigated during training, as the model can learn to assign lower probability to output combinations that violate the ground-truth constraints. %
All formulas in~\cref{tab:spl} achieve perfect consistency (see \cref{app:additional_results}). 
Failures are also unevenly distributed across models. \Cref{tab:spl} shows that \texttt{gpt-oss-120b} produces a valid formula on every task ($0/18$ failures), while \texttt{olmo3-32b-think} and \texttt{qwen3-8b} fail on $6/18$ and $7/18$ tasks respectively, roughly evenly split between degenerate ($\dagger$), unproduced ($\star$), and run-failure ($\S$) formulas (see~\cref{tab:spl-coverage-gaps} for details).

\section{Related Work}

\textbf{Constraint Modeling in LLMs}. Recent research has explored LLMs as a tool to translate natural language descriptions into formal constraint, logic, or CP models~\citep{tsouros2023holy, michailidis2024constraint, michailidis2025cp}. This direction has also been explored through in-context and retrieval-augmented strategies for example selection~\citep{michailidis2024constraint}, active-learning~\citep{mechqrane2024using, mechqrane2026active}, fine-tuning%
~\citep{shi2025constraintllm}, SAT solving~\citep{hazra2025have}, and anomaly detection~\citep{zhang2026llm}.
A parallel line uses LLMs as natural-language-to-symbolic translators paired with external solvers or provers, targeting propositional and first-order logic~\citep{liu2026veritrans, pan2025langsat, pan2025fine}, ontologies~\citep{strader2024indoor}, or executable programs~\citep{austin2021program}. Orthogonal approaches generate answer set programming (ASP) rules~\citep{kareem2024using}, synthesize inductive logic programming benchmarks~\citep{helff2026slr}, and ground symbols from natural language descriptions~\citep{stein2025road}.
However, these works evaluate end-task accuracy or formalization quality, and none targets auto-formalization for NeSy predictors. %

\textbf{Knowledge Acquisition in NeSy}. Knowledge acquisition is a well-studied problem in NeSy AI and statistical relational learning~\citep{de2021statistical}. Existing approaches fall into two settings: (i) learning explicit knowledge from data, and (ii) learning models that acquire explicit and interpretable knowledge. %
The former aims to recover symbolic knowledge from training examples using a predefined vocabulary, as in inductive logic programming and its probabilistic extensions~\citep{muggleton1994inductive,de2008probabilistic}. The latter includes a growing body of NeSy models~\citep{daniele2023deep,tang2023perception,shindo2023thinking,shindo2024learning,wust2024pix2code,debot2024interpretable}, which may be unreliable due to reasoning shortcuts~\citep{bortolotti2025shortcuts}.
Both settings assume that the predicate vocabulary is specified by a human expert or that the knowledge is learned from data. Instead, we study whether LLMs can formalize it from natural language descriptions.

\section{Conclusion}

To the best of our knowledge, 
ours is the first benchmark and end-to-end pipeline for LLM-based auto-formalization of NeSy predictors, using constraints either as a layer or training loss. %
We investigate its opportunities and limitations, showing that the best-performing LLMs can generate high-quality formulas that perform comparably to manually constructed ones in downstream NeSy predictors, and that their performance depends on the quality of the constraint specification.
Our results further highlight which LLMs, prompting strategies, and output formats are most suitable for auto-formalization, while showing that smaller models can struggle substantially.
More broadly, we see this as a step toward promoting NeSy AI adoption: our goal is not to replace domain experts, but to support them, since generated formulas remain interpretable and can be inspected and corrected. %

\textbf{Future Work}. 
Our evaluation is restricted to propositional logic formulas with \SPL and \SL as representative NeSy approaches. We focus on formulas compatible with current NeSy pipelines, as handling substantially more complex constraints remains a general challenge in NeSy AI~\citep{huang2021scallop}. We leave the effects of automatic symbol extraction~\citep{stein2025road} and their downstream consequences~\citep{marconato2026symbol} for future work.
Additional future directions include evaluating auto-formalization with a broader range of NeSy predictors, including architectures that support first order logic, such as \texttt{DeepProbLog}~\citep{manhaeve2018deepproblog}, \texttt{LogicTensorNetworks}~\citep{rinaldi2026sltnstructurallogictensor, badreddine2022logic}, and ASP-based models~\citep{aspis2022embed, yang2020neurasp, skryagin2023scalable, cunnington2024role}. Another direction is to study auto-formalization with more complex constraint formalisms, such as  satisfiability modulo theory over linear real arithmetic~\citep{kolb2018learnin, de2023neural, kurscheidt2025pal, kurscheidt2026theory}, as well as 
uncertainty estimation~\citep{marconato2024bears, DBLP:conf/aaai/RuggieriP25, bortolotti2026concise} and fine-tuning~\citep{calanzone2025logically}.

\subsubsection*{Author Contributions}

The original idea of auto-formalizing Neuro-Symbolic predictors was conceived by Antonio Vergari and later refined together with Stefano Teso, Andrea Passerini, and Samuele Bortolotti. It was subsequently discussed with Weixin Chen and Han Zhao, who were independently working on a similar topic. Samuele Bortolotti designed the pipeline and constructed the benchmark, which was later discussed and refined with Weixin Chen. The metrics were formalized by Stefano Teso in discussion with Samuele Bortolotti. Samuele Bortolotti and Weixin Chen conducted the experiments and wrote the manuscript. Antonio Vergari, Stefano Teso, Andrea Passerini, and Han Zhao supervised the work.

\subsubsection*{Acknowledgments}

The authors are grateful to Paolo Morettin, Emanuele Marconato, Emile van Krieken, and Nicolas Lazzari for useful discussion. 
The authors acknowledge the CINECA award under the ISCRA initiative for the availability of high performance computing resources and support. 
Funded by the European Union. The views and opinions expressed are however those of the author(s) only and do not necessarily reflect those of the European Union, the European Health and Digital Executive Agency (HaDEA) or the European Research Executive Agency. Neither the European Union nor the granting authority can be held responsible for them. Grant Agreement no. 101120763 - TANGO.
Antonio Vergari is supported by the ``UNREAL: Unified Reasoning Layer for Trustworthy ML'' project (EP/Y023838/1) selected by the ERC and funded by UKRI EPSRC.
Stefano Teso was partially supported by the Flemish Government under the Flemish research foundation (FWO) project ``Neurosymbolic AI for Constraint Learning'' (Project G047124N).
Weixin Chen and Han Zhao are partially supported by an NSF grant \#2504555.

\newpage
\bibliography{references, explanatory-supervision, arxiv}
\newpage

\appendix

\section{Implementation}

All the experiments are implemented using Python 3.12.4 with PyTorch 2.11.
LLM evaluation was performed using \texttt{vLLm}~\citep{kwon2023efficient} on different clusters:

\begin{table}[H]
\centering
\caption{LLM evaluation with the vLLM backend across the clusters used in our experiments.}
\scriptsize
\setlength{\tabcolsep}{5pt}
\renewcommand{\arraystretch}{1.35}

\begin{tabular}{@{}llccc@{}}
\toprule
\textsc{Model} & \textsc{Link} & \textsc{CUDA} & \textsc{vLLM} & \textsc{GPUs} \\
\midrule

\texttt{qwen3-32b} &
\HF{Qwen/Qwen3-32B} &
\multirow{3}{*}{12.5.0} &
\multirow{3}{*}{0.23.0} &
\multirow{3}{*}{\shortstack[c]{2$\times$ A100\\80\,GB}} \\

\texttt{mistral-nemo} &
\HF{mistralai/Mistral-Nemo-Instruct-2407} &
& & \\

\texttt{phi4-reasoning-plus} &
\HF{microsoft/Phi-4-reasoning-plus} &
& & \\

\midrule

\texttt{gemma3-27b} &
\HF{google/gemma-3-27b-it} &
\multirow{2}{*}{12.6} &
\multirow{2}{*}{0.23.0} &
\multirow{2}{*}{\shortstack[c]{1--2$\times$ RTX PRO\\6000 Blackwell\\96\,GB (sm\_120)}} \\

\texttt{gemma4-31b} &
\HF{google/gemma-4-31B-it} &
& & \\

\midrule

\texttt{olmo3-32b-think} &
\HF{allenai/Olmo-3-32B-Think} &
12.6 &
0.23.0 &
\begin{tabular}[c]{@{}c@{}}
4$\times$ A100 \\
\end{tabular} \\

\midrule

\texttt{qwen3-8b} &
\HF{Qwen/Qwen3-8B} &
\multirow{1}{*}{12.6} &
\multirow{1}{*}{0.19.0} &
\multirow{1}{*}{\shortstack[c]{2$\times$ RTX 4090 24\,GB}} \\

\midrule

\texttt{gpt-oss-120b} &
\HF{openai/gpt-oss-120b} &
\multirow{2}{*}{12.6} &
\multirow{2}{*}{0.23.0} &
\multirow{2}{*}{\shortstack[c]{1$\times$ H200 NVL\\141\,GB}} \\

\texttt{gpt-oss-20b} &
\HF{openai/gpt-oss-20b} &
& & \\

\midrule

\texttt{qwen3-coder-next} &
\HF{Qwen/Qwen3-Coder-Next} &
12.6 &
0.23.0 &
\begin{tabular}[c]{@{}c@{}}
2$\times$ H200 NVL \\
141\,GB
\end{tabular} \\

\midrule

\texttt{deepseek-r1-llama-70b} &
\HF{deepseek-ai/DeepSeek-R1-Distill-Llama-70B} &
12.6 &
0.23.0 &
\begin{tabular}[c]{@{}c@{}}
4$\times$ RTX A6000 \\
48\,GB
\end{tabular} \\

\bottomrule
\end{tabular}

\label{tab:llm-eval-clusters-vllm}
\end{table}

\SPL and \SL experiments were done using two RTX 4090s. 
The implementations of \SPL were taken verbatim from \texttt{cirkit}~\citep{The_APRIL_Lab_cirkit_2024}. Datasets were either taken from their original repository or generated. All the details are provided in~\cref{app:datasets}.
All the tested models are open-weight, directly downloaded from \href{https://huggingface.co/}{HuggingFace}. Hyperparameters are in~\cref{app:llm-hyp} and~\cref{app:spl-model-selection}.

\section{Metrics}\label{app:metrics}

In this section, we go through all the metrics we used in our experiments.

\subsection{Auto-formalization}

Let $\hat{\phi}$ be the LLM-generated formula and $\phi^*$ the ground-truth formula over $N$ propositional variables. The syntax error rate is computed (after the self-verification step, if present) as
\[
    \text{SyntaxErr} = \frac{|\{i : \hat{\phi}_i \text{ fails to parse}\}|}{n}.
\]

Confusion-matrix counts are obtained via model counting ($\text{MC}$)~\citep{chavira2008wmc}: exactly when the task has $\le 20$ variables, and with \texttt{ApproxMC}~\citep{chakraborty2016approxmc} otherwise. Here, $\text{TP}$, $\text{FP}$, $\text{FN}$, and $\text{TN}$ denote true positives, false positives, false negatives, and true negatives, respectively. For formulas without auxiliary variables, all four cells are counted independently:
\[
    \text{TP} = \text{MC}(\phi^*\!\land\hat{\phi}),\ \text{FP} = \text{MC}(\neg \phi^*\!\land\hat{\phi}),\ \text{FN} = \text{MC}(\phi^*\!\land\neg\hat{\phi}),\ \text{TN} = \text{MC}(\neg \phi^*\!\land\neg\hat{\phi}),
\]
where $\neg\phi$ is encoded by \pysat Tseitin negation with auxiliary variables kept disjoint between the two negated formulas.

Formats such as \cpmpy and \sympy may introduce Tseitin auxiliary variables $A$ during conversion. Let $\hat\phi(X,A)$ denote the predicted encoding over primary variables $X$ and auxiliary variables $A$, and let $\hat\phi_X \equiv \exists A.,\hat\phi$ denote its projection onto $X$. Negating the encoding clause-by-clause is \emph{not} equivalent to negating this projection: in general,
$\neg\exists A.,\hat\phi \not\equiv \exists A.,\neg\hat\phi$,
and the latter may admit spurious assignments. We therefore \textit{never negate an encoding containing auxiliary variables}. Instead, all reported quantities are computed from $\hat\phi_X$, with the projection performed \emph{before} any negation or conjunction.

We compute only
\[
    T=\text{MC}_X(\phi^*),\quad P=\text{MC}_X(\hat\phi),\quad \text{TP}=\text{MC}_X(\phi^*\!\land\hat\phi),
\]
and derive $\text{FP}=P-\text{TP}$, $\text{FN}=T-\text{TP}$, $\text{TN}=2^N-T-P+\text{TP}$.
When \texttt{ApproxMC} is used, each of the three calls is run at confidence $\delta/3$, so all $T$, $P$ and $\text{TP}$ counts hold jointly with probability $\ge 1-\delta$ (union bound).

\subsection{Downstream NeSy predictor}

During our \SPL and \SL evaluation pipeline, we rely on standard supervised learning metrics such as F1 score, accuracy, precision and recall. Here, we describe additional metrics we used for a qualitative analysis of the downstream NeSy predictor. 
All the metrics are computed on a hold-out test set.

\textbf{Consistency (\con)}. This metric measures whether the final maximum a posteriori prediction of the NeSy predictor complies with the ground-truth formula. 

\begin{equation}
\con = \frac{1}{N} \sum_{i=1}^{N} \Ind{\hat{\vy}_i \models \phi^*},
\end{equation}

If the consistency is not $1.00$, it indicates that some predicted assignments do not satisfy the constraints. Such predictions can be harmful to the downstream application.

\textbf{Ground-truth vs predicted formula relationship (\frel)}. 
For each (model, task) pair, we classify the LLM-generated formula $\hat\phi$ against the task ground-truth formula $\phi^*$ into one of four exact relations. We test the two subset relations exactly:
\[
  \phi^* \subseteq \hat\phi \iff \phi^* \wedge \lnot \hat\phi \equiv \bot,
  \qquad\qquad
  \hat\phi \subseteq \phi^* \iff \hat\phi \wedge \lnot \phi^* \equiv \bot .
\]
The pair $(\hat\phi,\phi^*)$ is then assigned one of four relations:
\[
  \mathrm{rel}(\hat\phi,\phi^*) =
  \begin{cases}
    \texttt{equal}        & \text{if } \phi^*\subseteq \hat\phi \text{ and } \hat\phi\subseteq \phi^* \quad(\hat\phi = \phi^*)\\
    \texttt{permissive}  & \text{if } \phi^*\subseteq \hat\phi \text{ and } \hat\phi\not\subseteq \phi^* \quad(\hat\phi \Leftarrow \phi^*)\\
    \texttt{strict}  & \text{if } \hat\phi\subseteq \phi^* \text{ and } \phi^*\not\subseteq \hat\phi \quad(\hat\phi \Rightarrow \phi^*)\\
    \texttt{incomparable} & \text{otherwise} \quad(\hat\phi \parallel \phi^*)
  \end{cases}
\]
\texttt{permissive} means the LLM formula admits every valid assignment plus some invalid ones (false positives only); \texttt{strict} means it excludes some valid assignments (false negatives, which permanently remove support to valid assignments from the predictive distribution); \texttt{incomparable} means both kinds of error occur simultaneously.

\textbf{Model-count ratio (\mcr)}. We additionally report the ratio of the two formulas model counts
\[
  \mcr(\hat\phi,\phi^*) \;=\; \frac{\mc(\hat\phi)}{\max(\mc(\phi^*),\,1)},
\]
which quantifies \emph{how much} more (or less) permissive $\hat\phi$ is even when the qualitative relation is not \texttt{equal}, \eg $\mathrm{mc\_ratio}=1.75$ for a \texttt{permissive} formula means it admits 75\% more satisfying assignments than the ground-truth.

\section{Error Taxonomy} 
\label{app:errors}

When evaluating auto-formalization, we use error not only as a means of analysis, but also as feedback for self-correction~\citep{teso2019explanatory}. We partition errors into two categories: \emph{general}, comprising errors inherent to the resulting \DIMACS representation regardless of the source format, and \emph{generation}, comprising errors introduced by the particular process used to generate the \DIMACS.

\textbf{General} (applies to \DIMACS directly, and to \nat, and to \pysat/
\cpmpy/\sympy executed output):
\begin{center}
\begin{tabular}{ll}
\toprule
\textsc{Code} & \textsc{Trigger} \\
\midrule
\texttt{no\_cnf\_tags} & Output missing \texttt{<cnf>...</cnf>} tags \\
\texttt{invalid\_header} & \texttt{p cnf N M} line absent or malformed \\
\texttt{clause\_no\_zero} & Clause does not terminate with \texttt{0} \\
\texttt{empty\_clause} & A clause contains only \texttt{0} \\
\texttt{invalid\_content} & Non-integer content in a clause body \\
\texttt{missing\_num\_vars} & Header present but $N$ missing or zero \\
\texttt{clause\_count\_mismatch} & Declared clause count $M$ disagrees with body \\
\texttt{literal\_out\_of\_range} & A literal exceeds $[-N,N]\setminus\{0\}$ \\
\texttt{formula\_too\_complex} & \nat to CNF conversion blows up \\
\texttt{zero\_literal\_in\_clause} & \DIMACS $0$ delimiter not at the end of the clause \\
\bottomrule
\end{tabular}
\end{center}

\textbf{Generation} (format-specific, precedes or replaces the \DIMACS layer):
\begin{center}
\scalebox{0.9}{
\begin{tabular}{lll}
\toprule
\textsc{Format} & \textsc{Code} & \textsc{Trigger} \\
\midrule
\pysat & \texttt{execution\_error} & Program raised an error when executed \\
 & \texttt{timeout\_error} & Program execution exceeded the time limit \\
 & \texttt{encoder\_aux\_collision} & PySAT encoder auxiliary-variable index collision risk \\
\cpmpy & \texttt{execution\_error} & Program raised an error when executed \\
 & \texttt{timeout\_error} & Program execution exceeded the time limit \\
 & \texttt{unknown\_variable} & Referenced variable outside the declared mapping \\
 & \texttt{non\_boolean\_variable} & Declared variable is not Boolean \\
\sympy & \texttt{execution\_error} & Program raised an error when executed \\
 & \texttt{timeout\_error} & Program execution exceeded the time limit \\
 & \texttt{unknown\_variable} & Referenced variable outside the declared mapping \\
\nat & \texttt{no\_cnf\_tags} & Output missing \texttt{<cnf>...</cnf>} tags \\
 & \texttt{empty\_formula} & Nothing between the tags \\
 & \texttt{parse\_error} & Expression fails to parse \\
 & \texttt{invalid\_operator} & Operator other than AND/OR/NOT (any case) \\
 & \texttt{empty\_clause} & Empty clause \texttt{()} in the expression \\
 & \texttt{unknown\_variable} & A formula symbol outside the expected mapping \\
\bottomrule
\end{tabular}
}
\end{center}
A single output may trigger more than one code. The \emph{preamble auto-fix} step repairs \texttt{missing\_num\_vars} and \texttt{clause\_count\_mismatch} by recomputing the header from the clause bodies directly, without an additional LLM query; it applies wherever the \DIMACS- syntax layer applies (\DIMACS and the compiled output of \pysat/\cpmpy/ \sympy), not to \nat. Such \emph{triggers} will be contained in the self-verification step for guiding the LLM in the correction.

\section{Datasets}\label{app:datasets}

In this section, we describe all the datasets used during our evaluation. A comparison about dataset statistics, specically, clause and variable counts is provided in~\cref{tab:datasets-summary}.

\begin{table*}[!ht]
    \caption{\textbf{\autobench is NeSy-ready, while existing benchmarks are not}. \satbench contains many small problems, whereas \dcpbench contains fewer but substantially larger problems. \autobench offers a balanced mix and, importantly, includes input-output annotations for use with NeSy predictors. Variables and clauses are reported as {\sc min}~/~{\sc max}~/~{\sc mean}.
    }
    \scriptsize
    \centering
    \setlength{\tabcolsep}{4pt}
    \begin{tabular}{lcccccccc}
        \toprule
            \multirow{2}{*}{\sc Dataset}
            & \multirow{2}{*}{\sc Problems}
            & \multicolumn{3}{c}{\sc Variables}
            & \multicolumn{3}{c}{\sc Clauses}
            & \multirow{2}{*}{\shortstack{\sc NeSy-Ready}}
        \\
        \cmidrule(lr){3-5}\cmidrule(lr){6-8}
            &
            & {\sc Min} & {\sc Max} & {\sc Mean}
            & {\sc Min} & {\sc Max} & {\sc Mean}
            &
        \\
        \midrule
        \satbench~\citep{wei2025satbench}
            & $1{,}041$
            & $5$ & $90$ & $36$
            & $4$ & $50$ & $20$
            & \XMARK
        \\
        \dcpbench~\citep{michailidis2025cp}
            & $85$
            & $8$ & $1{,}102{,}080$ & $58{,}673$
            & $20$ & $7{,}095{,}776$ & $394{,}204$
            & \XMARK
        \\
        \autobench (ours)%
            & $18$
            & $9$ & $102$ & $32$
            & $17$ & $129{,}648$ & $7{,}966$
            & \CMARK
        \\
        \bottomrule
    \end{tabular}
    \label{tab:datasets-summary}
\end{table*}

\subsection{SATBench}\label{app:datasets_satbench}

\satbench~\citep{wei2025satbench} is a benchmark for evaluating the logical reasoning capabilities of LLMs through natural language descriptions of problems generated from CNF formulas. Each instance pairs a Boolean formula with a natural language scenario and descriptions of its variables, allowing models to reason about the underlying logical constraints expressed in natural language and determine whether the constraints are satisfiable.

For our auto-formalization setting, we repurpose these paired natural language descriptions and reference CNF formulas to evaluate whether models can recover the underlying logical constraints from natural language. We use the LLM4Code/SATBench Hugging Face dataset pinned at revision \texttt{186740e}. The original dataset contains $2{,}100$ instances, evenly split between SAT ($1{,}050$) and UNSAT ($1{,}050$) formulas. We discard the latter because an UNSAT formula has no valid use for a NeSy predictor. Additionally, we remove 9 SAT malformed instances (mismatch between the declared variable/clause count). This leaves $1{,}041$ instances for evaluation.

Across these instances, the number of clauses ranges from 4 to 50, and the number of variables ranges from 5 to 90. Each instance specifies Boolean variables indexed by coordinates (e.g., \texttt{x\_0\_0}), together with a natural language problem description and variable descriptions. An example is provided in \Cref{tab:satbench_example}.

\begin{table}[h!]
\centering
\caption{
\textbf{Example from \satbench.}
Each \satbench instance pairs a natural language scenario and variable descriptions with a reference Boolean formula.
Here, five community-garden volunteers, Ana, Ben, Carla, Dave, and Emma, decide whether to sign up for weekend watering duty, where $x_i$ indicates whether volunteer $i$ signs up, in the order just given. Each row shows one natural-language constraint and its corresponding CNF clause
and DIMACS encoding. The reference formula is the conjunction of all clauses. In our auto-formalization setting, the natural language scenario and constraints serve as input, and the conjunction of all constraints serves as the reference formalization.
}
\label{tab:satbench_example}

\begin{tabular}{lll}
\toprule
\textsc{Natural-language constraint}
& \textsc{CNF clause}
& \textsc{DIMACS encoding} \\
\midrule
Ana signs up for watering duty.
& $x_0$
& \texttt{1 0} \\

Either Ben does not sign up, or Ana does not sign up.
& $\neg x_1 \lor \neg x_0$
& \texttt{-2 -1 0} \\

Either Ana does not sign up, or Emma signs up.
& $\neg x_0 \lor x_4$
& \texttt{-1 5 0} \\

Either Carla does not sign up, or Ben signs up.
& $\neg x_2 \lor x_1$
& \texttt{-3 2 0} \\
\bottomrule
\end{tabular}
\end{table}

\subsection{DCPBench-Open}

DCP-Bench-Open~\citep{michailidis2025cp, dcpbenchopen} contains $164$ CP problems, including puzzles, scheduling tasks, and CSPLib instances. An input instance consists of a natural language description, a CP instance, a reference \cpmpy model, and an example solution. Unlike \satbench, these reference models can contain arithmetic constraints. 

To make this benchmark suitable for an auto-formalization task, we instantiate the reference \cpmpy models with their instance data and convert the constraints into CNF using the \texttt{to\_cnf}\footnote{\url{https://cpmpy.readthedocs.io/en/latest/api/transformations/to_cnf.html}} \cpmpy function, encoding the variables as one-hot. 
As they cannot be used by a NeSy predictor, we exclude optimization problems, constraints that result in an error during CNF conversion, and conversions that exceed a $90$ s timeout due to scalability issues~\citep{huang2021scallop}. The experiments reported in the main paper use a subset of 85 problems with usable \DIMACS targets. To avoid introducing additional bias, we remove the problem name from the natural language description.
To make the instances compatible with \satbench and \autobench, we explicitly specify the Boolean variables by enumerating them and stating their value ranges, \eg \texttt{1-101: men: integer, domain [0, 100]}.

\textbf{Example (\texttt{room\_assignment}).} The natural language description presents a room assignment problem. Specifically, we are asked to assign four room requests to five rooms, numbered $0$ to $4$, with one room per request and no overlapping requests assigned to the same room. The concrete instance has start dates May $1$, $2$, $3$, and $4$, and end dates May $5$, $6$, $7$, and $8$, respectively, all in $2024$. The first request is preassigned to room $3$. We write $r\_i$ to denote the room assigned to request $i$, the CP constraints are:
\[
    r_i \in \{0,\ldots,4\}\quad (i=0,\ldots,3),
    \qquad r_0=3,
    \qquad \operatorname{AllDifferent}(r_0,r_1,r_2,r_3).
\]
The actual variable declaration supplied in the prompt is:
\begin{MyVerbatim}[fontsize=\small]
1-5: room_assignments_0: integer, domain [0, 4]
6-10: room_assignments_1: integer, domain [0, 4]
11-15: room_assignments_2: integer, domain [0, 4]
16-20: room_assignments_3: integer, domain [0, 4]
\end{MyVerbatim}
Let $b_{i,v}$ mean $r_i=v$, an equivalent CNF over these $20$ variables is
\begin{align*}
    &\bigwedge_{i=0}^{3}\left[
        \left(\bigvee_{v=0}^{4} b_{i,v}\right)
        \land \bigwedge_{0\leq v<w\leq4}(\neg b_{i,v}\lor\neg b_{i,w})
    \right] \\
    &\qquad {}\land b_{0,3}
    \land \bigwedge_{0\leq i<j\leq3}\bigwedge_{v=0}^{4}
        (\neg b_{i,v}\lor\neg b_{j,v}).
\end{align*}
The first line enforces exactly one room per request; the second enforces the pre-assignment and prevents room sharing. 

\subsection{Auto-NeSy-Bench} 

\autobench consists of a set of 18 NeSy tasks. These are established NeSy visual-reasoning benchmarks and vary from single-digit to hundred-plus-variable formulas. \Cref{tab:programs} lists every task with its ground-truth formula size.

\begin{table}[h]
\centering
\caption{NeSy tasks and their ground-truth \DIMACS sizes.}
\label{tab:programs}
\small
\begin{tabular}{lrr}
\toprule
\textsc{Task} & \textsc{Variables} & \textsc{Clauses} \\
\midrule
\texttt{chx}~\citep{cohen2022torch} & 9 & 27 \\
\texttt{fashion}~\citep{xiao2017fashion} & 10 & 46 \\
\texttt{bdd-oia-2}~\citep{xu2020boia} & 11 & 17 \\
\texttt{cle4evr}~\citep{johnson2017clevr} & 12 & 28 \\
\texttt{cifar10}~\citep{krizhevsky2009cifar} & 15 & 157 \\
\texttt{mn-add-bin}~\citep{manhaeve2018deepproblog} & 13 & 512 \\
\texttt{mn-mul-bin}~\citep{bortolotti2024benchmark} & 15 & 712 \\
\texttt{sushi}~\citep{kamishima2003nantonac} & 16 & 56 \\
\texttt{kand-logic-2}~\citep{mueller2021kandinsky} & 24 & 1328 \\
\texttt{warcraft}~\citep{ahmed2022semantic} & 24 & 289 \\
\texttt{bdd-oia}~\citep{xu2020boia} & 25 & 31 \\
\texttt{cebab}~\citep{bortolotti2026concise} & 25 & 1070 \\
\texttt{kand-logic}~\citep{mueller2021kandinsky} & 36 & 129648 \\
\texttt{mn-add}~\citep{manhaeve2018deepproblog} & 39 & 364 \\
\texttt{road-r}~\citep{giunchiglia2023road} & 41 & 243 \\
\texttt{sudoku}~\citep{augustine2022visual} & 64 & 400 \\
\texttt{cifar100}~\citep{krizhevsky2009cifar} & 100 & 4951 \\
\texttt{mn-mul}~\citep{bortolotti2024benchmark} & 102 & 3514 \\
\bottomrule
\end{tabular}
\end{table}

\textbf{\texttt{chx}}. We use \texttt{ChestX-Rays}~\citep{cohen2022torch}, following the preprocessing of~\citet{pugnana2025ask}. The dataset consists of images of X-ray scans, annotated with four binary symptoms: \texttt{Fracture}, \texttt{Pneumothorax}, \texttt{Airspace opacity}, and \texttt{Nodule/mass}, forming the concept vector $\vc \in \{0,1\}^4$. We adopt the logical formalization introduced in~\citet{bortolotti2026concise}, where the target label is determined solely by the total number of present symptoms, \ie $\sum_{i=1}^{4} \vc_i$, mapping to five severity levels (\texttt{healthy}, \texttt{green\_code}, \texttt{yellow\_code}, \texttt{orange\_code}, and \texttt{red\_code}). The resulting dataset contains $3,150$ training, $350$ validation, and $876$ test samples.

\begin{figure}[H]
    \centering
    \includegraphics[width=0.2\linewidth]{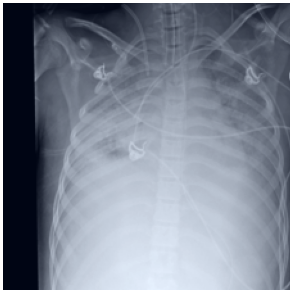}
    \caption{A sample from the \texttt{chx} dataset.}
    \label{fig:chx}
\end{figure}

\textbf{\texttt{fashion} \& \texttt{cifar100}}. Following~\citep{xu2018semantic}, we formalized \texttt{FashionMNIST}~\citep{xiao2017fashion} and \texttt{CIFAR100}~\citep{krizhevsky2009cifar} as one-hot prediction task. Formally, 
\[
    \vy_k = \Ind{\vk = \vy}.
\]
\texttt{FashionMNIST} consists of $70{,}000$ $28{\times}28$ grayscale images of garments in $10$ classes ($60{,}000$ train / $10{,}000$ test); \texttt{CIFAR100} consists of $60{,}000$ $32{\times}32$ colored images evenly distributed over $100$ fine-grained classes ($600$ images per class, standard $50{,}000$/$10{,}000$ train/test split). In both cases, we hold out $10{,}000$ training images for validation.

\begin{figure}[H]
    \centering

    \begin{subfigure}{0.40\linewidth}
        \centering
        \includegraphics[width=0.4\linewidth]{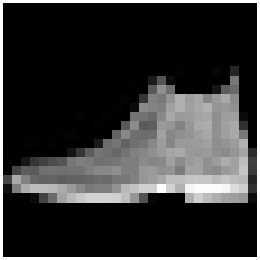}
        \caption{\texttt{fashion}}
    \end{subfigure}\hfill
    \begin{subfigure}{0.40\linewidth}
        \centering
        \includegraphics[width=0.4\linewidth]{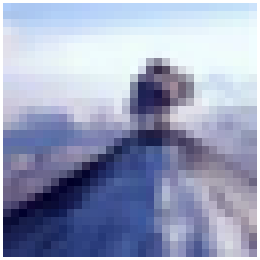}
        \caption{\texttt{cifar100}}
    \end{subfigure}
    \caption{Samples from \texttt{fashion} and \texttt{cifar100} datasets.}
    \label{fig:mutual}
\end{figure}

\textbf{\texttt{cifar10}}. We use \texttt{CIFAR-10}~\citep{krizhevsky2009cifar}, following the preprocessing and logical formalization of~\citet{bortolotti2026concise}. The original dataset contains $60,000$ images from $10$ classes; we use $35,000$ training, $15,000$ validation, and $10,000$ test samples. Each image is associated with a binary concept vector $\mathbf{c}\in \{0,1\}^7$, encoding the semantic attributes \texttt{ani}, \texttt{wng}, \texttt{met}, \texttt{whl}, \texttt{hai}, \texttt{hrn}, and \texttt{snt}. Class labels are deterministically inferred from these concepts through the logical rules of~\citet{bortolotti2026concise}. The concept-label mapping is intentionally non-injective, making the pairs \texttt{automobile}/\texttt{truck} and \texttt{dog}/\texttt{horse} indistinguishable from concepts alone.

\begin{figure}[H] 
\centering \includegraphics[width=0.2\linewidth]{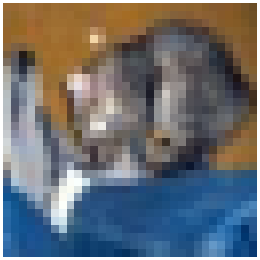} \caption{A sample from the \texttt{cifar10} dataset.} \label{fig:cifar} 
\end{figure}

\textbf{\texttt{bdd-oia} \& \texttt{bdd-oia-2}}. \texttt{BDD-OIA} is an autonomous driving task consisting of frames retrieved from driving scene videos~\citep{xu2020boia, bortolotti2024benchmark}. We provide two versions of this dataset in order to encode different levels of granularity. Originally, each frame comes with four binary labels (\texttt{move\_forward}, \texttt{stop}, \texttt{turn\_left}, and \texttt{turn\_right}) and $21$ binary concepts that provide an explanation for the selected actions. 
The first version adopts a similar prior knowledge defined in~\citet{marconato2023not}, which we briefly summarize here for completeness. For the \texttt{move\_forward} and \texttt{stop} actions, the following logical rules are used:
\[
\begin{aligned}
&O := \texttt{car} \lor \texttt{person}
      \lor \texttt{rider} \lor \texttt{other\_obstacle},\\
&F := \texttt{green\_light} \lor \texttt{follow}
      \lor \texttt{road\_clear},\\
&S := \texttt{red\_light} \lor \texttt{traffic\_sign} \lor O,\\
&\texttt{move\_forward} \Longleftrightarrow F \land \neg S,\\
&\texttt{stop} \Longleftrightarrow S.
\end{aligned}
\]
For the turning actions, the following rules define the conditions for executing a left turn:
\[
\begin{aligned}
&L^{+} := \texttt{left\_lane}
          \lor \texttt{left\_green\_light}
          \lor \texttt{left\_follow},\\
&L^{-} := \texttt{no\_left\_lane}
          \lor \texttt{left\_obstacle}
          \lor \texttt{left\_solid\_line},\\
&\texttt{turn\_left} \Longleftrightarrow L^{+} \land \neg L^{-}.
\end{aligned}
\]

The rules for \texttt{turn\_right} are defined analogously by replacing the left-related concepts with their corresponding right-related counterparts. 
The symbols $O$, $F$, $S$, $L^{+}$, and $L^{-}$ are abbreviations, not additional benchmark variables.
The second version of the dataset instead simply filters the concepts and action maintaining only those related to the \texttt{move\_forward} and \texttt{stop} actions.
The dataset itself consists of 16k fully annotated training frames, along with 2k and 4.5k labeled frames for validation and testing, respectively. Instead of processing the full frames, we rely on the embeddings of a pretrained Faster-RCNN~\citep{ren2016faster} provided by~\citet{bortolotti2024benchmark}.

\textbf{\texttt{mn-add} \& \texttt{mn-add-bin}}.  \MNISTAdd, originally introduced in~\citet{manhaeve2018deepproblog}, is a canonical benchmark for NeSy AI. Each sample consists of two \MNIST~\citep{lecun1998mnist} digit images $(\vx_1,\vx_2)$ representing digits $(\vc_1,\vc_2)\in[0,9]^2$, with the target defined as their arithmetic sum:
\[
    \vy = \vc_1 + \vc_2.
\]
The dataset contains $54$k training, $6$k validation, and $10$k test examples. We provide two versions of this task, one in which the digits are encoded as one-hot, and another in which the digits are encoded as a single binary value.

\begin{figure}[H] 
\centering \includegraphics[width=0.2\linewidth]{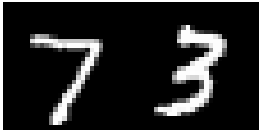} \caption{A sample from the \texttt{mn-add}/\texttt{mn-mul} dataset.} 
\label{fig:mnist}
\end{figure}

\textbf{\texttt{mn-mul} \& \texttt{mn-mul-bin}}. \texttt{MN-Mul} is a variant of \MNISTAdd~\citep{manhaeve2018deepproblog}. Each sample consists of two \MNIST~\citep{lecun1998mnist} digit images $(\vx_1,\vx_2)$ representing digits $(\vc_1,\vc_2)\in[0,9]^2$, with the target defined as their arithmetic product:
\[
    \vy = \vc_1 * \vc_2.
\]
The dataset contains $54$k training, $6$k validation, and $10$k test examples. We provide two versions of this task, one in which the digits are encoded as one-hot, and another in which the digits are encoded as a single binary value.

\textbf{\texttt{sushi}}. We use the \texttt{SUSHI3} tabular preference dataset~\citep{kamishima2003nantonac}, in which $5{,}000$ respondents each provide a total order over $10$ sushi varieties, and adopt the ranking formalization of~\citet{xu2018semantic}. Each order is split into a $6$-item context sub-ranking (conditioning input) and a $4$-item label sub-ranking. The label is encoded as a $4\times4$ permutation matrix: variable $\vx_{ij}$ is true iff item $i$ is assigned rank $j$, for $i,j\in\{0,1,2,3\}$ ($16$ Boolean variables). A valid assignment is exactly a permutation matrix, \ie every item receives exactly one rank and every rank is taken by exactly one item. We split the $5{,}000$ respondents into $4{,}000$/$500$/$500$ for train/validation/test.

\textbf{\texttt{kand-logic} \& \texttt{kand-logic-2}}. \texttt{Kand-Logic} is a two-image relational variant of the Kandinsky-pattern benchmark~\citep{mueller2021kandinsky}, provided through \rsbench~\citep{bortolotti2024benchmark}. Each sample consists of two images, each containing three geometric primitives; every primitive carries a shape (\texttt{square}, \texttt{triangle}, \texttt{circle}) and a color (\texttt{red}, \texttt{blue}, \texttt{yellow}), each being an one-hot vector, \ie $6$ variables per primitive, $18$ per image, $36$ in total. The pair is positive iff both images realize the same \emph{pattern category} on shape or on color, where a category is one of \texttt{same} (all three primitives equal), \texttt{pair} (exactly two equal), or \texttt{diff} (all three distinct). We also include \texttt{kand-logic-2}, which uses two primitives per image ($24$ variables): with only two primitives, the \texttt{pair} category is undefined and \texttt{same}/\texttt{diff} become complementary. Training data is produced by running \rsbench \texttt{SyntheticKandinskyGenerator} to emit single-figure images, pairing images that share a pattern category, and verifying every sample against the ground-truth formula before writing it; we generate $800$/$100$/$100$ train/validation/test samples per variant.

\begin{figure}[H]
    \centering

    \begin{subfigure}{0.40\linewidth}
        \centering
        \includegraphics[width=0.8\linewidth]{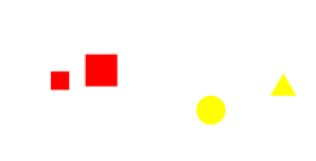}
        \caption{\texttt{kand-logic-2}}
    \end{subfigure}\hfill
    \begin{subfigure}{0.40\linewidth}
        \centering
        \includegraphics[width=0.8\linewidth]{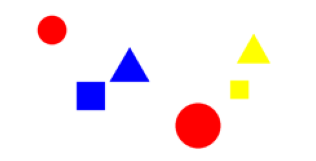}
        \caption{\texttt{kand-logic}}
    \end{subfigure}
    \caption{Samples from \texttt{kand-logic-2}/\texttt{kand-logic} datasets.}
    \label{fig:kand}
\end{figure}

\textbf{\texttt{sudoku}}. We use a $4\times4$ \texttt{VisualSudoku} task in the spirit of~\citet{augustine2022visual}: each cell of a $4\times4$ grid is rendered with an \MNIST image ($1$ to $4$), and the model must decide whether the depicted grid is a valid solution. The one-hot encoding uses variable $\vx_{ijv}$ for ``cell $(i,j)$ holds digit $v$'' with $i,j\in\{0,\dots,3\}$ and $v\in\{1,\dots,4\}$ ($64$ Boolean variables). An assignment is valid iff it is a complete solution: exactly one digit per cell and each digit appearing exactly once in every row, every column, and every $2\times2$ box. We use $4{,}000$/$500$/$500$ train/validation/test grids drawn from the $288$ valid boards (with random digit relabelings).

\begin{figure}[H]
    \centering
    \includegraphics[width=0.3\linewidth]{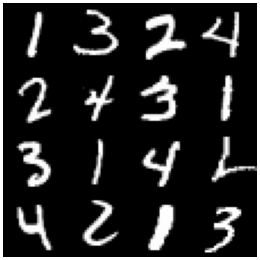}
    \caption{A sample from the \texttt{sudoku} dataset.}
    \label{fig:sudoku}
\end{figure}

\textbf{\texttt{road-r}}. \texttt{ROAD-R}~\citep{giunchiglia2023road} is an extension of the ROAD autonomous driving dataset with a set of hand-written logical requirements over the label space. Each frame is annotated with an agent class (one of $10$), a traffic-light color (\texttt{red}/\texttt{amber}/\texttt{green}, at most one), a set of actions (\texttt{16}), and a set of locations (\texttt{12}), which result in $41$ Boolean variables ($1$-$10$ agent, $11$-$13$ color, $14$-$29$ action, $30$-$41$ location). Unlike the other tasks, the target CNF is not compiled from a Python program: it is the requirements are provided by the authors ($41$ variables, $243$ clauses). Samples are $8$-frame clips ending at each annotated key-frame. It consists of $\sim\!47$k samples, split $80$/$10$/$10$ by source video.

\textbf{\texttt{warcraft}}. We use a $4\times4$ \texttt{Warcraft} shortest-path task in the spirit of~\citet{vlastelica2019differentiation}. For simplicity, maps are procedurally generated: each of the $16$ grid cells is assigned one of $5$ terrain classes i.i.d.\ uniformly at random, then rendered with a real \texttt{Warcraft}~II terrain tile of the matching class. The model must decide whether the depicted set of grid edges is the minimum-cost simple path from the top-left cell $(0,0)$ to the bottom-right cell $(3,3)$. The encoding uses one Boolean variable per grid edge, \ie $24$ Boolean variables in total. An assignment is valid iff it is a complete solution: the source and sink each have exactly one selected incident edge and they trace a single simple path between the two corners. Each of the $5$ terrain classes $k\in\{0,\dots,4\}$ carries a per-cell traversal cost $c_k = 4^k$ (\ie \ $1,4,16,64,256$). The ground-truth optimal path is computed in advance. We use $8{,}000$/$1{,}000$/$1{,}000$ train/validation/test grids, each rendered as a $64\times64$ image.

\begin{figure}[H]
    \centering
    \includegraphics[width=0.3\linewidth]{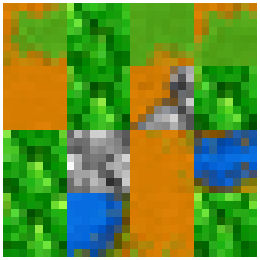}
    \caption{A sample from the \texttt{warcraft} dataset.}
    \label{fig:warcraft}
\end{figure}

\textbf{\texttt{cebab}.} We use \texttt{CEBaB}~\citep{abraham2022cebab}, a dataset of restaurant reviews annotated via human-mediated counterfactual edits along four aspects: \texttt{food}, \texttt{service}, \texttt{noise}, and \texttt{ambiance}, each labeled with a majority sentiment (\texttt{Positive}, \texttt{Negative}, \texttt{Neutral}, or \texttt{unknown}), forming four one-hot arrays. We adopt the logical formalization of~\citet{bortolotti2026concise}, which adds a fifth concept, \texttt{review\_majority}, obtained from the original $1$ to $5$ star rating into the same four values, giving $\vc\in\{0,1\}^{20}$. The target label is a one-hot vote over the five concepts (\texttt{positive}, \texttt{negative}, \texttt{neutral}, \texttt{unknown}, \texttt{conflict}): the majority class among the five wins, higher-priority classes break ties in their favor, and a \texttt{positive}/\texttt{negative} tie at the overall maximum yields \texttt{conflict}. We use the CEBaB v$1.1$ observational split: $1{,}755$ training, $1{,}673$ validation, and $1{,}689$ test reviews.

\textbf{Sample from \texttt{cebab}}.

\begin{verbatim}
Review text:
    "Limoncello has excellent Italian food. So good, 
    that it tastes homemade. The only drawback is that 
    it is so loud in the restaurant."

GROUND-TRUTH ANNOTATION:
    food     aspect majority = Positive
    service  aspect majority = unknown
    noise    aspect majority = Negative
    ambiance aspect majority = unknown
    review   majority        = Positive   (raw star rating: "4")
\end{verbatim}

\textbf{\texttt{cle4evr}.} \texttt{CLE4EVR} is a two-object relational variant of the \texttt{CLEVR} visual reasoning benchmark~\citep{johnson2017clevr}, provided through \rsbench~\citep{bortolotti2024benchmark}. Each image depicts exactly two 3D-rendered objects; every object carries a shape (\texttt{cube}, \texttt{sphere}, \texttt{cylinder}) and a color (\texttt{red}, \texttt{blue}, \texttt{green}), each a $3$-way one-hot vector with $6$ variables per object, $12$ in total. An assignment is valid iff the two objects share the same shape \emph{and} the same color. Scenes are rendered via Blender/Cycles, with both objects constructed to match by construction so every sample satisfies the ground-truth formula by design. We generate $5{,}000$/$2{,}000$/$5{,}000$ train/validation/test images.

\begin{figure}[H]
    \centering
    \includegraphics[width=0.3\linewidth]{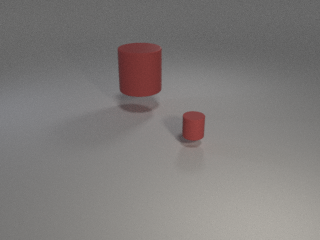}
    \caption{A sample from the \texttt{cle4evr} dataset.}
    \label{fig:cle4evr}
\end{figure}

\textbf{Construction of the target \DIMACS}. We use \pysat~\citep{imms-sat18} to compile each task validity predicate into an equivalent \DIMACS CNF over the variable layout described above. Every compiled formula is cross-checked against a Python oracle, exhaustively when the variable count permits ($\le 2^{20}$ assignments).\footnote{All formulas were independently written and consistently checked by two of the authors.} \texttt{road-r} is the only exception: its CNF is taken unchanged from its dataset rather than compiled.

\textbf{Constructing the task descriptions.} Each task is presented to the LLM using a fixed prompt scaffold (\cref{app:prompts}), instantiated with two hand-written components: a \emph{constraint description} and a \emph{variable list}. 
We provide the constraint description in two variants with different levels of granularity.
The \emph{detailed} description presents each rule and explicitly lists the background facts that should \emph{not} be encoded.
The \emph{not detailed} description details the same intended task more concisely in ordinary prose, leaving more of the rule decomposition to the model and, in some cases, leaving constraints or modeling conventions implicit (\eg ``the car may move forward iff it is allowed to proceed and the road ahead is clear''). Full prompts are shown in~\cref{app:problemtexts}.
Both descriptions, as well as the per-variable descriptions, were independently written by two of the authors, who first impersonated an expert and then a non-expert who at least knew the task they wanted to solve. The authors then compared their versions clause by clause and reconciled all discrepancies until reaching full agreement; the released texts are the resulting consensus versions.
Thus, the constraint-description explicitness is the only factor that differs between the natural and formalized conditions: the surrounding prompt template, variable numbering, and injected variable list are byte-identical across conditions.

\subsubsection{Licensing and Redistribution}

We plan to release the benchmark under a two-license split, following common practice for mixed-provenance NeSy and vision benchmarks: one license for our own code, and one for the data artifacts we redistribute. We plan to distribute the code under the \texttt{BSD-3-Clause}\footnote{\url{https://opensource.org/license/bsd-3-clause}} license.

\textbf{Redistributed data artifacts.} We plan to release labels, CNF targets, and any bundled images or features under \texttt{CC~BY-NC-SA~4.0}\footnote{\url{https://creativecommons.org/licenses/by-nc-sa/4.0/deed.en}}. This is the most restrictive license among the datasets in \autobench and is inherited from \texttt{ROAD-R}~\citep{giunchiglia2023road}, which is released under \texttt{CC~BY-NC-SA~4.0}. All other bundled datasets are distributed under licenses that are equally or more permissive (\texttt{MIT}\footnote{\url{https://opensource.org/license/mit}}, \texttt{BSD-3-Clause}, \texttt{CC~BY~4.0}, or unrestricted). The datasets \texttt{mn-add}, \texttt{mn-mul} (and their \texttt{-bin} variants), \texttt{kand-logic}, \texttt{kand-logic-2}, \texttt{bdd-oia}, and \texttt{bdd-oia-2} are sourced from \texttt{rsbench}\footnote{\url{https://unitn-sml.github.io/rsbench}} through its companion codebase, \texttt{rsbench-code}\footnote{\url{https://github.com/unitn-sml/rsbench-code}}~\citep{bortolotti2024benchmark}. Detailed licensing information for all datasets is provided in~\cref{tab:licensing}.

\begin{table}[h!]
\centering\scriptsize
\caption{\textbf{License and redistribution status of every \autobench datasets}. ``Redistributed'' indicates whether the data
itself can be shipped in a new benchmark, as opposed to being fetched from the original source.}
\label{tab:licensing}
\begin{tabularx}{\textwidth}{@{}l l X l@{}}
\toprule
\textsc{Dataset} & \textsc{Task(s)} & \textsc{License} & \textsc{Redistributed} \\
\midrule
\texttt{MNIST} \citep{lecun1998mnist} &
  \texttt{mn-add(mul)(-bin)} &
  Unrestricted (NIST-derived) & \CMARK \\
\addlinespace[2pt]
\texttt{Fashion-MNIST} \citep{xiao2017fashion} &
  \texttt{fashion} &
  MIT & \CMARK \\
\addlinespace[2pt]
\texttt{CIFAR-based} \citep{krizhevsky2009cifar} &
  \texttt{cifar10(100)} &
  No explicit license & \XMARK \\
\addlinespace[2pt]
\texttt{BDD-OIA}~\citep{xu2020boia} &
  \texttt{bdd-oia(-2)} &
  See~\citep{bortolotti2024benchmark} &  \CMARK \\
\addlinespace[2pt]
\texttt{ROAD-R} \citep{giunchiglia2023road} &
  \texttt{road-r} &
  CC~BY-NC-SA~4.0 & \CMARK \\
\addlinespace[2pt]
\texttt{SUSHI3} \citep{kamishima2003nantonac} &
  \texttt{sushi} &
  Research use permitted; redistribution forbidden. & \XMARK \\
\addlinespace[2pt]
\texttt{CEBaB} \citep{abraham2022cebab} &
  \texttt{cebab} &
  CC~BY~4.0 & \CMARK \\
\addlinespace[2pt]
\texttt{ChestX-Rays} \citep{cohen2022torch} &
  \texttt{chx} &
  Unrestricted (NIH), attribution requested & \CMARK \\
\addlinespace[2pt]
\texttt{Warcraft tiles} \citep{vlastelica2019differentiation} &
  \texttt{warcraft} &
  MIT & \CMARK \\
\addlinespace[2pt]
\texttt{Kandinsky/CLEVR} \citep{bortolotti2024benchmark} &
  \texttt{kand-logic(-2)}, \texttt{cle4evr} &
  BSD-3-Clause & \CMARK \\
\bottomrule
\end{tabularx}
\end{table}

\textbf{Datasets not redistributed.} For the datasets marked \XMARK above, the released codebase cannot bundle the data. Instead, it will provide a download script that retrieves each dataset directly from its original source when first needed.

\texttt{SUSHI3.} This dataset is not redistributed with the benchmark and must be downloaded directly from Kamishima's archive (\url{https://www.kamishima.net/sushi/}). Its license explicitly states that ``You must not redistribute.'' While research use is permitted, redistribution of the dataset is expressly prohibited.

\texttt{CIFAR-10 / CIFAR-100.} These datasets cannot be redistributed with a new benchmark and can be instead downloaded on demand via \texttt{torchvision} (\url{https://www.cs.toronto.edu/~kriz/cifar.html}). We follow this approach because the datasets do not provide an explicit license or redistribution grant from their creators.

\subsubsection{Constraint descriptions (detailed vs. non-detailed)}
\label{app:problemtexts}

For each task, we give the \texttt{\{problem\}} text in both phrasings. Here we list all the prompts used for \autobench.

\textbf{\texttt{mn-add-bin}.}
\begin{MyVerbatim}
[DETAILED]
We want to formalize the single-digit addition task (full range: digits 0-9).

Logical constraints:
- The first digit is binary encoded (value between 0 and 9).
- The second digit is binary encoded (value between 0 and 9).
- The sum is binary encoded (value between 0 and 18).
- The sum is equal to the first digit plus the second digit in binary.

[NOT DETAILED]
We want to formalize the single-digit addition task (full range: digits 0-9).

Logical constraints:
- The sum is equal to the first digit plus the second digit in binary.
\end{MyVerbatim}

\textbf{\texttt{mn-add}.}
\begin{MyVerbatim}
[DETAILED]
We want to formalize the single-digit addition task (full range: digits 0-9).

Logical constraints:
- The first digit must have exactly one value from 0 to 9.
- The second digit must have exactly one value from 0 to 9.
- The sum must have exactly one value from 0 to 18.
- The sum must equal the first digit plus the second digit.

[NOT DETAILED]
We want to formalize the single-digit addition task (full range: digits 0-9).

Logical constraints:
- The sum must equal the first digit plus the second digit.
\end{MyVerbatim}

\textbf{\texttt{mn-mul-bin}.}
\begin{MyVerbatim}
[DETAILED]
We want to formalize the single-digit multiplication task (full range: digits 0-9).

Logical constraints:
- The first digit is binary encoded (value between 0 and 9).
- The second digit is binary encoded (value between 0 and 9).
- The product is binary encoded (value between 0 and 81).
- The product is equal to the first digit multiplied by the second digit in binary.

[NOT DETAILED]
We want to formalize the single-digit multiplication task (full range: digits 0-9).

Logical constraints:
- The product is equal to the first digit multiplied by the second digit in binary.
\end{MyVerbatim}

\textbf{\texttt{mn-mul}.}
\begin{MyVerbatim}
[DETAILED]
We want to formalize the single-digit multiplication task (full range: digits 0-9).

Logical constraints:
- The first digit must have exactly one value from 0 to 9.
- The second digit must have exactly one value from 0 to 9.
- The product must have exactly one value from 0 to 81.
- The product is equal to the first digit multiplied by the second digit.

[NOT DETAILED]
We want to formalize the single-digit multiplication task (full range: digits 0-9).

Logical constraints:
- The product is equal to the first digit multiplied by the second digit.
\end{MyVerbatim}

\textbf{\texttt{chx}.}
\begin{MyVerbatim}
[DETAILED]
We want to formalize a clinical severity-scoring rule for chest X-ray findings.

Logical constraints:
The severity code is determined by the number of findings that are present:
- no findings correspond to healthy
- one finding to green_code
- two findings to yellow_code
- three findings to orange_code
- four findings to red_code.

Exactly one of the emergency codes must be true.

[NOT DETAILED]
We want to formalize a clinical severity-scoring rule for chest X-ray findings.

Logical constraints:
- The severity code is determined by the number of findings that are present
  (from 0 - healthy, to 4 - red code).
- Exactly one of the emergency codes must be true.
\end{MyVerbatim}

\textbf{\texttt{fashion}.}
\begin{MyVerbatim}
[DETAILED]
We want to formalize a constraint on a dress classification task: every clothing image
belongs to exactly one of the classes.

Logical constraints:
- The item must belong to exactly one class.
- No two classes can be assigned to the same item.

[NOT DETAILED]
We want to formalize a constraint on a dress classification task: every clothing image
belongs to exactly one of the classes.

Logical constraints:
- The item must belong to exactly one class.
\end{MyVerbatim}

\textbf{\texttt{cifar10}.}
\begin{MyVerbatim}
[DETAILED]
We want to formalize an object classification task.

Logical constraints:
- The object is an airplane if it has wings and is metallic, but has no wheels and is not an
  animal, hairy, horned, or snouted.
- The object is either a car or truck if it has wheels and is metallic, but has no wings and
  is not an animal, hairy, horned, or snouted.
- The object is a bird if it is an animal with wings, but has no wheels and is not metallic,
  hairy, horned, or snouted.
- The object is a frog if it is an animal with no wheels, wings, metallic surface, hair,
  horns, or snout.
- The object is a deer if it is an animal that is hairy and has horns and a snout, but has no
  wheels, wings, or metallic surface.
- The object is a cat if it is an animal that is hairy, but has no wheels, wings, metallic
  surface, horns, or snout.
- The object is either a dog or horse if it is an animal that is hairy and has a snout, but
  has no wheels, wings, metallic surface, or horns.
- The object is a ship if it is metallic, but has no wheels or wings and is not an animal,
  hairy, horned, or snouted.
- Exactly one of the object class must be true.

If the object does not exactly match any of these descriptions, it cannot be classified as
any of the above.

[NOT DETAILED]
We want to formalize an object classification task.

Logical constraints:
- An airplane has wings and is metallic.
- A car or truck has wheels and is metallic.
- A bird is an animal with wings.
- A frog is an animal with none of the listed physical features.
- A deer is an animal with hair, horns, and a snout.
- A cat is an animal with hair.
- A dog or horse is an animal with hair and a snout.
- A ship is metallic but has no animal features.
- Exactly one class must be true.

If the object does not exactly match any of these descriptions, it cannot be classified as
any of the above.
\end{MyVerbatim}

\textbf{\texttt{cifar100}.}
\begin{MyVerbatim}
[DETAILED]
We want to formalize a constraint on a general object classification task: every image
belongs to exactly one of the classes.

Logical constraints:
- The object belongs to exactly one class.
- No two classes can be assigned to the same object.

[NOT DETAILED]
We want to formalize a constraint on a general object classification task: every image
belongs to exactly one of the classes.

Logical constraints:
- The object belongs to exactly one class.
\end{MyVerbatim}

\textbf{\texttt{sushi}.}
\begin{MyVerbatim}
[DETAILED]
We want to formalize the validity of a ranking of sushi types across preference ranks.

Logical constraints:
- Each sushi item must be assigned to exactly one preference rank.
- Every rank must be assigned to exactly one sushi item.
- Each sushi item cannot be assigned to more than one rank.
- No two sushi items can share the same rank.

[NOT DETAILED]
We want to formalize the validity of a ranking of sushi types across preference ranks.

Logical constraints:
- Encode a valid permutation matrix representing a complete ranking of the four sushi items.
\end{MyVerbatim}

\textbf{\texttt{sudoku}.}
\begin{MyVerbatim}
[DETAILED]
We want to formalize a grid validity task (4x4 grid).

Logical constraints:
- Each cell must contain exactly one digit.
- Every row must contain each digit exactly once.
- Every column must contain each digit exactly once.
- Every 2x2 box must contain each digit exactly once.
- The four 2x2 boxes are:
    - Rows 0-1 and Columns 0-1
    - Rows 0-1 and Columns 2-3
    - Rows 2-3 and Columns 0-1
    - Rows 2-3 and Columns 2-3
- The constraints apply simultaneously to all 16 cells of the grid

[NOT DETAILED]
We want to formalize a grid validity task (4x4 grid).

Logical constraints:
- Each cell must contain exactly one digit.
- Each digit 1-4 appears exactly once in each row, each column, and every 2x2 box.
\end{MyVerbatim}

\textbf{\texttt{warcraft}.}
\begin{MyVerbatim}
[DETAILED]
We want to formalize a grid-pathfinding validity task on a 4x4 grid.

Logical constraints:
- The selected variables represent the edges belonging to a path on the 4x4 grid.
- The path must start at the top-left cell (0,0).
- Exactly one of the two edges incident to the start cell (right or down) must be selected.
- The path must end at the bottom-right cell (3,3).
- Exactly one of the two edges incident to the end cell (left or up) must be selected.
- Every other cell in the grid must have either:
    - no incident selected edges (the path does not pass through the cell), or
    - exactly two incident selected edges (one edge enters the cell and one edge leaves it).
- No intermediate cell may have exactly one or more than two incident selected edges.
- The selected edges must form a continuous path connecting the start cell to the end cell.
- The selected edges must not contain any closed cycle.
- There must not be any disconnected loop or additional cycle elsewhere in the grid.
- The selected edges must form exactly one simple path from cell (0,0) to cell (3,3).

[NOT DETAILED]
We want to formalize a grid-pathfinding validity task on a 4x4 grid.

Logical constraints:
- The selected edges should form one simple path on the 4x4 grid starting from cell (0,0)
  to cell (3,3).
- No loops, cycles, or disconnected paths are allowed.
\end{MyVerbatim}

\textbf{\texttt{kand-logic}.}
\begin{MyVerbatim}
[DETAILED]
We want to formalize a pattern-recognition task: whether two images share a common
shape/color pattern.

Logical constraints:
- Each primitive has exactly one shape and one color.
- The two images must share either the same color pattern or the same shape pattern:
    - a color pattern is defined by whether the three primitives in an image all have the
      same color, exactly two share the same color, or all three have different colors.
    - a shape pattern is defined by whether the three primitives all have the same shape,
      exactly two share the same shape, or all three have different shapes.

[NOT DETAILED]
We want to formalize a pattern-recognition task: whether two images share a common
shape/color pattern.

Logical constraints:
- The two images must share a pattern. This happens if the same color/shape pattern applies
  to all three objects, exactly two objects, or none in both images.
\end{MyVerbatim}

\textbf{\texttt{kand-logic-2}.}
\begin{MyVerbatim}
[DETAILED]
We want to formalize a pattern-recognition task: whether two images share a common
shape/color pattern.

Logical constraints:
- Each primitive has exactly one shape and one color.
- The two images must share either the same color pattern or the same shape pattern:
    - a color pattern is defined by whether the two primitives in an image have the same
      color or different colors.
    - a shape pattern is defined by whether the two primitives have the same shape or
      different shapes.

[NOT DETAILED]
We want to formalize a pattern-recognition task: whether two images share a common
shape/color pattern.

Logical constraints:
- The two images must share a pattern. This happens if the same color/shape pattern applies
  to both objects or to neither in both images.
\end{MyVerbatim}

\textbf{\texttt{bdd-oia}.}
\begin{MyVerbatim}
[DETAILED]
We want to formalize a decision task: choosing driving actions from a set of binary
road-condition variables.

Domain notes (background only, do NOT encode these as CNF clauses; the concept variables
above are free inputs, not constrained against each other):
  - red_light and green_light are not physically both true at once.
  - road_clear is true exactly when none of car/person/rider/other_obstacle hold.

Logical constraints:
- The car can move forward if and only if there is a green light, it can follow the vehicle
  ahead, or the road is clear, provided that there is no red light and nothing is blocking
  the way, such as a traffic sign, a vehicle blocking the passage, a pedestrian, a cyclist,
  or any other obstacle.
- The car must stop if and only if there is a red light indicating that it must stop, a
  traffic sign requiring it to stop, or something is blocking the way, such as a vehicle
  blocking the passage, a pedestrian, a cyclist, or any other obstacle.
- The car can turn left if and only if there is a left lane available, the left traffic light
  is green, or it can follow another vehicle turning left, provided that there is no
  indication that no left lane is available, no obstacle is blocking the left turn, and there
  is no solid line preventing the maneuver.
- The car can turn right if and only if there is a right lane available, the right traffic
  light is green, or it can follow another vehicle turning right, provided that there is no
  indication that no right lane is available, no obstacle is blocking the right turn, and
  there is no solid line preventing the maneuver.

[NOT DETAILED]
We want to formalize a decision task: choosing driving actions from a set of binary
road-condition variables.

Logical constraints:
- The car can move forward if and only if it is allowed to proceed and the road ahead is
  clear.
- The car must stop if and only if a traffic signal, sign, or obstacle requires it to stop.
- The car can turn left if and only if the maneuver is permitted and the left lane is clear.
- The car can turn right if and only if the maneuver is permitted and the right lane is
  clear.

Note: red_light and green_light are not physically both true at once, and road_clear is true
exactly when none of car/person/rider/other_obstacle hold, but these are background facts
about how the inputs behave in practice, not additional rules to encode.
\end{MyVerbatim}

\textbf{\texttt{bdd-oia-2}.}
\begin{MyVerbatim}
[DETAILED]
We want to formalize a decision task: choosing driving actions from a set of binary
road-condition variables.

Domain notes (background only, do NOT encode these as CNF clauses; the concept variables
above are free inputs, not constrained against each other):
  - red_light and green_light are not physically both true at once.
  - road_clear is true exactly when none of car/person/rider/other_obstacle hold.

Logical constraints:
- The car can move forward if and only if there is a green light, it can follow the vehicle
  ahead, or the road is clear, provided that there is no red light and nothing is blocking
  the way, such as a traffic sign, a vehicle blocking the passage, a pedestrian, a cyclist,
  or any other obstacle.
- The car must stop if and only if there is a red light indicating that it must stop, a
  traffic sign requiring it to stop, or something is blocking the way, such as a vehicle
  blocking the passage, a pedestrian, a cyclist, or any other obstacle.

[NOT DETAILED]
We want to formalize a decision task: choosing driving actions from a set of binary
road-condition variables.

Logical constraints:
- The car can move forward if and only if it is allowed to proceed and the road ahead is
  clear.
- The car must stop if and only if a traffic signal, sign, or obstacle requires it to stop.

Note: red_light and green_light are not physically both true at once, and road_clear is true
exactly when none of car/person/rider/other_obstacle hold, but these are background facts
about how the inputs behave in practice, not additional rules to encode.
\end{MyVerbatim}

\textbf{\texttt{cle4evr}.}
\begin{MyVerbatim}
[DETAILED]
We want to formalize a co-occurrence matching task between two objects in a scene.

Logical constraints:
- Both objects must have the same shape and the same color.
- Each object must have exactly one shape and one color.

[NOT DETAILED]
We want to formalize a co-occurrence matching task between two objects in a scene.

Logical constraints:
- Both objects must have the same shape and the same color.
\end{MyVerbatim}

\textbf{\texttt{cebab}.}
\begin{MyVerbatim}
[DETAILED]
We want to formalize an overall-sentiment aggregation rule for restaurant reviews.

Logical constraints:
- Exactly one output label must be true.
- The label is positive if and only if positive sentiment is the most common sentiment
  across the five sentiments, including cases where it ties with neutral or unknown, but not
  with negative.
- The label is negative if and only if negative sentiment is the most common sentiment
  across the five sentiments, including cases where it ties with neutral or unknown, but not
  with positive.
- The label is neutral if and only if neutral sentiment is strictly more common than both
  positive and negative sentiment, including cases where it ties with unknown.
- The label is unknown if and only if unknown is strictly more common than positive,
  negative, and neutral sentiment.
- The label is conflict if and only if positive and negative sentiment are tied as the most
  common sentiments.

[NOT DETAILED]
We want to formalize an overall-sentiment aggregation rule for restaurant reviews.

Logical constraints:
- The output label is determined by a majority vote over the five input sentiments:
  positive, negative, neutral, and unknown.
- In case of a tie, use the priority order: Positive > Negative > Neutral > Unknown.
- If positive and negative are tied for the highest count, the output is Conflict.
\end{MyVerbatim}

\textbf{\texttt{road-r}.}
\begin{MyVerbatim}
[DETAILED]
We want to formalize a consistency-checking task from autonomous driving. Picture a single
frame of dash-cam footage from a self-driving car, with a box drawn around one detected road
user or traffic light. An annotator has labeled that one detection with a whole set of tags
at once, i.e., what kind of agent it is, what it's doing, where it is, and, if it's a
traffic light, what color it's showing.

Logical constraints:
- Exactly one agent class must be true: pedestrian, car, cyclist, motorbike, medium vehicle,
  large vehicle, bus, emergency vehicle, traffic light, or other traffic light.
- At least one location must be true among vehicle lane, outgoing lane, outgoing cycle lane,
  incoming lane, incoming cycle lane, pavement, left pavement, right pavement, junction,
  crossing location, bus stop, or parking area. Traffic-light locations are also permitted
  by the stated requirement.
- If the detection is pushing an object, it must be a pedestrian.
- A pedestrian must have at least one of: pushing an object, moving away, moving towards,
  moving, stopped, turning left, turning right, waiting to cross, crossing from the left,
  crossing from the right, or crossing.
- Crossing, crossing from the left, crossing from the right, moving, moving towards, moving
  away, or stopped requires the detection to be a road user: pedestrian, car, cyclist,
  motorbike, medium vehicle, large vehicle, bus, or emergency vehicle.
- Waiting to cross requires the detection to be a pedestrian or cyclist.
- An emergency vehicle, bus, large vehicle, motorbike, or car must have at least one of the
  specified driving actions: overtaking, moving away, moving towards, moving, braking,
  stopped, indicating left, indicating right, hazard lights on, turning left, turning right,
  crossing from the left, crossing from the right, or crossing.
- A cyclist must have at least one action from crossing, moving away, moving towards, moving,
  braking, stopped, indicating left, indicating right, turning left, turning right,
  overtaking, waiting to cross, crossing from the left, or crossing from the right.
- Hazard lights require the detection to be an emergency vehicle, car, medium vehicle, large
  vehicle, bus, or motorbike.
- Overtaking, indicating left, indicating right, or braking requires the detection to be a
  cyclist, medium vehicle, large vehicle, bus, motorbike, emergency vehicle, or car.
- Turning left or turning right requires the detection to be a car, cyclist, motorbike,
  medium vehicle, large vehicle, bus, or emergency vehicle.
- If crossing is true, the detection must be a pedestrian or cyclist.
- The following agent classes are mutually exclusive:
    - car with cyclist, motorbike, medium vehicle, large vehicle, bus, emergency vehicle,
      traffic light, other traffic light, or pedestrian
    - cyclist with motorbike, medium vehicle, large vehicle, bus, emergency vehicle, traffic
      light, other traffic light, or pedestrian
    - motorbike with medium vehicle, large vehicle, bus, emergency vehicle, traffic light,
      other traffic light, or pedestrian
    - medium vehicle with large vehicle, bus, emergency vehicle, traffic light, other traffic
      light, pedestrian
    - large vehicle with bus, emergency vehicle, traffic light, other traffic light,
      pedestrian
    - bus with emergency vehicle, traffic light, other traffic light, or pedestrian
    - emergency vehicle with traffic light, other traffic light, or pedestrian
    - traffic light with other traffic light
- Traffic-light colours are mutually exclusive: red, amber, and green cannot occur together.
- Red, amber, or green requires the detection to be a traffic light or another traffic light.
- Traffic-light colours are incompatible with the road-user classes and actions specified in
  the requirements. In particular, red/amber/green cannot occur together with pedestrian,
  car, cyclist, motorbike, medium vehicle, large vehicle, bus, or emergency vehicle, and
  cannot occur with the prohibited movement/action combinations listed in the requirements.
- Traffic lights and other traffic lights cannot have the prohibited actions such as moving
  away, moving towards, moving, braking, stopped, indicating left/right, hazard lights,
  turning left/right, overtaking, waiting to cross, crossing from either side, crossing, or
  pushing an object.
- Traffic lights and other traffic lights are also incompatible with pedestrian.
- The following action pairs cannot occur together:
    - moving away and moving
    - moving towards and moving
    - moving and stopped
    - indicating left and indicating right
    - braking and waiting to cross
    - braking and pushing an object
    - crossing and braking
    - waiting to cross and indicating left
    - waiting to cross and indicating right
    - waiting to cross and overtaking
    - waiting to cross and crossing from the left
    - waiting to cross and crossing from the right
    - waiting to cross and crossing
    - pedestrian and overtaking
    - pedestrian and indicating left
    - pedestrian and indicating right
    - pedestrian and hazard lights
    - pedestrian and braking
    - pedestrian and pushing an object
- The following location pairs cannot occur together:
    - outgoing lane and outgoing cycle lane
    - vehicle lane and incoming cycle lane
    - outgoing lane and incoming cycle lane
    - incoming lane and outgoing cycle lane
    - outgoing lane and pavement
    - incoming cycle lane and outgoing cycle lane
    - outgoing lane and right pavement
    - incoming cycle lane and pavement
    - crossing location and outgoing lane
    - vehicle lane and parking area
    - bus stop and outgoing lane
    - outgoing lane and parking area
    - bus stop and outgoing cycle lane
    - parking area and outgoing cycle lane
    - crossing location and incoming cycle lane
    - left pavement and right pavement
    - incoming lane and parking area
    - incoming cycle lane and bus stop
    - incoming cycle lane and parking area
    - pavement and parking area
    - left pavement and parking area
    - right pavement and parking area
    - junction and parking area
    - crossing location and bus stop
    - crossing location and parking area
    - bus stop and parking area
- All of the above constraints apply simultaneously to the same detection.

[NOT DETAILED]
We want to formalize a consistency-checking task from autonomous driving. Picture a single
frame of dash-cam footage from a self-driving car, with a box drawn around one detected road
user or traffic light. An annotator has labeled that one detection with a whole set of tags
at once, i.e., what kind of agent it is, what it's doing, where it is, and, if it's a
traffic light, what color it's showing.

Logical constraints:
- Exactly one agent class must be true.
- At least one location must be true.
- Each agent class can only have its valid actions.
- Each action can only apply to valid agent classes.
- Each agent class can only occur in valid locations.
- Traffic lights can only have a valid traffic-light colour and no movement actions.
- Red, amber, and green are mutually exclusive.
- Pushing an object is only allowed for pedestrians.
- Crossing and waiting-to-cross are only allowed for valid road users.
- Invalid agent-class combinations are not allowed.
- Invalid agent-action combinations are not allowed.
- Invalid agent-location combinations are not allowed.
- Invalid action-action combinations are not allowed.
- Invalid location-location combinations are not allowed.
- All constraints apply to the same detection.
\end{MyVerbatim}

\subsubsection{Variable lists}
\label{app:varlists}

The block below is substituted verbatim for \texttt{\{variables\}}. Each line is
``\texttt{<index>: <name>: <description>}''.

\textbf{\texttt{chx} (9).}
\begin{MyVerbatim}
  1: fracture: a bone fracture is visible
  2: pneumothorax: a pneumothorax (collapsed lung) is visible
  3: airspace_opacity: airspace opacity (e.g. consolidation or infiltrate) is visible
  4: nodule_mass: a nodule or mass is visible
  5: healthy: no findings are present
  6: green_code: green emergency code
  7: yellow_code: yellow emergency code
  8: orange_code: orange emergency code
  9: red_code: red emergency code
\end{MyVerbatim}

\textbf{\texttt{fashion} (10).}
\begin{MyVerbatim}
  1: class_0_tshirt_top: the clothing item is a t-shirt
  2: class_1_trouser: the clothing item is a pair of trousers
  3: class_2_pullover: the clothing item is a pullover
  4: class_3_dress: the clothing item is a dress
  5: class_4_coat: the clothing item is a coat
  6: class_5_sandal: the footwear is a sandal
  7: class_6_shirt: the clothing item is a shirt
  8: class_7_sneaker: the footwear is a sneaker
  9: class_8_bag: the item is a bag
 10: class_9_ankle_boot: the footwear is an ankle boot
\end{MyVerbatim}

\textbf{\texttt{bdd-oia-2} (11).}
\begin{MyVerbatim}
  1: green_light: the traffic light ahead is green
  2: follow: a lead vehicle is present and can be followed
  3: road_clear: the road ahead is clear (no obstacles)
  4: red_light: the traffic light ahead is red
  5: traffic_sign: a stop/yield sign is visible
  6: car: there is a car blocking the way
  7: person: there is a pedestrian crossing the street
  8: rider: there is a rider blocking the way
  9: other_obstacle: there is an obstacle blocking the way
 10: move_forward: proceed straight ahead
 11: stop_action: come to a stop
\end{MyVerbatim}

\textbf{\texttt{cle4evr} (12).}
\begin{MyVerbatim}
  1: obj1_shape_cube: the first object is a cube
  2: obj1_shape_sphere: the first object is a sphere
  3: obj1_shape_cylinder: the first object is a cylinder
  4: obj1_color_red: the first object is red
  5: obj1_color_blue: the first object is blue
  6: obj1_color_green: the first object is green
  7: obj2_shape_cube: the second object is a cube
  8: obj2_shape_sphere: the second object is a sphere
  9: obj2_shape_cylinder: the second object is a cylinder
 10: obj2_color_red: the second object is red
 11: obj2_color_blue: the second object is blue
 12: obj2_color_green: the second object is green
\end{MyVerbatim}

\textbf{\texttt{cifar10} (15).}
\begin{MyVerbatim}
  1: whl: the object has wheels
  2: met: the object is metallic
  3: wng: the object has wings
  4: ani: the object is an animal
  5: hai: the object is hairy
  6: hrn: the object has horns
  7: snt: the object has a snout
  8: label_airplane: the object is an airplane
  9: label_car_or_truck: the object is either a car or a truck
 10: label_bird: the object is a bird
 11: label_frog: the object is a frog
 12: label_deer: the object is a deer
 13: label_cat: the object is a cat
 14: label_dog_or_horse: the object is either a dog or a horse
 15: label_ship: the object is a ship
\end{MyVerbatim}

\textbf{\texttt{mn-mul-bin} (15).}
\begin{MyVerbatim}
  1: d1_bit0: the least significant bit of the first digit
  2: d1_bit1: the second bit of the first digit
  3: d1_bit2: the third bit of the first digit
  4: d1_bit3: the most significant bit of the first digit
  5: d2_bit0: the least significant bit of the second digit
  6: d2_bit1: the second bit of the second digit
  7: d2_bit2: the third bit of the second digit
  8: d2_bit3: the most significant bit of the second digit
  9: prod_bit0: the least significant bit of the product
 10: prod_bit1: the second bit of the product
 11: prod_bit2: the third bit of the product
 12: prod_bit3: the fourth bit of the product
 13: prod_bit4: the fifth bit of the product
 14: prod_bit5: the sixth bit of the product
 15: prod_bit6: the most significant bit of the product
\end{MyVerbatim}

\textbf{\texttt{sushi} (16).}
\begin{MyVerbatim}
  1..4:   item_0_rank_{0..3}: sushi item 0 is assigned to rank {0..3}
  5..8:   item_1_rank_{0..3}: sushi item 1 is assigned to rank {0..3}
  9..12:  item_2_rank_{0..3}: sushi item 2 is assigned to rank {0..3}
 13..16:  item_3_rank_{0..3}: sushi item 3 is assigned to rank {0..3}
(rendered as 16 explicit lines "item_i_rank_j: sushi item i is assigned to rank j")
\end{MyVerbatim}

\textbf{\texttt{mn-add-bin} (13).}
\begin{MyVerbatim}
  1: d1_bit0: the least significant bit of the first digit
  2: d1_bit1: the second bit of the first digit
  3: d1_bit2: the third bit of the first digit
  4: d1_bit3: the most significant bit of the first digit
  5: d2_bit0: the least significant bit of the second digit
  6: d2_bit1: the second bit of the second digit
  7: d2_bit2: the third bit of the second digit
  8: d2_bit3: the most significant bit of the second digit
  9: sum_bit0: the least significant bit of the sum
 10: sum_bit1: the second bit of the sum
 11: sum_bit2: the third bit of the sum
 12: sum_bit3: the fourth bit of the sum
 13: sum_bit4: the most significant bit of the sum
\end{MyVerbatim}

\textbf{\texttt{kand-logic-2} (24).}
\begin{MyVerbatim}
For im in {1,2}, primitive p in {1,2}:
  im{im}_p{p}_shape_square   : primitive p in image im is a square
  im{im}_p{p}_shape_triangle : primitive p in image im is a triangle
  im{im}_p{p}_shape_circle   : primitive p in image im is a circle
  im{im}_p{p}_color_red      : primitive p in image im is red
  im{im}_p{p}_color_blue     : primitive p in image im is blue
  im{im}_p{p}_color_yellow   : primitive p in image im is yellow
Indices 1..24 in the order (im1 p1, im1 p2, im2 p1, im2 p2), 6 bits each.
\end{MyVerbatim}

\textbf{\texttt{warcraft} (24).}
\begin{MyVerbatim}
  1..12:  edge_right_r_c (r=0..3, c=0..2):
          the edge between cells (r,c) and (r,c+1) is part of the path
 13..24:  edge_down_r_c  (r=0..2, c=0..3):
          the edge between cells (r,c) and (r+1,c) is part of the path
\end{MyVerbatim}

\textbf{\texttt{bdd-oia} (25).}
\begin{MyVerbatim}
  1: green_light: the traffic light ahead is green
  2: follow: a lead vehicle is present and can be followed
  3: road_clear: the road ahead is clear (no obstacles)
  4: red_light: the traffic light ahead is red
  5: traffic_sign: a stop/yield sign is visible
  6: car: there is a car blocking the way
  7: person: there is a pedestrian crossing the street
  8: rider: there is a rider blocking the way
  9: other_obstacle: there is an obstacle blocking the way
 10: left_lane: there is a left lane enabling a left turn
 11: left_green_light: there is a green light enabling a left turn
 12: left_follow: a lead vehicle on the left lane is present and can be followed
 13: no_left_lane: there is no left lane
 14: left_obstacle: there is an obstacle blocking the left lane
 15: left_solid_line: there is a solid line preventing a left turn
 16: right_lane: there is a right lane enabling a left turn
 17: right_green_light: there is a green light enabling a right turn
 18: right_follow: a lead vehicle on the right lane is present and can be followed
 19: no_right_lane: there is no right lane
 20: right_obstacle: there is an obstacle blocking the right lane
 21: right_solid_line: there is a solid line preventing a right turn
 22: move_forward: proceed straight ahead
 23: stop_action: come to a stop
 24: turn_left: turn left
 25: turn_right: turn right
\end{MyVerbatim}

\textbf{\texttt{cebab} (25).}
\begin{MyVerbatim}
  1: food_aspect_majority_positive: positive sentiment toward the food
  2: food_aspect_majority_negative: negative sentiment toward the food
  3: food_aspect_majority_neutral: neutral sentiment toward the food
  4: food_aspect_majority_unknown: unknown sentiment toward the food
  5: service_aspect_majority_positive: positive sentiment toward the service
  6: service_aspect_majority_negative: negative sentiment toward the service
  7: service_aspect_majority_neutral: neutral sentiment toward the service
  8: service_aspect_majority_unknown: unknown sentiment toward the service
  9: noise_aspect_majority_positive: positive sentiment toward the noise
 10: noise_aspect_majority_negative: negative sentiment toward the noise
 11: noise_aspect_majority_neutral: neutral sentiment toward the noise
 12: noise_aspect_majority_unknown: unknown sentiment toward the noise
 13: ambiance_aspect_majority_positive: positive sentiment toward the ambiance
 14: ambiance_aspect_majority_negative: negative sentiment toward the ambiance
 15: ambiance_aspect_majority_neutral: neutral sentiment toward the ambiance
 16: ambiance_aspect_majority_unknown: unknown sentiment toward the ambiance
 17: review_majority_positive: positive overall review sentiment
 18: review_majority_negative: negative overall review sentiment
 19: review_majority_neutral: neutral overall review sentiment
 20: review_majority_unknown: unknown overall review sentiment
 21: label_positive: final assessment of the restaurant is positive
 22: label_negative: final assessment of the restaurant is negative
 23: label_neutral: final assessment of the restaurant is neutral
 24: label_unknown: final assessment of the restaurant is unknown
 25: label_conflict: final assessment of the restaurant is conflicting
\end{MyVerbatim}

\textbf{\texttt{kand-logic} (36).}
\begin{MyVerbatim}
Same schema as kand-logic-2 but with 3 primitives per image:
For im in {1,2}, primitive p in {1,2,3}:
  im{im}_p{p}_shape_{square,triangle,circle}, im{im}_p{p}_color_{red,blue,yellow}
Indices 1..36 in the order (im1 p1..p3, im2 p1..p3), 6 bits each.
\end{MyVerbatim}

\textbf{\texttt{mn-add} (39).}
\begin{MyVerbatim}
  1..10:  d1_{0..9}: the first digit is {0..9}
 11..20:  d2_{0..9}: the second digit is {0..9}
 21..39:  s_{0..18}:  the sum is {0..18}
\end{MyVerbatim}

\textbf{\texttt{road-r} (41).}
\begin{MyVerbatim}
Agent (one-hot):
  1: ped     the detection is a pedestrian
  2: car     the detection is a car
  3: cyc     the detection is a cyclist
  4: mobike  the detection is a motorbike
  5: med_veh the detection is a medium vehicle
  6: lar_veh the detection is a large vehicle
  7: bus     the detection is a bus
  8: em_veh  the detection is an emergency vehicle
  9: tl      the detection is a traffic light
 10: oth_tl  the detection is another traffic light
Traffic-light colour (at most one):
 11: red     the traffic light is red
 12: amber   the traffic light is amber
 13: green   the traffic light is green
Action:
 14: mov_away     the detected agent is moving away
 15: mov_tow      the detected agent is moving towards the ego vehicle
 16: mov          the detected agent is moving
 17: brake        the detected agent is braking
 18: stop         the detected agent is stopped
 19: incat_left   the agent is indicating left
 20: incat_rht    the agent is indicating right
 21: haz_lit      the agent has its hazard lights on
 22: tur_lft      the agent is turning left
 23: tur_rht      the agent is turning right
 24: ovtak        the agent is overtaking
 25: wait_2x      the agent is waiting to cross
 26: xing_fm_lft  the agent is crossing from the left
 27: xing_fm_rht  the agent is crossing from the right
 28: xing         the agent is crossing
 29: push_obj     the agent is pushing an object
Location:
 30: veh_lane        the detection is in a vehicle lane
 31: outgo_lane      the detection is in an outgoing lane
 32: outgo_cyc_lane  the detection is in an outgoing cycle lane
 33: incom_lane      the detection is in an incoming lane
 34: incom_cyc_lane  the detection is in an incoming cycle lane
 35: pav             the detection is on the pavement
 36: lft_pav         the detection is on the left pavement
 37: rht_pav         the detection is on the right pavement
 38: jun             the detection is at a junction
 39: xing_loc        the detection is at a crossing location
 40: bus_stop        the detection is at a bus stop
 41: parking         the detection is in a parking area
\end{MyVerbatim}

\textbf{\texttt{sudoku} (64).}
\begin{MyVerbatim}
  index i*16 + j*4 + v  ->  cell_i_j_val_v (i,j in 0..3, v in 1..4):
  "cell at row i, column j contains digit v"
\end{MyVerbatim}

\textbf{\texttt{cifar100} (100).}
\begin{MyVerbatim}
  1..100: class_{0..99}: the image belongs to class {0..99}
\end{MyVerbatim}

\textbf{\texttt{mn-mul} (102).}
\begin{MyVerbatim}
  1..10:  d1_{0..9}: the first digit is {0..9}
 11..20:  d2_{0..9}: the second digit is {0..9}
 21..102: p_{0..81}:  the product is {0..81}
\end{MyVerbatim}

\section{Large Language Models}\label{app:llms}
Backends span eight open-weight model families Qwen(-Coder)~\citep{yang2025qwen3}, 
Gemma~\citep{gemma_2025, team2026gemma}, Mistral-Nemo~\citep{jiang2023mistral7b}, DeepSeek-R1-Distill~\citep{guo2025deepseek, liu2024deepseek}, Olmo~\citep{groeneveld2024olmo}, Phi~\citep{abdin2025phi}, GPT-OSS~\citep{openai2025gptoss120bgptoss20bmodel}) at parameter counts from 8.2B to 117B. \Cref{tab:llm-eval-clusters-vllm} lists every checkpoint. We load each checkpoint \textit{unquantized} in bfloat16. \texttt{gpt-oss-20b} and \texttt{gpt-oss-120b} are the only exceptions to this policy: both ship with their own native low-bit format (MXFP4 under \texttt{vLLM}) and are loaded as-is. We choose instruction-tuned models when possible.

\section{Prompts}\label{app:prompts}

Each (dataset, output format, prompting mode) combination has its own template (\texttt{fonesy/prompts/<dataset>\_<format>[\_cot].txt}); modes are ZS and ZS-CoT. We show the representative ZS \DIMACS template for each of the three datasets below.

\textbf{\autobench}.

\begin{MyVerbatim}
You are an expert in SAT solving and constraint modeling.

Your task is to convert the following logical problem into a CNF formula in DIMACS format.

The Boolean variables are numbered as follows:
{variables}

DIMACS format rules:
- First line: p cnf <num_vars> <num_clauses>
- Each subsequent line is one clause: space-separated integers ending with 0
- Positive integer i = variable i is True; negative integer -i = variable i is False
- Variable numbers MUST match exactly the mapping above

Problem:
{problem}

Strict output rules:
- Start your answer with <cnf> and end with </cnf>
- Every clause line MUST end with 0
- NO markdown, NO code fences, NO backticks, NO explanation outside the tags
\end{MyVerbatim}

\textbf{\satbench}. 

\begin{MyVerbatim}
{scenario}
{variable_mapping}

Use only the mapped variables; do not introduce auxiliary variables.

Logical constraints (encode EXACTLY these as CNF clauses, and nothing else):
{conditions}
\end{MyVerbatim}

\textbf{\dcpbench}.

\begin{MyVerbatim}
You are an expert in SAT solving and constraint modeling.

Your task is to convert the following logical problem into a CNF formula in DIMACS format.

This problem's declared variables are:
{variables}

DIMACS format rules:
- First line: p cnf <num_vars> <num_clauses>
- Each subsequent line is one clause: space-separated integers ending with 0
- Positive integer i = variable i is True; negative integer -i = variable i is False
- Each declared variable's numbered range above IS its one-hot expansion, in
  ascending domain-value order (e.g. "1-101: men: integer, domain [0, 100]"
  means variable 1 is true iff men=0, variable 2 iff men=1, ..., variable 101
  iff men=100); exactly one variable in each range must be true
- If you need additional Boolean variables for auxiliary sub-expressions,
  number them consecutively starting right after the highest number shown above

Problem:
{problem}

Strict output rules:
- Start your answer with <cnf> and end with </cnf>
- Every clause line MUST end with 0
- NO markdown, NO code fences, NO backticks, NO explanation outside the tags
\end{MyVerbatim}

The self-correction prompt (shared by all three datasets) supplies the original problem, the proposed CNF, and the detected error message, and requires the model to answer \texttt{[[OK]]} or \texttt{[[FIXED]]} followed by a corrected \texttt{<cnf>...</cnf>} block.

\textbf{Other output formats.}
The templates described above correspond to the \DIMACS variants. The remaining four formats, \ie \pysat, \cpmpy, \sympy, and \nat, retain the same role sentence, the \texttt{{variables}} and \texttt{{problem}} slots, and the \texttt{<cnf>...</cnf>} wrapper. They differ only in the format-specific instructions and in how the variable list is to be interpreted.

\begin{itemize}
\item \pysat: the model generates a Python program rather than raw clauses. The program imports \texttt{CNF} from \texttt{pysat.formula} and \texttt{sys}, appends clauses as lists of non-zero integers, and writes the resulting formula using \texttt{cnf.to\_fp(sys.stdout)}. Variable indices must match the provided mapping. When using \texttt{CardEnc} or \texttt{PBEnc}, the model must specify \texttt{top\_id=N} or \texttt{vpool=IDPool(start\_from=N+1)} to ensure that auxiliary variable IDs do not overlap with the mapped variables.

\item \cpmpy: the model generates a Python program that declares one \texttt{cp.boolvar(name='<name>')} for each listed variable. Variables are therefore identified by name rather than by integer index. The model combines these Boolean variables using \texttt{\&}, \texttt{|}, and \texttt{\textasciitilde}, and passes the resulting constraint list to the pre-injected \texttt{\_print\_dimacs([...])} function, which must not be redefined. For \dcpbench, variable names follow the convention \texttt{<name>\_eq<value>}, with one Boolean variable for each value in the domain.

\item \sympy: the model generates a Python program using \texttt{Symbol('<name>')} to represent variables and \texttt{to\_cnf(expr, simplify=False)} to convert the resulting expression to CNF. The resulting formula is passed to the pre-injected \texttt{\_print\_dimacs(cnf\_expr)} function. Variable naming follows the same convention as in \cpmpy.

\item \textbf{Natural-language operators} (\texttt{nat}): the model produces no code and does not use variable numbering. Variables are referred to by name, and the output must have the form \texttt{(lit OR lit ...) AND (...)}. The operators \texttt{AND}, \texttt{OR}, and \texttt{NOT} are accepted in any letter case. XOR, NAND, NOR, XNOR, and the symbols \texttt{\&}, \texttt{|}, and \texttt{\textasciitilde} are not permitted. Each clause must be enclosed in parentheses, and clauses must be joined by a top-level \texttt{AND}.
\end{itemize}

The output constraints are adapted accordingly. The requirement that every clause end with \texttt{0} applies only to the DIMACS formats, while the code-based formats require the generated program to print a valid DIMACS formula to standard output. In the ZS-CoT variants, the strict output constraints are replaced by an instruction to perform the reasoning inside \texttt{<think></think>} before emitting the final \texttt{<cnf>} block. Self-correction uses a dedicated template for each format while retaining the same \texttt{[[OK]]}/\texttt{[[FIXED]]} protocol.

\section{LLMs Hyperparameters and Model Selection}
\label{app:llm-hyp}

Each evaluation has a budget of $5$ self-verification steps.

We do \emph{not} tune any decoding hyperparameter. For every model, the sampling configuration, whether to sample at all, the temperature $\tau$, the nucleus threshold $\rho$ (\texttt{top\_p}), $\kappa$ (\texttt{top\_k}), and any repetition penalty, is taken straight from the model HuggingFace model card / \texttt{generation\_config.json}, \ie the provider's own recommended inference settings. Almost all values in \cref{tab:llm-decoding} come directly from that source; the few exceptions (\eg \texttt{repetition\_penalty}$=1.3$ for the Gemma checkpoints, eager attention for the Qwen family) are documented mitigations for known degenerate-decoding or numerical issues.

The only decoding value we set ourselves is the generation-length cap. Because a target CNF can span thousands of clauses, we use $32{,}768$ new tokens across all settings; this budget is identical for every model and every cluster. Reasoning-tuned checkpoints (\texttt{qwen3-8b}, \texttt{qwen3-32b}, the \texttt{deepseek-r1-llama-70b}, \texttt{phi4-reasoning-plus}, \texttt{olmo3-32b-think}, and the \texttt{gpt-oss} family) keep their internal ``thinking'' traces enabled; all other models follow their card greedy-vs.-sampling recommendation. 

\begin{table}[h]
    \centering
    \caption{\textbf{Decoding hyperparameters for the evaluated LLMs}. ``sample'' is \texttt{do\_sample}; a dash means the parameter is unset (provider default);     ``think'' marks models whose reasoning trace is always active. All values are the model-card / \texttt{generation\_config.json} defaults unless noted. 
    }
    \label{tab:llm-decoding}
    \scalebox{0.82}{
    \begin{tabular}{lccccccc}
        \toprule
        \textsc{Model} & \textsc{sample} & $\tau$ & $\rho$ & $\kappa$ & \textsc{rep.\ pen.} & \textsc{think} & \textsc{precision} \\
        \midrule
        \texttt{qwen3-32b}                      & \checkmark & 0.6 & 0.95 & 20 & --   & \checkmark & bf16 \\
        \rowcolor[HTML]{EFEFEF}
        \texttt{qwen3-8b}                       & \checkmark & 0.6 & 0.95 & 20 & --   & \checkmark & bf16 \\
        \texttt{qwen3-coder-next}               & \checkmark & 1.0 & 0.95 & 40 & --   & --         & bf16 \\
        \rowcolor[HTML]{EFEFEF}
        \texttt{mistral-nemo}     & --         & --  & --   & -- & --   & --         & bf16 \\
        \texttt{phi4-reasoning-plus}           & \checkmark & 0.8 & 0.95 & 50 & --   & \checkmark & bf16 \\
        \rowcolor[HTML]{EFEFEF}
        \texttt{gemma3-27b-it}                 & \checkmark & 0.7 & 0.95 & 64 & 1.3  & --         & bf16 \\
        \texttt{gemma4-31B-it}                 & \checkmark & 0.7 & 0.95 & 64 & 1.3  & --         & bf16 \\
        \rowcolor[HTML]{EFEFEF}
        \texttt{olmo3-32b-think}        
        & \checkmark & 0.6 & 0.95 & -- & --   & \checkmark & bf16 \\
        \texttt{deepseek-r1-llama-70b}  & \checkmark & 0.6 & 0.95 & -- & --   & \checkmark & bf16 \\
        \rowcolor[HTML]{EFEFEF}
        \texttt{gpt-oss-20b}                    & --         & --  & --   & -- & --   & (native)   & MXFP4 \\
        \texttt{gpt-oss-120b}                   & --         & --  & --   & -- & --   & (native)   & MXFP4 \\
        \bottomrule
    \end{tabular}
    }
\end{table}

\section{\SPL and \SL Model Selection}
\label{app:spl-model-selection}

All \SPL and \SL experiments use the \texttt{Adam}~\citep{kingma2014adam} optimizer. Hyperparameters were selected through Bayesian optimization with \texttt{Optuna}~\citep{akiba2019optuna}, using a \texttt{TPE} sampler~\citep{watanabe2023tree} (all with seed $32$) together with a median pruner that early-stops clearly bad trials from their intermediate per-epoch validation F1. Each study maximizes the best validation macro-F1 that a trial reaches over its own epochs. Studies ran $30$ trials of $50$ epochs each for \texttt{mn-add-bin}, \texttt{mn-add}, \texttt{mn-mul-bin}, \texttt{mn-mul}, \texttt{sudoku}, \texttt{bdd-oia}, \texttt{bdd-oia-2}, \texttt{sushi}, \texttt{cle4evr} and \texttt{kand-logic}; $20$ trials of $30$ epochs each for \texttt{fashion}, \texttt{warcraft}, \texttt{cifar10}, \texttt{cifar100}, \texttt{chx} and \texttt{cebab}; and $12$ trials of $6$ epochs each for \texttt{road-r} due to time constraints.

Only three quantities are tuned: the learning rate $\gamma$, searched log-uniformly over $[10^{-5}, 10^{-1}]$; the \SPL head width $h$, over integers in $[32, 512]$; and the batch size $\nu$, searched over a categorical grid that depends on the backbone and on the formula source. 
For the ground-truth formulas, $\nu$ spans over $\{16, 32, 64, 128, 256\}$ by default, restricted to $\{16, 32, 64\}$ for the \texttt{ResNet-18}~\citep{he2016deep} and \texttt{BERT}~\citep{devlin2019bert} backbones for memory reasons. For the LLM-predicted formulas the $\nu$ grid is $\{16, 32, 64, 128, 256\}$ for all \texttt{Conv4}, \texttt{MLP}, \texttt{ResNet-18} and \texttt{BERT} tasks, with three exceptions: $\{8, 16, 32, 64\}$ for \texttt{chx} and \texttt{cebab} under \texttt{gpt-oss-120b}, and $\{16, 32, 64, 128\}$ for \texttt{cifar10}, \texttt{cifar100} and \texttt{chx} under \texttt{qwen3-8b}. For the \texttt{Faster-RCNN}~\citep{ren2016faster} backbones (\texttt{cle4evr}, \texttt{kand-logic}, \texttt{kand-logic-2}) the grid is $\{2, 4, 8\}$ in all settings. For \texttt{road-r} with the \texttt{SlowFast}~\citep{feichtenhofer2019slowfast} backbone the search space is narrower: $\gamma \in [10^{-5}, 3\!\times\!10^{-3}]$, $h \in [64, 512]$ and, for the ground-truth formula, $\nu \in \{32, 64, 96\}$; the \texttt{gpt-oss-120b} search used $\nu \in \{16, 32, 64\}$.

Final numbers are obtained by retraining the selected configuration for a \emph{fixed} $50$ epochs\footnote{$10$ for \texttt{road-r} due to time constraints.} with no early stopping; the reported test metrics use the \emph{final-epoch} model. Every task is trained with the same $5$ seeds for consistency, specifically $\{1011, 2021, 3031, 4041, 5051\}$; we report mean $\pm$ std.

The circuit has no trainable parameters of its own: it is compiled offline from each task CNF into a smooth structured-probabilistic layer whose sum layers are conditionally parameterized by a softmax gate function. A small \texttt{MLP} head $\texttt{Linear}(d_f, h) \to \texttt{ReLU} \to \texttt{Linear}(h, G)$ maps the backbone feature vector (dimension $d_f$) to the $G$ gate-function weights with which is parametrized.

Even when two formulas are logically equivalent ($\frel$ is $=$), \textit{they do not} necessarily yield the same circuit. The generated formula may use a different syntactic encoding of the same constraint; in particular, \cpmpy and \sympy CNF lowering introduces auxiliary Tseitin variables. We existentially quantify these variables out during compilation, so the resulting circuit is defined over the same primary variables as the ground-truth circuit. However, the compiled structure, such as the vtree and SDD size, and consequently the number of sum-layer parameters produced by the network, depends on the specific CNF and \textit{may therefore differ} from that of the ground-truth circuit.\footnote{This affects only the \SPL circuit and not the neural backbone.} As a result, optimal hyperparameters \textit{do not} transfer across formulas, even when the formulas are logically equivalent. We therefore perform a separate hyperparameter search for each formula.

Below we list the configurations that performed best on each \autobench dataset ($\gamma$ / $\nu$ / $h$; optimizer \texttt{Adam}; backbone $b$) for ground-truth \SPL formulas.

\begin{itemize}
    \item \texttt{mn-add-bin} / \texttt{mn-add}: $\gamma = 6.4\!\times\!10^{-3}$ /
          $2.6\!\times\!10^{-3}$, $\nu = 32$ / $16$, $h = 498$ / $188$, $b = $ \texttt{Conv4}.
    \item \texttt{mn-mul-bin} / \texttt{mn-mul}: $\gamma = 1.6\!\times\!10^{-3}$ / $1.6\!\times\!10^{-3}$,
          $\nu = 256$ / $256$, $h = 370$ / $370$, $b = $ \texttt{Conv4}.
    \item \texttt{sudoku}: $\gamma = 2.9\!\times\!10^{-4}$, $\nu = 16$, $h = 455$, $b = $ \texttt{Conv4}.
    \item \texttt{fashion}: $\gamma = 4.7\!\times\!10^{-3}$, $\nu = 128$, $h = 52$, $b = $ \texttt{Conv4}.
    \item \texttt{warcraft}: $\gamma = 1.2\!\times\!10^{-2}$, $\nu = 64$, $h = 303$, $b = $ \texttt{Conv4}.
    \item \texttt{cifar10}: $\gamma = 1.0\!\times\!10^{-3}$, $\nu = 32$, $h = 223$, $b = $ \texttt{ResNet-18}.
    \item \texttt{cifar100}: $\gamma = 2.8\!\times\!10^{-4}$, $\nu = 32$, $h = 444$, $b = $ \texttt{ResNet-18}.
    \item \texttt{chx}: $\gamma = 9.5\!\times\!10^{-5}$, $\nu = 16$, $h = 462$, $b = $ \texttt{ResNet-18}.
    \item \texttt{bdd-oia} / \texttt{bdd-oia-2}: $\gamma = 1.4\!\times\!10^{-4}$ /
          $4.4\!\times\!10^{-5}$, $\nu = 16$ / $64$, $h = 262$ / $441$, $b = $ \texttt{MLP} (over 2048-d \texttt{Faster-RCNN} features).
    \item \texttt{sushi}: $\gamma = 1.6\!\times\!10^{-4}$, $\nu = 16$, $h = 254$, $b = $ \texttt{MLP} (over 36-d one-hot context).
    \item \texttt{cle4evr}: $\gamma = 8.4\!\times\!10^{-5}$, $\nu = 2$, $h = 512$, $b = $ \texttt{Faster-RCNN}.
    \item \texttt{kand-logic} / \texttt{kand-logic-2}: $\gamma = 2.5\!\times\!10^{-5}$, $\nu = 4$, $h = 327$, $b = $ \texttt{Faster-RCNN}.
    \item \texttt{cebab}: $\gamma = 1.0\!\times\!10^{-3}$, $\nu = 32$, $h = 223$, $b = $ \texttt{BERT}.
    \item \texttt{road-r}: $\gamma = 1.74\!\times\!10^{-4}$, $\nu = 64$, $h = 242$, $b = $ \texttt{SlowFast-50}.
\end{itemize}

Below, instad, we list the configurations that performed best on each \autobench dataset ($\gamma$ / $\nu$ / $h$; optimizer \texttt{Adam}; backbone $b$) for the LLM-predicted \SPL formulas. We omit listing datasets for which the LLM did not produce a valid formula.

\textbf{\texttt{gpt-oss-120b}.}
\begin{itemize}
    \item \texttt{mn-add-bin}: $\gamma = 2.6\!\times\!10^{-3}$, $\nu = 16$, $h = 188$, $b = $ \texttt{Conv4}.
    \item \texttt{mn-add}: $\gamma = 1.1\!\times\!10^{-2}$, $\nu = 64$, $h = 272$, $b = $ \texttt{Conv4}.
    \item \texttt{mn-mul-bin}: $\gamma = 2.3\!\times\!10^{-3}$, $\nu = 16$, $h = 190$, $b = $ \texttt{Conv4}.
    \item \texttt{mn-mul}: $\gamma = 7.8\!\times\!10^{-4}$, $\nu = 16$, $h = 162$, $b = $ \texttt{Conv4}.
    \item \texttt{sudoku}: $\gamma = 4.0\!\times\!10^{-4}$, $\nu = 16$, $h = 227$, $b = $ \texttt{Conv4}.
    \item \texttt{fashion}: $\gamma = 1.7\!\times\!10^{-3}$, $\nu = 128$, $h = 116$, $b = $ \texttt{Conv4}.
    \item \texttt{warcraft}: $\gamma = 3.4\!\times\!10^{-3}$, $\nu = 16$, $h = 185$, $b = $ \texttt{Conv4}.
    \item \texttt{cifar10}: $\gamma = 1.6\!\times\!10^{-3}$, $\nu = 256$, $h = 370$, $b = $ \texttt{ResNet-18}.
    \item \texttt{cifar100}: $\gamma = 2.8\!\times\!10^{-4}$, $\nu = 64$, $h = 190$, $b = $ \texttt{ResNet-18}.
    \item \texttt{chx}: $\gamma = 2.5\!\times\!10^{-5}$, $\nu = 8$, $h = 478$, $b = $ \texttt{ResNet-18}.
    \item \texttt{bdd-oia}: $\gamma = 9.0\!\times\!10^{-5}$, $\nu = 16$, $h = 356$, $b = $ \texttt{MLP}.
    \item \texttt{bdd-oia-2}: $\gamma = 4.5\!\times\!10^{-5}$, $\nu = 64$, $h = 34$, $b = $ \texttt{MLP}.
    \item \texttt{sushi}: $\gamma = 3.2\!\times\!10^{-3}$, $\nu = 256$, $h = 505$, $b = $ \texttt{MLP}.
    \item \texttt{cle4evr}: $\gamma = 9.0\!\times\!10^{-5}$, $\nu = 2$, $h = 356$, $b = $ \texttt{Faster-RCNN}.
    \item \texttt{kand-logic}: $\gamma = 1.4\!\times\!10^{-5}$, $\nu = 4$, $h = 289$, $b = $ \texttt{Faster-RCNN}.
    \item \texttt{kand-logic-2}: $\gamma = 3.1\!\times\!10^{-5}$, $\nu = 4$, $h = 375$, $b = $ \texttt{Faster-RCNN}.
    \item \texttt{cebab}: $\gamma = 1.8\!\times\!10^{-3}$, $\nu = 8$, $h = 141$, $b = $ \texttt{BERT}.
    \item \texttt{road-r}: $\gamma = 9.5\!\times\!10^{-5}$, $\nu = 16$, $h = 462$, $b = $ \texttt{Slowfast-50}.
\end{itemize}

\textbf{\texttt{olmo3-32b-think}.}
\begin{itemize}
    \item \texttt{mn-add}: $\gamma = 3.0\!\times\!10^{-3}$, $\nu = 16$, $h = 59$, $b = $ \texttt{Conv4}.
    \item \texttt{mn-mul-bin}: $\gamma = 1.4\!\times\!10^{-3}$, $\nu = 16$, $h = 153$, $b = $ \texttt{Conv4}.
    \item \texttt{mn-mul}: $\gamma = 8.5\!\times\!10^{-4}$, $\nu = 16$, $h = 321$, $b = $ \texttt{Conv4}.
    \item \texttt{fashion}: $\gamma = 2.1\!\times\!10^{-3}$, $\nu = 64$, $h = 474$, $b = $ \texttt{Conv4}.
    \item \texttt{warcraft}: $\gamma = 1.2\!\times\!10^{-2}$, $\nu = 16$, $h = 185$, $b = $ \texttt{Conv4}.
    \item \texttt{cifar10}: $\gamma = 1.6\!\times\!10^{-3}$, $\nu = 256$, $h = 370$, $b = $ \texttt{ResNet-18}.
    \item \texttt{cifar100}: $\gamma = 9.5\!\times\!10^{-4}$, $\nu = 256$, $h = 60$, $b = $ \texttt{ResNet-18}.
    \item \texttt{bdd-oia-2}: $\gamma = 2.1\!\times\!10^{-5}$, $\nu = 16$, $h = 216$, $b = $ \texttt{MLP}.
    \item \texttt{sushi}: $\gamma = 1.1\!\times\!10^{-3}$, $\nu = 64$, $h = 164$, $b = $ \texttt{MLP}.
    \item \texttt{cle4evr}: $\gamma = 9.0\!\times\!10^{-5}$, $\nu = 2$, $h = 356$, $b = $ \texttt{Faster-RCNN}.
    \item \texttt{kand-logic-2}: $\gamma = 1.7\!\times\!10^{-5}$, $\nu = 2$, $h = 506$, $b = $ \texttt{Faster-RCNN}.
    \item \texttt{cebab}: $\gamma = 6.1\!\times\!10^{-4}$, $\nu = 16$, $h = 90$, $b = $ \texttt{BERT}.
\end{itemize}

\textbf{\texttt{qwen3-8b}.}
\begin{itemize}
    \item \texttt{mn-add-bin}: $\gamma = 2.8\!\times\!10^{-3}$, $\nu = 16$, $h = 146$, $b = $ \texttt{Conv4}.
    \item \texttt{mn-add}: $\gamma = 2.6\!\times\!10^{-3}$, $\nu = 16$, $h = 254$, $b = $ \texttt{Conv4}.
    \item \texttt{mn-mul}: $\gamma = 2.6\!\times\!10^{-3}$, $\nu = 16$, $h = 188$, $b = $ \texttt{Conv4}.
    \item \texttt{sudoku}: $\gamma = 1.9\!\times\!10^{-3}$, $\nu = 16$, $h = 274$, $b = $ \texttt{Conv4}.
    \item \texttt{fashion}: $\gamma = 1.6\!\times\!10^{-3}$, $\nu = 16$, $h = 190$, $b = $ \texttt{Conv4}.
    \item \texttt{cifar10}: $\gamma = 4.4\!\times\!10^{-4}$, $\nu = 64$, $h = 135$, $b = $ \texttt{ResNet-18}.
    \item \texttt{cifar100}: $\gamma = 4.8\!\times\!10^{-4}$, $\nu = 64$, $h = 191$, $b = $ \texttt{ResNet-18}.
    \item \texttt{bdd-oia}: $\gamma = 2.2\!\times\!10^{-4}$, $\nu = 16$, $h = 300$, $b = $ \texttt{MLP}.
    \item \texttt{sushi}: $\gamma = 4.7\!\times\!10^{-4}$, $\nu = 32$, $h = 512$, $b = $ \texttt{MLP}.
    \item \texttt{cle4evr}: $\gamma = 9.0\!\times\!10^{-5}$, $\nu = 2$, $h = 356$, $b = $ \texttt{Faster-RCNN}.
    \item \texttt{kand-logic-2}: $\gamma = 1.8\!\times\!10^{-5}$, $\nu = 2$, $h = 499$, $b = $ \texttt{Faster-RCNN}.
    \item \texttt{chx}: $\gamma = 5.4\!\times\!10^{-5}$, $\nu = 16$, $h = 414$, $b = $ \texttt{ResNet-18}.
\end{itemize}

Due to time constraints, we reused the same hyperparameters for \SL (\cref{sec:semantic-loss}).

\subsection{\SPL Model Architectures}
\label{sec:spl-architecture-details}

Six architectures $f$ are used as backbones across the \SPL and \SL experiments. \texttt{Conv4}~\citep{meng2023few} and \texttt{ResNet-18}~\citep{he2016deep} are trained \emph{from scratch} (random initialization); \texttt{Faster-RCNN}~\citep{ren2016faster} is \texttt{COCO}-pretrained~\citep{lin2014microsoft} and fine-tuned jointly with the head; \texttt{BERT}~\citep{devlin2019bert} is frozen with a trainable head; \texttt{Slowfast-50}~\citep{feichtenhofer2019slowfast} is trained from scratch. We assign: \texttt{Conv4} for the \MNIST-based tasks, \texttt{sudoku}, \texttt{fashion} and \texttt{warcraft}; \texttt{ResNet-18} for \texttt{cifar10}, \texttt{cifar100} and \texttt{chx}; a 2-3 layer \texttt{MLP} over precomputed feature vectors for \texttt{bdd-oia}, \texttt{bdd-oia-2} and \texttt{sushi}; \texttt{Faster-RCNN} for \texttt{cle4evr}, \texttt{kand-logic} and \texttt{kand-logic-2}; \texttt{bert-base-uncased} for \CEBAB; and \texttt{Slowfast-50} for \texttt{road-r}. The \texttt{ResNet-18} and \texttt{BERT} stacks are the standard \texttt{torchvision}/\texttt{transformers} definitions (see \cref{tab:resnet-arch,tab:bert-arch}). The remaining three are detailed below.

\textbf{\texttt{Conv4}}. A four-block CNN used for the \MNIST-based tasks, \ie \texttt{sudoku}, \texttt{fashion} and \texttt{warcraft}. \cref{tab:conv4-arch} details the layer configuration.

\begin{table}[H]
    \centering
    \caption{\textbf{\texttt{Conv4} architecture} (\MNIST-based tasks, \texttt{sudoku}, \texttt{fashion} and \texttt{warcraft}).
    $C_{\text{in}}$ is $1$ for grayscale and $3$ for RGB inputs.}
    \label{tab:conv4-arch}
    \scalebox{0.9}{
    \begin{tabular}{llll}
        \toprule
        \textsc{Input} & \textsc{Layer Type} & \textsc{Parameter} & \textsc{Activation} \\
        \midrule
        $(C_{\text{in}}, H, W)$ & Conv2d + BatchNorm2d & depth$=64$, kernel$=3$, stride$=1$, padding$=1$ & ReLU \\
        $(64, H, W)$            & MaxPool2d            & kernel$=2$, stride$=2$                          & \\
        \multicolumn{4}{c}{$\times\,3$ more identical $(\text{Conv2d}+\text{BN}+\text{ReLU}+\text{MaxPool2d})$ blocks, depth $64\!\to\!64$} \\
        $(64, H', W')$          & AdaptiveAvgPool2d    & output$=(1,1)$                                  & \\
        $(64, 1, 1)$            & Flatten              & dim$=64$                                        & \\
        $(64)$                  & Linear              & $64 \to \mathtt{output\_size}$                 & \\
        \bottomrule
    \end{tabular}
    }
\end{table}

\textbf{\texttt{ResNet-18}}. The standard~\citet{he2016deep} \texttt{ResNet-18}, randomly initialized (\texttt{weights=None}) and trained jointly with the \SPL head from scratch, used for \texttt{cifar10}, \texttt{cifar100} and \texttt{chx}. The stem convolution is swapped only when the input is not 3-channel; every other layer is the unmodified \texttt{torchvision} definition. \cref{tab:resnet-arch} details the layer configuration.

\begin{table}[H]
    \centering
    \caption{\textbf{\texttt{ResNet-18} architecture} (\texttt{cifar10}, \texttt{cifar100}, \texttt{chx}). $C_{\text{in}}$ is $3$ for all three tasks; $(H,W)$ is $32\!\times\!32$ for \texttt{cifar10}, \texttt{cifar100} and $224\!\times\!224$ for \texttt{chx}.}
    \label{tab:resnet-arch}
    \scalebox{0.85}{
    \begin{tabular}{llll}
        \toprule
        \textsc{Input} & \textsc{Layer Type} & \textsc{Parameter} & \textsc{Activation} \\
        \midrule
        $(C_{\text{in}}, H, W)$        & Conv2d + BatchNorm2d & depth$=64$, kernel$=7$, stride$=2$, padding$=3$ & ReLU \\
        $(64, H/2, W/2)$                & MaxPool2d            & kernel$=3$, stride$=2$, padding$=1$             & \\
        $(64, H/4, W/4)$                & BasicBlock $\times\,2$ (\texttt{layer1}) & depth $64\!\to\!64$, stride$=1$      & ReLU \\
        $(64, H/4, W/4)$                & BasicBlock $\times\,2$ (\texttt{layer2}) & depth $64\!\to\!128$, stride$=2$ (block 1) & ReLU \\
        $(128, H/8, W/8)$               & BasicBlock $\times\,2$ (\texttt{layer3}) & depth $128\!\to\!256$, stride$=2$ (block 1) & ReLU \\
        $(256, H/16, W/16)$             & BasicBlock $\times\,2$ (\texttt{layer4}) & depth $256\!\to\!512$, stride$=2$ (block 1) & ReLU \\
        $(512, H/32, W/32)$             & AdaptiveAvgPool2d    & output$=(1,1)$                                  & \\
        $(512, 1, 1)$                    & Flatten              & dim$=512$                                       & \\
        $(512)$                          & Linear               & $512 \to \mathtt{output\_size}$                & \\
        \bottomrule
    \end{tabular}
    }
\end{table}

\textbf{\texttt{MLP}}.
For tasks whose inputs are pre-extracted feature vectors rather than images, \ie \texttt{bdd-oia} (2048-d \texttt{Faster-RCNN} embeddings) and Sushi (36-d one-hot context ranking), a small $2$-$3$ layer \texttt{MLP} is trained from scratch. \cref{tab:mlp-arch} details the configuration.

\begin{table}[H]
    \centering
    \caption{\textbf{\texttt{MLP} backbone} (\texttt{bdd-oia}, \texttt{sushi}). $d_{\text{in}}$ is $2048$ for BDD-OIA and
    $36$ for \texttt{sushi}.}
    \label{tab:mlp-arch}
    \scalebox{0.9}{
    \begin{tabular}{llll}
        \toprule
        \textsc{Input} & \textsc{Layer Type} & \textsc{Parameter} & \textsc{Activation} \\
        \midrule
        $(d_{\text{in}})$ & Linear & $d_{\text{in}} \to h$ & ReLU \\
        $(h)$             & Linear & $h \to h$ \quad ($\times\,(L-1)$) & ReLU \\
        $(h)$             & Linear & $h \to \mathtt{output\_size}$ & \\
        \bottomrule
    \end{tabular}
    }
\end{table}

\textbf{\texttt{BERT}}. For \texttt{cebab}, review text is tokenized (max length $n_{\text{tok}}=128$) and encoded with a frozen \texttt{bert-base-uncased}~\citep{devlin2019bert}; only a small \texttt{Linear}$\to$\texttt{ReLU} projection stack on top of the pooled \texttt{[CLS]} token is trained, sharing its $h$/$L$ configuration with the \texttt{MLP} backbone (\cref{tab:mlp-arch}). \cref{tab:bert-arch} details the configuration.

\begin{table}[H]
    \centering
    \caption{\textbf{\texttt{BERT} backbone} (\texttt{cebab}). $n_{\text{tok}}=128$ is the fixed max token length; $L \in \{2,3\}$ and $h$ configure the trainable projection (same search parameters as \cref{tab:mlp-arch}). \texttt{BERT} itself is frozen; only the projection and \SPL head receive gradients.}
    \label{tab:bert-arch}
    \small
    \begin{tabular}{p{0.16\linewidth} p{0.27\linewidth} p{0.32\linewidth} p{0.15\linewidth}}
        \toprule
        \textsc{Input} &
        \textsc{Layer Type} &
        \textsc{Parameter} &
        \textsc{Activation} \\
        \midrule

        $(n_{\text{tok}}, 2)$ &
        tokenized input (\texttt{input\_ids}, \texttt{attention\_mask}) &
        max length $n_{\text{tok}}=128$ &
        \\

        $(n_{\text{tok}}, 2)$ &
        BertModel (\texttt{bert-base-uncased}, frozen) &
        $12$ Transformer encoder layers,\\
        & & hidden$=768$, heads$=12$,\\
        & & FFN$=3072$ &
        GELU (internal) \\

        $(768)$ &
        \texttt{[CLS]} pooled token &
        — &
        \\

        $(768)$ &
        Linear &
        $768 \to h$ &
        ReLU \\

        $(h)$ &
        Linear \quad ($\times\,(L-1)$) &
        $h \to h$ &
        ReLU \\

        $(h)$ &
        Linear &
        $h \to \mathtt{output\_size}$ &
        \\

        \bottomrule
    \end{tabular}
\end{table}

\textbf{\texttt{Slowfast-50}}.
For \texttt{road-r}, whose samples are short video clips, we adopt the \texttt{Slowfast-50} backbone (\citealp{feichtenhofer2019slowfast}; two-pathway 3D ResNet, bottleneck block counts $[3, 4, 6, 3]$), vendored from the official \texttt{ROAD-R} codebase and trained from scratch. Each sample is an $8$-frame box-crop clip of shape $(3, 8, 96, 96)$. The \emph{fast} pathway processes all $8$ frames at a reduced channel width ($\beta = 0.125$); the \emph{slow} pathway processes every second frame ($\alpha = 2$) at full width; lateral connections fuse fast features into the slow pathway after each stage. We global-average-pool the final (\texttt{c5}) feature map over time and space and concatenate the two pathways ($2048$ slow $+$ $256$ fast $= 2304$ channels), which feeds the \SPL head. \cref{tab:slowfast-arch} details the stage configuration.

\begin{table}[H]
    \centering
    \caption{\textbf{\texttt{Slowfast-50} backbone} used for \texttt{road-r}, for an $8$-frame $(3, 8, 96, 96)$ clip.
    Channel counts are given as slow\,/\,fast; the slow pathway runs at
    $T/\alpha = 4$ frames, the fast pathway at $T = 8$.}
    \label{tab:slowfast-arch}
    \scalebox{0.82}{
    \begin{tabular}{lllc}
        \toprule
        \textsc{Stage} & \textsc{Slow pathway} & \textsc{Fast pathway} & \textsc{Output (slow\,/\,fast ch.)} \\
        \midrule
        data    & sample $4$ frames ($\alpha=2$) & all $8$ frames & $3$ / $3$ \\
        conv1   & Conv3d $k{=}(1,7,7)$, $s{=}(1,2,2)$ & Conv3d $k{=}(5,7,7)$, $s{=}(1,2,2)$ & $64$ / $8$ \\
        pool1   & MaxPool3d $k{=}(1,3,3)$, $s{=}(1,2,2)$ & MaxPool3d $k{=}(1,3,3)$, $s{=}(1,2,2)$ & $64$ / $8$ \\
        res2    & $3\times$ bottleneck & $3\times$ bottleneck & $256$ / $32$ \\
        res3    & $4\times$ bottleneck, stride $2$ & $4\times$ bottleneck, stride $2$ & $512$ / $64$ \\
        res4    & $6\times$ bottleneck, stride $2$ & $6\times$ bottleneck, stride $2$ & $1024$ / $128$ \\
        res5    & $3\times$ bottleneck, dilation $2$ & $3\times$ bottleneck, dilation $2$ & $2048$ / $256$ \\
        fusion  & \multicolumn{2}{l}{lateral Conv3d ($s{=}(\alpha,1,1)$) fast$\to$slow after pool1 / res2 / res3 / res4} & \\
        head    & \multicolumn{2}{l}{global avg-pool over $(T,H,W)$, concat slow$\,\|\,$fast} & $2304$ \\
                & \multicolumn{2}{l}{$\texttt{Linear}(2304, h) \to \texttt{ReLU} \to \texttt{Linear}(h, G)$ (\SPL head)} & \\
        \bottomrule
    \end{tabular}
    }
\end{table}

\section{Additional Results} \label{app:additional_results}

In this section we discuss additional results and ablation for the two stages of the auto-formalization pipeline.

\subsection{Auto-Formalization}

In this section, we present additional results along the three axes of inputs, models, and outputs. We also provide an error analysis to better understand the errors of the generated formalizations. This investigation is divided into four components, \ie input-level analysis, model-level analysis, output-level analysis, error analysis.

\subsubsection{Input-Level Analysis}

\begin{figure*}[!htbp]
    \centering
    \includegraphics[width=\textwidth]{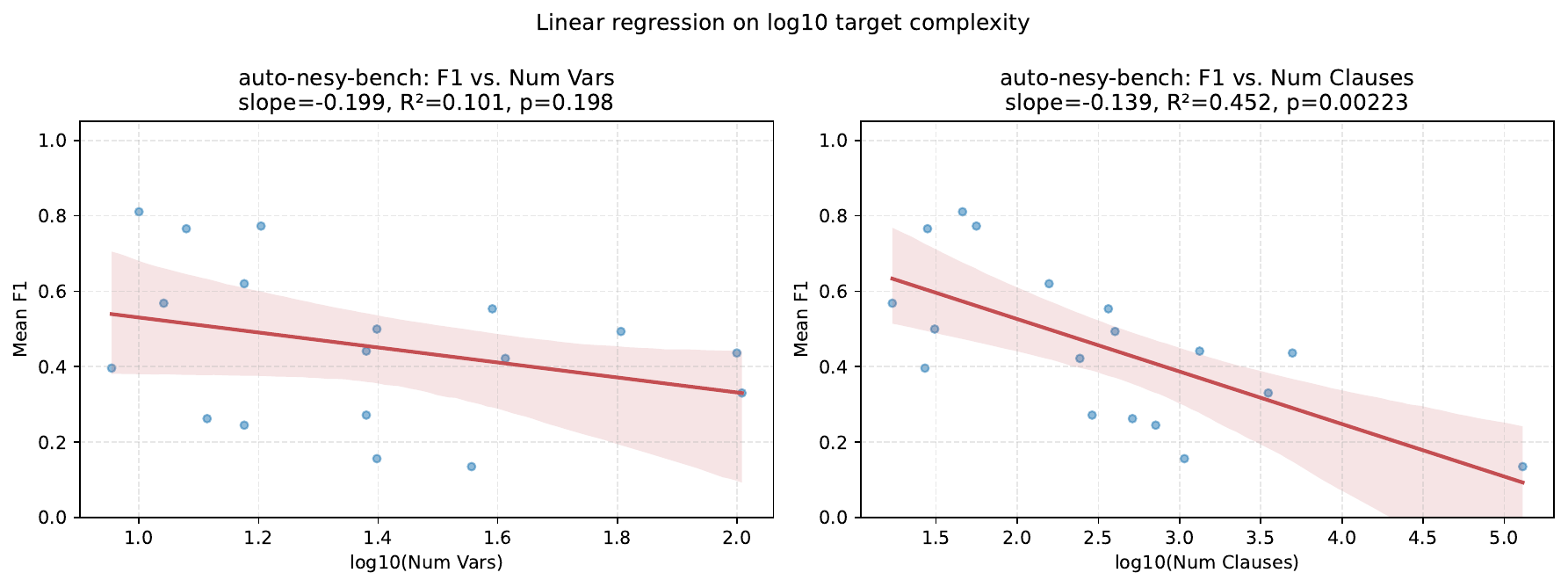}
    \includegraphics[width=\textwidth]{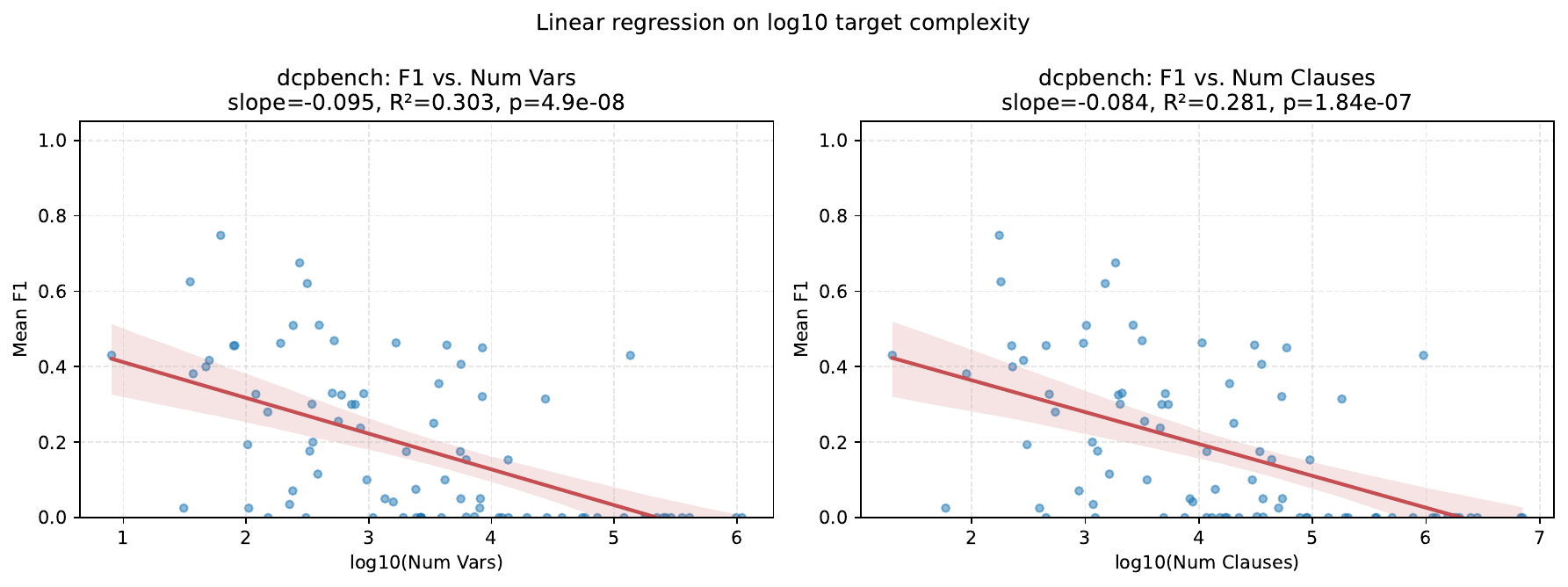}
    \includegraphics[width=\textwidth]{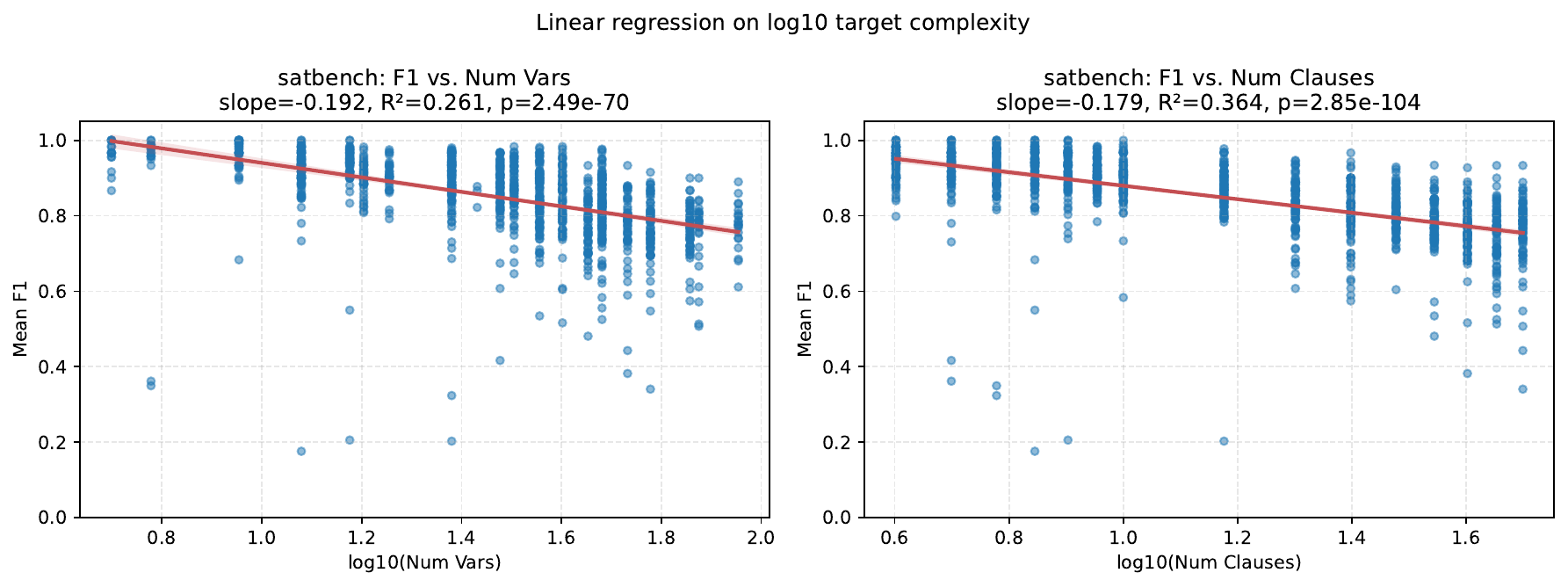}
    \caption{
    \textbf{Larger formulas are generally more difficult to formalize, although the magnitude of this effect is dataset-dependent.} We report mean F1 against the number of variables or clauses on a log scale for \autobench (top), \dcpbench (middle), and \satbench (bottom). Each point represents the mean F1 for a specific benchmark instance. The red line and shaded region indicate the ordinary least-squares fit and its $95\%$ confidence interval. 
    }
    \label{fig:problem-complexity}
\end{figure*}

\textbf{How does problem complexity influence auto-formalization performance?}
Beyond the prompt granularity studied in the main text (\cref{sec:results-auto-form}), auto-formalization performance may also depend on problem complexity. We consider two quantifiable proxies for complexity: the number of variables and clauses in the ground-truth \DIMACS formula. Intuitively, larger formulas impose a greater formalization burden, as the model must track more variables and consistently encode more logical interactions, increasing the risk of omissions and inconsistencies.
To conduct this analysis, for each benchmark, we average the F1 scores across model configurations for problems of different sizes, considering the number of variables and clauses separately. As formula sizes span several orders of magnitude, we use their base-10 logarithms and fit a separate univariate linear regression for each dataset and complexity measure. The resulting slope represents the expected change in mean F1 for a tenfold increase in the number of variables or clauses. We also report the $R^2$ metric, which measures the fraction of between-problem variation in F1 explained by the corresponding complexity measure. \Cref{fig:problem-complexity} shows the problem-level observations, fitted regression lines, and $95\%$ confidence intervals.

All six fitted slopes are negative: across all three datasets, mean F1 decreases as either the number of variables or clauses increases, suggesting that larger target formulas are generally more difficult to formalize.
On \autobench, a tenfold increase in the number of variables is associated with an F1 decrease of $0.1986$ ($R^2=0.101$, $p=0.198$), while a tenfold increase in the number of clauses corresponds to a decrease of $0.1391$ ($R^2=0.452$, $p=0.00223$). Thus, among the 18 \autobench problems, clause count is more informative than variable count, although the small sample size warrants caution.
On \dcpbench, the slopes for variable and clause count are $-0.0947$ and $-0.0845$, respectively. The corresponding fits explain roughly $28$-$30\%$ of the problem-level variation ($R^2=0.303$, $p=4.9\times10^{-8}$ for variables; $R^2=0.281$, $p=1.84\times10^{-7}$ for clauses).

On \satbench, the slopes are $-0.1921$ for variables ($R^2=0.261$, $p=2.49\times10^{-70}$) and $-0.1789$ for clauses ($R^2=0.364$, $p=2.85\times10^{-104}$), indicating that mean F1 decreases with formula size on this benchmark as well.

Overall, larger target formulas are associated with lower auto-formalization performance, but the magnitude of this effect is dataset-dependent. The results also suggest that problem complexity may not be fully captured by the formula size.

\subsubsection{Model-Level Analysis}
\textbf{How well can each model perform under its best-performing setting?}
In the main text, we analyze model performance on each dataset across the ten configurations, averaging over two prompting strategies and five output formats. Here, we further examine the performance of each model by selecting its best-performing configuration. To make the comparison fair, we selected the detailed version of the prompt for \autobench. For each model, we select the prompting-format configuration that achieves the highest F1 score.
As shown in \cref{tab:best-model-configurations}, selecting the best configuration for each model yields substantial improvements over averaging all configurations equally.
On \autobench, the mean F1 across models increases from $0.455$ to $0.692$; on \dcpbench, it increases from $0.183$ to $0.297$; and on \satbench, it increases from $0.849$ to $0.942$.
These gaps indicate that matching the prompting strategy and output format to each model is important for auto-formalization performance.
The selected configurations also reveal clear preferences across datasets. On \autobench, ten of the eleven models select ZS-CoT, with \texttt{deepseek-r1-llama-70b} as the only exception. Seven of the eleven selected configurations use \cpmpy, three use \sympy, and one uses \nat, indicating a strong preference for high-level symbolic programming formats. On \dcpbench, three of the four models select ZS, while three select \cpmpy and one selects \sympy. On \satbench, all three evaluated models select ZS-CoT with \nat. Together, these results suggest that the preferred prompting strategy and output format depend on the dataset, although high-level code formats are predominant on \autobench and \dcpbench.
Several models achieve very strong performance under their best configurations. On \autobench, \texttt{gemma4-31b} achieves the highest F1 and precision, at $0.991$ and $0.986$, respectively, followed by \texttt{gpt-oss-120b}, with an F1 of $0.984$ and precision of $0.974$. Both also attain near-perfect recall, reaching $0.997$ and $1.000$, respectively.
On \dcpbench, \texttt{gpt-oss-120b} achieves the highest F1 ($0.433$) with ZS and \cpmpy, followed by \texttt{qwen3-coder-next} ($0.302$) with ZS-CoT and \cpmpy, \texttt{gpt-oss-20b} ($0.286$) with ZS and \cpmpy, and \texttt{deepseek-r1-llama-70b} ($0.165$) with ZS and \sympy. On \satbench, all three models select ZS-CoT and \nat, with \texttt{gemma4-31b} achieving the highest F1 of $0.992$, followed by \texttt{gpt-oss-120b} at $0.976$ and \texttt{qwen3-coder-next} at $0.859$.
These results show that, with an appropriate prompting strategy and output format, some current open-weight LLMs can formalize logical constraints from natural language descriptions with very high accuracy on \autobench and \satbench. 
At the same time, the substantially lower best-configuration performance on \dcpbench indicates that reliable auto-formalization remains challenging for more difficult instances.

\begin{table*}[htbp]
\centering
\scriptsize
\setlength{\tabcolsep}{4.5pt}
\caption{\textbf{Selecting an appropriate prompting strategy and output format substantially improves best-case formalization performance.}
Best formalization performance of each model on \autobench, \dcpbench, and \satbench. For each model, we select the best of ten configurations formed by two prompting strategies and five output formats according to F1 score, breaking ties by accuracy and then by fixed prompt and format order, and report the corresponding precision, recall, and syntax error rate. Average rows give each model equal weight. Within each dataset, best results are bolded, and second-best results are underlined; averages are excluded from the ranking.}
\label{tab:best-model-configurations}
\scalebox{0.9}{
\begin{tabular}{llllrrrr}
\toprule
\textsc{Dataset} & \textsc{Model} & \textsc{Prompt} & \textsc{Format} & \textsc{F1} ($\uparrow$) & \textsc{Precision} ($\uparrow$) & \textsc{Recall} ($\uparrow$) & \textsc{Syn. ER} ($\downarrow$) \\
\midrule
\multirow[c]{12}{*}{\autobench}
 & \texttt{deepseek-r1-llama-70b} & ZS & \sympy & $0.676$ & $0.640$ & $0.810$ & $\underline{11.1\%}$ \\
 & \texttt{gemma3-27b} & ZS-CoT & \nat & $0.135$ & $0.192$ & $0.449$ & $22.2\%$ \\
 & \texttt{gemma4-31b} & ZS-CoT & \cpmpy & $\mathbf{0.991}$ & $\mathbf{0.986}$ & $\underline{0.997}$ & $\mathbf{0.0\%}$ \\
 & \texttt{gpt-oss-120b} & ZS-CoT & \cpmpy & $\underline{0.984}$ & $\underline{0.974}$ & $\mathbf{1.000}$ & $\mathbf{0.0\%}$ \\
 & \texttt{gpt-oss-20b} & ZS-CoT & \sympy & $0.769$ & $0.756$ & $0.833$ & $16.7\%$ \\
 & \texttt{mistral-nemo} & ZS-CoT & \cpmpy & $0.259$ & $0.256$ & $0.325$ & $50.0\%$ \\
 & \texttt{olmo3-32b-think} & ZS-CoT & \cpmpy & $0.733$ & $0.711$ & $0.833$ & $\underline{11.1\%}$ \\
 & \texttt{phi4-reasoning-plus} & ZS-CoT & \sympy & $0.721$ & $0.720$ & $0.722$ & $27.8\%$ \\
 & \texttt{qwen3-32b} & ZS-CoT & \cpmpy & $0.935$ & $0.922$ & $\mathbf{1.000}$ & $\mathbf{0.0\%}$ \\
 & \texttt{qwen3-8b} & ZS-CoT & \cpmpy & $0.649$ & $0.694$ & $0.664$ & $\underline{11.1\%}$ \\
 & \texttt{qwen3-coder-next} & ZS-CoT & \cpmpy & $0.755$ & $0.802$ & $0.770$ & $16.7\%$ \\
 & \textsc{Average} & --- & --- & $0.692$ & $0.696$ & $0.764$ & $15.2\%$ \\
\midrule
\multirow[c]{5}{*}{\dcpbench}
 & \texttt{deepseek-r1-llama-70b} & ZS & \sympy & $0.165$ & $0.162$ & $0.271$ & $37.6\%$ \\
 & \texttt{gpt-oss-120b} & ZS & \cpmpy & $\mathbf{0.433}$ & $\mathbf{0.441}$ & $\mathbf{0.506}$ & $\mathbf{5.9\%}$ \\
 & \texttt{gpt-oss-20b} & ZS & \cpmpy & $0.286$ & $0.286$ & $\underline{0.333}$ & $36.5\%$ \\
 & \texttt{qwen3-coder-next} & ZS-CoT & \cpmpy & $\underline{0.302}$ & $\underline{0.318}$ & $0.312$ & $\underline{29.4\%}$ \\
 & \textsc{Average} & --- & --- & $0.297$ & $0.302$ & $0.355$ & $27.4\%$ \\
\midrule
\multirow[c]{4}{*}{\satbench}
 & \texttt{gemma4-31b} & ZS-CoT & \nat & $\mathbf{0.992}$ & $\mathbf{0.991}$ & $\mathbf{0.994}$ & $\mathbf{0.0\%}$ \\
 & \texttt{gpt-oss-120b} & ZS-CoT & \nat & $\underline{0.976}$ & $\underline{0.973}$ & $\underline{0.988}$ & $\mathbf{0.0\%}$ \\
 & \texttt{qwen3-coder-next} & ZS-CoT & \nat & $0.859$ & $0.849$ & $0.907$ & $\underline{2.0\%}$ \\
 & \textsc{Average} & --- & --- & $0.942$ & $0.937$ & $0.963$ & $0.7\%$ \\
\bottomrule
\end{tabular}}
\end{table*}

\textbf{Complete~\cref{sec:results-auto-form} results}. \Cref{tab:config-results-autobench,tab:config-results-dcpbench,tab:config-results-satbench} report the underlying results for every available model, prompting strategy, and output format. Each entry is averaged across the problems in the corresponding dataset.

\begingroup
\scriptsize
\setlength{\tabcolsep}{4pt}
\begin{longtable}{@{}lllrrrr@{}}
\caption{Complete per-configuration formalization results on \autobench. ZS and ZS-CoT denote zero-shot and zero-shot chain-of-thought prompting. Missing task scores are assigned zero. Syntax ER is taken directly from the selected results JSON.}
\label{tab:config-results-autobench}\\
\toprule
\textsc{Model} & \textsc{Prompt} & \textsc{Format} & \textsc{F1} ($\uparrow$) & \textsc{Precision} ($\uparrow$) & \textsc{Recall} ($\uparrow$) & \textsc{Syntax ER} ($\downarrow$) \\
\midrule
\endfirsthead
\multicolumn{7}{c}{\tablename\ \thetable{} -- continued} \\
\toprule
\textsc{Model} & \textsc{Prompt} & \textsc{Format} & \textsc{F1} ($\uparrow$) & \textsc{Precision} ($\uparrow$) & \textsc{Recall} ($\uparrow$) & \textsc{Syntax ER} ($\downarrow$) \\
\midrule
\endhead
\midrule
\multicolumn{7}{r}{Continued on next page} \\
\endfoot
\bottomrule
\endlastfoot
\texttt{deepseek-r1-llama-70b} & ZS & \cpmpy & $0.503$ & $0.532$ & $0.551$ & $33.3\%$ \\
\rowcolor[HTML]{EFEFEF}
\texttt{deepseek-r1-llama-70b} & ZS & \DIMACS & $0.523$ & $0.522$ & $0.587$ & $16.7\%$ \\
\texttt{deepseek-r1-llama-70b} & ZS & \nat & $0.363$ & $0.409$ & $0.619$ & $5.6\%$ \\
\rowcolor[HTML]{EFEFEF}
\texttt{deepseek-r1-llama-70b} & ZS & \pysat & $0.541$ & $0.695$ & $0.604$ & $16.7\%$ \\
\texttt{deepseek-r1-llama-70b} & ZS & \sympy & $0.676$ & $0.640$ & $0.810$ & $11.1\%$ \\
\rowcolor[HTML]{EFEFEF}
\texttt{deepseek-r1-llama-70b} & ZS-CoT & \cpmpy & $0.111$ & $0.111$ & $0.167$ & $83.3\%$ \\
\texttt{deepseek-r1-llama-70b} & ZS-CoT & \DIMACS & $0.388$ & $0.360$ & $0.484$ & $11.1\%$ \\
\rowcolor[HTML]{EFEFEF}
\texttt{deepseek-r1-llama-70b} & ZS-CoT & \nat & $0.390$ & $0.451$ & $0.578$ & $16.7\%$ \\
\texttt{deepseek-r1-llama-70b} & ZS-CoT & \pysat & $0.316$ & $0.325$ & $0.405$ & $38.9\%$ \\
\rowcolor[HTML]{EFEFEF}
\texttt{deepseek-r1-llama-70b} & ZS-CoT & \sympy & $0.313$ & $0.300$ & $0.462$ & $50.0\%$ \\
\midrule
\texttt{gemma3-27b} & ZS & \cpmpy & $0.049$ & $0.043$ & $0.167$ & $83.3\%$ \\
\rowcolor[HTML]{EFEFEF}
\texttt{gemma3-27b} & ZS & \DIMACS & $0.006$ & $0.020$ & $0.062$ & $11.1\%$ \\
\texttt{gemma3-27b} & ZS & \nat & $0.064$ & $0.060$ & $0.361$ & $33.3\%$ \\
\rowcolor[HTML]{EFEFEF}
\texttt{gemma3-27b} & ZS & \pysat & $0.056$ & $0.065$ & $0.077$ & $66.7\%$ \\
\texttt{gemma3-27b} & ZS & \sympy & $0.029$ & $0.017$ & $0.222$ & $77.8\%$ \\
\rowcolor[HTML]{EFEFEF}
\texttt{gemma3-27b} & ZS-CoT & \cpmpy & $0.056$ & $0.056$ & $0.111$ & $88.9\%$ \\
\texttt{gemma3-27b} & ZS-CoT & \DIMACS & $0.002$ & $0.022$ & $0.061$ & $16.7\%$ \\
\rowcolor[HTML]{EFEFEF}
\texttt{gemma3-27b} & ZS-CoT & \nat & $0.135$ & $0.192$ & $0.449$ & $22.2\%$ \\
\texttt{gemma3-27b} & ZS-CoT & \pysat & $0.003$ & $0.002$ & $0.017$ & $66.7\%$ \\
\rowcolor[HTML]{EFEFEF}
\texttt{gemma3-27b} & ZS-CoT & \sympy & $0.001$ & $0.000$ & $0.111$ & $88.9\%$ \\
\midrule
\texttt{gemma4-31b} & ZS & \cpmpy & $0.886$ & $0.887$ & $0.941$ & $5.6\%$ \\
\rowcolor[HTML]{EFEFEF}
\texttt{gemma4-31b} & ZS & \DIMACS & $0.254$ & $0.320$ & $0.365$ & $5.6\%$ \\
\texttt{gemma4-31b} & ZS & \nat & $0.334$ & $0.461$ & $0.462$ & $0.0\%$ \\
\rowcolor[HTML]{EFEFEF}
\texttt{gemma4-31b} & ZS & \pysat & $0.770$ & $0.863$ & $0.767$ & $5.6\%$ \\
\texttt{gemma4-31b} & ZS & \sympy & $0.748$ & $0.728$ & $0.859$ & $5.6\%$ \\
\rowcolor[HTML]{EFEFEF}
\texttt{gemma4-31b} & ZS-CoT & \cpmpy & $0.991$ & $0.986$ & $0.997$ & $0.0\%$ \\
\texttt{gemma4-31b} & ZS-CoT & \DIMACS & $0.715$ & $0.731$ & $0.807$ & $5.6\%$ \\
\rowcolor[HTML]{EFEFEF}
\texttt{gemma4-31b} & ZS-CoT & \nat & $0.586$ & $0.591$ & $0.791$ & $0.0\%$ \\
\texttt{gemma4-31b} & ZS-CoT & \pysat & $0.900$ & $0.889$ & $0.917$ & $5.6\%$ \\
\rowcolor[HTML]{EFEFEF}
\texttt{gemma4-31b} & ZS-CoT & \sympy & $0.873$ & $0.863$ & $0.915$ & $5.6\%$ \\
\midrule
\texttt{gpt-oss-120b} & ZS & \cpmpy & $0.925$ & $0.971$ & $0.939$ & $0.0\%$ \\
\rowcolor[HTML]{EFEFEF}
\texttt{gpt-oss-120b} & ZS & \DIMACS & $0.640$ & $0.664$ & $0.719$ & $11.1\%$ \\
\texttt{gpt-oss-120b} & ZS & \nat & $0.562$ & $0.568$ & $0.650$ & $16.7\%$ \\
\rowcolor[HTML]{EFEFEF}
\texttt{gpt-oss-120b} & ZS & \pysat & $0.813$ & $0.856$ & $0.828$ & $0.0\%$ \\
\texttt{gpt-oss-120b} & ZS & \sympy & $0.812$ & $0.804$ & $0.826$ & $16.7\%$ \\
\rowcolor[HTML]{EFEFEF}
\texttt{gpt-oss-120b} & ZS-CoT & \cpmpy & $0.984$ & $0.974$ & $1.000$ & $0.0\%$ \\
\texttt{gpt-oss-120b} & ZS-CoT & \DIMACS & $0.687$ & $0.720$ & $0.792$ & $5.6\%$ \\
\rowcolor[HTML]{EFEFEF}
\texttt{gpt-oss-120b} & ZS-CoT & \nat & $0.645$ & $0.629$ & $0.796$ & $16.7\%$ \\
\texttt{gpt-oss-120b} & ZS-CoT & \pysat & $0.924$ & $0.914$ & $0.941$ & $0.0\%$ \\
\rowcolor[HTML]{EFEFEF}
\texttt{gpt-oss-120b} & ZS-CoT & \sympy & $0.871$ & $0.863$ & $0.885$ & $11.1\%$ \\
\midrule
\texttt{gpt-oss-20b} & ZS & \cpmpy & $0.703$ & $0.695$ & $0.719$ & $27.8\%$ \\
\rowcolor[HTML]{EFEFEF}
\texttt{gpt-oss-20b} & ZS & \DIMACS & $0.563$ & $0.551$ & $0.641$ & $22.2\%$ \\
\texttt{gpt-oss-20b} & ZS & \nat & $0.328$ & $0.423$ & $0.375$ & $50.0\%$ \\
\rowcolor[HTML]{EFEFEF}
\texttt{gpt-oss-20b} & ZS & \pysat & $0.403$ & $0.584$ & $0.485$ & $11.1\%$ \\
\texttt{gpt-oss-20b} & ZS & \sympy & $0.616$ & $0.665$ & $0.614$ & $33.3\%$ \\
\rowcolor[HTML]{EFEFEF}
\texttt{gpt-oss-20b} & ZS-CoT & \cpmpy & $0.634$ & $0.666$ & $0.625$ & $33.3\%$ \\
\texttt{gpt-oss-20b} & ZS-CoT & \DIMACS & $0.366$ & $0.400$ & $0.472$ & $38.9\%$ \\
\rowcolor[HTML]{EFEFEF}
\texttt{gpt-oss-20b} & ZS-CoT & \nat & $0.283$ & $0.391$ & $0.334$ & $38.9\%$ \\
\texttt{gpt-oss-20b} & ZS-CoT & \pysat & $0.383$ & $0.511$ & $0.458$ & $27.8\%$ \\
\rowcolor[HTML]{EFEFEF}
\texttt{gpt-oss-20b} & ZS-CoT & \sympy & $0.769$ & $0.756$ & $0.833$ & $16.7\%$ \\
\midrule
\texttt{mistral-nemo} & ZS & \cpmpy & $0.094$ & $0.085$ & $0.222$ & $66.7\%$ \\
\rowcolor[HTML]{EFEFEF}
\texttt{mistral-nemo} & ZS & \DIMACS & $0.036$ & $0.026$ & $0.202$ & $33.3\%$ \\
\texttt{mistral-nemo} & ZS & \nat & $0.000$ & $0.013$ & $0.000$ & $77.8\%$ \\
\rowcolor[HTML]{EFEFEF}
\texttt{mistral-nemo} & ZS & \pysat & $0.020$ & $0.012$ & $0.184$ & $66.7\%$ \\
\texttt{mistral-nemo} & ZS & \sympy & $0.000$ & $0.000$ & $0.056$ & $94.4\%$ \\
\rowcolor[HTML]{EFEFEF}
\texttt{mistral-nemo} & ZS-CoT & \cpmpy & $0.259$ & $0.256$ & $0.325$ & $50.0\%$ \\
\texttt{mistral-nemo} & ZS-CoT & \DIMACS & $0.058$ & $0.054$ & $0.184$ & $33.3\%$ \\
\rowcolor[HTML]{EFEFEF}
\texttt{mistral-nemo} & ZS-CoT & \nat & $0.018$ & $0.011$ & $0.108$ & $77.8\%$ \\
\texttt{mistral-nemo} & ZS-CoT & \pysat & $0.039$ & $0.051$ & $0.206$ & $33.3\%$ \\
\rowcolor[HTML]{EFEFEF}
\texttt{mistral-nemo} & ZS-CoT & \sympy & $0.007$ & $0.003$ & $0.056$ & $94.4\%$ \\
\midrule
\texttt{olmo3-32b-think} & ZS & \cpmpy & $0.547$ & $0.553$ & $0.598$ & $33.3\%$ \\
\rowcolor[HTML]{EFEFEF}
\texttt{olmo3-32b-think} & ZS & \DIMACS & $0.580$ & $0.572$ & $0.676$ & $5.6\%$ \\
\texttt{olmo3-32b-think} & ZS & \nat & $0.416$ & $0.431$ & $0.639$ & $16.7\%$ \\
\rowcolor[HTML]{EFEFEF}
\texttt{olmo3-32b-think} & ZS & \pysat & $0.650$ & $0.722$ & $0.676$ & $16.7\%$ \\
\texttt{olmo3-32b-think} & ZS & \sympy & $0.698$ & $0.735$ & $0.758$ & $11.1\%$ \\
\rowcolor[HTML]{EFEFEF}
\texttt{olmo3-32b-think} & ZS-CoT & \cpmpy & $0.733$ & $0.711$ & $0.833$ & $11.1\%$ \\
\texttt{olmo3-32b-think} & ZS-CoT & \DIMACS & $0.372$ & $0.569$ & $0.574$ & $5.6\%$ \\
\rowcolor[HTML]{EFEFEF}
\texttt{olmo3-32b-think} & ZS-CoT & \nat & $0.415$ & $0.449$ & $0.504$ & $27.8\%$ \\
\texttt{olmo3-32b-think} & ZS-CoT & \pysat & $0.567$ & $0.675$ & $0.631$ & $16.7\%$ \\
\rowcolor[HTML]{EFEFEF}
\texttt{olmo3-32b-think} & ZS-CoT & \sympy & $0.716$ & $0.676$ & $0.878$ & $5.6\%$ \\
\midrule
\texttt{phi4-reasoning-plus} & ZS & \cpmpy & $0.499$ & $0.554$ & $0.611$ & $33.3\%$ \\
\rowcolor[HTML]{EFEFEF}
\texttt{phi4-reasoning-plus} & ZS & \DIMACS & $0.167$ & $0.222$ & $0.167$ & $72.2\%$ \\
\texttt{phi4-reasoning-plus} & ZS & \nat & $0.249$ & $0.299$ & $0.253$ & $55.6\%$ \\
\rowcolor[HTML]{EFEFEF}
\texttt{phi4-reasoning-plus} & ZS & \pysat & $0.333$ & $0.333$ & $0.333$ & $66.7\%$ \\
\texttt{phi4-reasoning-plus} & ZS & \sympy & $0.613$ & $0.612$ & $0.667$ & $27.8\%$ \\
\rowcolor[HTML]{EFEFEF}
\texttt{phi4-reasoning-plus} & ZS-CoT & \cpmpy & $0.548$ & $0.544$ & $0.552$ & $38.9\%$ \\
\texttt{phi4-reasoning-plus} & ZS-CoT & \DIMACS & $0.222$ & $0.222$ & $0.222$ & $66.7\%$ \\
\rowcolor[HTML]{EFEFEF}
\texttt{phi4-reasoning-plus} & ZS-CoT & \nat & $0.333$ & $0.333$ & $0.333$ & $66.7\%$ \\
\texttt{phi4-reasoning-plus} & ZS-CoT & \pysat & $0.542$ & $0.541$ & $0.543$ & $38.9\%$ \\
\rowcolor[HTML]{EFEFEF}
\texttt{phi4-reasoning-plus} & ZS-CoT & \sympy & $0.721$ & $0.720$ & $0.722$ & $27.8\%$ \\
\midrule
\texttt{qwen3-32b} & ZS & \cpmpy & $0.838$ & $0.863$ & $0.868$ & $0.0\%$ \\
\rowcolor[HTML]{EFEFEF}
\texttt{qwen3-32b} & ZS & \DIMACS & $0.574$ & $0.615$ & $0.692$ & $11.1\%$ \\
\texttt{qwen3-32b} & ZS & \nat & $0.530$ & $0.555$ & $0.610$ & $16.7\%$ \\
\rowcolor[HTML]{EFEFEF}
\texttt{qwen3-32b} & ZS & \pysat & $0.644$ & $0.630$ & $0.667$ & $22.2\%$ \\
\texttt{qwen3-32b} & ZS & \sympy & $0.690$ & $0.694$ & $0.687$ & $22.2\%$ \\
\rowcolor[HTML]{EFEFEF}
\texttt{qwen3-32b} & ZS-CoT & \cpmpy & $0.935$ & $0.922$ & $1.000$ & $0.0\%$ \\
\texttt{qwen3-32b} & ZS-CoT & \DIMACS & $0.468$ & $0.488$ & $0.485$ & $38.9\%$ \\
\rowcolor[HTML]{EFEFEF}
\texttt{qwen3-32b} & ZS-CoT & \nat & $0.536$ & $0.595$ & $0.680$ & $11.1\%$ \\
\texttt{qwen3-32b} & ZS-CoT & \pysat & $0.717$ & $0.726$ & $0.741$ & $16.7\%$ \\
\rowcolor[HTML]{EFEFEF}
\texttt{qwen3-32b} & ZS-CoT & \sympy & $0.618$ & $0.689$ & $0.693$ & $5.6\%$ \\
\midrule
\texttt{qwen3-8b} & ZS & \cpmpy & $0.498$ & $0.483$ & $0.548$ & $11.1\%$ \\
\rowcolor[HTML]{EFEFEF}
\texttt{qwen3-8b} & ZS & \DIMACS & $0.291$ & $0.380$ & $0.431$ & $11.1\%$ \\
\texttt{qwen3-8b} & ZS & \nat & $0.292$ & $0.285$ & $0.403$ & $33.3\%$ \\
\rowcolor[HTML]{EFEFEF}
\texttt{qwen3-8b} & ZS & \pysat & $0.122$ & $0.181$ & $0.151$ & $61.1\%$ \\
\texttt{qwen3-8b} & ZS & \sympy & $0.600$ & $0.596$ & $0.604$ & $11.1\%$ \\
\rowcolor[HTML]{EFEFEF}
\texttt{qwen3-8b} & ZS-CoT & \cpmpy & $0.649$ & $0.694$ & $0.664$ & $11.1\%$ \\
\texttt{qwen3-8b} & ZS-CoT & \DIMACS & $0.435$ & $0.510$ & $0.475$ & $11.1\%$ \\
\rowcolor[HTML]{EFEFEF}
\texttt{qwen3-8b} & ZS-CoT & \nat & $0.272$ & $0.272$ & $0.356$ & $44.4\%$ \\
\texttt{qwen3-8b} & ZS-CoT & \pysat & $0.000$ & $0.056$ & $0.000$ & $88.9\%$ \\
\rowcolor[HTML]{EFEFEF}
\texttt{qwen3-8b} & ZS-CoT & \sympy & $0.564$ & $0.563$ & $0.601$ & $22.2\%$ \\
\midrule
\texttt{qwen3-coder-next} & ZS & \cpmpy & $0.460$ & $0.498$ & $0.480$ & $38.9\%$ \\
\rowcolor[HTML]{EFEFEF}
\texttt{qwen3-coder-next} & ZS & \DIMACS & $0.135$ & $0.126$ & $0.481$ & $5.6\%$ \\
\texttt{qwen3-coder-next} & ZS & \nat & $0.235$ & $0.231$ & $0.305$ & $27.8\%$ \\
\rowcolor[HTML]{EFEFEF}
\texttt{qwen3-coder-next} & ZS & \pysat & $0.491$ & $0.540$ & $0.582$ & $16.7\%$ \\
\texttt{qwen3-coder-next} & ZS & \sympy & $0.538$ & $0.558$ & $0.527$ & $16.7\%$ \\
\rowcolor[HTML]{EFEFEF}
\texttt{qwen3-coder-next} & ZS-CoT & \cpmpy & $0.755$ & $0.802$ & $0.770$ & $16.7\%$ \\
\texttt{qwen3-coder-next} & ZS-CoT & \DIMACS & $0.519$ & $0.584$ & $0.703$ & $11.1\%$ \\
\rowcolor[HTML]{EFEFEF}
\texttt{qwen3-coder-next} & ZS-CoT & \nat & $0.303$ & $0.295$ & $0.444$ & $44.4\%$ \\
\texttt{qwen3-coder-next} & ZS-CoT & \pysat & $0.729$ & $0.827$ & $0.751$ & $0.0\%$ \\
\rowcolor[HTML]{EFEFEF}
\texttt{qwen3-coder-next} & ZS-CoT & \sympy & $0.611$ & $0.608$ & $0.718$ & $22.2\%$ \\
\end{longtable}
\endgroup

\begingroup
\scriptsize
\setlength{\tabcolsep}{4pt}
\begin{longtable}{@{}lllrrrr@{}}
\caption{Complete per-configuration formalization results on \dcpbench. ZS and ZS-CoT denote zero-shot and zero-shot chain-of-thought prompting. Missing task scores are assigned zero. Syntax ER is taken directly from the selected results JSON.}
\label{tab:config-results-dcpbench}\\
\toprule
\textsc{Model} & \textsc{Prompt} & \textsc{Format} & \textsc{F1 ($\uparrow$)} & \textsc{Precision ($\uparrow$)} & \textsc{Recall ($\uparrow$)} & \textsc{Syntax ER ($\downarrow$)} \\
\midrule
\endfirsthead
\multicolumn{7}{c}{\tablename\ \thetable{} -- continued} \\
\toprule
\textsc{Model} & \textsc{Prompt} & \textsc{Format} & \textsc{F1 ($\uparrow$)} & \textsc{Precision ($\uparrow$)} & \textsc{Recall ($\uparrow$)} & \textsc{Syntax ER ($\downarrow$)} \\
\midrule
\endhead
\midrule
\multicolumn{7}{r}{Continued on next page} \\
\endfoot
\bottomrule
\endlastfoot
\texttt{deepseek-r1-llama-70b} & ZS & \cpmpy & $0.121$ & $0.119$ & $0.200$ & $40.0\%$ \\
\rowcolor[HTML]{EFEFEF}
\texttt{deepseek-r1-llama-70b} & ZS & \DIMACS & $0.078$ & $0.075$ & $0.319$ & $25.9\%$ \\
\texttt{deepseek-r1-llama-70b} & ZS & \nat & $0.061$ & $0.063$ & $0.210$ & $29.4\%$ \\
\rowcolor[HTML]{EFEFEF}
\texttt{deepseek-r1-llama-70b} & ZS & \pysat & $0.085$ & $0.088$ & $0.174$ & $30.6\%$ \\
\texttt{deepseek-r1-llama-70b} & ZS & \sympy & $0.165$ & $0.162$ & $0.271$ & $37.6\%$ \\
\rowcolor[HTML]{EFEFEF}
\texttt{deepseek-r1-llama-70b} & ZS-CoT & \cpmpy & $0.067$ & $0.071$ & $0.124$ & $44.7\%$ \\
\texttt{deepseek-r1-llama-70b} & ZS-CoT & \DIMACS & $0.030$ & $0.030$ & $0.270$ & $18.8\%$ \\
\rowcolor[HTML]{EFEFEF}
\texttt{deepseek-r1-llama-70b} & ZS-CoT & \nat & $0.071$ & $0.073$ & $0.270$ & $30.6\%$ \\
\texttt{deepseek-r1-llama-70b} & ZS-CoT & \pysat & $0.049$ & $0.046$ & $0.076$ & $69.4\%$ \\
\rowcolor[HTML]{EFEFEF}
\texttt{deepseek-r1-llama-70b} & ZS-CoT & \sympy & $0.052$ & $0.048$ & $0.106$ & $68.2\%$ \\
\midrule
\texttt{gpt-oss-120b} & ZS & \cpmpy & $0.433$ & $0.441$ & $0.506$ & $5.9\%$ \\
\rowcolor[HTML]{EFEFEF}
\texttt{gpt-oss-120b} & ZS & \DIMACS & $0.266$ & $0.297$ & $0.383$ & $5.9\%$ \\
\texttt{gpt-oss-120b} & ZS & \nat & $0.184$ & $0.200$ & $0.263$ & $45.9\%$ \\
\rowcolor[HTML]{EFEFEF}
\texttt{gpt-oss-120b} & ZS & \pysat & $0.410$ & $0.421$ & $0.465$ & $3.5\%$ \\
\texttt{gpt-oss-120b} & ZS & \sympy & $0.356$ & $0.367$ & $0.469$ & $17.6\%$ \\
\rowcolor[HTML]{EFEFEF}
\texttt{gpt-oss-120b} & ZS-CoT & \cpmpy & $0.391$ & $0.395$ & $0.467$ & $12.9\%$ \\
\texttt{gpt-oss-120b} & ZS-CoT & \DIMACS & $0.266$ & $0.301$ & $0.344$ & $21.2\%$ \\
\rowcolor[HTML]{EFEFEF}
\texttt{gpt-oss-120b} & ZS-CoT & \nat & $0.141$ & $0.167$ & $0.210$ & $55.3\%$ \\
\texttt{gpt-oss-120b} & ZS-CoT & \pysat & $0.429$ & $0.439$ & $0.458$ & $3.5\%$ \\
\rowcolor[HTML]{EFEFEF}
\texttt{gpt-oss-120b} & ZS-CoT & \sympy & $0.360$ & $0.371$ & $0.433$ & $8.2\%$ \\
\midrule
\texttt{gpt-oss-20b} & ZS & \cpmpy & $0.286$ & $0.286$ & $0.333$ & $36.5\%$ \\
\rowcolor[HTML]{EFEFEF}
\texttt{gpt-oss-20b} & ZS & \DIMACS & $0.183$ & $0.194$ & $0.261$ & $40.0\%$ \\
\texttt{gpt-oss-20b} & ZS & \nat & $0.122$ & $0.120$ & $0.180$ & $64.7\%$ \\
\rowcolor[HTML]{EFEFEF}
\texttt{gpt-oss-20b} & ZS & \pysat & $0.123$ & $0.136$ & $0.177$ & $21.2\%$ \\
\texttt{gpt-oss-20b} & ZS & \sympy & $0.261$ & $0.260$ & $0.312$ & $40.0\%$ \\
\rowcolor[HTML]{EFEFEF}
\texttt{gpt-oss-20b} & ZS-CoT & \cpmpy & $0.214$ & $0.213$ & $0.235$ & $40.0\%$ \\
\texttt{gpt-oss-20b} & ZS-CoT & \DIMACS & $0.157$ & $0.156$ & $0.227$ & $51.8\%$ \\
\rowcolor[HTML]{EFEFEF}
\texttt{gpt-oss-20b} & ZS-CoT & \nat & $0.106$ & $0.106$ & $0.172$ & $69.4\%$ \\
\texttt{gpt-oss-20b} & ZS-CoT & \pysat & $0.244$ & $0.247$ & $0.276$ & $25.9\%$ \\
\rowcolor[HTML]{EFEFEF}
\texttt{gpt-oss-20b} & ZS-CoT & \sympy & $0.232$ & $0.237$ & $0.271$ & $41.2\%$ \\
\midrule
\texttt{qwen3-coder-next} & ZS & \cpmpy & $0.164$ & $0.167$ & $0.182$ & $69.4\%$ \\
\rowcolor[HTML]{EFEFEF}
\texttt{qwen3-coder-next} & ZS & \DIMACS & $0.005$ & $0.003$ & $0.594$ & $0.0\%$ \\
\texttt{qwen3-coder-next} & ZS & \nat & $0.045$ & $0.041$ & $0.142$ & $35.3\%$ \\
\rowcolor[HTML]{EFEFEF}
\texttt{qwen3-coder-next} & ZS & \pysat & $0.118$ & $0.122$ & $0.151$ & $51.8\%$ \\
\texttt{qwen3-coder-next} & ZS & \sympy & $0.053$ & $0.051$ & $0.118$ & $72.9\%$ \\
\rowcolor[HTML]{EFEFEF}
\texttt{qwen3-coder-next} & ZS-CoT & \cpmpy & $0.302$ & $0.318$ & $0.312$ & $29.4\%$ \\
\texttt{qwen3-coder-next} & ZS-CoT & \DIMACS & $0.188$ & $0.200$ & $0.467$ & $7.1\%$ \\
\rowcolor[HTML]{EFEFEF}
\texttt{qwen3-coder-next} & ZS-CoT & \nat & $0.143$ & $0.142$ & $0.300$ & $32.9\%$ \\
\texttt{qwen3-coder-next} & ZS-CoT & \pysat & $0.202$ & $0.200$ & $0.229$ & $21.2\%$ \\
\rowcolor[HTML]{EFEFEF}
\texttt{qwen3-coder-next} & ZS-CoT & \sympy & $0.170$ & $0.181$ & $0.197$ & $48.2\%$ \\
\end{longtable}
\endgroup

\begingroup
\scriptsize
\setlength{\tabcolsep}{4pt}
\begin{longtable}{@{}lllrrrr@{}}
\caption{Complete per-configuration formalization results on \satbench. ZS and ZS-CoT denote zero-shot and zero-shot chain-of-thought prompting. Missing task scores are assigned zero. Syntax ER is taken directly from the selected results JSON.}
\label{tab:config-results-satbench}\\
\toprule
\textsc{Model} & \textsc{Prompt} & \textsc{Format} & \textsc{F1} ($\uparrow$) & \textsc{Precision} ($\uparrow$) & \textsc{Recall} ($\uparrow$) & \textsc{Syntax ER} ($\downarrow$) \\
\midrule
\endfirsthead
\multicolumn{7}{c}{\tablename\ \thetable{} -- continued} \\
\toprule
\textsc{Model} & \textsc{Prompt} & \textsc{Format} & \textsc{F1} ($\uparrow$) & \textsc{Precision} ($\uparrow$) & \textsc{Recall} ($\uparrow$) & \textsc{Syntax ER} ($\downarrow$) \\
\midrule
\endhead
\midrule
\multicolumn{7}{r}{Continued on next page} \\
\endfoot
\bottomrule
\endlastfoot
\texttt{gemma4-31b} & ZS & \cpmpy & $0.988$ & $0.987$ & $0.991$ & $0.1\%$ \\
\rowcolor[HTML]{EFEFEF}
\texttt{gemma4-31b} & ZS & \DIMACS & $0.530$ & $0.520$ & $0.568$ & $0.0\%$ \\
\texttt{gemma4-31b} & ZS & \nat & $0.965$ & $0.962$ & $0.980$ & $0.0\%$ \\
\rowcolor[HTML]{EFEFEF}
\texttt{gemma4-31b} & ZS & \pysat & $0.982$ & $0.981$ & $0.986$ & $0.0\%$ \\
\texttt{gemma4-31b} & ZS & \sympy & $0.970$ & $0.970$ & $0.976$ & $1.0\%$ \\
\rowcolor[HTML]{EFEFEF}
\texttt{gemma4-31b} & ZS-CoT & \cpmpy & $0.991$ & $0.990$ & $0.995$ & $0.0\%$ \\
\texttt{gemma4-31b} & ZS-CoT & \DIMACS & $0.974$ & $0.973$ & $0.978$ & $0.0\%$ \\
\rowcolor[HTML]{EFEFEF}
\texttt{gemma4-31b} & ZS-CoT & \nat & $0.992$ & $0.991$ & $0.994$ & $0.0\%$ \\
\texttt{gemma4-31b} & ZS-CoT & \pysat & $0.975$ & $0.974$ & $0.981$ & $0.0\%$ \\
\rowcolor[HTML]{EFEFEF}
\texttt{gemma4-31b} & ZS-CoT & \sympy & $0.988$ & $0.988$ & $0.994$ & $0.0\%$ \\
\midrule
\texttt{gpt-oss-120b} & ZS & \cpmpy & $0.967$ & $0.963$ & $0.985$ & $0.0\%$ \\
\rowcolor[HTML]{EFEFEF}
\texttt{gpt-oss-120b} & ZS & \DIMACS & $0.966$ & $0.962$ & $0.978$ & $0.0\%$ \\
\texttt{gpt-oss-120b} & ZS & \nat & $0.970$ & $0.966$ & $0.985$ & $0.2\%$ \\
\rowcolor[HTML]{EFEFEF}
\texttt{gpt-oss-120b} & ZS & \pysat & $0.951$ & $0.945$ & $0.971$ & $0.0\%$ \\
\texttt{gpt-oss-120b} & ZS & \sympy & $0.960$ & $0.954$ & $0.984$ & $0.0\%$ \\
\rowcolor[HTML]{EFEFEF}
\texttt{gpt-oss-120b} & ZS-CoT & \cpmpy & $0.972$ & $0.968$ & $0.989$ & $0.0\%$ \\
\texttt{gpt-oss-120b} & ZS-CoT & \DIMACS & $0.961$ & $0.957$ & $0.976$ & $0.0\%$ \\
\rowcolor[HTML]{EFEFEF}
\texttt{gpt-oss-120b} & ZS-CoT & \nat & $0.976$ & $0.973$ & $0.988$ & $0.0\%$ \\
\texttt{gpt-oss-120b} & ZS-CoT & \pysat & $0.954$ & $0.948$ & $0.971$ & $0.1\%$ \\
\rowcolor[HTML]{EFEFEF}
\texttt{gpt-oss-120b} & ZS-CoT & \sympy & $0.958$ & $0.953$ & $0.981$ & $0.0\%$ \\
\midrule
\texttt{qwen3-coder-next} & ZS & \cpmpy & $0.632$ & $0.631$ & $0.639$ & $32.4\%$ \\
\rowcolor[HTML]{EFEFEF}
\texttt{qwen3-coder-next} & ZS & \DIMACS & $0.123$ & $0.116$ & $0.156$ & $1.8\%$ \\
\texttt{qwen3-coder-next} & ZS & \nat & $0.756$ & $0.744$ & $0.807$ & $3.7\%$ \\
\rowcolor[HTML]{EFEFEF}
\texttt{qwen3-coder-next} & ZS & \pysat & $0.614$ & $0.610$ & $0.627$ & $31.5\%$ \\
\texttt{qwen3-coder-next} & ZS & \sympy & $0.704$ & $0.701$ & $0.753$ & $13.6\%$ \\
\rowcolor[HTML]{EFEFEF}
\texttt{qwen3-coder-next} & ZS-CoT & \cpmpy & $0.700$ & $0.696$ & $0.720$ & $24.2\%$ \\
\texttt{qwen3-coder-next} & ZS-CoT & \DIMACS & $0.639$ & $0.631$ & $0.702$ & $0.5\%$ \\
\rowcolor[HTML]{EFEFEF}
\texttt{qwen3-coder-next} & ZS-CoT & \nat & $0.859$ & $0.849$ & $0.907$ & $2.0\%$ \\
\texttt{qwen3-coder-next} & ZS-CoT & \pysat & $0.764$ & $0.757$ & $0.791$ & $10.6\%$ \\
\rowcolor[HTML]{EFEFEF}
\texttt{qwen3-coder-next} & ZS-CoT & \sympy & $0.687$ & $0.682$ & $0.716$ & $14.9\%$ \\
\end{longtable}
\endgroup

\textbf{How does auto-formalization performance vary with model size?}
Understanding how model capabilities change with scale is a central question in the study of LLMs. 
Comparing models with different parameters can reveal whether a capability improves with scale and how consistent this relationship is. We therefore study how auto-formalization performance varies with model size.
For each model, we average F1 across the ten configurations formed by two prompting strategies and five output formats. We then fit an ordinary least-squares regression of the mean F1 compared to the LLM parameter count on a log-10 scale. 
The slope measures the expected change in mean F1 associated with a tenfold increase in model size. Our comparison includes eleven models on \autobench, ranging from 8.2B to 117B total parameters, four models on \dcpbench, and three models on \satbench. 

\begin{figure*}[!htbp]
    \centering
    \includegraphics[width=0.80\textwidth]{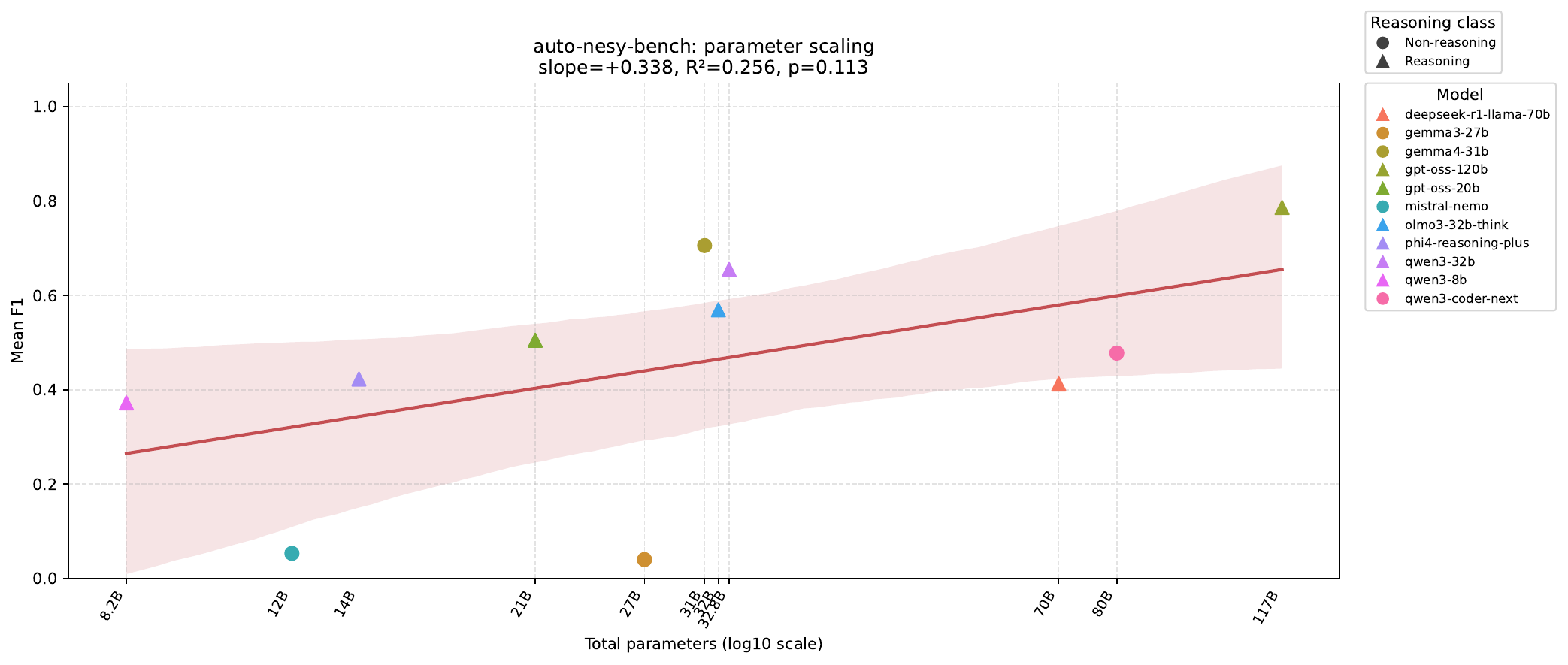}
    \includegraphics[width=0.80\textwidth]{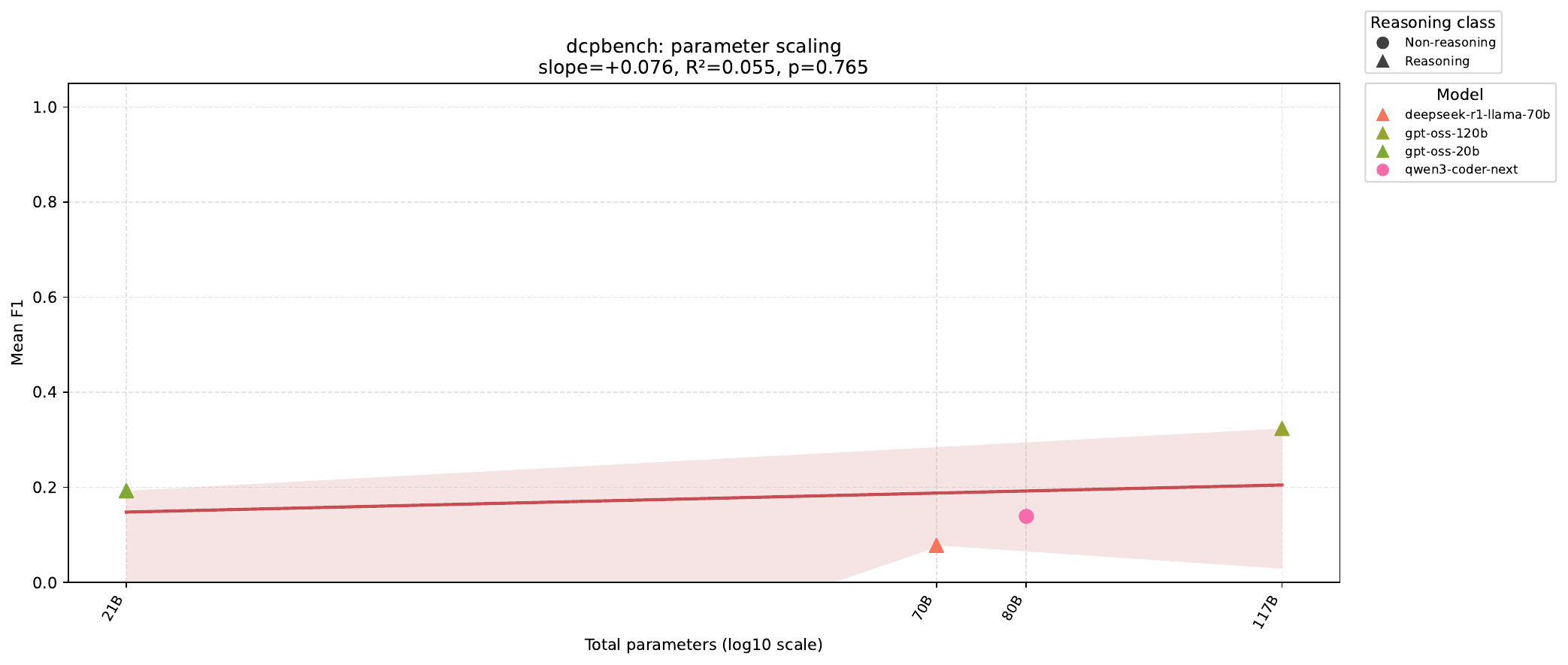}
    \includegraphics[width=0.80\textwidth]{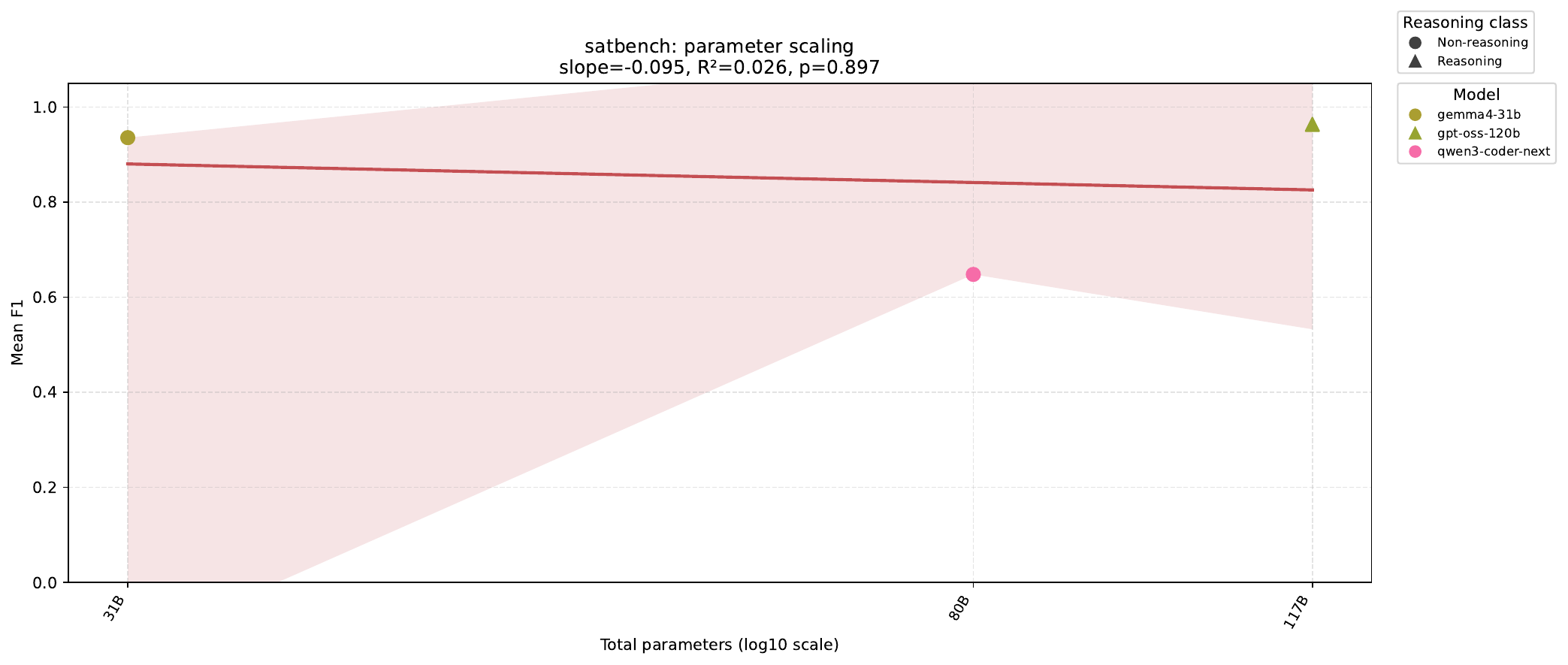}
    \caption{\textbf{The relationship between model size and auto-formalization performance varies across datasets. Larger models tend to perform better on \autobench and \dcpbench, but this trend does not extend to \satbench.} We report mean F1 as a function of total model parameter count on \autobench (top), \dcpbench (middle), and \satbench (bottom). Each point represents a model mean F1 across matched runs. Circles denote non-reasoning models, while triangles denote reasoning models. The red line and shaded region show the ordinary least-squares fit and its 95\% confidence interval, respectively. Parameter count is shown on a base-10 logarithmic scale. None of the fitted associations is statistically significant. 
    }
    \label{fig:model-size}
\end{figure*}

As shown in \cref{fig:model-size}, the connection between auto-formalization performance and model size varies across datasets. On \autobench, the fitted slope is $+0.3380$, corresponding to an estimated F1 increase of $0.3380$ for a tenfold increase in parameter count. The correlation is moderate ($r=0.506$), with model size explaining $25.6\%$ of the between-model variation ($R^2=0.256$); however, the association is not statistically significant ($p=0.113$). With only eleven models spanning different architectures, model families, and training procedures, this result does not establish a general scaling law.
On \dcpbench, the fitted slope is smaller ($+0.0764$), with $R^2=0.055$ and $p=0.765$. This association is also not statistically significant, and the regression is based on only four models; it should therefore be interpreted as a descriptive trend rather than evidence of general scaling. On \satbench, the fitted slope is weakly negative ($-0.0948$), with $r=-0.161$, $R^2=0.026$, and $p=0.897$. The 117B \texttt{gpt-oss-120b} achieves higher F1 than the 80B \texttt{qwen3-coder-next} ($0.963$ vs.\ $0.648$), but the 31B \texttt{gemma4-31b} also substantially outperforms the latter ($0.936$ vs.\ $0.648$). With only three models, this comparison cannot establish a scaling relationship.
Model size also does not fully determine performance. On \autobench, \texttt{gpt-oss-20b} and \texttt{olmo3-32b-think} achieve F1 scores of $0.505$ and $0.569$, respectively, outperforming the larger 70B \texttt{deepseek-r1-llama-70b} ($0.412$) and 80B \texttt{qwen3-coder-next} ($0.478$). Even \texttt{qwen3-8b} achieves an F1 of $0.372$.
Overall, larger models tend to perform better on \autobench and \dcpbench, but this trend does not extend to \satbench, and parameter count explains only part of the observed variation. The deviations from the fitted trends suggest that architecture, training, and other model characteristics also contribute substantially to auto-formalization performance.

\subsubsection{Output-Level Analysis}
\begin{figure*}[!htbp]
    \centering
    \includegraphics[width=0.75\textwidth]{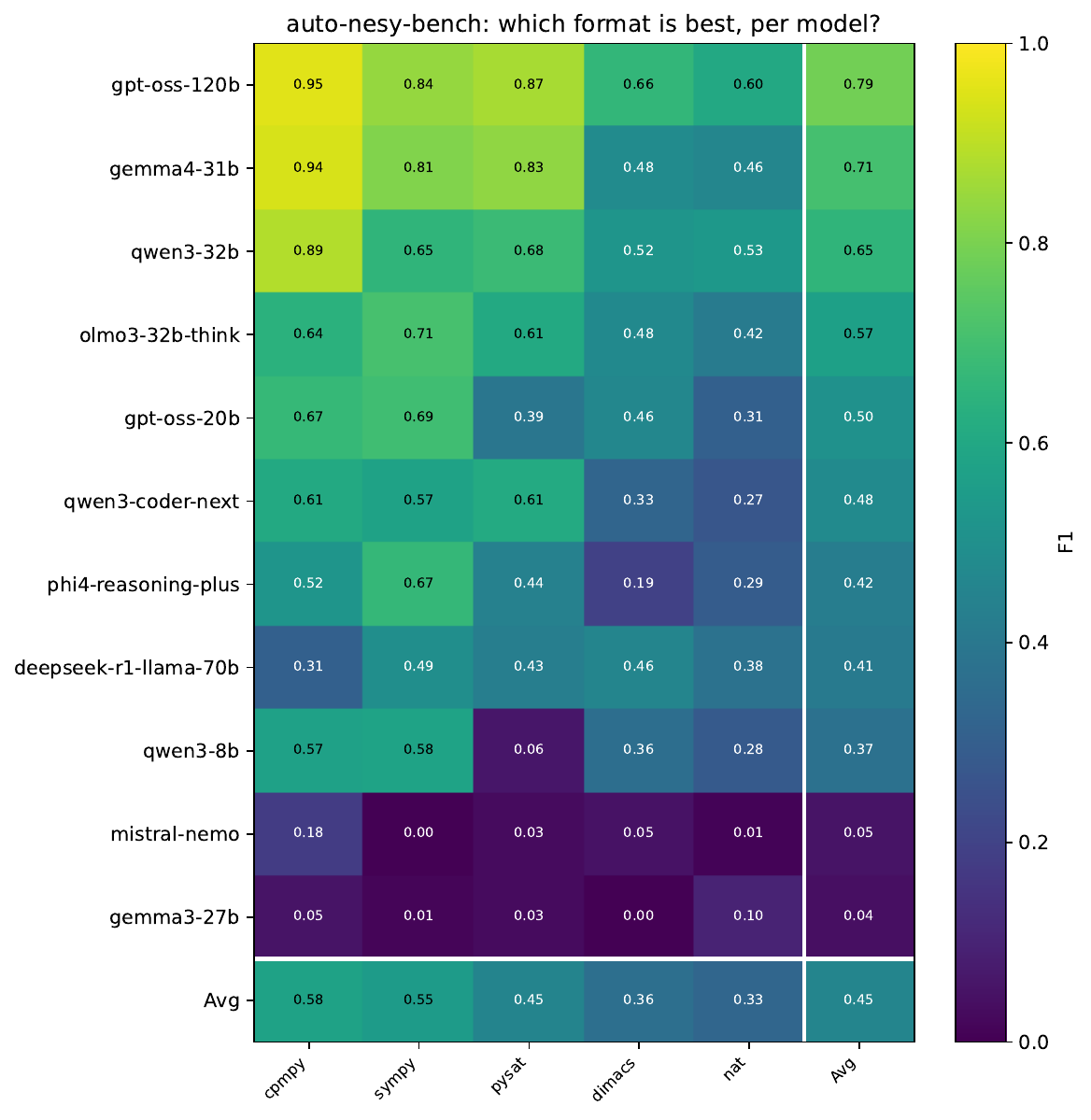}
    \includegraphics[width=0.65\textwidth]{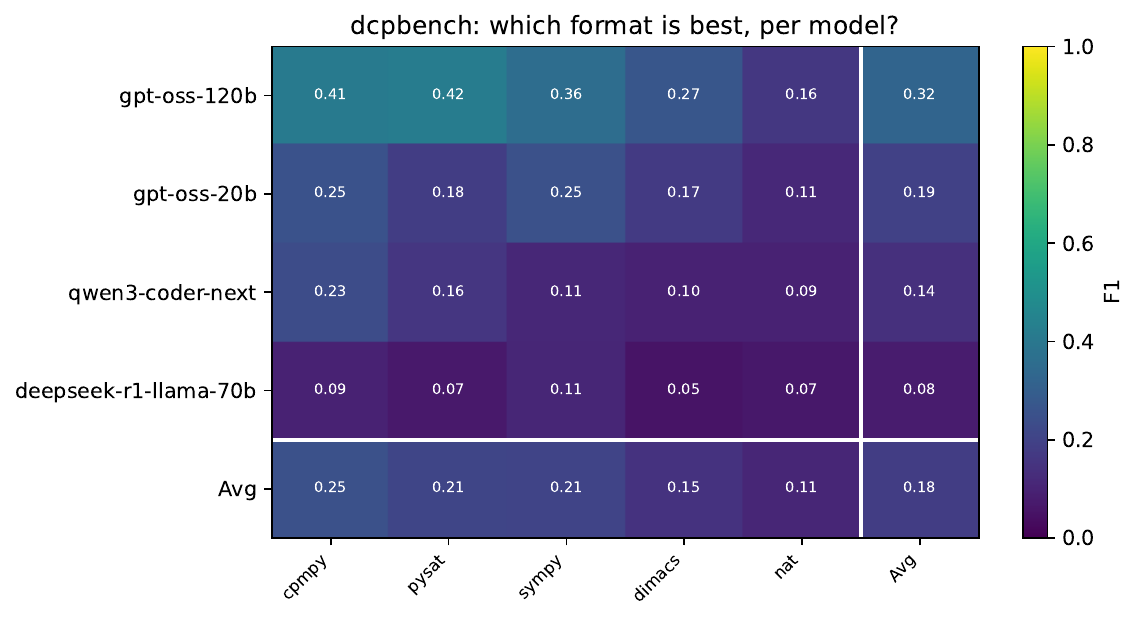}
    \includegraphics[width=0.60\textwidth]{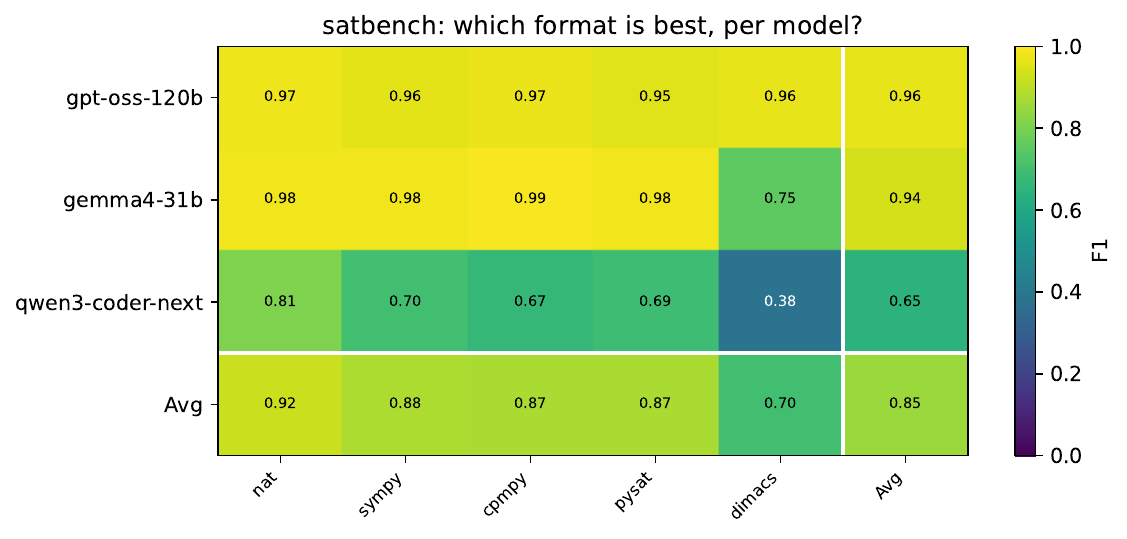}
    \caption{\textbf{Code representations, particularly \cpmpy and \sympy, are generally effective on \autobench and \dcpbench, while \nat performs best on \satbench.} We report mean F1 by model and output format on \autobench (top), \dcpbench (middle), and \satbench (bottom). Each cell reports F1 averaged across the two prompting strategies for a given model-format pair. The rightmost column shows the mean across formats for each model, while the bottom row shows the mean across models for each format. Missing F1 scores are treated as zero.}
    \label{fig:output-format}
\end{figure*}

\textbf{Which output format achieves the best performance?}
The proposed pipeline allows an LLM to generate a CNF in one of five output formats: \DIMACS, \nat, \pysat, \cpmpy, or \sympy code. Each representation is then deterministically compiled into \DIMACS, providing a common representation for evaluation.
The choice of output format substantially affects performance. In the main paper, we report the average performance of each output format across matched runs. Here, we break down these results by model to examine how individual models perform across output formats. \Cref{fig:output-format} shows the F1 score for each model-format pair, averaged over the two prompting strategies (ZS and ZS-CoT).
Output formats are ordered from highest to lowest according to their mean F1 across models. Models are similarly ordered according to their mean F1 across output formats.

Code-based formats perform best on average on \autobench. \cpmpy achieves the highest across-model mean F1 at $0.575$, followed by \sympy at $0.549$. \pysat ranks third at $0.453$, while direct \DIMACS generation reaches $0.364$ and \nat performs worst at $0.331$. Thus, compared with generating the final representation directly, the two strongest higher-level formats improve mean F1 by approximately $0.19$-$0.21$. This suggests that allowing models to generate a high-level symbolic program may make formalization easier than requiring them to construct the low-level \DIMACS encoding directly, potentially because code is more prevalent in LLM training data.

However, the best output format varies across models. The strongest model overall on \autobench, \texttt{gpt-oss-120b}, achieves its highest F1 with \cpmpy ($0.954$), compared with $0.869$ for \pysat, $0.841$ for \sympy, $0.664$ for \DIMACS, and $0.604$ for \nat. In contrast, \sympy performs best for several other competitive models, including \texttt{olmo3-32b-think} ($0.707$), \texttt{gpt-oss-20b} ($0.693$), and \texttt{qwen-8b} ($0.582$). \texttt{qwen-coder-next} performs similarly with \pysat ($0.610$) and \cpmpy ($0.608$). Thus, while \cpmpy and \sympy are the strongest formats on average, choosing the format for a specific model can yield substantial additional gains.

A similar preference for code-based formats emerges on \dcpbench, although the best format varies more substantially across models. Across models, \cpmpy achieves the highest mean F1 ($0.247$), followed by \pysat ($0.207$) and \sympy ($0.206$). At the model level, however, \texttt{deepseek-r1-llama-70b} performs best with \sympy ($0.109$), \texttt{gpt-oss-120b} with \pysat ($0.419$), \texttt{gpt-oss-20b} with \cpmpy ($0.250$), and \texttt{qwen-coder-next} with \cpmpy ($0.233$). No single code-based format is therefore consistently best across models.

The pattern is different on \satbench. Here, \nat is the strongest format on average across models, achieving an F1 of $0.920$. It is also the strongest format for \texttt{gpt-oss-120b} ($0.973$) and \texttt{qwen-coder-next} ($0.808$), while \texttt{gemma4-31b} performs best with \cpmpy ($0.989$), compared with $0.979$ for \nat. Thus, the advantage of code-based representations observed on \autobench and \dcpbench does not transfer uniformly across datasets.

Overall, these results show that no single output format consistently dominates across models and datasets. Higher-level code representations, particularly \cpmpy and \sympy, perform well on \autobench and \dcpbench, whereas \nat is substantially stronger on \satbench. The choice of output format should therefore be viewed as a model- and dataset-dependent factor in formalization performance.

\subsubsection{Syntactic Error Analysis}
Finally, we examine which errors most often prevent the auto-formalization pipeline from producing valid formulas and whether these errors differ across models and output formats.
We focus on syntactic errors, including missing output delimiters, parsing errors, malformed \DIMACS content, program-execution failures, and timeouts. The complete error taxonomy is provided in \cref{app:errors}.

\begin{figure*}[!htbp]
    \centering
    \includegraphics[width=\textwidth]{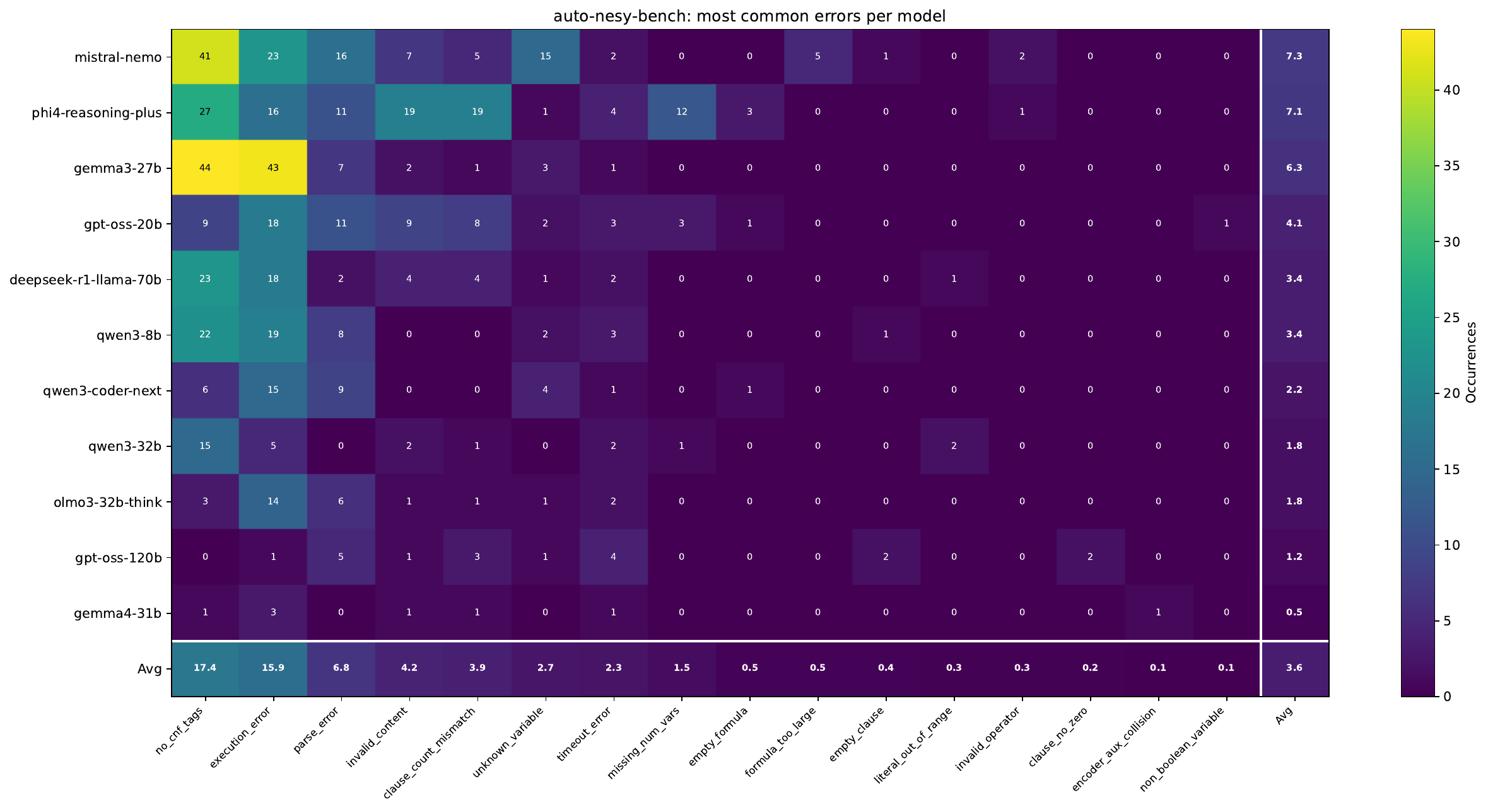}
    \includegraphics[width=\textwidth]{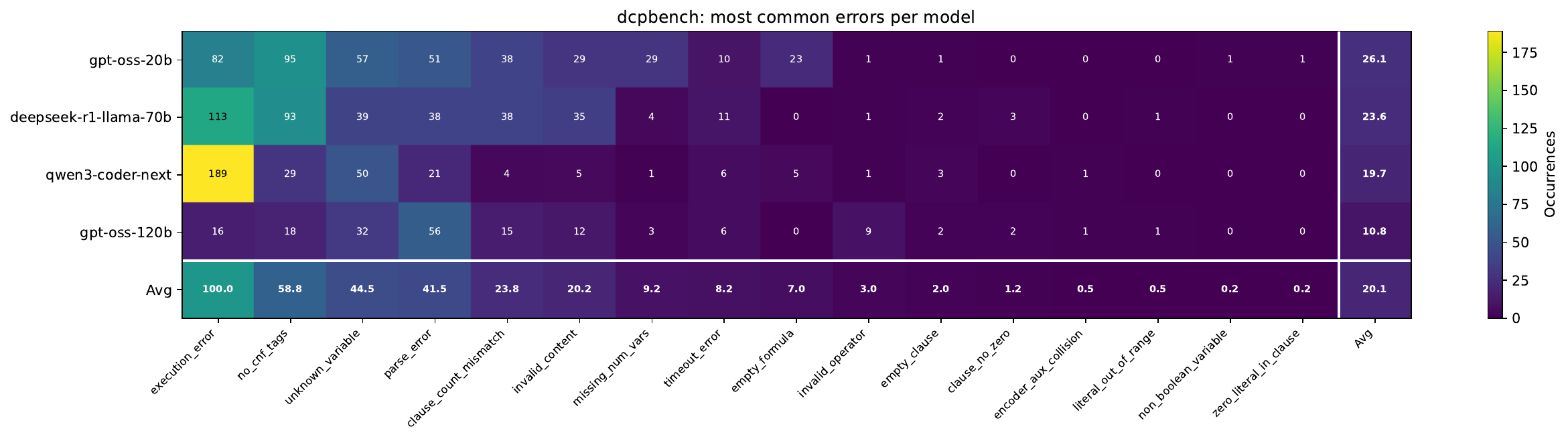}
    \includegraphics[width=\textwidth]{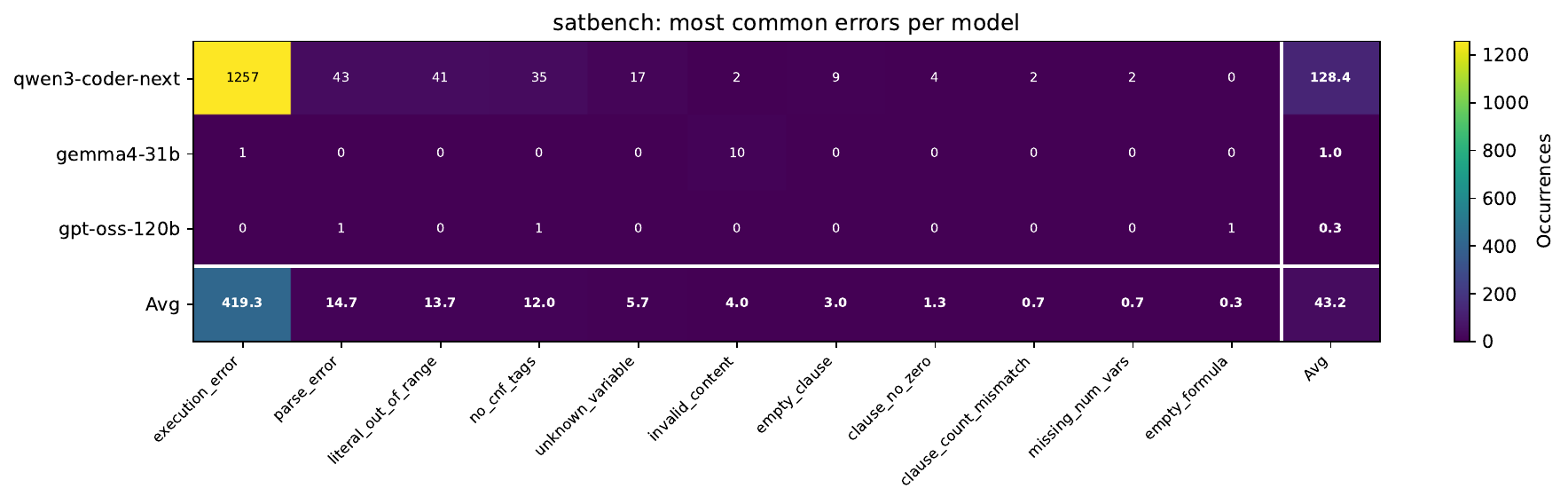}
    \caption{\textbf{Syntactic errors are concentrated in a small number of categories and models.} Occurrences of each error code in the final model outputs on \autobench (top), \dcpbench (middle), and \satbench (bottom), aggregated across output formats and prompting strategies. A single output can contribute to multiple error codes. The rightmost column reports the mean across error types, and the bottom row reports the mean across models.}
    \label{fig:error-per-model}
\end{figure*}

\Cref{fig:error-per-model} shows that errors are concentrated in a small number of categories and models. On \autobench, \texttt{no\_cnf\_tags} is the most common error category, with $17.4$ occurrences, followed by \texttt{execution\_error} ($15.9$), \texttt{parse\_error} ($6.8$), \texttt{invalid\_content} ($4.2$), and \texttt{clause\_count\_mismatch} ($3.9$). \texttt{mistral-nemo} has the most errors overall, with an average of $7.3$ errors per error type. This includes $23$ execution errors and $41$ missing-tag errors. \texttt{phi4-reasoning-plus} follows with an average of $7.1$ errors per type, while \texttt{gemma3-27b} averages $6.3$, with $43$ execution errors and $44$ missing-tag errors. In contrast, \texttt{gemma4-31b} has only $0.5$ errors per type, and \texttt{gpt-oss-120b} has $1.2$, with one execution error and no missing-tag errors. These results show that auto-formalization depends not only on the LLM reasoning ability, but also on its ability to follow formatting instructions and produce outputs that can be parsed and executed correctly.
On \dcpbench, execution errors are the most common category, averaging $100.0$ occurrences per model, followed by missing tags ($58.8$), unknown variables ($44.5$), and parsing errors ($41.5$). \texttt{gpt-oss-20b} has the largest mean count across error types ($26.1$), followed by \texttt{deepseek-r1-llama-70b} ($23.6$), \texttt{qwen3-coder-next} ($19.7$), and \texttt{gpt-oss-120b} ($10.8$).
On \satbench, \texttt{qwen3-coder-next} has the most errors, with $1{,}257$ execution errors and an average of $128.4$ errors per type. In contrast, \texttt{gpt-oss-120b} has only three errors in total, with no execution errors. \texttt{gemma4-31b} has eleven errors: ten invalid-content errors and one execution error.

\begin{figure*}[!htbp]
    \centering
    \includegraphics[width=\textwidth]{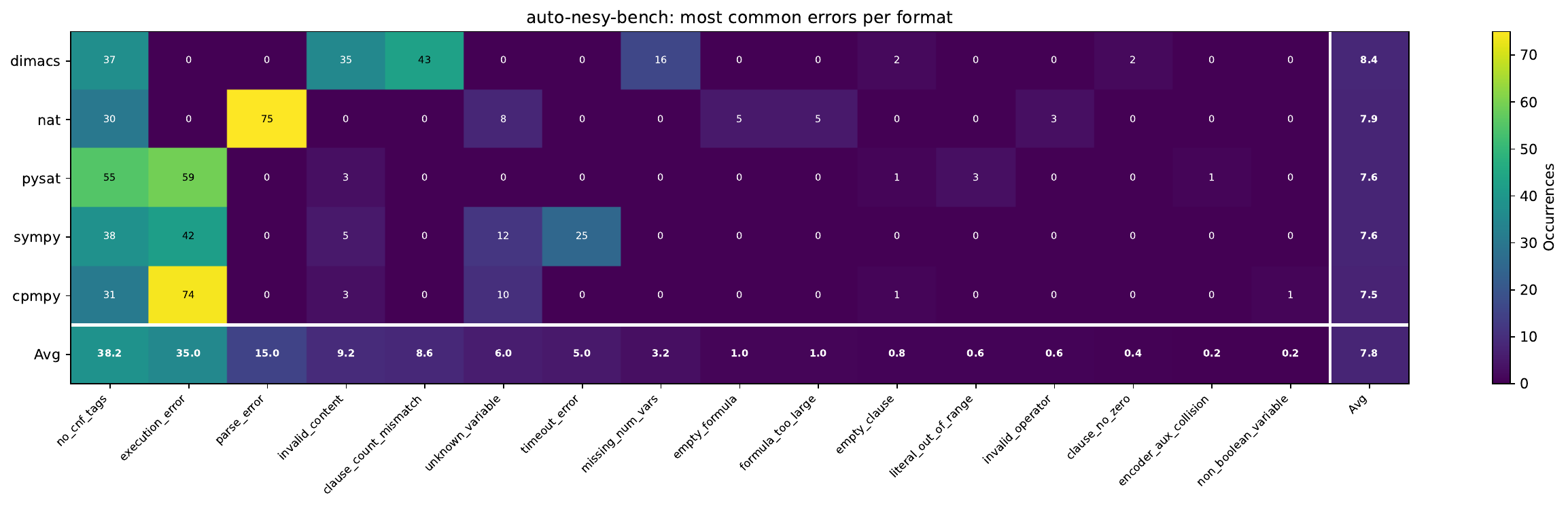}
    \includegraphics[width=\textwidth]{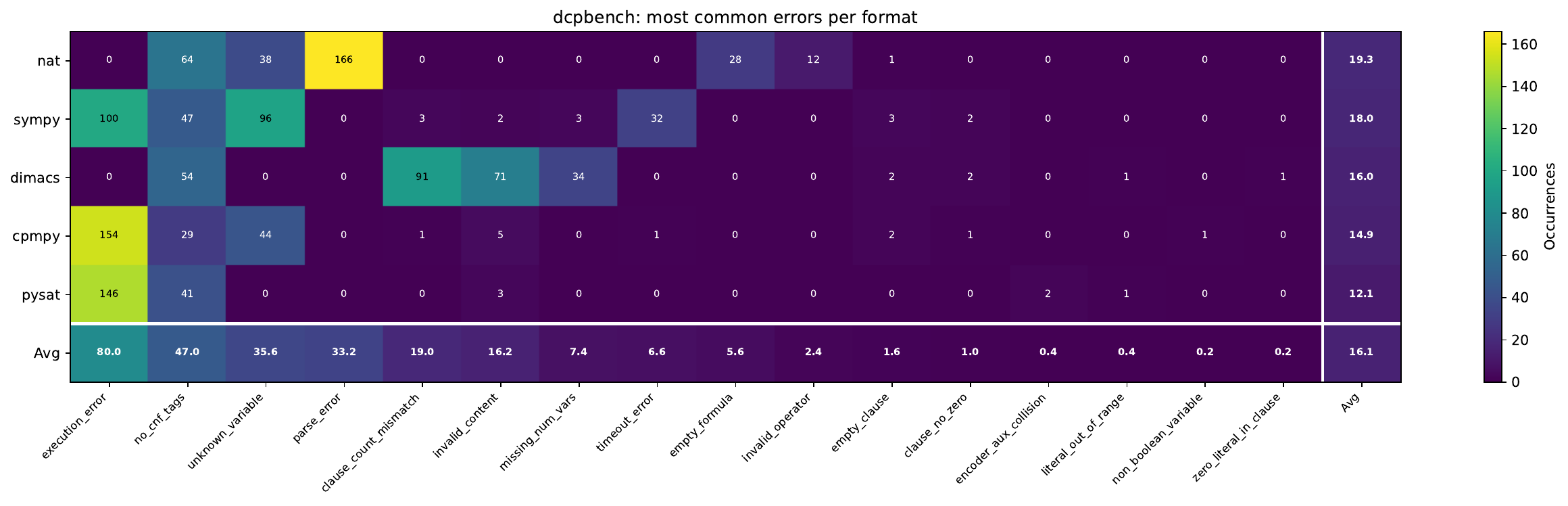}
    \includegraphics[width=\textwidth]{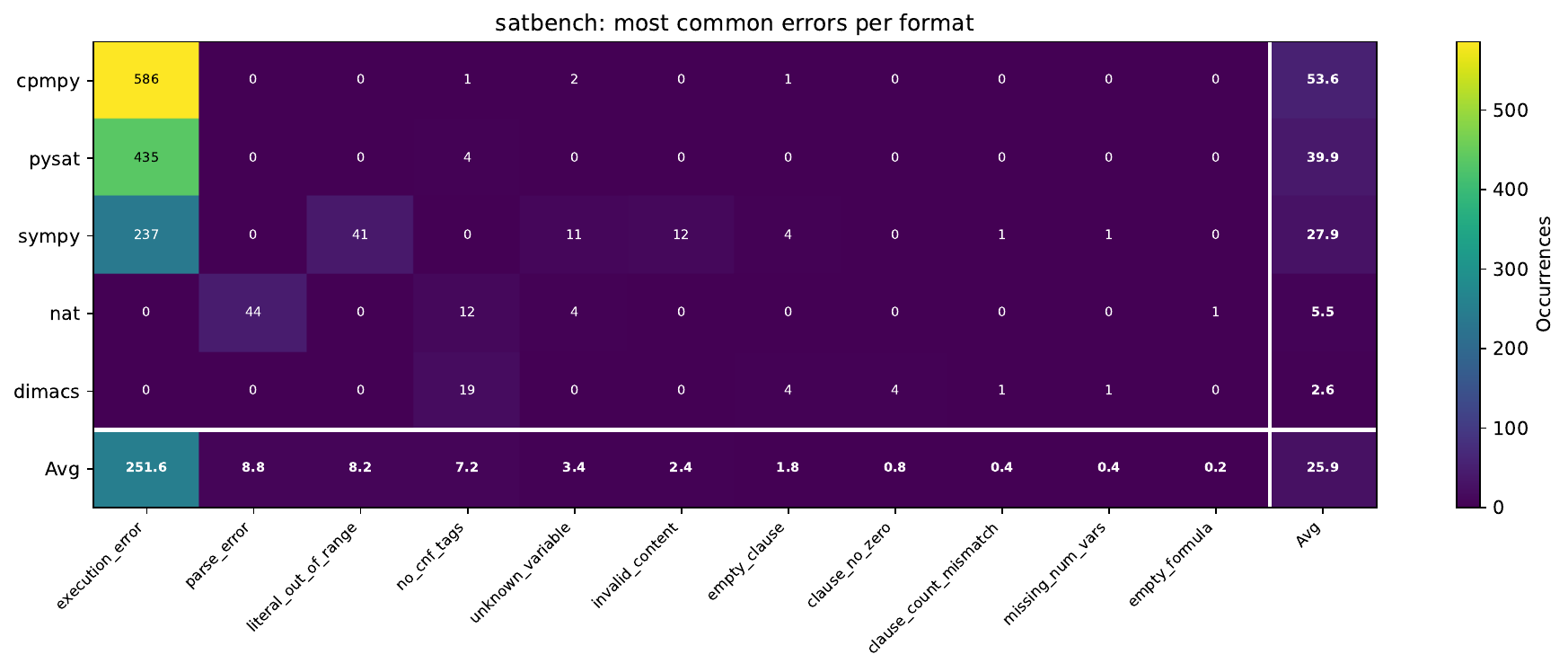}
    \caption{\textbf{Output formats exhibit distinct syntactic errors.} Occurrences of each error code in the final model outputs on \autobench (top), \dcpbench (middle), and \satbench (bottom), aggregated over models and prompting strategies. A single output can contribute to multiple error codes. The rightmost column averages over error types, and the bottom row averages over formats.}
    \label{fig:error-per-format}
\end{figure*}

The format-level analysis in \cref{fig:error-per-format} shows that the types of errors vary substantially across output formats. On \autobench, direct \DIMACS generation has the highest mean count ($8.4$ occurrences per error type), despite having the lowest syntax error rate, as a single output can contribute to multiple error codes. Its main errors are clause-count mismatches ($43$), missing \texttt{<cnf>} tags ($37$), and invalid content ($35$). In contrast, \nat mainly fails during parsing ($75$ errors), followed by missing tags ($30$). Code-based formats largely avoid these errors but introduce execution failures: \pysat, \cpmpy, and \sympy produce $59$, $74$, and $42$ execution errors, respectively. \cpmpy has the lowest mean count overall ($7.5$), while \sympy produces $25$ timeouts, suggesting that symbolic conversion can become a computational bottleneck. Missing \texttt{<cnf>} tags occur across all formats, ranging from $30$ to $55$ instances, indicating a general instruction-following issue.
On \dcpbench, \pysat has the lowest average number of errors ($12.1$), while \DIMACS has a higher average of $16.0$. Clause-count mismatches ($91$) and invalid content ($71$) are still common errors. The \nat format has the highest mean count ($19.3$), including $166$ parsing errors, while \pysat and \cpmpy produce $146$ and $154$ execution errors, respectively. \sympy averages $18.0$, including $100$ execution errors, $96$ unknown-variable errors, and $32$ timeouts.
On \satbench, \DIMACS has the lowest mean count ($2.6$), followed by \nat ($5.5$). Execution errors dominate the code-based formats, with $586$ for \cpmpy, $435$ for \pysat, and $237$ for \sympy; all are produced by \texttt{qwen3-coder-next} except for one \cpmpy execution error from \texttt{gemma4-31b}.
Overall, the types of errors depend on the output format. Higher-level formats reduce manual encoding errors but can cause more errors considering code execution, symbolic conversion, and computational cost.

\clearpage

\subsection{Downstream Results of LLM-Generated Formulas on NeSy Predictors}

In the following tables, we report the full evaluation results for the ground-truth formulas and the formulas generated by \texttt{gpt-oss-120b}, \texttt{olmo3-32b-think}, and \texttt{qwen3-8b}, as discussed in~\cref{sec:results-spl}.

\subsubsection{Downstream Results with Detailed Prompts on \SPL}

\begin{table}[H]
    \centering
    \scriptsize
    \caption{\textbf{\SPL full results with \texttt{olmo3-32b-think} formulas using \cpmpy, CoT and detailed prompt}. The table reports consistency, accuracy, precision, F1, \frel, and \mcr. Mean $\pm$ std over 5 seeds.}
    \scalebox{0.9}{
\begin{tabular}{lrrrrrrr}
\toprule
\textsc{Task} & \textsc{Consistency} $(\uparrow)$ & \textsc{Accuracy} $(\uparrow)$ & \textsc{Precision} $(\uparrow)$ & \textsc{Recall} $(\uparrow)$ & \textsc{F1} $(\uparrow)$ & \frel & \mcr \\
\midrule
\texttt{bdd-oia-2} & $1.00 \pm 0.00$ & $0.76 \pm 0.00$ & $0.75 \pm 0.00$ & $0.75 \pm 0.00$ & $0.75 \pm 0.00$ & $=$ & $1.00$ \\
\rowcolor[HTML]{EFEFEF}
\texttt{cebab} & $1.00 \pm 0.00$ & $0.63 \pm 0.01$ & $0.48 \pm 0.01$ & $0.45 \pm 0.01$ & $0.46 \pm 0.01$ & $=$ & $1.00$ \\
\texttt{cifar10} & $1.00 \pm 0.00$ & $0.94 \pm 0.00$ & $0.81 \pm 0.01$ & $0.81 \pm 0.01$ & $0.81 \pm 0.01$ & $=$ & $1.00$ \\
\rowcolor[HTML]{EFEFEF}
\texttt{cifar100} & $1.00 \pm 0.00$ & $0.99 \pm 0.00$ & $0.25 \pm 0.01$ & $0.31 \pm 0.01$ & $0.26 \pm 0.01$ & $=$ & $1.00$ \\
\texttt{cle4evr} & $1.00 \pm 0.00$ & $1.00 \pm 0.00$ & $0.99 \pm 0.01$ & $0.99 \pm 0.01$ & $0.99 \pm 0.01$ & $=$ & $1.00$ \\
\rowcolor[HTML]{EFEFEF}
\texttt{fashion} & $1.00 \pm 0.00$ & $0.98 \pm 0.00$ & $0.92 \pm 0.00$ & $0.92 \pm 0.01$ & $0.91 \pm 0.01$ & $=$ & $1.00$ \\
\texttt{kand-logic-2} & $1.00 \pm 0.00$ & $0.84 \pm 0.01$ & $0.77 \pm 0.01$ & $0.77 \pm 0.01$ & $0.77 \pm 0.01$ & $=$ & $1.00$ \\
\rowcolor[HTML]{EFEFEF}
\texttt{mn-add} & $1.00 \pm 0.00$ & $0.98 \pm 0.00$ & $0.98 \pm 0.00$ & $0.98 \pm 0.00$ & $0.98 \pm 0.00$ & $=$ & $1.00$ \\
\texttt{mn-mul-bin} & $1.00 \pm 0.00$ & $0.98 \pm 0.00$ & $0.98 \pm 0.00$ & $0.98 \pm 0.00$ & $0.98 \pm 0.00$ & $=$ & $1.00$ \\
\rowcolor[HTML]{EFEFEF}
\texttt{mn-mul} & $1.00 \pm 0.00$ & $0.98 \pm 0.00$ & $0.98 \pm 0.00$ & $0.98 \pm 0.00$ & $0.98 \pm 0.00$ & $=$ & $1.00$ \\
\texttt{sushi} & $1.00 \pm 0.00$ & $0.72 \pm 0.01$ & $0.36 \pm 0.02$ & $0.31 \pm 0.01$ & $0.30 \pm 0.01$ & $=$ & $1.00$ \\
\rowcolor[HTML]{EFEFEF}
\texttt{warcraft} & $1.00 \pm 0.00$ & $0.94 \pm 0.00$ & $0.87 \pm 0.01$ & $0.86 \pm 0.02$ & $0.86 \pm 0.01$ & $\Leftarrow$ & $1.75$ \\
\bottomrule
\end{tabular}
}

    \label{tab:spl-consistency-olmo3}
\end{table}

\begin{table}[H]
    \centering
    \scriptsize
    \caption{\textbf{\SPL full results with \texttt{qwen3-8b} formulas using \cpmpy, CoT and detailed prompt}. The table reports consistency, accuracy, precision, F1, \frel, and \mcr. Mean $\pm$ std over 5 seeds.}
    \scalebox{0.9}{
\begin{tabular}{lrrrrrrr}
\toprule
\textsc{Task} & \textsc{Consistency} $(\uparrow)$ & \textsc{Accuracy} $(\uparrow)$ & \textsc{Precision} $(\uparrow)$ & \textsc{Recall} $(\uparrow)$ & \textsc{F1} $(\uparrow)$ & \frel & \mcr \\
\midrule
\texttt{bdd-oia} & $1.00 \pm 0.00$ & $0.78 \pm 0.00$ & $0.58 \pm 0.01$ & $0.58 \pm 0.01$ & $0.58 \pm 0.01$ & $=$ & $1.00$ \\
\rowcolor[HTML]{EFEFEF}
\texttt{cifar10} & $1.00 \pm 0.00$ & $0.94 \pm 0.00$ & $0.82 \pm 0.01$ & $0.81 \pm 0.01$ & $0.81 \pm 0.01$ & $=$ & $1.00$ \\
\texttt{cifar100} & $1.00 \pm 0.00$ & $0.99 \pm 0.00$ & $0.25 \pm 0.00$ & $0.32 \pm 0.00$ & $0.27 \pm 0.00$ & $=$ & $1.00$ \\
\rowcolor[HTML]{EFEFEF}
\texttt{cle4evr} & $1.00 \pm 0.00$ & $1.00 \pm 0.00$ & $1.00 \pm 0.00$ & $1.00 \pm 0.00$ & $1.00 \pm 0.00$ & $=$ & $1.00$ \\
\texttt{fashion} & $1.00 \pm 0.00$ & $0.98 \pm 0.00$ & $0.92 \pm 0.00$ & $0.92 \pm 0.00$ & $0.92 \pm 0.01$ & $=$ & $1.00$ \\
\rowcolor[HTML]{EFEFEF}
\texttt{kand-logic-2} & $1.00 \pm 0.00$ & $0.80 \pm 0.00$ & $0.71 \pm 0.00$ & $0.70 \pm 0.00$ & $0.70 \pm 0.00$ & $\Rightarrow$ & $0.02$ \\
\texttt{mn-add-bin} & $1.00 \pm 0.00$ & $0.98 \pm 0.00$ & $0.98 \pm 0.00$ & $0.98 \pm 0.00$ & $0.98 \pm 0.00$ & $=$ & $1.00$ \\
\rowcolor[HTML]{EFEFEF}
\texttt{mn-add} & $1.00 \pm 0.00$ & $0.98 \pm 0.00$ & $0.98 \pm 0.00$ & $0.98 \pm 0.00$ & $0.98 \pm 0.00$ & $=$ & $1.00$ \\
\texttt{mn-mul} & $1.00 \pm 0.00$ & $0.98 \pm 0.00$ & $0.98 \pm 0.00$ & $0.98 \pm 0.00$ & $0.98 \pm 0.00$ & $=$ & $1.00$ \\
\rowcolor[HTML]{EFEFEF}
\texttt{sudoku} & $1.00 \pm 0.00$ & $0.92 \pm 0.03$ & $0.85 \pm 0.05$ & $0.84 \pm 0.05$ & $0.84 \pm 0.05$ & $=$ & $1.00$ \\
\texttt{sushi} & $1.00 \pm 0.00$ & $0.72 \pm 0.00$ & $0.36 \pm 0.02$ & $0.31 \pm 0.01$ & $0.30 \pm 0.01$ & $=$ & $1.00$ \\
\bottomrule
\end{tabular}
}

    \label{tab:spl-consistency-qwen3}
\end{table}

\begin{table}[H]
    \centering
    \scriptsize
    \caption{\textbf{\SPL full results with \texttt{gpt-oss-120b} formulas using \cpmpy, CoT and detailed prompt}. The table reports consistency, accuracy, precision, F1, \frel, and \mcr. Mean $\pm$ std over 5 seeds.}
    \scalebox{0.9}{
\begin{tabular}{lrrrrrrr}
\toprule
\textsc{Task} & \textsc{Consistency} $(\uparrow)$ & \textsc{Accuracy} $(\uparrow)$ & \textsc{Precision} $(\uparrow)$ & \textsc{Recall} $(\uparrow)$ & \textsc{F1} $(\uparrow)$ & \frel & \mcr \\
\midrule
\texttt{bdd-oia} & $1.00 \pm 0.00$ & $0.78 \pm 0.00$ & $0.60 \pm 0.01$ & $0.58 \pm 0.01$ & $0.59 \pm 0.01$ & $=$ & $1.00$ \\
\rowcolor[HTML]{EFEFEF}
\texttt{bdd-oia-2} & $1.00 \pm 0.00$ & $0.76 \pm 0.00$ & $0.75 \pm 0.00$ & $0.75 \pm 0.01$ & $0.75 \pm 0.00$ & $=$ & $1.00$ \\
\texttt{cebab} & $1.00 \pm 0.00$ & $0.64 \pm 0.01$ & $0.48 \pm 0.01$ & $0.47 \pm 0.02$ & $0.47 \pm 0.01$ & $=$ & $1.00$ \\
\rowcolor[HTML]{EFEFEF}
\texttt{chx} & $1.00 \pm 0.00$ & $0.64 \pm 0.02$ & $0.59 \pm 0.07$ & $0.47 \pm 0.02$ & $0.48 \pm 0.02$ & $=$ & $1.00$ \\
\texttt{cifar10} & $1.00 \pm 0.00$ & $0.94 \pm 0.00$ & $0.81 \pm 0.01$ & $0.80 \pm 0.01$ & $0.80 \pm 0.01$ & $=$ & $1.00$ \\
\rowcolor[HTML]{EFEFEF}
\texttt{cifar100} & $1.00 \pm 0.00$ & $0.99 \pm 0.00$ & $0.25 \pm 0.00$ & $0.32 \pm 0.01$ & $0.27 \pm 0.00$ & $=$ & $1.00$ \\
\texttt{cle4evr} & $1.00 \pm 0.00$ & $1.00 \pm 0.00$ & $0.99 \pm 0.01$ & $0.99 \pm 0.01$ & $0.99 \pm 0.01$ & $=$ & $1.00$ \\
\rowcolor[HTML]{EFEFEF}
\texttt{fashion} & $1.00 \pm 0.00$ & $0.98 \pm 0.00$ & $0.92 \pm 0.01$ & $0.92 \pm 0.01$ & $0.92 \pm 0.01$ & $=$ & $1.00$ \\
\texttt{kand-logic} & $1.00 \pm 0.00$ & $0.75 \pm 0.00$ & $0.62 \pm 0.00$ & $0.62 \pm 0.00$ & $0.62 \pm 0.00$ & $=$ & $1.00$ \\
\rowcolor[HTML]{EFEFEF}
\texttt{kand-logic-2} & $1.00 \pm 0.00$ & $0.84 \pm 0.00$ & $0.76 \pm 0.01$ & $0.76 \pm 0.01$ & $0.76 \pm 0.01$ & $=$ & $1.00$ \\
\texttt{mn-add-bin} & $1.00 \pm 0.00$ & $0.98 \pm 0.00$ & $0.98 \pm 0.00$ & $0.98 \pm 0.00$ & $0.98 \pm 0.00$ & $=$ & $1.00$ \\
\rowcolor[HTML]{EFEFEF}
\texttt{mn-add} & $1.00 \pm 0.00$ & $0.98 \pm 0.00$ & $0.98 \pm 0.00$ & $0.98 \pm 0.00$ & $0.98 \pm 0.00$ & $=$ & $1.00$ \\
\texttt{mn-mul-bin} & $1.00 \pm 0.00$ & $0.98 \pm 0.00$ & $0.98 \pm 0.00$ & $0.98 \pm 0.00$ & $0.98 \pm 0.00$ & $=$ & $1.00$ \\
\rowcolor[HTML]{EFEFEF}
\texttt{mn-mul} & $1.00 \pm 0.00$ & $0.98 \pm 0.00$ & $0.98 \pm 0.00$ & $0.98 \pm 0.00$ & $0.98 \pm 0.00$ & $=$ & $1.00$ \\
\texttt{road-r} & $1.00 \pm 0.00$ & $0.92 \pm 0.00$ & $0.26 \pm 0.02$ & $0.23 \pm 0.01$ & $0.22 \pm 0.01$ & $\parallel$ & $1.03$ \\
\rowcolor[HTML]{EFEFEF}
\texttt{sudoku} & $1.00 \pm 0.00$ & $0.89 \pm 0.05$ & $0.78 \pm 0.10$ & $0.78 \pm 0.11$ & $0.77 \pm 0.11$ & $=$ & $1.00$ \\
\texttt{sushi} & $1.00 \pm 0.00$ & $0.72 \pm 0.00$ & $0.32 \pm 0.03$ & $0.30 \pm 0.01$ & $0.27 \pm 0.01$ & $=$ & $1.00$ \\
\rowcolor[HTML]{EFEFEF}
\texttt{warcraft} & $1.00 \pm 0.00$ & $0.93 \pm 0.01$ & $0.84 \pm 0.02$ & $0.84 \pm 0.02$ & $0.84 \pm 0.02$ & $\Leftarrow$ & $1.75$ \\
\bottomrule
\end{tabular}
}

    \label{tab:spl-consistency-gptoss}
\end{table}

\begin{table}[H]
    \centering
    \scriptsize
    \caption{\textbf{\SPL full results with ground-truth formulas using \cpmpy, CoT and detailed prompt}. The table reports consistency, accuracy, precision, F1, \frel, and \mcr. Mean $\pm$ std over 5 seeds. As \con and \mcr are defined against ground-truth formulas, they are left empty.}
    \scalebox{0.9}{
\begin{tabular}{lrrrrrrr}
\toprule
\textsc{Task} & \textsc{Consistency} $(\uparrow)$ & \textsc{Accuracy} $(\uparrow)$ & \textsc{Precision} $(\uparrow)$ & \textsc{Recall} $(\uparrow)$ & \textsc{F1} $(\uparrow)$ & \frel & \mcr \\
\midrule
\texttt{bdd-oia} & $-$ & $0.78 \pm 0.00$ & $0.59 \pm 0.00$ & $0.57 \pm 0.01$ & $0.58 \pm 0.00$ & $-$ & $-$ \\
\rowcolor[HTML]{EFEFEF}
\texttt{bdd-oia-2} & $-$ & $0.75 \pm 0.00$ & $0.75 \pm 0.00$ & $0.74 \pm 0.00$ & $0.74 \pm 0.00$ & $-$ & $-$ \\
\texttt{cebab} & $-$ & $0.64 \pm 0.00$ & $0.48 \pm 0.02$ & $0.46 \pm 0.02$ & $0.47 \pm 0.01$ & $-$ & $-$ \\
\rowcolor[HTML]{EFEFEF}
\texttt{chx} & $-$ & $0.68 \pm 0.01$ & $0.61 \pm 0.10$ & $0.52 \pm 0.04$ & $0.54 \pm 0.05$ & $-$ & $-$ \\
\texttt{cifar10} & $-$ & $0.96 \pm 0.00$ & $0.88 \pm 0.00$ & $0.87 \pm 0.01$ & $0.87 \pm 0.01$ & $-$ & $-$ \\
\rowcolor[HTML]{EFEFEF}
\texttt{cifar100} & $-$ & $0.99 \pm 0.00$ & $0.32 \pm 0.00$ & $0.41 \pm 0.00$ & $0.35 \pm 0.00$ & $-$ & $-$ \\
\texttt{cle4evr} & $-$ & $1.00 \pm 0.00$ & $1.00 \pm 0.00$ & $1.00 \pm 0.00$ & $1.00 \pm 0.00$ & $-$ & $-$ \\
\rowcolor[HTML]{EFEFEF}
\texttt{fashion} & $-$ & $0.98 \pm 0.00$ & $0.92 \pm 0.00$ & $0.92 \pm 0.00$ & $0.92 \pm 0.00$ & $-$ & $-$ \\
\texttt{kand-logic} & $-$ & $0.80 \pm 0.00$ & $0.71 \pm 0.01$ & $0.70 \pm 0.01$ & $0.70 \pm 0.01$ & $-$ & $-$ \\
\rowcolor[HTML]{EFEFEF}
\texttt{kand-logic-2} & $-$ & $0.87 \pm 0.00$ & $0.81 \pm 0.01$ & $0.81 \pm 0.01$ & $0.81 \pm 0.01$ & $-$ & $-$ \\
\texttt{mn-add-bin} & $-$ & $0.98 \pm 0.00$ & $0.98 \pm 0.00$ & $0.98 \pm 0.00$ & $0.98 \pm 0.00$ & $-$ & $-$ \\
\rowcolor[HTML]{EFEFEF}
\texttt{mn-add} & $-$ & $0.98 \pm 0.00$ & $0.98 \pm 0.00$ & $0.98 \pm 0.00$ & $0.98 \pm 0.00$ & $-$ & $-$ \\
\texttt{mn-mul-bin} & $-$ & $0.99 \pm 0.00$ & $0.99 \pm 0.00$ & $0.98 \pm 0.00$ & $0.98 \pm 0.00$ & $-$ & $-$ \\
\rowcolor[HTML]{EFEFEF}
\texttt{mn-mul} & $-$ & $0.99 \pm 0.00$ & $0.98 \pm 0.00$ & $0.98 \pm 0.00$ & $0.98 \pm 0.00$ & $-$ & $-$ \\
\texttt{road-r} & $-$ & $0.94 \pm 0.00$ & $0.39 \pm 0.02$ & $0.30 \pm 0.01$ & $0.31 \pm 0.01$ & $-$ & $-$ \\
\rowcolor[HTML]{EFEFEF}
\texttt{sudoku} & $-$ & $0.90 \pm 0.04$ & $0.80 \pm 0.07$ & $0.79 \pm 0.08$ & $0.79 \pm 0.08$ & $-$ & $-$ \\
\texttt{sushi} & $-$ & $0.70 \pm 0.00$ & $0.31 \pm 0.01$ & $0.30 \pm 0.01$ & $0.29 \pm 0.01$ & $-$ & $-$ \\
\rowcolor[HTML]{EFEFEF}
\texttt{warcraft} & $-$ & $0.94 \pm 0.00$ & $0.87 \pm 0.01$ & $0.86 \pm 0.01$ & $0.87 \pm 0.01$ & $-$ & $-$ \\
\bottomrule
\end{tabular}
}

    \label{tab:spl-consistency-gt}
\end{table}

\textbf{Failed attempts}. \Cref{tab:spl-coverage-gaps} accounts for missing results in~\cref{tab:spl} and~\cref{tab:semantic-loss-exp}, all under the \cpmpy+CoT configuration used throughout the \SPL downstream evaluation. 
We distinguish three categories. \emph{Formula degenerate} means a CNF was produced and compiles, but is unusable: either it is unsatisfiable (0 models, confirmed directly via model counting), or it declares variables the encoded constraint never actually uses, \eg \texttt{olmo3-32b-think}/\texttt{mn-add-bin} formula, for instance, omits 4 of its 13 declared variables entirely, \ie (\texttt{sum\_bit1}-\texttt{sum\_bit4}). \emph{No formula produced} means the auto-formalization run for that (model, task, format, prompt) combination never emitted a usable CNF to begin with. \emph{Run failure} means that a valid formula exist, but the run fails due to SDD compilation.
Five of the thirteen errors are caused by incorrect formulas: \texttt{olmo3-32b-think}/\texttt{bdd-oia}, \texttt{olmo3-32b-think}/\texttt{mn-add-bin}, \texttt{qwen3-8b}/\texttt{kand-logic}, \texttt{qwen3-8b}/\texttt{chx}, and \texttt{qwen3-8b}/\texttt{road-r}.

\begin{table}[H]
\centering
\caption{\textbf{Reasons behind the missing results in~\cref{tab:spl}, \cpmpy + CoT.}}
\label{tab:spl-coverage-gaps}
\scriptsize
\scalebox{1.0}{
\begin{tabular}{llll}
\toprule
\textsc{Model} & \textsc{Task} & \textsc{Category} & \textsc{Reason} (\textsc{CPMpy}+\textsc{CoT}) \\
\midrule
\texttt{olmo3-32b-think} & \texttt{bdd-oia}         & formula degenerate & UNSAT (0 models) \\
\rowcolor[HTML]{EFEFEF}
\texttt{olmo3-32b-think} & \texttt{kand-logic}      & no formula produced & no CNF generated \\
\texttt{olmo3-32b-think} & \texttt{sudoku}          & no formula produced & no CNF generated \\
\rowcolor[HTML]{EFEFEF}
\texttt{olmo3-32b-think} & \texttt{chx}             & run failure & CUDA device-side assert \\
\texttt{olmo3-32b-think} & \texttt{road-r}          & run failure & process stalled \\
\rowcolor[HTML]{EFEFEF}
\texttt{olmo3-32b-think} & \texttt{mn-add-bin}  & \textbf{formula degenerate} & \textbf{4 of 13 declared variables are missing} \\
\midrule
\texttt{qwen3-8b} & \texttt{kand-logic}          & formula degenerate & UNSAT (0 models) \\
\rowcolor[HTML]{EFEFEF}
\texttt{qwen3-8b} & \texttt{chx}                 & formula degenerate & formula omits 1 of 9 oracle concept variables \\
\texttt{qwen3-8b} & \texttt{bdd-oia-2}      & no formula produced & no CNF generated \\
\rowcolor[HTML]{EFEFEF}
\texttt{qwen3-8b} & \texttt{cebab}               & no formula produced & no CNF generated \\
\texttt{qwen3-8b} & \texttt{warcraft}       & run failure & oracle/formula output-shape mismatch \\
\rowcolor[HTML]{EFEFEF}
\texttt{qwen3-8b} & \texttt{mn-mul-bin} & run failure & SDD compilation error \\
\texttt{qwen3-8b} & \texttt{road-r}              & \textbf{formula degenerate} & UNSAT (0 models) \\
\bottomrule
\end{tabular}
}
\end{table}

\subsubsection{What Happens When \SPL Is Equipped with LLM-Generated Formulas from Less Detailed Prompts?}
\label{sec:spl-less-detailed}

To study this ablation, we run the same experiment as for \SPL in~\cref{sec:results-spl}, but use a less fine-grained prompt description. This is possible because \autobench is the \textit{only} auto-formalization benchmark that supports this setting. Below, we compare the ground-truth model (which is the same as in~\cref{sec:results-spl}) with the models trained under this setting.

\textbf{Discussion}.
\Cref{tab:spl-results-natural-prompt} shows the \SPL results with \cpmpy output using CoT and non-detailed prompt. In this setting, LLM-generated formulas are still usable in most cases, but they are much less likely to be exactly correct than when using detailed prompts (\cref{tab:spl}).
Only about half of the usable formulas are equal to the ground-truth, \ie \frel is $=$ ($9/16$ for \texttt{gpt-oss-120b}, $7/12$ for \texttt{olmo3-32b-think}, and $6/12$ for \texttt{qwen3-8b}), whereas the remaining ones are mostly \texttt{incomparable} ($\parallel$) or \texttt{strict} ($\Rightarrow$).
As a result, the generated formulas are farther from the ground-truth formula. The average $\Delta$\textsc{F1} and $\Delta$ recall are $-0.12$ and $-0.10$ for \texttt{gpt-oss-120b}, $-0.15$ and $-0.14$ for \texttt{olmo3-32b-think}, and $-0.14$ and $-0.12$ for \texttt{qwen3-8b}. With detailed prompts, these differences are less than $0.03$ and $0.02$, respectively.
This gap is mainly caused by non-equivalent formulas, which reduce F1 by $0.19$--$0.28$ on average. As with detailed prompts, overly strict formulas are the most harmful. On \texttt{kand-logic-2}, the formulas produced by \texttt{gpt-oss-120b} and \texttt{olmo3-32b-think} capture only $2\%$ of the valid assignments (\mcr $=0.02$), reducing F1 by $0.36$ and $0.26$, respectively.
\texttt{incomparable} formulas can be just as harmful. For example, \texttt{gpt-oss-120b} on \texttt{cifar10} loses $0.55$ F1, while \texttt{olmo3-32b-think} loses $0.54$ on \texttt{chx} and $0.43$ on \texttt{mn-add-bin}. In contrast, over-approximations ($\Leftarrow$) have little effect. On \texttt{cifar10}, the formulas produced by \texttt{olmo3-32b-think} and \texttt{qwen3-8b} allow about $19\times$ more assignments than the ground truth, yet reduce F1 by only $0.06$-$0.07$. This suggests that the model can learn to down-weight these spurious output combinations during training.
\texttt{equal} formulas perform almost as well as the ground truth on nearly all tasks. The main exception is \texttt{sudoku}, where F1 drops by $0.23$-$0.49$ even though the formula is equivalent to the ground truth.
Unlike with detailed prompts, predictions are no longer always consistent with the ground-truth constraints. \texttt{incomparable} and \texttt{unknown} formulas can allow assignments that the ground truth rejects, which lowers \SPL consistency. For example, on \texttt{road-r}, consistency drops by $0.12$, $0.18$, and $1.00$ for \texttt{gpt-oss-120b}, \texttt{olmo3-32b-think}, and \texttt{qwen3-8b}, respectively. On \texttt{kand-logic}, it drops by $0.16$ for \texttt{qwen3-8b}. In contrast, \texttt{strict} and \texttt{permissive} formulas never produce inconsistent predictions.
Failures are also more common for the strongest model: \texttt{gpt-oss-120b} fails on $2/18$ tasks, compared with $0/18$ with detailed prompts. \texttt{olmo3-32b-think} and \texttt{qwen3-8b} each fail on $6/18$ tasks. Most failures are caused by degenerate formulas that leave some task variables unconstrained or are unsatisfiable (see \cref{tab:spl-coverage-gaps-natural}).
Overall, these results show that the task specification matters as much as the choice of LLM. Detailed prompts are important for obtaining correct formulas and, consequently, \SPL performance close to that achieved with expert-provided knowledge.

\begin{table}[H]
    \centering
    \scriptsize
     \caption{\textbf{With less-detailed prompts, generated formulas still match expert-provided ones on most tasks, but non-equivalent formulas degrade downstream \SPL performance.} Formulas are generated with \cpmpy + CoT from the non-detailed prompt. We report mean $\Delta$\textsc{F1}/$\Delta$ Recall (\textsc{Rec})/$\Delta$ consistency (\textsc{Cons}) relative to the ground-truth (\textsc{G-T}) formula; standard deviations across $5$ seeds are omitted (see below for the full results). \frel denotes the relation between predicted and ground-truth formulas: \texttt{equal} ($=$), \texttt{strict} ($\Rightarrow$), \texttt{permissive} ($\Leftarrow$), or \texttt{incomparable} ($\parallel$), indicating whether the predicted formula satisfying assignments are equal to, a subset of, a superset of, or neither; \texttt{unknown} marks the two \texttt{kand-logic} formulas, for which the exact comparison, which is based on model counting, did not terminate. \mcr measures the ratio of satisfying assignments of the predicted formula to that of the ground-truth formula. Consistency drops below that of \textsc{G-T} only for \texttt{incomparable} or \texttt{unknown} formulas, which allow predictions the ground truth rejects. Interestingly, this does not happen for \texttt{strict} or \texttt{permissive} formulas. Missing cells indicate that the formula is either missing or was not executable (see~\cref{tab:spl-coverage-gaps-natural} for details).}
    \scalebox{0.8}{
\begin{tabular}{l|ccc|ccccc|ccccc}
\toprule
& \multicolumn{3}{c|}{\textsc{G-T}} & \multicolumn{5}{c|}{\texttt{gpt-oss-120b}} & \multicolumn{5}{c}{\texttt{olmo3-32b-think}} \\
\midrule
\textsc{Task} & \textsc{F1} $(\uparrow)$ & \textsc{Rec} $(\uparrow)$ & \textsc{Cons} $(\uparrow)$ & $\Delta$\textsc{F1} & $\Delta$\textsc{Rec} & $\Delta$\textsc{Cons} & \frel & \mcr & $\Delta$\textsc{F1} & $\Delta$\textsc{Rec} & $\Delta$\textsc{Cons} & \frel & \mcr \\
\midrule
\texttt{bdd-oia} & $0.58$ & $0.57$ & $1.00$ & -- & -- & -- & -- & -- & -- & -- & -- & -- & -- \\
\rowcolor[HTML]{EFEFEF}
\texttt{bdd-oia-2} & $0.74$ & $0.74$ & $1.00$ & $-0.08$ & $-0.08$ & $+0.00$ & $\parallel$ & $1.00$ & -- & -- & -- & -- & -- \\
\texttt{cebab} & $0.47$ & $0.46$ & $1.00$ & -- & -- & -- & -- & -- & -- & -- & -- & -- & -- \\
\rowcolor[HTML]{EFEFEF}
\texttt{chx} & $0.54$ & $0.52$ & $1.00$ & $-0.27$ & $-0.16$ & $+0.00$ & $\Rightarrow$ & $0.94$ & $-0.54$ & $-0.52$ & $-0.35$ & $\parallel$ & $1.00$ \\
\texttt{cifar10} & $0.87$ & $0.87$ & $1.00$ & $-0.55$ & $-0.47$ & $+0.00$ & $\parallel$ & $3.38$ & $-0.07$ & $-0.07$ & $+0.00$ & $\Leftarrow$ & $19.25$ \\
\rowcolor[HTML]{EFEFEF}
\texttt{cifar100} & $0.35$ & $0.41$ & $1.00$ & $-0.08$ & $-0.09$ & $+0.00$ & $=$ & $1.00$ & $-0.09$ & $-0.10$ & $+0.00$ & $=$ & $1.00$ \\
\texttt{cle4evr} & $1.00$ & $1.00$ & $1.00$ & $-0.00$ & $-0.00$ & $+0.00$ & $=$ & $1.00$ & $-0.01$ & $-0.01$ & $+0.00$ & $=$ & $1.00$ \\
\rowcolor[HTML]{EFEFEF}
\texttt{fashion} & $0.92$ & $0.92$ & $1.00$ & $-0.00$ & $+0.00$ & $+0.00$ & $=$ & $1.00$ & $-0.00$ & $-0.00$ & $+0.00$ & $=$ & $1.00$ \\
\texttt{kand-logic} & $0.70$ & $0.70$ & $1.00$ & $-0.09$ & $-0.09$ & $+0.00$ & unknown & -- & -- & -- & -- & -- & -- \\
\rowcolor[HTML]{EFEFEF}
\texttt{kand-logic-2} & $0.81$ & $0.81$ & $1.00$ & $-0.36$ & $-0.32$ & $+0.00$ & $\Rightarrow$ & $0.02$ & $-0.26$ & $-0.26$ & $+0.00$ & $\Rightarrow$ & $0.02$ \\
\texttt{mn-add-bin} & $0.98$ & $0.98$ & $1.00$ & $-0.00$ & $-0.00$ & $+0.00$ & $=$ & $1.00$ & $-0.43$ & $-0.40$ & $-0.01$ & $\parallel$ & $2.56$ \\
\rowcolor[HTML]{EFEFEF}
\texttt{mn-add} & $0.98$ & $0.98$ & $1.00$ & $-0.00$ & $-0.00$ & $+0.00$ & $=$ & $1.00$ & $-0.00$ & $-0.00$ & $+0.00$ & $=$ & $1.00$ \\
\texttt{mn-mul-bin} & $0.98$ & $0.98$ & $1.00$ & $-0.00$ & $-0.01$ & $+0.00$ & $=$ & $1.00$ & -- & -- & -- & -- & -- \\
\rowcolor[HTML]{EFEFEF}
\texttt{mn-mul} & $0.98$ & $0.98$ & $1.00$ & $-0.00$ & $-0.00$ & $+0.00$ & $=$ & $1.00$ & $-0.00$ & $-0.00$ & $+0.00$ & $=$ & $1.00$ \\
\texttt{road-r} & $0.31$ & $0.30$ & $1.00$ & $-0.09$ & $-0.07$ & $-0.12$ & $\parallel$ & $76.64$ & $-0.09$ & $-0.07$ & $-0.18$ & $\parallel$ & $242.22$ \\
\rowcolor[HTML]{EFEFEF}
\texttt{sudoku} & $0.79$ & $0.79$ & $1.00$ & $-0.36$ & $-0.35$ & $+0.00$ & $=$ & $1.00$ & $-0.23$ & $-0.23$ & $+0.00$ & $=$ & $1.00$ \\
\texttt{sushi} & $0.29$ & $0.30$ & $1.00$ & $-0.02$ & $+0.01$ & $+0.00$ & $=$ & $1.00$ & $-0.00$ & $+0.01$ & $+0.00$ & $=$ & $1.00$ \\
\rowcolor[HTML]{EFEFEF}
\texttt{warcraft} & $0.87$ & $0.86$ & $1.00$ & $-0.03$ & $-0.03$ & $+0.00$ & $\Leftarrow$ & $1.75$ & -- & -- & -- & -- & -- \\
\bottomrule
\end{tabular}%
}

\scalebox{0.85}{
\begin{tabular}{l|ccc|ccccc}
\toprule
& \multicolumn{3}{c|}{\textsc{G-T}} & \multicolumn{5}{c}{\texttt{qwen3-8b}} \\
\midrule
\textsc{Task} & \textsc{F1} $(\uparrow)$ & \textsc{Rec} $(\uparrow)$ & \textsc{Cons} $(\uparrow)$ & $\Delta$\textsc{F1} & $\Delta$\textsc{Rec} & $\Delta$\textsc{Cons} & \frel & \mcr \\
\midrule
\texttt{bdd-oia} & $0.58$ & $0.57$ & $1.00$ & -- & -- & -- & -- & -- \\
\rowcolor[HTML]{EFEFEF}
\texttt{bdd-oia-2} & $0.74$ & $0.74$ & $1.00$ & -- & -- & -- & -- & -- \\
\texttt{cebab} & $0.47$ & $0.46$ & $1.00$ & -- & -- & -- & -- & -- \\
\rowcolor[HTML]{EFEFEF}
\texttt{chx} & $0.54$ & $0.52$ & $1.00$ & -- & -- & -- & -- & -- \\
\texttt{cifar10} & $0.87$ & $0.87$ & $1.00$ & $-0.06$ & $-0.07$ & $+0.00$ & $\Leftarrow$ & $19.62$ \\
\rowcolor[HTML]{EFEFEF}
\texttt{cifar100} & $0.35$ & $0.41$ & $1.00$ & $-0.08$ & $-0.09$ & $+0.00$ & $=$ & $1.00$ \\
\texttt{cle4evr} & $1.00$ & $1.00$ & $1.00$ & $-0.00$ & $-0.00$ & $+0.00$ & $=$ & $1.00$ \\
\rowcolor[HTML]{EFEFEF}
\texttt{fashion} & $0.92$ & $0.92$ & $1.00$ & $-0.00$ & $-0.00$ & $+0.00$ & $=$ & $1.00$ \\
\texttt{kand-logic} & $0.70$ & $0.70$ & $1.00$ & $-0.29$ & $-0.26$ & $-0.16$ & unknown & -- \\
\rowcolor[HTML]{EFEFEF}
\texttt{kand-logic-2} & $0.81$ & $0.81$ & $1.00$ & $-0.24$ & $-0.23$ & $-0.15$ & $\parallel$ & $0.19$ \\
\texttt{mn-add-bin} & $0.98$ & $0.98$ & $1.00$ & $+0.00$ & $+0.00$ & $-0.00$ & $\Leftarrow$ & $2.56$ \\
\rowcolor[HTML]{EFEFEF}
\texttt{mn-add} & $0.98$ & $0.98$ & $1.00$ & $-0.00$ & $-0.00$ & $+0.00$ & $=$ & $1.00$ \\
\texttt{mn-mul-bin} & $0.98$ & $0.98$ & $1.00$ & $-0.24$ & $-0.20$ & $-0.01$ & $\parallel$ & $2.56$ \\
\rowcolor[HTML]{EFEFEF}
\texttt{mn-mul} & $0.98$ & $0.98$ & $1.00$ & -- & -- & -- & -- & -- \\
\texttt{road-r} & $0.31$ & $0.30$ & $1.00$ & $-0.28$ & $-0.15$ & $-1.00$ & $\parallel$ & $7.81$ \\
\rowcolor[HTML]{EFEFEF}
\texttt{sudoku} & $0.79$ & $0.79$ & $1.00$ & $-0.49$ & $-0.47$ & $+0.00$ & $=$ & $1.00$ \\
\texttt{sushi} & $0.29$ & $0.30$ & $1.00$ & $+0.01$ & $+0.01$ & $+0.00$ & $=$ & $1.00$ \\
\rowcolor[HTML]{EFEFEF}
\texttt{warcraft} & $0.87$ & $0.86$ & $1.00$ & -- & -- & -- & -- & -- \\
\bottomrule
\end{tabular}
}

    \label{tab:spl-results-natural-prompt}
\end{table}

\begin{table}[H]
    \centering
    \scriptsize
    \caption{\textbf{\SPL results for non detailed prompt using \texttt{gpt-oss-120b} formulas}. The table reports consistency, accuracy, precision, F1, \frel, and \mcr. Mean $\pm$ std over 5 seeds.}
    \scalebox{0.9}{
\begin{tabular}{lrrrrrrr}
\toprule
\textsc{Task} & \textsc{Consistency} $(\uparrow)$ & \textsc{Accuracy} $(\uparrow)$ & \textsc{Precision} $(\uparrow)$ & \textsc{Recall} $(\uparrow)$ & \textsc{F1} $(\uparrow)$ & \frel & \mcr \\
\midrule
\texttt{bdd-oia-2} & $1.00 \pm 0.00$ & $0.70 \pm 0.01$ & $0.71 \pm 0.00$ & $0.66 \pm 0.02$ & $0.67 \pm 0.02$ & $\parallel$ & $1.00$ \\
\rowcolor[HTML]{EFEFEF}
\texttt{chx} & $1.00 \pm 0.00$ & $0.42 \pm 0.03$ & $0.27 \pm 0.07$ & $0.36 \pm 0.01$ & $0.26 \pm 0.03$ & $\Rightarrow$ & $0.94$ \\
\texttt{cifar10} & $1.00 \pm 0.00$ & $0.82 \pm 0.00$ & $0.29 \pm 0.01$ & $0.40 \pm 0.01$ & $0.32 \pm 0.01$ & $\parallel$ & $3.38$ \\
\rowcolor[HTML]{EFEFEF}
\texttt{cifar100} & $1.00 \pm 0.00$ & $0.99 \pm 0.00$ & $0.25 \pm 0.00$ & $0.32 \pm 0.01$ & $0.27 \pm 0.01$ & $=$ & $1.00$ \\
\texttt{cle4evr} & $1.00 \pm 0.00$ & $1.00 \pm 0.00$ & $0.99 \pm 0.01$ & $0.99 \pm 0.01$ & $0.99 \pm 0.01$ & $=$ & $1.00$ \\
\rowcolor[HTML]{EFEFEF}
\texttt{fashion} & $1.00 \pm 0.00$ & $0.98 \pm 0.00$ & $0.92 \pm 0.01$ & $0.92 \pm 0.01$ & $0.92 \pm 0.01$ & $=$ & $1.00$ \\
\texttt{kand-logic} & $1.00 \pm 0.00$ & $0.74 \pm 0.00$ & $0.61 \pm 0.00$ & $0.61 \pm 0.00$ & $0.61 \pm 0.00$ & unknown & -- \\
\rowcolor[HTML]{EFEFEF}
\texttt{kand-logic-2} & $1.00 \pm 0.00$ & $0.66 \pm 0.06$ & $0.45 \pm 0.19$ & $0.49 \pm 0.09$ & $0.45 \pm 0.16$ & $\Rightarrow$ & $0.02$ \\
\texttt{mn-add-bin} & $1.00 \pm 0.00$ & $0.98 \pm 0.00$ & $0.98 \pm 0.00$ & $0.98 \pm 0.00$ & $0.98 \pm 0.00$ & $=$ & $1.00$ \\
\rowcolor[HTML]{EFEFEF}
\texttt{mn-add} & $1.00 \pm 0.00$ & $0.98 \pm 0.00$ & $0.98 \pm 0.00$ & $0.98 \pm 0.00$ & $0.98 \pm 0.00$ & $=$ & $1.00$ \\
\texttt{mn-mul-bin} & $1.00 \pm 0.00$ & $0.98 \pm 0.00$ & $0.98 \pm 0.00$ & $0.98 \pm 0.00$ & $0.98 \pm 0.00$ & $=$ & $1.00$ \\
\rowcolor[HTML]{EFEFEF}
\texttt{mn-mul} & $1.00 \pm 0.00$ & $0.98 \pm 0.00$ & $0.98 \pm 0.00$ & $0.98 \pm 0.00$ & $0.98 \pm 0.00$ & $=$ & $1.00$ \\
\texttt{road-r} & $0.88 \pm 0.03$ & $0.93 \pm 0.00$ & $0.28 \pm 0.02$ & $0.23 \pm 0.01$ & $0.22 \pm 0.01$ & $\parallel$ & $76.64$ \\
\rowcolor[HTML]{EFEFEF}
\texttt{sudoku} & $1.00 \pm 0.00$ & $0.72 \pm 0.04$ & $0.45 \pm 0.07$ & $0.44 \pm 0.07$ & $0.43 \pm 0.07$ & $=$ & $1.00$ \\
\texttt{sushi} & $1.00 \pm 0.00$ & $0.72 \pm 0.00$ & $0.32 \pm 0.04$ & $0.30 \pm 0.01$ & $0.27 \pm 0.01$ & $=$ & $1.00$ \\
\rowcolor[HTML]{EFEFEF}
\texttt{warcraft} & $1.00 \pm 0.00$ & $0.93 \pm 0.01$ & $0.84 \pm 0.02$ & $0.84 \pm 0.01$ & $0.83 \pm 0.01$ & $\Leftarrow$ & $1.75$ \\
\bottomrule
\end{tabular}
}

    \label{tab:spl-natural-consistency-gptoss}
\end{table}

\begin{table}[H]
    \centering
    \scriptsize
    \caption{\textbf{\SPL results for non detailed prompt using \texttt{olmo-32b-think}} formulas. The table reports consistency, accuracy, precision, F1, \frel, and \mcr. Mean $\pm$ std over 5 seeds.}
    \label{tab:spl-natural-consistency-olmo3}
    \scalebox{0.9}{
\begin{tabular}{lrrrrrrr}
\toprule
\textsc{Task} & \textsc{Consistency} $(\uparrow)$ & \textsc{Accuracy} $(\uparrow)$ & \textsc{Precision} $(\uparrow)$ & \textsc{Recall} $(\uparrow)$ & \textsc{F1} $(\uparrow)$ & \frel & \mcr \\
\midrule
\texttt{chx} & $0.65 \pm 0.08$ & $0.00 \pm 0.00$ & $0.00 \pm 0.00$ & $0.00 \pm 0.00$ & $0.00 \pm 0.00$ & $\parallel$ & $1.00$ \\
\rowcolor[HTML]{EFEFEF}
\texttt{cifar10} & $1.00 \pm 0.00$ & $0.94 \pm 0.00$ & $0.81 \pm 0.01$ & $0.80 \pm 0.01$ & $0.80 \pm 0.01$ & $\Leftarrow$ & $19.25$ \\
\texttt{cifar100} & $1.00 \pm 0.00$ & $0.99 \pm 0.00$ & $0.25 \pm 0.01$ & $0.31 \pm 0.01$ & $0.26 \pm 0.01$ & $=$ & $1.00$ \\
\rowcolor[HTML]{EFEFEF}
\texttt{cle4evr} & $1.00 \pm 0.00$ & $0.99 \pm 0.01$ & $0.99 \pm 0.02$ & $0.99 \pm 0.02$ & $0.99 \pm 0.02$ & $=$ & $1.00$ \\
\texttt{fashion} & $1.00 \pm 0.00$ & $0.98 \pm 0.00$ & $0.92 \pm 0.00$ & $0.92 \pm 0.01$ & $0.91 \pm 0.01$ & $=$ & $1.00$ \\
\rowcolor[HTML]{EFEFEF}
\texttt{kand-logic-2} & $1.00 \pm 0.00$ & $0.70 \pm 0.00$ & $0.56 \pm 0.01$ & $0.55 \pm 0.00$ & $0.54 \pm 0.00$ & $\Rightarrow$ & $0.02$ \\
\texttt{mn-add-bin} & $0.99 \pm 0.00$ & $0.53 \pm 0.00$ & $0.65 \pm 0.02$ & $0.58 \pm 0.00$ & $0.55 \pm 0.01$ & $\parallel$ & $2.56$ \\
\rowcolor[HTML]{EFEFEF}
\texttt{mn-add} & $1.00 \pm 0.00$ & $0.98 \pm 0.00$ & $0.98 \pm 0.00$ & $0.98 \pm 0.00$ & $0.98 \pm 0.00$ & $=$ & $1.00$ \\
\texttt{mn-mul} & $1.00 \pm 0.00$ & $0.98 \pm 0.00$ & $0.98 \pm 0.00$ & $0.98 \pm 0.00$ & $0.98 \pm 0.00$ & $=$ & $1.00$ \\
\rowcolor[HTML]{EFEFEF}
\texttt{road-r} & $0.82 \pm 0.03$ & $0.93 \pm 0.00$ & $0.25 \pm 0.01$ & $0.23 \pm 0.01$ & $0.21 \pm 0.01$ & $\parallel$ & $242.22$ \\
\texttt{sudoku} & $1.00 \pm 0.00$ & $0.78 \pm 0.02$ & $0.58 \pm 0.03$ & $0.56 \pm 0.04$ & $0.56 \pm 0.04$ & $=$ & $1.00$ \\
\rowcolor[HTML]{EFEFEF}
\texttt{sushi} & $1.00 \pm 0.00$ & $0.71 \pm 0.00$ & $0.35 \pm 0.03$ & $0.31 \pm 0.01$ & $0.29 \pm 0.02$ & $=$ & $1.00$ \\
\bottomrule
\end{tabular}
}

\end{table}

\begin{table}[H]
    \centering
    \scriptsize
    \caption{\textbf{\SPL results for non detailed prompt using \texttt{qwen3-8b}} formulas. The table reports consistency, accuracy, precision, F1, \frel, and \mcr. Mean $\pm$ std over 5 seeds.}
    \scalebox{0.9}{
\begin{tabular}{lrrrrrrr}
\toprule
\textsc{Task} & \textsc{Consistency} $(\uparrow)$ & \textsc{Accuracy} $(\uparrow)$ & \textsc{Precision} $(\uparrow)$ & \textsc{Recall} $(\uparrow)$ & \textsc{F1} $(\uparrow)$ & \frel & \mcr \\
\midrule
\texttt{cifar10} & $1.00 \pm 0.00$ & $0.94 \pm 0.00$ & $0.82 \pm 0.00$ & $0.81 \pm 0.01$ & $0.81 \pm 0.01$ & $\Leftarrow$ & $19.62$ \\
\rowcolor[HTML]{EFEFEF}
\texttt{cifar100} & $1.00 \pm 0.00$ & $0.99 \pm 0.00$ & $0.25 \pm 0.00$ & $0.32 \pm 0.01$ & $0.27 \pm 0.00$ & $=$ & $1.00$ \\
\texttt{cle4evr} & $1.00 \pm 0.00$ & $1.00 \pm 0.00$ & $1.00 \pm 0.00$ & $1.00 \pm 0.00$ & $1.00 \pm 0.00$ & $=$ & $1.00$ \\
\rowcolor[HTML]{EFEFEF}
\texttt{fashion} & $1.00 \pm 0.00$ & $0.98 \pm 0.00$ & $0.92 \pm 0.00$ & $0.92 \pm 0.00$ & $0.91 \pm 0.00$ & $=$ & $1.00$ \\
\texttt{kand-logic} & $0.84 \pm 0.07$ & $0.62 \pm 0.00$ & $0.47 \pm 0.01$ & $0.44 \pm 0.00$ & $0.41 \pm 0.01$ & unknown & -- \\
\rowcolor[HTML]{EFEFEF}
\texttt{kand-logic-2} & $0.85 \pm 0.04$ & $0.72 \pm 0.01$ & $0.61 \pm 0.01$ & $0.58 \pm 0.02$ & $0.57 \pm 0.02$ & $\parallel$ & $0.19$ \\
\texttt{mn-add-bin} & $1.00 \pm 0.00$ & $0.98 \pm 0.00$ & $0.98 \pm 0.00$ & $0.98 \pm 0.00$ & $0.98 \pm 0.00$ & $\Leftarrow$ & $2.56$ \\
\rowcolor[HTML]{EFEFEF}
\texttt{mn-add} & $1.00 \pm 0.00$ & $0.98 \pm 0.00$ & $0.98 \pm 0.00$ & $0.98 \pm 0.00$ & $0.98 \pm 0.00$ & $=$ & $1.00$ \\
\texttt{mn-mul-bin} & $0.99 \pm 0.01$ & $0.85 \pm 0.00$ & $0.72 \pm 0.01$ & $0.79 \pm 0.00$ & $0.74 \pm 0.01$ & $\parallel$ & $2.56$ \\
\rowcolor[HTML]{EFEFEF}
\texttt{road-r} & $0.00 \pm 0.00$ & $0.80 \pm 0.01$ & $0.02 \pm 0.01$ & $0.15 \pm 0.01$ & $0.03 \pm 0.00$ & $\parallel$ & $7.81$ \\
\texttt{sudoku} & $1.00 \pm 0.00$ & $0.66 \pm 0.02$ & $0.32 \pm 0.04$ & $0.32 \pm 0.03$ & $0.30 \pm 0.04$ & $=$ & $1.00$ \\
\rowcolor[HTML]{EFEFEF}
\texttt{sushi} & $1.00 \pm 0.00$ & $0.72 \pm 0.01$ & $0.36 \pm 0.02$ & $0.31 \pm 0.01$ & $0.30 \pm 0.01$ & $=$ & $1.00$ \\
\bottomrule
\end{tabular}
}

    \label{tab:spl-natural-consistency-qwen3}
\end{table}

\textbf{Failed attempts (non-detailed prompts)}. \Cref{tab:spl-coverage-gaps-natural} accounts for the missing results in~\cref{tab:spl-results-natural-prompt}, using the same \cpmpy+CoT configuration but with non-detailed prompts. We use the same three categories as for detailed prompts (\cref{tab:spl-coverage-gaps}).
Most errors are now \emph{formula-degenerate} ($10$ out of $14$). In \texttt{bdd-oia}, all three models omit part of the scene description: \texttt{gpt-oss-120b} never uses $5$ of the $25$ variables (\texttt{follow}, \texttt{left\_follow}, \texttt{no\_left\_lane}, \texttt{right\_follow}, \texttt{no\_right\_lane}), while \texttt{olmo3-32b-think} and \texttt{qwen3-8b} omit $6$. On \texttt{bdd-oia-2}, both smaller models also drop the \texttt{follow} concept. The \texttt{qwen3-8b} formula for \texttt{mn-mul} omits $45$ of its $102$ variables, corresponding exactly to the product classes that no pair of digits can produce (\eg \texttt{p\_11}, \texttt{p\_13}, and \texttt{p\_17}). Rather than explicitly constraining these variables to false, the formula leaves them unconstrained. Two formulas are unsatisfiable: \texttt{olmo3-32b-think}/\texttt{warcraft} and \texttt{qwen3-8b}/\texttt{cebab}. Finally, the \texttt{qwen3-8b} formula for \texttt{chx} admits only a single assignment, leaving no parameters in the compiled circuit for \SPL to learn.
The remaining four gaps are \emph{no formula produced}: \texttt{cebab} for \texttt{gpt-oss-120b} and \texttt{olmo3-32b-think}, \texttt{kand-logic} and \texttt{mn-mul-bin} for \texttt{olmo3-32b-think}, and \texttt{warcraft} for \texttt{qwen3-8b}. Unlike with detailed prompts, there are no \emph{run failures}: every usable formula is trained successfully.
With detailed prompts, \texttt{gpt-oss-120b} produced a usable formula for every task. Without them, even the strongest model fails on $2/18$ tasks, and failures for all models are more often caused by incorrect formulas than by pipeline issues. This suggests that, without a detailed specification, LLMs tend to under-specify the constraints rather than fail to produce a formula altogether.

\begin{table}[H]
\centering
\caption{\textbf{Missing results in~\cref{tab:spl-results-natural-prompt}, \cpmpy + CoT with non-detailed prompts.}}
\label{tab:spl-coverage-gaps-natural}
\scriptsize
\scalebox{1.0}{
\begin{tabular}{llll}
\toprule
\textsc{Model} & \textsc{Task} & \textsc{Category} & \textsc{Reason} (\textsc{CPMpy}+\textsc{CoT}, non-detailed prompt) \\
\midrule
\texttt{gpt-oss-120b} & \texttt{bdd-oia}      & formula degenerate  & 5 of 25 task variables are missing \\
\rowcolor[HTML]{EFEFEF}
\texttt{gpt-oss-120b} & \texttt{cebab}        & no formula produced & no CNF generated \\
\midrule
\texttt{olmo3-32b-think} & \texttt{bdd-oia}    & formula degenerate  & 6 of 25 task variables are missing \\
\rowcolor[HTML]{EFEFEF}
\texttt{olmo3-32b-think} & \texttt{bdd-oia-2}  & formula degenerate  & 1 of 11 task variables is missing \\
\texttt{olmo3-32b-think} & \texttt{warcraft}   & formula degenerate  & UNSAT (0 models) \\
\rowcolor[HTML]{EFEFEF}
\texttt{olmo3-32b-think} & \texttt{cebab}      & no formula produced & no CNF generated \\
\texttt{olmo3-32b-think} & \texttt{kand-logic} & no formula produced & no CNF generated \\
\rowcolor[HTML]{EFEFEF}
\texttt{olmo3-32b-think} & \texttt{mn-mul-bin} & no formula produced & no CNF generated \\
\midrule
\texttt{qwen3-8b} & \texttt{bdd-oia}     & formula degenerate  & 6 of 25 task variables are missing \\
\rowcolor[HTML]{EFEFEF}
\texttt{qwen3-8b} & \texttt{bdd-oia-2}   & formula degenerate  & 1 of 11 task variables is missing \\
\texttt{qwen3-8b} & \texttt{cebab}       & formula degenerate  & UNSAT (0 models) \\
\rowcolor[HTML]{EFEFEF}
\texttt{qwen3-8b} & \texttt{mn-mul}      & formula degenerate  & 45 of 102 task variables are missing \\
\texttt{qwen3-8b} & \texttt{chx}         & formula degenerate  & no trainable circuit parameters \\
\rowcolor[HTML]{EFEFEF}
\texttt{qwen3-8b} & \texttt{warcraft}    & no formula produced & no CNF generated \\
\bottomrule
\end{tabular}
}
\end{table}

\subsubsection{What if the Constraints are Soft? Semantic Loss Downstream Results with Detailed Prompts}
\label{sec:semantic-loss}

In addition to the \SPL results, we further analyze the downstream performance under a different NeSy approach based on the \textit{semantic loss} (\SL)~\citep{xu2018semantic}, which incorporates logical constraints as a penalty loss rather than explicitly enforcing them during prediction. Specifically, we replicate the same \SPL analysis (\cpmpy+CoT and detailed prompt in~\cref{sec:results-spl}) using the \SL.

\textbf{Semantic Loss}. Given an input $\vx$ and an unconstrained predictive distribution $p_\theta(\vy\mid\vx)$, \SL~\citep{xu2018semantic} incorporates a logical constraint $\BK$ by maximizing the probability mass assigned to its satisfying assignments. Assuming conditional independence of the variables given the model outputs, it is defined as
\begin{equation}
\label{eq:semantic-loss}
\mathcal{L}_{\mathrm{SL}}(\BK,p_{\theta}) = -\log \sum_{\vy\models\BK} \prod_{i=1}^{n} p_\theta(Y_i\mid\vx),
\end{equation}
where $n$ is the number of primary Boolean output variables. For a Boolean variable, $p_\theta(Y_i\mid\vx)$ is a shorthand for $p_\theta(Y_i=1\mid\vx)$ if $y_i=1$ and $1-p_\theta(Y_i=1\mid\vx)$ otherwise. Thus, the sum computes the total probability assigned to all satisfying assignments, and the negative logarithm penalizes the model when this probability is low. The sum in~\cref{eq:semantic-loss} is exactly the \emph{weighted model count} (WMC)~\citep{chavira2008wmc} of $\BK$ under the weights induced by $p_\theta(\vy\mid\vx)$.

\textbf{Implementation}. For these experiments, we reuse the same SDDs compiled for the \SPL experiments. At training time, the SDD is loaded using \texttt{KLay}~\citep{maene2025klay}, which makes the WMC, and hence of $\mathcal{L}_{\mathrm{SL}}$, faster.
Unlike \SPL, which explicitly incorporates logical constraints into the prediction mechanism, \SL treats $\BK$ as a \emph{soft constraint} through an additional differentiable loss term. In our setting, it is combined with the downstream task objective as
\begin{equation}
\mathcal{L}_{\mathrm{total}}(\BK,p_{\theta}) = \mathcal{L}_{\mathrm{task}}(p_\theta) +
\lambda\mathcal{L}_{\mathrm{SL}}(\BK,p_{\theta}),
\end{equation}
where $\lambda$ controls the strength of the logical constraint. As all the benchmarks in \autobench are classification tasks, $\mathcal{L}_{\mathrm{task}}$ corresponds to the negative log-likelihood, while for all \SL experiments $\lambda$ is set to $1$.

\textbf{Architectures}. \SL uses the same per-task backbones as \SPL. We only replace the \SPL head, which outputs the parameters of the circuit, with a classification head: an MLP with three hidden layers with ReLU activations, whose width is set to the \SPL head width tuned for that task, and one output per variable of the CNF.
Each output is passed through a sigmoid, so that $p_\theta$ factorizes into independent Bernoulli distributions over the variables; this is the distribution on which $\mathcal{L}_{\mathrm{SL}}$ computes the probability of satisfying $\BK$, with the positive and negative literal weights given by $\log p_\theta$ and $\log(1-p_\theta)$.
At test time, each variable is predicted by thresholding its probability at $0.5$; for tasks with a single multi-class output, we predict the class whose output-bit pattern is most likely. 

\textbf{Discussion}. \Cref{tab:semantic-loss-exp} reports the \SL experiments in the same setting as \cref{sec:results-spl}, \ie using the formulas generated by the LLMs with detailed prompts, chain-of-thought, and \cpmpy as output format. Without \SL, the unconstrained network is often inconsistent with the ground-truth formula: its consistency is below $0.70$ on seven of the $18$ tasks, ranging from $0.06$ on \texttt{sushi} and $0.11$ on \texttt{sudoku} to $0.86$ on \texttt{warcraft}, whereas simpler tasks such as \texttt{cle4evr}, \texttt{fashion}, and the \texttt{mn} datasets are already almost fully consistent ($\geq 0.98$).
Adding the \SL with an equivalent formula ($=$) never decreases consistency and increases it on all tasks that leave room for improvement, most notably on \texttt{sushi}, where it rises from $0.06$ to $0.88$-$0.97$, and on \texttt{road-r} and \texttt{cebab} ($+0.11$ to $+0.17$).
Unlike \SPL (see \cref{sec:results-spl}), however, whose predictions satisfy the formula by construction, \SL only encourages consistency during training, and it rarely reaches $1$: on the hardest tasks it remains far from it (\texttt{kand-logic} $0.21$-$0.22$, \texttt{sudoku} $0.23$-$0.42$).
Downstream performance, for \texttt{equal} ($=$) formulas, is almost unaffected by the semantic loss (mean $\Delta$F1 between $-0.00$ and $+0.02$ across LLMs), with gains concentrated on the most constrained tasks (\texttt{sudoku}, $+0.11$ to $+0.15$ F1; \texttt{mn-mul-bin}, up to $+0.08$).
The only formula that reduces consistency is the one generated by \texttt{qwen3-8b} for \texttt{kand-logic-2} ($-0.23$ \texttt{Cons}, $-0.05$ F1). This formula is too \texttt{strict} ($\Rightarrow$): it admits only $2\%$ of the assignments allowed by the ground truth (\mcr $=0.02$), so it rules out most of the labels seen during training.
Interestingly, as for \SPL, \texttt{permissive} formulas ($\Leftarrow$) are much less harmful: on \texttt{warcraft}, the formulas of \texttt{gpt-oss-120b} and \texttt{olmo3-32b-think} still improve consistency by $+0.08$ and $+0.01$, respectively. Probably the classification loss also contributes by assigning low probability mass on invalid combinations.

\begin{table}[H]
    \scriptsize
    \centering
    \caption{\textbf{Downstream \SL performance using generated formulas, when available, is comparable to that of expert-provided formulas on \autobench. Semantic loss mainly improves consistency in almost all cases compared with an unconstrained baseline.}
    The top table reports \SL with the ground-truth (\textsc{G-T}) and \texttt{gpt-oss-120b} formulas, the bottom one with the \texttt{olmo3-32b-think} and \texttt{qwen3-8b} formulas against the unconstrained baseline.
    In both, the first group (Unconstrained) is the same backbone and classifier trained without the semantic loss, for which we report F1, Recall (\textsc{Rec}) and consistency (\textsc{Cons}, the fraction of test predictions that satisfy the ground-truth formula); every other group reports $\Delta$\textsc{F1}/$\Delta$\textsc{Rec}/$\Delta$\textsc{Cons} of \SL relative to Unconstrained (positive is better).
    Standard deviations across $5$ seeds are omitted ($\leq 0.02$ for F1 and \textsc{Rec} on all tasks except \texttt{sudoku}, \texttt{warcraft}, \texttt{mn-mul-bin} and \texttt{chx}); the full results, with standard deviations and all metrics, are attached below for reference.
    \frel denotes the relationship between the predicted and ground-truth formulas: \texttt{equal} ($=$), \texttt{strict} ($\Rightarrow$), \texttt{permissive} ($\Leftarrow$), or \texttt{incomparable} ($\parallel$), indicating whether the predicted formula satisfying assignments are equal to, a subset of, a superset of, or neither relative to the ground truth.
    \mcr measures the ratio of the number of satisfying assignments of the predicted formula to that of the ground-truth formula.
    Missing cells indicate the errors in the formula generation, formatting or during inference. Details about the missing results is provided in~\cref{tab:spl-coverage-gaps}.}

    \scalebox{0.8}{
\begin{tabular}{l|ccc|ccccc|ccccc}
\toprule
& \multicolumn{3}{c|}{\textsc{Unconstrained}} & \multicolumn{5}{c|}{\textsc{G-T}} & \multicolumn{5}{c}{\texttt{gpt-oss-120b}} \\
\midrule
\textsc{Task} & \textsc{F1} $(\uparrow)$ & \textsc{Rec} $(\uparrow)$ & \textsc{Cons} $(\uparrow)$ & $\Delta$\textsc{F1} & $\Delta$\textsc{Rec} & $\Delta$\textsc{Cons} & \frel & \mcr & $\Delta$\textsc{F1} & $\Delta$\textsc{Rec} & $\Delta$\textsc{Cons} & \frel & \mcr \\
\midrule
\texttt{bdd-oia} & $0.58$ & $0.58$ & $0.90$ & $-0.00$ & $+0.01$ & $+0.02$ & $=$ & $1.00$ & $+0.01$ & $+0.02$ & $+0.01$ & $=$ & $1.00$ \\
\rowcolor[HTML]{EFEFEF}
\texttt{bdd-oia-2} & $0.73$ & $0.71$ & $0.92$ & $-0.00$ & $+0.00$ & $+0.02$ & $=$ & $1.00$ & $+0.00$ & $+0.01$ & $+0.00$ & $=$ & $1.00$ \\
\texttt{cebab} & $0.46$ & $0.46$ & $0.63$ & $+0.00$ & $+0.00$ & $+0.13$ & $=$ & $1.00$ & $+0.01$ & $+0.01$ & $+0.17$ & $=$ & $1.00$ \\
\rowcolor[HTML]{EFEFEF}
\texttt{chx} & $0.57$ & $0.53$ & $0.91$ & $-0.00$ & $+0.00$ & $+0.07$ & $=$ & $1.00$ & $-0.02$ & $-0.02$ & $+0.05$ & $=$ & $1.00$ \\
\texttt{cifar10} & $0.87$ & $0.86$ & $0.94$ & $-0.02$ & $-0.01$ & $+0.05$ & $=$ & $1.00$ & $-0.02$ & $-0.01$ & $+0.05$ & $=$ & $1.00$ \\
\rowcolor[HTML]{EFEFEF}
\texttt{cifar100} & $0.52$ & $0.45$ & $0.63$ & $+0.01$ & $+0.03$ & $+0.07$ & $=$ & $1.00$ & $+0.01$ & $+0.02$ & $+0.09$ & $=$ & $1.00$ \\
\texttt{cle4evr} & $1.00$ & $1.00$ & $1.00$ & $+0.00$ & $+0.00$ & $+0.00$ & $=$ & $1.00$ & $+0.00$ & $+0.00$ & $+0.00$ & $=$ & $1.00$ \\
\rowcolor[HTML]{EFEFEF}
\texttt{fashion} & $0.92$ & $0.92$ & $0.99$ & $+0.00$ & $+0.00$ & $+0.00$ & $=$ & $1.00$ & $+0.00$ & $+0.00$ & $+0.00$ & $=$ & $1.00$ \\
\texttt{kand-logic} & $0.72$ & $0.71$ & $0.19$ & $+0.00$ & $+0.00$ & $+0.02$ & $=$ & $1.00$ & $+0.01$ & $+0.02$ & $+0.02$ & $=$ & $1.00$ \\
\rowcolor[HTML]{EFEFEF}
\texttt{kand-logic-2} & $0.79$ & $0.80$ & $0.66$ & $+0.00$ & $+0.00$ & $+0.04$ & $=$ & $1.00$ & $+0.00$ & $+0.00$ & $+0.08$ & $=$ & $1.00$ \\
\texttt{mn-add-bin} & $0.98$ & $0.98$ & $0.98$ & $+0.00$ & $+0.00$ & $+0.01$ & $=$ & $1.00$ & $+0.00$ & $+0.00$ & $+0.01$ & $=$ & $1.00$ \\
\rowcolor[HTML]{EFEFEF}
\texttt{mn-add} & $0.98$ & $0.98$ & $0.99$ & $+0.00$ & $+0.00$ & $+0.00$ & $=$ & $1.00$ & $-0.01$ & $-0.01$ & $-0.00$ & $=$ & $1.00$ \\
\texttt{mn-mul-bin} & $0.71$ & $0.71$ & $0.99$ & $+0.03$ & $+0.03$ & $+0.00$ & $=$ & $1.00$ & $+0.08$ & $+0.08$ & $+0.00$ & $=$ & $1.00$ \\
\rowcolor[HTML]{EFEFEF}
\texttt{mn-mul} & $0.98$ & $0.98$ & $0.99$ & $+0.00$ & $+0.00$ & $+0.00$ & $=$ & $1.00$ & $+0.00$ & $+0.00$ & $+0.01$ & $=$ & $1.00$ \\
\texttt{road-r} & $0.27$ & $0.25$ & $0.65$ & $+0.01$ & $+0.02$ & $+0.16$ & $=$ & $1.00$ & $+0.02$ & $+0.03$ & $+0.16$ & $\parallel$ & $1.03$ \\
\rowcolor[HTML]{EFEFEF}
\texttt{sudoku} & $0.74$ & $0.72$ & $0.11$ & $+0.11$ & $+0.12$ & $+0.12$ & $=$ & $1.00$ & $+0.11$ & $+0.12$ & $+0.16$ & $=$ & $1.00$ \\
\texttt{sushi} & $0.16$ & $0.14$ & $0.06$ & $+0.02$ & $+0.11$ & $+0.82$ & $=$ & $1.00$ & $+0.07$ & $+0.13$ & $+0.91$ & $=$ & $1.00$ \\
\rowcolor[HTML]{EFEFEF}
\texttt{warcraft} & $0.78$ & $0.78$ & $0.86$ & $-0.01$ & $-0.01$ & $+0.05$ & $=$ & $1.00$ & $+0.07$ & $+0.07$ & $+0.08$ & $\Leftarrow$ & $1.75$ \\
\bottomrule
\end{tabular}%
}

    \vskip 1em

    \scalebox{0.8}{
\begin{tabular}{l|ccc|ccccc|ccccc}
\toprule
& \multicolumn{3}{c|}{\textsc{Unconstrained}} & 
\multicolumn{5}{c|}{\texttt{olmo3-32b-think}} & 
\multicolumn{5}{c}{\texttt{qwen3-8b}} \\
\midrule
\textsc{Task} & \textsc{F1} $(\uparrow)$ & \textsc{Rec} $(\uparrow)$ & \textsc{Cons} $(\uparrow)$ & $\Delta$\textsc{F1} & $\Delta$\textsc{Rec} & $\Delta$\textsc{Cons} & \frel & \mcr & $\Delta$\textsc{F1} & $\Delta$\textsc{Rec} & $\Delta$\textsc{Cons} & \frel & \mcr \\
\midrule
\texttt{bdd-oia} & $0.58$ & $0.58$ & $0.90$ & -- & -- & -- & -- & -- & $+0.01$ & $+0.02$ & $+0.04$ & $=$ & $1.00$ \\
\rowcolor[HTML]{EFEFEF}
\texttt{bdd-oia-2} & $0.73$ & $0.71$ & $0.92$ & $+0.00$ & $+0.01$ & $+0.02$ & $=$ & $1.00$ & -- & -- & -- & -- & -- \\
\texttt{cebab} & $0.46$ & $0.46$ & $0.63$ & $+0.02$ & $+0.02$ & $+0.11$ & $=$ & $1.00$ & -- & -- & -- & -- & -- \\
\rowcolor[HTML]{EFEFEF}
\texttt{chx} & $0.57$ & $0.53$ & $0.91$ & -- & -- & -- & -- & -- & -- & -- & -- & -- & -- \\
\texttt{cifar10} & $0.87$ & $0.86$ & $0.94$ & $-0.02$ & $-0.01$ & $+0.05$ & $=$ & $1.00$ & $-0.01$ & $+0.00$ & $+0.05$ & $=$ & $1.00$ \\
\rowcolor[HTML]{EFEFEF}
\texttt{cifar100} & $0.52$ & $0.45$ & $0.63$ & $+0.00$ & $+0.02$ & $+0.09$ & $=$ & $1.00$ & $+0.02$ & $+0.04$ & $+0.10$ & $=$ & $1.00$ \\
\texttt{cle4evr} & $1.00$ & $1.00$ & $1.00$ & $-0.00$ & $-0.00$ & $+0.00$ & $=$ & $1.00$ & $-0.00$ & $-0.00$ & $+0.00$ & $=$ & $1.00$ \\
\rowcolor[HTML]{EFEFEF}
\texttt{fashion} & $0.92$ & $0.92$ & $0.99$ & $+0.00$ & $+0.00$ & $+0.00$ & $=$ & $1.00$ & $+0.00$ & $+0.00$ & $+0.00$ & $=$ & $1.00$ \\
\texttt{kand-logic} & $0.72$ & $0.71$ & $0.19$ & -- & -- & -- & -- & -- & -- & -- & -- & -- & -- \\
\rowcolor[HTML]{EFEFEF}
\texttt{kand-logic-2} & $0.79$ & $0.80$ & $0.66$ & $+0.00$ & $+0.00$ & $+0.04$ & $=$ & $1.00$ & $-0.05$ & $-0.05$ & $-0.23$ & $\Rightarrow$ & $0.02$ \\
\texttt{mn-add-bin} & $0.98$ & $0.98$ & $0.98$ & -- & -- & -- & -- & -- & $+0.00$ & $+0.00$ & $+0.01$ & $=$ & $1.00$ \\
\rowcolor[HTML]{EFEFEF}
\texttt{mn-add} & $0.98$ & $0.98$ & $0.99$ & $-0.00$ & $-0.00$ & $-0.00$ & $=$ & $1.00$ & $-0.00$ & $-0.00$ & $+0.00$ & $=$ & $1.00$ \\
\texttt{mn-mul-bin} & $0.71$ & $0.71$ & $0.99$ & $+0.07$ & $+0.07$ & $+0.00$ & $=$ & $1.00$ & -- & -- & -- & -- & -- \\
\rowcolor[HTML]{EFEFEF}
\texttt{mn-mul} & $0.98$ & $0.98$ & $0.99$ & $+0.00$ & $+0.00$ & $+0.01$ & $=$ & $1.00$ & $-0.00$ & $-0.00$ & $+0.00$ & $=$ & $1.00$ \\
\texttt{road-r} & $0.27$ & $0.25$ & $0.65$ & -- & -- & -- & -- & -- & -- & -- & -- & -- & -- \\
\rowcolor[HTML]{EFEFEF}
\texttt{sudoku} & $0.74$ & $0.72$ & $0.11$ & -- & -- & -- & -- & -- & $+0.15$ & $+0.16$ & $+0.32$ & $=$ & $1.00$ \\
\texttt{sushi} & $0.16$ & $0.14$ & $0.06$ & $+0.06$ & $+0.13$ & $+0.91$ & $=$ & $1.00$ & $+0.09$ & $+0.15$ & $+0.89$ & $=$ & $1.00$ \\
\rowcolor[HTML]{EFEFEF}
\texttt{warcraft} & $0.78$ & $0.78$ & $0.86$ & $-0.16$ & $-0.14$ & $+0.01$ & $\Leftarrow$ & $1.75$ & -- & -- & -- & -- & -- \\
\bottomrule
\end{tabular}%
}

    \label{tab:semantic-loss-exp}
\end{table}

\begin{table}[H]
    \centering
    \scriptsize
    \caption{\textbf{Unconstrained model}. The table reports consistency, accuracy, precision and F1. Mean $\pm$ std over 5 seeds.}
    \scalebox{0.9}{
\begin{tabular}{lrrrrr}
\toprule
\textsc{Task} & \textsc{Consistency} $(\uparrow)$ & \textsc{Accuracy} $(\uparrow)$ & \textsc{Precision} $(\uparrow)$ & \textsc{Recall} $(\uparrow)$ & \textsc{F1} $(\uparrow)$ \\
\midrule
\texttt{bdd-oia} & $0.90 \pm 0.00$ & $0.78 \pm 0.00$ & $0.59 \pm 0.01$ & $0.58 \pm 0.01$ & $0.58 \pm 0.00$\\
\rowcolor[HTML]{EFEFEF}
\texttt{bdd-oia-2} & $0.92 \pm 0.00$ & $0.75 \pm 0.00$ & $0.76 \pm 0.00$ & $0.71 \pm 0.01$ & $0.73 \pm 0.01$\\
\texttt{cebab} & $0.63 \pm 0.03$ & $0.64 \pm 0.01$ & $0.48 \pm 0.01$ & $0.46 \pm 0.02$ & $0.46 \pm 0.01$\\
\rowcolor[HTML]{EFEFEF}
\texttt{chx} & $0.91 \pm 0.01$ & $0.69 \pm 0.01$ & $0.68 \pm 0.04$ & $0.53 \pm 0.02$ & $0.57 \pm 0.02$\\
\texttt{cifar10} & $0.94 \pm 0.00$ & $0.96 \pm 0.00$ & $0.89 \pm 0.00$ & $0.86 \pm 0.00$ & $0.87 \pm 0.00$\\
\rowcolor[HTML]{EFEFEF}
\texttt{cifar100} & $0.63 \pm 0.00$ & $0.99 \pm 0.00$ & $0.66 \pm 0.00$ & $0.45 \pm 0.01$ & $0.52 \pm 0.00$\\
\texttt{cle4evr} & $1.00 \pm 0.00$ & $1.00 \pm 0.00$ & $1.00 \pm 0.00$ & $1.00 \pm 0.00$ & $1.00 \pm 0.00$\\
\rowcolor[HTML]{EFEFEF}
\texttt{fashion} & $0.99 \pm 0.00$ & $0.98 \pm 0.00$ & $0.92 \pm 0.00$ & $0.92 \pm 0.00$ & $0.92 \pm 0.00$\\
\texttt{kand-logic} & $0.19 \pm 0.01$ & $0.82 \pm 0.00$ & $0.73 \pm 0.01$ & $0.71 \pm 0.01$ & $0.72 \pm 0.01$\\
\rowcolor[HTML]{EFEFEF}
\texttt{kand-logic-2} & $0.66 \pm 0.05$ & $0.86 \pm 0.01$ & $0.79 \pm 0.01$ & $0.80 \pm 0.01$ & $0.79 \pm 0.01$\\
\texttt{mn-add-bin} & $0.98 \pm 0.01$ & $0.98 \pm 0.00$ & $0.97 \pm 0.00$ & $0.98 \pm 0.01$ & $0.98 \pm 0.00$\\
\rowcolor[HTML]{EFEFEF}
\texttt{mn-add} & $0.99 \pm 0.00$ & $0.98 \pm 0.00$ & $0.98 \pm 0.00$ & $0.98 \pm 0.00$ & $0.98 \pm 0.00$\\
\texttt{mn-mul-bin} & $0.99 \pm 0.00$ & $0.98 \pm 0.00$ & $0.71 \pm 0.04$ & $0.71 \pm 0.04$ & $0.71 \pm 0.04$\\
\rowcolor[HTML]{EFEFEF}
\texttt{mn-mul} & $0.99 \pm 0.00$ & $0.98 \pm 0.00$ & $0.98 \pm 0.00$ & $0.98 \pm 0.00$ & $0.98 \pm 0.00$\\
\texttt{road-r} & $0.65 \pm 0.03$ & $0.94 \pm 0.00$ & $0.37 \pm 0.02$ & $0.25 \pm 0.01$ & $0.27 \pm 0.01$\\
\rowcolor[HTML]{EFEFEF}
\texttt{sudoku} & $0.11 \pm 0.10$ & $0.88 \pm 0.04$ & $0.77 \pm 0.08$ & $0.72 \pm 0.11$ & $0.74 \pm 0.10$\\
\texttt{sushi} & $0.06 \pm 0.01$ & $0.76 \pm 0.00$ & $0.34 \pm 0.04$ & $0.14 \pm 0.01$ & $0.16 \pm 0.01$\\
\rowcolor[HTML]{EFEFEF}
\texttt{warcraft} & $0.86 \pm 0.05$ & $0.91 \pm 0.03$ & $0.80 \pm 0.06$ & $0.78 \pm 0.08$ & $0.78 \pm 0.07$\\
\bottomrule
\end{tabular}
}

    \label{tab:sl-consistency-unconstrained}
\end{table}

\begin{table}[H]
    \centering
    \scriptsize
    \caption{\textbf{\SL logical consistency and downstream metrics with the \textsc{Ground-Truth} formula}. The table reports consistency, accuracy, precision, F1, \frel, and \mcr. Mean $\pm$ std over 5 seeds. \frel and \mcr are trivially $=$/$1.00$.}
    \scalebox{0.9}{
\begin{tabular}{lrrrrrrr}
\toprule
\textsc{Task} & \textsc{Consistency} $(\uparrow)$ & \textsc{Accuracy} $(\uparrow)$ & \textsc{Precision} $(\uparrow)$ & \textsc{Recall} $(\uparrow)$ & \textsc{F1} $(\uparrow)$ & \frel & \mcr \\
\midrule
\texttt{bdd-oia} & $0.92 \pm 0.00$ & $0.78 \pm 0.00$ & $0.59 \pm 0.01$ & $0.58 \pm 0.01$ & $0.58 \pm 0.00$ & $=$ & $1.00$ \\
\rowcolor[HTML]{EFEFEF}
\texttt{bdd-oia-2} & $0.94 \pm 0.01$ & $0.75 \pm 0.01$ & $0.75 \pm 0.01$ & $0.71 \pm 0.01$ & $0.73 \pm 0.01$ & $=$ & $1.00$ \\
\texttt{cebab} & $0.77 \pm 0.02$ & $0.64 \pm 0.01$ & $0.48 \pm 0.01$ & $0.47 \pm 0.02$ & $0.47 \pm 0.01$ & $=$ & $1.00$ \\
\rowcolor[HTML]{EFEFEF}
\texttt{chx} & $0.97 \pm 0.01$ & $0.69 \pm 0.01$ & $0.68 \pm 0.02$ & $0.54 \pm 0.03$ & $0.57 \pm 0.03$ & $=$ & $1.00$ \\
\texttt{cifar10} & $0.99 \pm 0.00$ & $0.95 \pm 0.00$ & $0.86 \pm 0.00$ & $0.84 \pm 0.01$ & $0.85 \pm 0.00$ & $=$ & $1.00$ \\
\rowcolor[HTML]{EFEFEF}
\texttt{cifar100} & $0.70 \pm 0.00$ & $0.99 \pm 0.00$ & $0.64 \pm 0.00$ & $0.47 \pm 0.00$ & $0.53 \pm 0.00$ & $=$ & $1.00$ \\
\texttt{cle4evr} & $1.00 \pm 0.00$ & $1.00 \pm 0.00$ & $1.00 \pm 0.00$ & $1.00 \pm 0.00$ & $1.00 \pm 0.00$ & $=$ & $1.00$ \\
\rowcolor[HTML]{EFEFEF}
\texttt{fashion} & $1.00 \pm 0.00$ & $0.98 \pm 0.00$ & $0.92 \pm 0.00$ & $0.92 \pm 0.00$ & $0.92 \pm 0.00$ & $=$ & $1.00$ \\
\texttt{kand-logic} & $0.21 \pm 0.02$ & $0.82 \pm 0.01$ & $0.73 \pm 0.01$ & $0.72 \pm 0.01$ & $0.72 \pm 0.01$ & $=$ & $1.00$ \\
\rowcolor[HTML]{EFEFEF}
\texttt{kand-logic-2} & $0.70 \pm 0.06$ & $0.87 \pm 0.01$ & $0.80 \pm 0.01$ & $0.80 \pm 0.01$ & $0.80 \pm 0.01$ & $=$ & $1.00$ \\
\texttt{mn-add-bin} & $0.99 \pm 0.00$ & $0.98 \pm 0.00$ & $0.98 \pm 0.00$ & $0.98 \pm 0.00$ & $0.98 \pm 0.00$ & $=$ & $1.00$ \\
\rowcolor[HTML]{EFEFEF}
\texttt{mn-add} & $0.99 \pm 0.00$ & $0.98 \pm 0.00$ & $0.98 \pm 0.00$ & $0.98 \pm 0.00$ & $0.98 \pm 0.00$ & $=$ & $1.00$ \\
\texttt{mn-mul-bin} & $0.99 \pm 0.00$ & $0.98 \pm 0.00$ & $0.74 \pm 0.04$ & $0.74 \pm 0.04$ & $0.74 \pm 0.04$ & $=$ & $1.00$ \\
\rowcolor[HTML]{EFEFEF}
\texttt{mn-mul} & $0.99 \pm 0.00$ & $0.98 \pm 0.00$ & $0.98 \pm 0.00$ & $0.98 \pm 0.00$ & $0.98 \pm 0.00$ & $=$ & $1.00$ \\
\texttt{road-r} & $0.81 \pm 0.04$ & $0.94 \pm 0.00$ & $0.36 \pm 0.01$ & $0.27 \pm 0.01$ & $0.28 \pm 0.01$ & $=$ & $1.00$ \\
\rowcolor[HTML]{EFEFEF}
\texttt{sudoku} & $0.23 \pm 0.10$ & $0.93 \pm 0.02$ & $0.87 \pm 0.04$ & $0.84 \pm 0.05$ & $0.85 \pm 0.05$ & $=$ & $1.00$ \\
\texttt{sushi} & $0.88 \pm 0.07$ & $0.72 \pm 0.00$ & $0.19 \pm 0.05$ & $0.25 \pm 0.01$ & $0.18 \pm 0.02$ & $=$ & $1.00$ \\
\rowcolor[HTML]{EFEFEF}
\texttt{warcraft} & $0.91 \pm 0.01$ & $0.90 \pm 0.02$ & $0.79 \pm 0.06$ & $0.77 \pm 0.06$ & $0.77 \pm 0.06$ & $=$ & $1.00$ \\
\bottomrule
\end{tabular}
}

    \label{tab:sl-consistency-groundtruth}
\end{table}

\begin{table}[H]
    \centering
    \scriptsize
    \caption{\textbf{\SL logical consistency and exact LLM-formula-vs-oracle relationship for the \texttt{gpt-oss-120b} formulas}. The table reports consistency, accuracy, precision, F1, \frel, and \mcr. Mean $\pm$ std over 5 seeds.}
    \scalebox{0.9}{
\begin{tabular}{lrrrrrrr}
\toprule
\textsc{Task} & \textsc{Consistency} $(\uparrow)$ & \textsc{Accuracy} $(\uparrow)$ & \textsc{Precision} $(\uparrow)$ & \textsc{Recall} $(\uparrow)$ & \textsc{F1} $(\uparrow)$ & \frel & \mcr \\
\midrule
\texttt{bdd-oia} & $0.91 \pm 0.01$ & $0.78 \pm 0.00$ & $0.59 \pm 0.00$ & $0.60 \pm 0.01$ & $0.59 \pm 0.01$ & $=$ & $1.00$ \\
\texttt{bdd-oia-2} & $0.92 \pm 0.01$ & $0.75 \pm 0.00$ & $0.75 \pm 0.00$ & $0.72 \pm 0.01$ & $0.73 \pm 0.00$ & $=$ & $1.00$ \\
\texttt{cebab} & $0.80 \pm 0.02$ & $0.64 \pm 0.01$ & $0.48 \pm 0.01$ & $0.47 \pm 0.02$ & $0.47 \pm 0.01$ & $=$ & $1.00$ \\
\rowcolor[HTML]{EFEFEF}
\texttt{chx} & $0.96 \pm 0.01$ & $0.68 \pm 0.01$ & $0.65 \pm 0.04$ & $0.52 \pm 0.01$ & $0.55 \pm 0.01$ & $=$ & $1.00$ \\
\texttt{cifar10} & $0.99 \pm 0.00$ & $0.95 \pm 0.00$ & $0.86 \pm 0.00$ & $0.85 \pm 0.01$ & $0.85 \pm 0.01$ & $=$ & $1.00$ \\
\rowcolor[HTML]{EFEFEF}
\texttt{cifar100} & $0.72 \pm 0.01$ & $0.99 \pm 0.00$ & $0.62 \pm 0.00$ & $0.47 \pm 0.00$ & $0.53 \pm 0.00$ & $=$ & $1.00$ \\
\texttt{cle4evr} & $1.00 \pm 0.00$ & $1.00 \pm 0.00$ & $1.00 \pm 0.00$ & $1.00 \pm 0.00$ & $1.00 \pm 0.00$ & $=$ & $1.00$ \\
\rowcolor[HTML]{EFEFEF}
\texttt{fashion} & $1.00 \pm 0.00$ & $0.98 \pm 0.00$ & $0.92 \pm 0.00$ & $0.92 \pm 0.00$ & $0.92 \pm 0.00$ & $=$ & $1.00$ \\
\texttt{kand-logic} & $0.22 \pm 0.03$ & $0.82 \pm 0.01$ & $0.74 \pm 0.01$ & $0.73 \pm 0.02$ & $0.73 \pm 0.01$ & $=$ & $1.00$ \\
\rowcolor[HTML]{EFEFEF}
\texttt{kand-logic-2} & $0.73 \pm 0.05$ & $0.86 \pm 0.01$ & $0.80 \pm 0.02$ & $0.80 \pm 0.02$ & $0.80 \pm 0.02$ & $=$ & $1.00$ \\
\texttt{mn-add-bin} & $0.99 \pm 0.00$ & $0.98 \pm 0.00$ & $0.98 \pm 0.00$ & $0.98 \pm 0.00$ & $0.98 \pm 0.00$ & $=$ & $1.00$ \\
\rowcolor[HTML]{EFEFEF}
\texttt{mn-add} & $0.99 \pm 0.00$ & $0.98 \pm 0.00$ & $0.98 \pm 0.00$ & $0.98 \pm 0.00$ & $0.98 \pm 0.00$ & $=$ & $1.00$ \\
\texttt{mn-mul-bin} & $0.99 \pm 0.00$ & $0.98 \pm 0.00$ & $0.78 \pm 0.04$ & $0.78 \pm 0.04$ & $0.78 \pm 0.04$ & $=$ & $1.00$ \\
\rowcolor[HTML]{EFEFEF}
\texttt{mn-mul} & $0.99 \pm 0.00$ & $0.99 \pm 0.00$ & $0.98 \pm 0.00$ & $0.98 \pm 0.00$ & $0.98 \pm 0.00$ & $=$ & $1.00$ \\
\texttt{road-r} & $0.82 \pm 0.03$ & $0.94 \pm 0.00$ & $0.39 \pm 0.03$ & $0.28 \pm 0.01$ & $0.29 \pm 0.01$ & $\parallel$ & $1.03$ \\
\rowcolor[HTML]{EFEFEF}
\texttt{sudoku} & $0.26 \pm 0.09$ & $0.93 \pm 0.03$ & $0.86 \pm 0.05$ & $0.84 \pm 0.06$ & $0.85 \pm 0.06$ & $=$ & $1.00$ \\
\texttt{sushi} & $0.97 \pm 0.02$ & $0.72 \pm 0.00$ & $0.26 \pm 0.04$ & $0.27 \pm 0.00$ & $0.23 \pm 0.01$ & $=$ & $1.00$ \\
\rowcolor[HTML]{EFEFEF}
\texttt{warcraft} & $0.94 \pm 0.01$ & $0.94 \pm 0.01$ & $0.86 \pm 0.01$ & $0.85 \pm 0.02$ & $0.85 \pm 0.01$ & $\Leftarrow$ & $1.75$ \\
\bottomrule
\end{tabular}
}

    \label{tab:sl-consistency-gptoss}
\end{table}

\begin{table}[H]
    \centering
    \scriptsize
    \caption{\textbf{\SL logical consistency and exact LLM-formula-vs-oracle relationship for the \texttt{olmo3-32b-think} formulas}. The table reports consistency, accuracy, precision, F1, \frel, and \mcr. Mean $\pm$ std over 5 seeds.}
    \scalebox{0.9}{
\begin{tabular}{lrrrrrrr}
\toprule
\textsc{Task} & \textsc{Consistency} $(\uparrow)$ & \textsc{Accuracy} $(\uparrow)$ & \textsc{Precision} $(\uparrow)$ & \textsc{Recall} $(\uparrow)$ & \textsc{F1} $(\uparrow)$ & \frel & \mcr \\
\midrule
\texttt{bdd-oia-2} & $0.94 \pm 0.00$ & $0.75 \pm 0.00$ & $0.75 \pm 0.00$ & $0.71 \pm 0.01$ & $0.73 \pm 0.00$ & $=$ & $1.00$ \\
\rowcolor[HTML]{EFEFEF}
\texttt{cebab} & $0.74 \pm 0.02$ & $0.66 \pm 0.01$ & $0.49 \pm 0.01$ & $0.48 \pm 0.02$ & $0.48 \pm 0.02$ & $=$ & $1.00$ \\
\texttt{cifar10} & $0.99 \pm 0.00$ & $0.95 \pm 0.00$ & $0.86 \pm 0.01$ & $0.85 \pm 0.00$ & $0.85 \pm 0.01$ & $=$ & $1.00$ \\
\rowcolor[HTML]{EFEFEF}
\texttt{cifar100} & $0.72 \pm 0.02$ & $0.99 \pm 0.00$ & $0.61 \pm 0.01$ & $0.47 \pm 0.01$ & $0.52 \pm 0.01$ & $=$ & $1.00$ \\
\texttt{cle4evr} & $1.00 \pm 0.00$ & $1.00 \pm 0.00$ & $1.00 \pm 0.00$ & $1.00 \pm 0.00$ & $1.00 \pm 0.00$ & $=$ & $1.00$ \\
\rowcolor[HTML]{EFEFEF}
\texttt{fashion} & $1.00 \pm 0.00$ & $0.98 \pm 0.00$ & $0.92 \pm 0.00$ & $0.92 \pm 0.00$ & $0.92 \pm 0.00$ & $=$ & $1.00$ \\
\texttt{kand-logic-2} & $0.70 \pm 0.02$ & $0.86 \pm 0.00$ & $0.80 \pm 0.00$ & $0.80 \pm 0.00$ & $0.80 \pm 0.00$ & $=$ & $1.00$ \\
\rowcolor[HTML]{EFEFEF}
\texttt{mn-add} & $0.99 \pm 0.00$ & $0.98 \pm 0.00$ & $0.98 \pm 0.00$ & $0.98 \pm 0.00$ & $0.98 \pm 0.00$ & $=$ & $1.00$ \\
\texttt{mn-mul-bin} & $0.99 \pm 0.00$ & $0.98 \pm 0.00$ & $0.77 \pm 0.03$ & $0.77 \pm 0.03$ & $0.77 \pm 0.03$ & $=$ & $1.00$ \\
\rowcolor[HTML]{EFEFEF}
\texttt{mn-mul} & $0.99 \pm 0.00$ & $0.99 \pm 0.00$ & $0.99 \pm 0.00$ & $0.99 \pm 0.00$ & $0.99 \pm 0.00$ & $=$ & $1.00$ \\
\texttt{sushi} & $0.96 \pm 0.02$ & $0.72 \pm 0.00$ & $0.29 \pm 0.03$ & $0.27 \pm 0.00$ & $0.23 \pm 0.00$ & $=$ & $1.00$ \\
\rowcolor[HTML]{EFEFEF}
\texttt{warcraft} & $0.87 \pm 0.05$ & $0.86 \pm 0.07$ & $0.65 \pm 0.22$ & $0.64 \pm 0.16$ & $0.62 \pm 0.20$ & $\Leftarrow$ & $1.75$ \\
\bottomrule
\end{tabular}
}

    \label{tab:sl-consistency-olmo3}
\end{table}

\begin{table}[H]
    \centering
    \scriptsize
    \caption{\textbf{\SL logical consistency and exact LLM-formula-vs-oracle relationship for the \texttt{qwen3-8b} formulas}. The table reports consistency, accuracy, precision, F1, \frel, and \mcr. Mean $\pm$ std over 5 seeds.}
    \scalebox{0.9}{
\begin{tabular}{lrrrrrrr}
\toprule
\textsc{Task} & \textsc{Consistency} $(\uparrow)$ & \textsc{Accuracy} $(\uparrow)$ & \textsc{Precision} $(\uparrow)$ & \textsc{Recall} $(\uparrow)$ & \textsc{F1} $(\uparrow)$ & \frel & \mcr \\
\midrule
\texttt{bdd-oia} & $0.94 \pm 0.01$ & $0.78 \pm 0.00$ & $0.59 \pm 0.01$ & $0.60 \pm 0.01$ & $0.59 \pm 0.00$ & $=$ & $1.00$ \\
\rowcolor[HTML]{EFEFEF}
\texttt{cifar10} & $0.99 \pm 0.00$ & $0.96 \pm 0.00$ & $0.87 \pm 0.00$ & $0.86 \pm 0.00$ & $0.86 \pm 0.00$ & $=$ & $1.00$ \\
\texttt{cifar100} & $0.73 \pm 0.00$ & $0.99 \pm 0.00$ & $0.63 \pm 0.00$ & $0.49 \pm 0.00$ & $0.54 \pm 0.00$ & $=$ & $1.00$ \\
\rowcolor[HTML]{EFEFEF}
\texttt{cle4evr} & $1.00 \pm 0.00$ & $1.00 \pm 0.00$ & $1.00 \pm 0.00$ & $1.00 \pm 0.00$ & $1.00 \pm 0.00$ & $=$ & $1.00$ \\
\texttt{fashion} & $1.00 \pm 0.00$ & $0.98 \pm 0.00$ & $0.92 \pm 0.00$ & $0.92 \pm 0.00$ & $0.92 \pm 0.00$ & $=$ & $1.00$ \\
\rowcolor[HTML]{EFEFEF}
\texttt{kand-logic-2} & $0.42 \pm 0.15$ & $0.83 \pm 0.01$ & $0.76 \pm 0.01$ & $0.75 \pm 0.01$ & $0.75 \pm 0.01$ & $\Rightarrow$ & $0.02$ \\
\texttt{mn-add-bin} & $0.99 \pm 0.00$ & $0.98 \pm 0.00$ & $0.97 \pm 0.00$ & $0.98 \pm 0.00$ & $0.98 \pm 0.00$ & $=$ & $1.00$ \\
\rowcolor[HTML]{EFEFEF}
\texttt{mn-add} & $0.99 \pm 0.00$ & $0.98 \pm 0.00$ & $0.98 \pm 0.00$ & $0.98 \pm 0.00$ & $0.98 \pm 0.00$ & $=$ & $1.00$ \\
\texttt{mn-mul} & $0.99 \pm 0.00$ & $0.98 \pm 0.00$ & $0.98 \pm 0.00$ & $0.98 \pm 0.00$ & $0.98 \pm 0.00$ & $=$ & $1.00$ \\
\rowcolor[HTML]{EFEFEF}
\texttt{sudoku} & $0.42 \pm 0.09$ & $0.94 \pm 0.02$ & $0.89 \pm 0.03$ & $0.88 \pm 0.04$ & $0.89 \pm 0.04$ & $=$ & $1.00$ \\
\texttt{sushi} & $0.95 \pm 0.02$ & $0.72 \pm 0.00$ & $0.34 \pm 0.03$ & $0.29 \pm 0.00$ & $0.26 \pm 0.01$ & $=$ & $1.00$ \\
\bottomrule
\end{tabular}
}

    \label{tab:sl-consistency-qwen3}
\end{table}

\clearpage

\subsubsection{Soft Constraints with  Semantic Loss equipped with LLM-Generated Formulas from Less Detailed Prompts}

As for \SPL in~\cref{sec:spl-less-detailed}, we replicate the same exact analysis for \SL, \ie \cpmpy with CoT and less detailed prompts.

\begin{table}[H]
    \centering
    \scriptsize
    \caption{\textbf{With non-detailed prompts, \SL benefits from correct formulas, while non-equivalent constraints degrade logical consistency.}
    Formulas are generated with \cpmpy + CoT from non-detailed prompts; the setup is identical to \cref{tab:semantic-loss-exp}.
    The top table reports \SL with the ground-truth (\textsc{G-T}) and \texttt{gpt-oss-120b} formulas, the bottom one with the \texttt{olmo3-32b-think} and \texttt{qwen3-8b} generated formulas.
    For the \textsc{Unconstrained} we report F1, Recall (\textsc{Rec}) and consistency (\textsc{Cons}, the fraction of predictions that satisfy the ground-truth formula); every other method reports $\Delta$\textsc{F1}/$\Delta$\textsc{Rec}/$\Delta$\textsc{Cons} of \SL relative to \textsc{Unconstrained}.
    Standard deviations across $5$ seeds are omitted. Full results are reported in \cref{app:additional_results}.
    \frel denotes the relationship between the predicted and ground-truth formulas: \texttt{equal} ($=$), \texttt{strict} ($\Rightarrow$), \texttt{permissive} ($\Leftarrow$), or \texttt{incomparable} ($\parallel$), indicating whether the predicted formula satisfying assignments are equal to, a subset of, a superset of, or neither relative to the ground truth; \texttt{unknown} indicates cases where the model count did not terminate.
    \mcr measures the ratio of the number of satisfying assignments of the predicted formula to that of the ground-truth formula.
    Missing cells indicate a degenerate or no formula, as for \SPL (see \cref{tab:spl-coverage-gaps-natural} for the full explanation).}
    \scalebox{0.8}{
\begin{tabular}{l|ccc|ccccc|ccccc}
\toprule
& \multicolumn{3}{c|}{\textsc{Unconstrained}} & \multicolumn{5}{c|}{\textsc{G-T}} & \multicolumn{5}{c}{\texttt{gpt-oss-120b}} \\
\midrule
\textsc{Task} & \textsc{F1} $(\uparrow)$ & \textsc{Rec} $(\uparrow)$ & \textsc{Cons} $(\uparrow)$ & $\Delta$\textsc{F1} & $\Delta$\textsc{Rec} & $\Delta$\textsc{Cons} & \frel & \mcr & $\Delta$\textsc{F1} & $\Delta$\textsc{Rec} & $\Delta$\textsc{Cons} & \frel & \mcr \\
\midrule
\texttt{bdd-oia} & $0.58$ & $0.58$ & $0.90$ & $-0.00$ & $+0.01$ & $+0.02$ & $=$ & $1.00$ & -- & -- & -- & -- & -- \\
\rowcolor[HTML]{EFEFEF}
\texttt{bdd-oia-2} & $0.73$ & $0.71$ & $0.92$ & $-0.00$ & $+0.00$ & $+0.02$ & $=$ & $1.00$ & $-0.01$ & $-0.02$ & $-0.02$ & $\parallel$ & $1.00$ \\
\texttt{cebab} & $0.46$ & $0.46$ & $0.63$ & $+0.00$ & $+0.00$ & $+0.13$ & $=$ & $1.00$ & -- & -- & -- & -- & -- \\
\rowcolor[HTML]{EFEFEF}
\texttt{chx} & $0.57$ & $0.53$ & $0.91$ & $-0.00$ & $+0.00$ & $+0.07$ & $=$ & $1.00$ & $-0.21$ & $-0.14$ & $-0.17$ & $\Rightarrow$ & $0.94$ \\
\texttt{cifar10} & $0.87$ & $0.86$ & $0.94$ & $-0.02$ & $-0.01$ & $+0.05$ & $=$ & $1.00$ & $-0.20$ & $-0.17$ & $-0.22$ & $\parallel$ & $3.38$ \\
\rowcolor[HTML]{EFEFEF}
\texttt{cifar100} & $0.52$ & $0.45$ & $0.63$ & $+0.01$ & $+0.03$ & $+0.07$ & $=$ & $1.00$ & $+0.01$ & $+0.03$ & $+0.09$ & $=$ & $1.00$ \\
\texttt{cle4evr} & $1.00$ & $1.00$ & $1.00$ & $+0.00$ & $+0.00$ & $+0.00$ & $=$ & $1.00$ & $-0.01$ & $-0.01$ & $-0.01$ & $=$ & $1.00$ \\
\rowcolor[HTML]{EFEFEF}
\texttt{fashion} & $0.92$ & $0.92$ & $0.99$ & $+0.00$ & $+0.00$ & $+0.00$ & $=$ & $1.00$ & $+0.00$ & $+0.00$ & $+0.00$ & $=$ & $1.00$ \\
\texttt{kand-logic} & $0.72$ & $0.71$ & $0.19$ & $+0.00$ & $+0.00$ & $+0.02$ & $=$ & $1.00$ & $+0.01$ & $+0.01$ & $+0.02$ & unknown & -- \\
\rowcolor[HTML]{EFEFEF}
\texttt{kand-logic-2} & $0.79$ & $0.80$ & $0.66$ & $+0.00$ & $+0.00$ & $+0.04$ & $=$ & $1.00$ & $-0.09$ & $-0.11$ & $-0.31$ & $\Rightarrow$ & $0.02$ \\
\texttt{mn-add-bin} & $0.98$ & $0.98$ & $0.98$ & $+0.00$ & $+0.00$ & $+0.01$ & $=$ & $1.00$ & $+0.00$ & $+0.00$ & $+0.01$ & $=$ & $1.00$ \\
\rowcolor[HTML]{EFEFEF}
\texttt{mn-add} & $0.98$ & $0.98$ & $0.99$ & $+0.00$ & $+0.00$ & $+0.00$ & $=$ & $1.00$ & $-0.01$ & $-0.01$ & $-0.01$ & $=$ & $1.00$ \\
\texttt{mn-mul-bin} & $0.71$ & $0.71$ & $0.99$ & $+0.03$ & $+0.03$ & $+0.00$ & $=$ & $1.00$ & $+0.03$ & $+0.03$ & $+0.00$ & $=$ & $1.00$ \\
\rowcolor[HTML]{EFEFEF}
\texttt{mn-mul} & $0.98$ & $0.98$ & $0.99$ & $+0.00$ & $+0.00$ & $+0.00$ & $=$ & $1.00$ & $+0.00$ & $+0.00$ & $+0.01$ & $=$ & $1.00$ \\
\texttt{road-r} & $0.27$ & $0.25$ & $0.65$ & $+0.01$ & $+0.02$ & $+0.16$ & $=$ & $1.00$ & $+0.02$ & $+0.02$ & $+0.01$ & $\parallel$ & $76.64$ \\
\rowcolor[HTML]{EFEFEF}
\texttt{sudoku} & $0.74$ & $0.72$ & $0.11$ & $+0.11$ & $+0.12$ & $+0.12$ & $=$ & $1.00$ & $+0.08$ & $+0.09$ & $+0.10$ & $=$ & $1.00$ \\
\texttt{sushi} & $0.16$ & $0.14$ & $0.06$ & $+0.02$ & $+0.11$ & $+0.82$ & $=$ & $1.00$ & $+0.08$ & $+0.14$ & $+0.90$ & $=$ & $1.00$ \\
\rowcolor[HTML]{EFEFEF}
\texttt{warcraft} & $0.78$ & $0.78$ & $0.86$ & $-0.01$ & $-0.01$ & $+0.05$ & $=$ & $1.00$ & $+0.06$ & $+0.06$ & $+0.07$ & $\Leftarrow$ & $1.75$ \\
\bottomrule
\end{tabular}%
}

    \scalebox{0.8}{
\begin{tabular}{l|ccc|ccccc|ccccc}
\toprule
& \multicolumn{3}{c|}{\textsc{Unconstrained}} & \multicolumn{5}{c|}{\texttt{olmo3-32b-think}} & \multicolumn{5}{c}{\texttt{qwen3-8b}} \\
\midrule
\textsc{Task} & \textsc{F1} $(\uparrow)$ & \textsc{Rec} $(\uparrow)$ & \textsc{Cons} $(\uparrow)$ & $\Delta$\textsc{F1} & $\Delta$\textsc{Rec} & $\Delta$\textsc{Cons} & \frel & \mcr & $\Delta$\textsc{F1} & $\Delta$\textsc{Rec} & $\Delta$\textsc{Cons} & \frel & \mcr \\
\midrule
\texttt{bdd-oia} & $0.58$ & $0.58$ & $0.90$ & -- & -- & -- & -- & --  & -- & -- & -- & -- & --  \\
\rowcolor[HTML]{EFEFEF}
\texttt{bdd-oia-2} & $0.73$ & $0.71$ & $0.92$ & -- & -- & -- & -- & -- & -- & -- & -- & -- & --  \\
\texttt{cebab} & $0.46$ & $0.46$ & $0.63$ & -- & -- & -- & -- & --  & -- & -- & -- & -- & --  \\
\rowcolor[HTML]{EFEFEF}
\texttt{chx} & $0.57$ & $0.53$ & $0.91$ & $-0.09$ & $-0.00$ & $-0.88$ & $\parallel$ & $1.00$ & $-0.41$ & $-0.28$ & $-0.05$ & -- & -- \\
\texttt{cifar10} & $0.87$ & $0.86$ & $0.94$ & $-0.01$ & $+0.00$ & $+0.02$ & $\Leftarrow$ & $19.25$ & $+0.00$ & $+0.01$ & $+0.02$ & $\Leftarrow$ & $19.62$ \\
\rowcolor[HTML]{EFEFEF}
\texttt{cifar100} & $0.52$ & $0.45$ & $0.63$ & $+0.00$ & $+0.02$ & $+0.09$ & $=$ & $1.00$ & $+0.02$ & $+0.04$ & $+0.10$ & $=$ & $1.00$ \\
\texttt{cle4evr} & $1.00$ & $1.00$ & $1.00$ & $-0.00$ & $-0.00$ & $+0.00$ & $=$ & $1.00$ & $-0.00$ & $-0.00$ & $+0.00$ & $=$ & $1.00$ \\
\rowcolor[HTML]{EFEFEF}
\texttt{fashion} & $0.92$ & $0.92$ & $0.99$ & $+0.00$ & $+0.00$ & $+0.00$ & $=$ & $1.00$ & $-0.00$ & $-0.00$ & $-0.00$ & $=$ & $1.00$ \\
\texttt{kand-logic} & $0.72$ & $0.71$ & $0.19$ & -- & -- & -- & -- & -- & $-0.02$ & $-0.02$ & $-0.02$ & unknown & -- \\
\rowcolor[HTML]{EFEFEF}
\texttt{kand-logic-2} & $0.79$ & $0.80$ & $0.66$ & $-0.09$ & $-0.11$ & $-0.29$ & $\Rightarrow$ & $0.02$ & $-0.09$ & $-0.14$ & $-0.53$ & $\parallel$ & $0.19$ \\
\texttt{mn-add-bin} & $0.98$ & $0.98$ & $0.98$ & $-0.33$ & $-0.27$ & $-0.30$ & $\parallel$ & $2.56$ & $+0.00$ & $+0.00$ & $+0.01$ & $\Leftarrow$ & $2.56$ \\
\rowcolor[HTML]{EFEFEF}
\texttt{mn-add} & $0.98$ & $0.98$ & $0.99$ & $-0.00$ & $-0.00$ & $-0.00$ & $=$ & $1.00$ & $+0.00$ & $+0.00$ & $+0.00$ & $=$ & $1.00$ \\
\texttt{mn-mul-bin} & $0.71$ & $0.71$ & $0.99$ & \multicolumn{5}{c|}{--} & $-0.20$ & $-0.19$ & $-0.11$ & $\parallel$ & $2.56$ \\
\rowcolor[HTML]{EFEFEF}
\texttt{mn-mul} & $0.98$ & $0.98$ & $0.99$ & $+0.00$ & $+0.00$ & $+0.01$ & $=$ & $1.00$ & -- & -- & -- & -- & --  \\
\texttt{road-r} & $0.27$ & $0.25$ & $0.65$ & $+0.01$ & $+0.01$ & $+0.06$ & $\parallel$ & $242.22$ & $-0.01$ & $+0.01$ & $-0.28$ & $\parallel$ & $7.81$ \\
\rowcolor[HTML]{EFEFEF}
\texttt{sudoku} & $0.74$ & $0.72$ & $0.11$ & $+0.11$ & $+0.12$ & $+0.13$ & $=$ & $1.00$ & $+0.15$ & $+0.16$ & $+0.31$ & $=$ & $1.00$ \\
\texttt{sushi} & $0.16$ & $0.14$ & $0.06$ & $+0.07$ & $+0.13$ & $+0.91$ & $=$ & $1.00$ & $+0.08$ & $+0.14$ & $+0.89$ & $=$ & $1.00$ \\
\rowcolor[HTML]{EFEFEF}
\texttt{warcraft} & $0.78$ & $0.78$ & $0.86$ & -- & -- & -- & -- & --  & -- & -- & -- & -- & -- \\
\bottomrule
\end{tabular}%
}

    \label{tab:sl-non-detailed}
\end{table}

\textbf{Discussion}. \Cref{tab:sl-non-detailed} propose the same \SL experiments from \cref{tab:semantic-loss-exp} but using formulas generated from non-detailed prompts.
\texttt{equal} formulas ($=$) behave similarly to those generated from detailed prompts: they never reduce consistency by more than $0.01$ and provide the same gains on all the tasks, \eg \texttt{sushi} ($+0.89$ to $+0.91$ \textsc{Cons}) and \texttt{sudoku} ($+0.10$ to $+0.31$ \textsc{Cons}, $+0.08$ to $+0.15$ F1), while leaving downstream performance mostly unaffected.
However, only about half of the produced formulas are exactly correct ($9/16$, $7/12$, and $6/13$ for \texttt{gpt-oss-120b}, \texttt{olmo3-32b-think}, and \texttt{qwen3-8b}, respectively), and the non-\texttt{equal} formulas are often harmful. With detailed prompts, only a single formula reduced consistency; here, the \texttt{strict} and \texttt{incomparable} formulas reduce consistency on most of the tasks, often by a large margin: \texttt{kand-logic-2} ($-0.29$ to $-0.53$ Cons for all three models), \texttt{chx} ($-0.88$ for \texttt{olmo3-32b-think}), \texttt{mn-add-bin} ($-0.30$ for \texttt{olmo3-32b-think}), \texttt{road-r} ($-0.28$ for \texttt{qwen3-8b}), and \texttt{cifar10} ($-0.22$ for \texttt{gpt-oss-120b}).
These formulas also reduce F1, \eg by up to $0.41$ for \texttt{qwen3-8b}/\texttt{chx}, whose formula admits only a single assignment of the concepts. Unlike \SPL, however, \SL can still be trained with such a formula as it is used as a loss function.
The reason is the same as for the too-\texttt{strict} formula in \cref{tab:semantic-loss-exp}. If the formula excludes some assignments that are actually present in the training data, the two losses give conflicting signals: the classification loss encourages the model to predict the ground-truth assignments, while the semantic loss pushes it away from them because they violate the formula. The model therefore ends up in a compromise that may satisfy neither the formula nor the ground-truth labels.
As a result, the average consistency gain of \SL over the unconstrained model drops from about $+0.10$ with detailed prompts to $+0.03$ for \texttt{gpt-oss-120b} and \texttt{qwen3-8b}, and becomes negative for \texttt{olmo3-32b-think} ($-0.02$). The mean $\Delta$F1 ranges from $-0.04$ to $-0.01$.
Permissive formulas ($\Leftarrow$) remain harmless, as with detailed prompts and for \SPL. On \texttt{cifar10}, formulas admitting about $19\times$ more assignments than the ground truth still slightly improve consistency ($+0.02$). Similarly, the permissive \texttt{gpt-oss-120b} formula for \texttt{warcraft} improves both F1 ($+0.06$) and consistency ($+0.07$), since a superset of the valid assignments can still exclude some invalid ones.
Overall, \SL is more robust than \SPL to imperfect formulas because it uses the formula as a soft training signal rather than enforcing it on every prediction. As a result, incorrect formulas do not necessarily prevent training, although they can still hurt performance when they conflict with the ground truth.

\begin{table}[H]
    \centering
    \scriptsize
    \caption{\textbf{\SL results with \texttt{qwen3-8b} generated formulas using \cpmpy output, CoT and detailed prompt}. The table reports consistency, accuracy, precision, F1, \frel, and \mcr. Mean $\pm$ std over 5 seeds.}
    \label{tab:sl-natural-consistency-qwen3}
    \scalebox{0.9}{
\begin{tabular}{lrrrrrrr}
\toprule
\textsc{Task} & \textsc{Consistency} $(\uparrow)$ & \textsc{Accuracy} $(\uparrow)$ & \textsc{Precision} $(\uparrow)$ & \textsc{Recall} $(\uparrow)$ & \textsc{F1} $(\uparrow)$ & \frel & \mcr \\
\midrule
\texttt{chx} & $0.86 \pm 0.10$ & $0.43 \pm 0.01$ & $0.30 \pm 0.11$ & $0.26 \pm 0.01$ & $0.16 \pm 0.02$ & -- & -- \\
\rowcolor[HTML]{EFEFEF}
\texttt{cifar10} & $0.96 \pm 0.00$ & $0.96 \pm 0.00$ & $0.87 \pm 0.00$ & $0.87 \pm 0.00$ & $0.87 \pm 0.00$ & $\Leftarrow$ & $19.62$ \\
\texttt{cifar100} & $0.73 \pm 0.00$ & $0.99 \pm 0.00$ & $0.63 \pm 0.00$ & $0.49 \pm 0.00$ & $0.54 \pm 0.00$ & $=$ & $1.00$ \\
\rowcolor[HTML]{EFEFEF}
\texttt{cle4evr} & $1.00 \pm 0.00$ & $1.00 \pm 0.00$ & $1.00 \pm 0.00$ & $1.00 \pm 0.00$ & $1.00 \pm 0.00$ & $=$ & $1.00$ \\
\texttt{fashion} & $0.99 \pm 0.00$ & $0.98 \pm 0.00$ & $0.92 \pm 0.00$ & $0.91 \pm 0.00$ & $0.92 \pm 0.00$ & $=$ & $1.00$ \\
\rowcolor[HTML]{EFEFEF}
\texttt{kand-logic} & $0.17 \pm 0.02$ & $0.80 \pm 0.01$ & $0.75 \pm 0.01$ & $0.70 \pm 0.01$ & $0.70 \pm 0.01$ & unknown & -- \\
\texttt{kand-logic-2} & $0.13 \pm 0.02$ & $0.82 \pm 0.00$ & $0.80 \pm 0.01$ & $0.66 \pm 0.02$ & $0.70 \pm 0.01$ & $\parallel$ & $0.19$ \\
\rowcolor[HTML]{EFEFEF}
\texttt{mn-add-bin} & $0.99 \pm 0.00$ & $0.98 \pm 0.00$ & $0.98 \pm 0.00$ & $0.98 \pm 0.01$ & $0.98 \pm 0.00$ & $\Leftarrow$ & $2.56$ \\
\texttt{mn-add} & $0.99 \pm 0.00$ & $0.98 \pm 0.00$ & $0.98 \pm 0.00$ & $0.98 \pm 0.00$ & $0.98 \pm 0.00$ & $=$ & $1.00$ \\
\rowcolor[HTML]{EFEFEF}
\texttt{mn-mul-bin} & $0.88 \pm 0.01$ & $0.87 \pm 0.01$ & $0.54 \pm 0.03$ & $0.51 \pm 0.02$ & $0.51 \pm 0.02$ & $\parallel$ & $2.56$ \\
\texttt{road-r} & $0.38 \pm 0.05$ & $0.93 \pm 0.00$ & $0.36 \pm 0.03$ & $0.26 \pm 0.01$ & $0.25 \pm 0.02$ & $\parallel$ & $7.81$ \\
\rowcolor[HTML]{EFEFEF}
\texttt{sudoku} & $0.42 \pm 0.16$ & $0.95 \pm 0.03$ & $0.90 \pm 0.06$ & $0.88 \pm 0.06$ & $0.89 \pm 0.06$ & $=$ & $1.00$ \\
\texttt{sushi} & $0.95 \pm 0.01$ & $0.72 \pm 0.00$ & $0.30 \pm 0.01$ & $0.28 \pm 0.00$ & $0.24 \pm 0.01$ & $=$ & $1.00$ \\
\bottomrule
\end{tabular}
}

\end{table}

\begin{table}[H]
    \centering
    \scriptsize
    \caption{\textbf{\SL results with \texttt{olmo-32b-think} generated formulas using \cpmpy output, CoT and detailed prompt}. The table reports consistency, accuracy, precision, F1, \frel, and \mcr. Mean $\pm$ std over 5 seeds.}
    \label{tab:sl-natural-consistency-olmo3}
    \scalebox{0.9}{
\begin{tabular}{lrrrrrrr}
\toprule
\textsc{Task} & \textsc{Consistency} $(\uparrow)$ & \textsc{Accuracy} $(\uparrow)$ & \textsc{Precision} $(\uparrow)$ & \textsc{Recall} $(\uparrow)$ & \textsc{F1} $(\uparrow)$ & \frel & \mcr \\
\midrule
\texttt{chx} & $0.02 \pm 0.01$ & $0.53 \pm 0.02$ & $0.62 \pm 0.02$ & $0.53 \pm 0.01$ & $0.48 \pm 0.01$ & $\parallel$ & $1.00$ \\
\rowcolor[HTML]{EFEFEF}
\texttt{cifar10} & $0.97 \pm 0.00$ & $0.96 \pm 0.00$ & $0.87 \pm 0.00$ & $0.86 \pm 0.01$ & $0.86 \pm 0.01$ & $\Leftarrow$ & $19.25$ \\
\texttt{cifar100} & $0.72 \pm 0.01$ & $0.99 \pm 0.00$ & $0.61 \pm 0.01$ & $0.47 \pm 0.00$ & $0.52 \pm 0.00$ & $=$ & $1.00$ \\
\rowcolor[HTML]{EFEFEF}
\texttt{cle4evr} & $1.00 \pm 0.00$ & $1.00 \pm 0.00$ & $1.00 \pm 0.00$ & $1.00 \pm 0.00$ & $1.00 \pm 0.00$ & $=$ & $1.00$ \\
\texttt{fashion} & $1.00 \pm 0.00$ & $0.98 \pm 0.00$ & $0.92 \pm 0.00$ & $0.92 \pm 0.00$ & $0.92 \pm 0.00$ & $=$ & $1.00$ \\
\rowcolor[HTML]{EFEFEF}
\texttt{kand-logic-2} & $0.36 \pm 0.01$ & $0.81 \pm 0.00$ & $0.72 \pm 0.01$ & $0.69 \pm 0.01$ & $0.70 \pm 0.01$ & $\Rightarrow$ & $0.02$ \\
\texttt{mn-add-bin} & $0.69 \pm 0.03$ & $0.68 \pm 0.03$ & $0.67 \pm 0.01$ & $0.70 \pm 0.03$ & $0.65 \pm 0.03$ & $\parallel$ & $2.56$ \\
\rowcolor[HTML]{EFEFEF}
\texttt{mn-add} & $0.99 \pm 0.00$ & $0.98 \pm 0.00$ & $0.98 \pm 0.00$ & $0.98 \pm 0.00$ & $0.98 \pm 0.00$ & $=$ & $1.00$ \\
\texttt{mn-mul} & $0.99 \pm 0.00$ & $0.99 \pm 0.00$ & $0.99 \pm 0.00$ & $0.98 \pm 0.00$ & $0.98 \pm 0.00$ & $=$ & $1.00$ \\
\rowcolor[HTML]{EFEFEF}
\texttt{road-r} & $0.72 \pm 0.02$ & $0.94 \pm 0.00$ & $0.39 \pm 0.03$ & $0.27 \pm 0.00$ & $0.28 \pm 0.01$ & $\parallel$ & $242.22$ \\
\texttt{sudoku} & $0.24 \pm 0.09$ & $0.92 \pm 0.03$ & $0.86 \pm 0.06$ & $0.83 \pm 0.07$ & $0.84 \pm 0.07$ & $=$ & $1.00$ \\
\rowcolor[HTML]{EFEFEF}
\texttt{sushi} & $0.96 \pm 0.02$ & $0.72 \pm 0.00$ & $0.29 \pm 0.03$ & $0.27 \pm 0.01$ & $0.23 \pm 0.01$ & $=$ & $1.00$ \\
\bottomrule
\end{tabular}
}

\end{table}

\begin{table}[H]
    \centering
    \scriptsize
    \caption{\textbf{\SL results with \texttt{gpt-oss-120b} generated formulas using \cpmpy output, CoT and detailed prompt}. The table reports consistency, accuracy, precision, F1, \frel, and \mcr. Mean $\pm$ std over 5 seeds.}
    \label{tab:sl-natural-consistency-gptoss}
    \scalebox{0.9}{
\begin{tabular}{lrrrrrrr}
\toprule
\textsc{Task} & \textsc{Consistency} $(\uparrow)$ & \textsc{Accuracy} $(\uparrow)$ & \textsc{Precision} $(\uparrow)$ & \textsc{Recall} $(\uparrow)$ & \textsc{F1} $(\uparrow)$ & \frel & \mcr \\
\midrule
\texttt{bdd-oia-simple} & $0.91 \pm 0.02$ & $0.74 \pm 0.00$ & $0.75 \pm 0.00$ & $0.69 \pm 0.01$ & $0.72 \pm 0.00$ & $\parallel$ & $1.00$ \\
\rowcolor[HTML]{EFEFEF}
\texttt{chx} & $0.74 \pm 0.10$ & $0.45 \pm 0.02$ & $0.48 \pm 0.11$ & $0.39 \pm 0.02$ & $0.36 \pm 0.03$ & $\Rightarrow$ & $0.94$ \\
\texttt{cifar10} & $0.73 \pm 0.09$ & $0.91 \pm 0.01$ & $0.83 \pm 0.02$ & $0.68 \pm 0.01$ & $0.67 \pm 0.02$ & $\parallel$ & $3.38$ \\
\rowcolor[HTML]{EFEFEF}
\texttt{cifar100} & $0.72 \pm 0.01$ & $0.99 \pm 0.00$ & $0.62 \pm 0.00$ & $0.48 \pm 0.00$ & $0.53 \pm 0.00$ & $=$ & $1.00$ \\
\texttt{cle4evr} & $0.99 \pm 0.02$ & $0.99 \pm 0.02$ & $0.99 \pm 0.02$ & $0.99 \pm 0.03$ & $0.99 \pm 0.03$ & $=$ & $1.00$ \\
\rowcolor[HTML]{EFEFEF}
\texttt{fashion-} & $1.00 \pm 0.00$ & $0.98 \pm 0.00$ & $0.92 \pm 0.00$ & $0.92 \pm 0.00$ & $0.92 \pm 0.00$ & $=$ & $1.00$ \\
\texttt{kand-logic} & $0.21 \pm 0.03$ & $0.82 \pm 0.01$ & $0.74 \pm 0.01$ & $0.73 \pm 0.01$ & $0.73 \pm 0.01$ & unknown & -- \\
\rowcolor[HTML]{EFEFEF}
\texttt{kand-logic-2} & $0.35 \pm 0.01$ & $0.81 \pm 0.00$ & $0.72 \pm 0.01$ & $0.69 \pm 0.01$ & $0.70 \pm 0.00$ & $\Rightarrow$ & $0.02$ \\
\texttt{mn-add-bin} & $0.99 \pm 0.00$ & $0.98 \pm 0.00$ & $0.98 \pm 0.00$ & $0.98 \pm 0.00$ & $0.98 \pm 0.00$ & $=$ & $1.00$ \\
\rowcolor[HTML]{EFEFEF}
\texttt{mn-add} & $0.98 \pm 0.00$ & $0.98 \pm 0.00$ & $0.98 \pm 0.00$ & $0.98 \pm 0.00$ & $0.98 \pm 0.00$ & $=$ & $1.00$ \\
\texttt{mn-mul-bin} & $0.99 \pm 0.00$ & $0.98 \pm 0.00$ & $0.74 \pm 0.04$ & $0.74 \pm 0.04$ & $0.74 \pm 0.04$ & $=$ & $1.00$ \\
\rowcolor[HTML]{EFEFEF}
\texttt{mn-mul} & $0.99 \pm 0.00$ & $0.99 \pm 0.00$ & $0.98 \pm 0.00$ & $0.98 \pm 0.00$ & $0.98 \pm 0.00$ & $=$ & $1.00$ \\
\texttt{road-r} & $0.67 \pm 0.03$ & $0.95 \pm 0.00$ & $0.40 \pm 0.02$ & $0.27 \pm 0.01$ & $0.29 \pm 0.01$ & $\parallel$ & $76.64$ \\
\rowcolor[HTML]{EFEFEF}
\texttt{sudoku} & $0.21 \pm 0.06$ & $0.91 \pm 0.02$ & $0.84 \pm 0.05$ & $0.81 \pm 0.05$ & $0.82 \pm 0.05$ & $=$ & $1.00$ \\
\texttt{sushi} & $0.95 \pm 0.02$ & $0.72 \pm 0.00$ & $0.29 \pm 0.02$ & $0.28 \pm 0.01$ & $0.24 \pm 0.02$ & $=$ & $1.00$ \\
\rowcolor[HTML]{EFEFEF}
\texttt{warcraft} & $0.93 \pm 0.01$ & $0.93 \pm 0.01$ & $0.86 \pm 0.01$ & $0.84 \pm 0.02$ & $0.84 \pm 0.02$ & $\Leftarrow$ & $1.75$ \\
\bottomrule
\end{tabular}
}

\end{table}

\end{document}